%% file: main.tex
\pdfoutput=1
\documentclass{article}
\usepackage{ijcai25}

\usepackage{times}
\usepackage{soul}
\usepackage{url}
\usepackage[hidelinks]{hyperref}
\usepackage[utf8]{inputenc}
\usepackage[small]{caption}
\usepackage{graphicx}
\usepackage{amssymb}
\usepackage{amsmath}
\usepackage{amsthm}
\usepackage{booktabs}
\usepackage{enumitem}
\usepackage{acronym}
\usepackage{subcaption}
\usepackage{thmtools, thm-restate}
\usepackage{tikz}
\usetikzlibrary{arrows,arrows.meta,automata,backgrounds,calc,patterns,positioning,shapes,shadows}
\usepackage{pgfplots}
\pgfplotsset{compat=1.18}
\usepackage{algorithm}
\usepackage[noend]{algpseudocode}
\algnewcommand{\algorithmicbreak}{\textbf{break}}
\algnewcommand{\algorithmiccontinue}{\textbf{continue}}
\algnewcommand{\algorithmicforeach}{\textbf{for each}}
\algnewcommand\Break{\State \algorithmicbreak}
\algnewcommand\Continue{\State \algorithmiccontinue}
\algdef{SE}[FOR]{ForEach}{EndForEach}[1]
	{\algorithmicforeach\ #1\ \algorithmicdo}
	{\algorithmicend\ \algorithmicforeach}
\algnewcommand{\LeftComment}[1]{\Statex \(\triangleright\) #1}
\algrenewcommand{\algorithmicindent}{1.15em}
\makeatletter
	\newcounter{HALG@line}
	\renewcommand{\theHALG@line}{\thealgorithm.\arabic{ALG@line}}
\makeatother
\usepackage{natbib}
\usepackage[capitalise]{cleveref}
\usepackage[switch]{lineno}
\usepackage{csquotes}

\input{defs.tex}

\newtheorem{theorem}{Theorem}
\newtheorem{corollary}[theorem]{Corollary}
\newtheorem{definition}{Definition}
\newtheorem{example}{Example}

\newtheorem{lemma}[theorem]{Lemma}

\title{Approximate Lifted Model Construction\thanks{Extended
version of paper accepted to the Proceedings of the 34th International Joint Conference on Artificial Intelligence (IJCAI-2025).}}

\author{
	Malte Luttermann$^1$ \and
	Jan Speller$^2$ \and
	Marcel Gehrke$^3$ \and
	Tanya Braun$^2$ \and
	Ralf Möller$^3$ \And
	Mattis Hartwig$^1$ \\
	\affiliations
	$^1$German Research Center for Artificial Intelligence (DFKI), Lübeck, Germany \\
	$^2$Data Science Group, University of Münster, Germany \\
	$^3$Institute for Humanities-Centered Artificial Intelligence, University of Hamburg, Germany \\
	\emails
	\{malte.luttermann,mattis.hartwig\}@dfki.de,
	\{marcel.gehrke,ralf.moeller\}@uni-hamburg.de,
	\{jan.speller,tanya.braun\}@uni-muenster.de
}

\begin{document}
\maketitle
\begin{abstract}
	Probabilistic relational models such as \aclp{pfg} enable efficient (lifted) inference by exploiting the indistinguishability of objects.
	In lifted inference, a representative of indistinguishable objects is used for computations.
	To obtain a relational (i.e., lifted) representation, the \ac{acp} algorithm is the state of the art.
	The \ac{acp} algorithm, however, requires underlying distributions, encoded as potential-based factorisations, to exactly match to identify and exploit indistinguishabilities.
	Hence, \ac{acp} is unsuitable for practical applications where potentials learned from data inevitably deviate even if associated objects are indistinguishable.
	To mitigate this problem, we introduce the \ac{eacp} algorithm, which allows for a deviation of potentials depending on a hyperparameter $\varepsilon$.
	\ac{eacp} efficiently uncovers and exploits indistinguishabilities that are not exact.
	We prove that the approximation error induced by \ac{eacp} is strictly bounded and our experiments show that the approximation error is close to zero in practice.
\end{abstract}

\acresetall

\section{Introduction}
Probabilistic relational models, denoted as \acp{pfg}, combine probabilistic modelling with relational logic (that is, first-order logic with known universes).
By introducing \acp{lv} to represent sets of indistinguishable objects, \acp{pfg} allow lifted inference algorithms to use a representative of indistinguishable objects for efficient computations.
In practice, however, when learning the underlying probability distribution of a \ac{pfg} from data, indistinguishable objects are often not recognised.
In particular, considering a potential-based factorisation of the probability distribution, learned potentials inevitably deviate even for indistinguishable objects due to estimates from data.
To mitigate this issue and ensure the practical applicability of obtaining a compact representation for lifted inference, we solve the problem of constructing a lifted representation while taking into account small deviations of potentials for indistinguishable objects.
In particular, we ensure that the obtained lifted representation is approximately equivalent to a given propositional (ground) representation by solving an optimisation problem to minimise the approximation error.
Allowing for small deviations between potentials is essential for practical applications, where potentials, for instance, are learned from data and hence are subject to inaccuracies.
For example, consider the probabilities $p_1 = 0.501$ and $p_2 = 0.499$.
In case $p_1$ and $p_2$ are estimates from data, it is likely that $p_1$ and $p_2$ should actually be considered equal.

\citet{Poole2003a} first introduces \acp{pfg}, which combine relational logic and probabilistic models, and \acl{lve} as a lifted inference algorithm to perform lifted probabilistic inference in \acp{pfg}.
In probabilistic inference, lifting exploits indistinguishabilities in a probabilistic model, allowing to carry out query answering more efficiently while maintaining exact answers~\citep{Niepert2014a}.
Since its first introduction by \citet{Poole2003a}, \acl{lve} has continuously been refined by many researchers to reach its current form~\citep{DeSalvoBraz2005a,DeSalvoBraz2006a,Milch2008a,Kisynski2009a,Taghipour2013a,Braun2018a}.
More recently, \citet{Luttermann2024g,Luttermann2024b} extend \acp{pfg} to incorporate causal knowledge and thereby allow to perform lifted causal inference.
To perform lifted probabilistic (or causal) inference, the lifted representation (e.g., a \ac{pfg}) has to be constructed first.
The \ac{acp} algorithm~\citep{Luttermann2024a,Luttermann2024f,Luttermann2024d,Luttermann2025b}, which generalises the CompressFactorGraph algorithm~\citep{Kersting2009a,Ahmadi2013a}, is the current state of the art to construct a \ac{pfg} from a propositional model with equivalent semantics.
\Ac{acp} employs a colour passing procedure to detect symmetric subgraphs, similar to the \acl{wl} algorithm~\citep{Weisfeiler1968a}, which is a well-known algorithm to test for graph isomorphism.
While \ac{acp} is able to construct a \ac{pfg} entailing equivalent semantics as a given propositional model, \ac{acp} requires potentials to exactly match, which is a significant limitation for practical applications.

In this paper, we contribute the \ac{eacp} algorithm, which solves the problem of constructing an approximate lifted representation with a minimal approximation error and thereby makes the construction of a lifted model applicable in practice.
The \ac{eacp} algorithm allows for potentials to deviate by a factor of $\varepsilon$ to still be considered identical, where $\varepsilon$ is a hyperparameter controlling the required agreement between potentials.
Thus, the hyperparameter $\varepsilon$ controls the trade-off between the exactness and the compactness of the lifted representation obtained by \ac{eacp}.
We further prove that the approximation error induced by \ac{eacp} is strictly bounded.
In addition to the theoretical bounds, we empirically show that \ac{eacp} significantly reduces run times for inference while at the same time keeping the approximation error close to zero.

The remaining part of this paper is structured as follows.
We begin by introducing background information and notations in \cref{sec:eacp_background}.
Thereafter, we introduce the \ac{eacp} algorithm to solve the problem of constructing an approximate lifted representation with a minimal approximation error.
We then prove that the approximation error induced by \ac{eacp} is strictly bounded and show that the given bound is optimal.
Finally, we empirically demonstrate that in practice, the actual approximation error induced by \ac{eacp} is well below the theoretical bounds before we conclude the paper.

\section{Background} \label{sec:eacp_background}
We first define \acp{fg} as propositional models and afterwards introduce the idea of lifted representations such as \acp{pfg}.
An \ac{fg} is a probabilistic graphical model to compactly represent a probability distribution over a set of \acp{rv} by factorising the distribution~\citep{Frey1997a,Kschischang2001a}.
\begin{definition}[Factor Graph]
	An \emph{\ac{fg}} $M = (\boldsymbol V, \boldsymbol E)$ is an undirected bipartite graph consisting of a node set $\boldsymbol V = \boldsymbol R \cup \boldsymbol \Phi$, where $\boldsymbol R = \{R_1, \ldots, R_n\}$ is a set of variable nodes (\acp{rv}) and $\boldsymbol \Phi = \{\phi_1, \ldots, \phi_m\}$ is a set of factor nodes (functions), as well as a set of edges $\boldsymbol E \subseteq \boldsymbol R \times \boldsymbol \Phi$.
	There is an edge between a variable node $R_i$ and a factor node $\phi_j$ in $\boldsymbol E$ if $R_i$ appears in the argument list of $\phi_j$.
	A factor $\phi_j(\mathcal R_j)$ defines a function $\phi_j \colon \times_{R \in \mathcal R_j} \range{R} \mapsto \mathbb{R}^+$ that maps the ranges of its arguments $\mathcal R_j$ (a sequence of \acp{rv} from $\boldsymbol R$) to a positive real number, called potential.
	The term $\range{R}$ denotes the possible values a \ac{rv} $R$ can take.
	We define the joint potential for an assignment $\boldsymbol r$ (where $\boldsymbol r$ is a shorthand notation for $\boldsymbol R = \boldsymbol r$) as
	\begin{align}
		\psi(\boldsymbol r) = \prod_{j=1}^m \phi_j(\boldsymbol r_j), \label{eq:eacp_joint_potential}
	\end{align}
	where $\boldsymbol r_j$ is a projection of $\boldsymbol r$ to the argument list of $\phi_j$.
	The full joint probability distribution encoded by $M$ is then given by the normalised joint potential
	\begin{align}
		P_M(\boldsymbol r) = \frac{1}{Z} \prod_{j=1}^m \phi_j(\boldsymbol r_j) = \frac{1}{Z} \psi(\boldsymbol r), \label{eq:eacp_full_joint_distribution}
	\end{align}
	where $Z = \sum_{\boldsymbol r} \prod_{j=1}^{m} \phi_j(\boldsymbol r_j)$ is the normalisation constant.
\end{definition}
\begin{figure}
	\centering
	\input{files/example_fg.tex}
	\caption{An \ac{fg} modelling the interplay between the revenue of a company ($Rev$) and the salaries of two employees ($SalA$, $SalB$). The potential tables of the factors are shown on the right.}
	\label{fig:eacp_example_fg}
\end{figure}
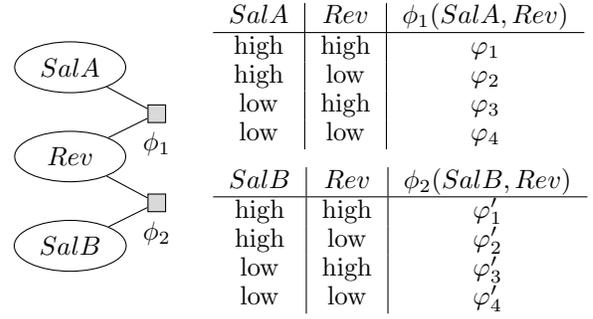
\begin{example} \label{ex:eacp_example_fg}
	Consider the \ac{fg} depicted in \cref{fig:eacp_example_fg}, which models the interplay between the revenue $Rev$ of a company and the salary of two employees, denoted as $SalA$ and $SalB$.
	We have $\boldsymbol R = \{SalA, \allowbreak SalB, \allowbreak Rev\}$, $\boldsymbol \Phi = \{\phi_1, \allowbreak \phi_2\}$, and $\boldsymbol E = \{ \{SalA, \allowbreak \phi_1\}, \allowbreak \{Rev, \allowbreak \phi_1\}, \allowbreak \{Rev, \allowbreak \phi_2\}, \allowbreak \{SalB, \allowbreak \phi_2\} \}$.
	For the sake of this example, let $\range{SalA} = \allowbreak \range{SalB} = \allowbreak \range{Rev} = \allowbreak \{\low, \allowbreak \high\}$.
	The potential tables of $\phi_1$ and $\phi_2$ are shown on the right.
	In particular, it holds that $\phi_1(\high, \high) = \varphi_1$, $\phi_1(\high, \low) = \varphi_2$, and so on, where all $\varphi_i, \varphi'_i \in \mathbb{R}^+$ are arbitrary positive real numbers.
\end{example}
Probabilistic inference describes the task of computing marginal distributions of \acp{rv} given observations for other \acp{rv}.
In other words, probabilistic inference refers to query answering, where a query is defined as follows.
\begin{definition}[Query]
	A \emph{query} $P(Q \mid E_1 = e_1, \ldots, E_k = e_k)$ consists of a query term $Q$ and a set of events $\{E_j = e_j\}_{j=1}^{k}$ (called evidence), where $Q$ and each $E_j$ are \acp{rv}.
	To query a specific probability instead of a probability distribution, the query term is an event $Q = q$.
\end{definition}
\begin{example} \label{ex:eacp_example_query}
	Take a look at the \ac{fg} shown in \cref{fig:eacp_example_fg}.
	The query $P(SalA \mid Rev = \high)$ asks for the probability distribution of $A$'s salary given that the company has a high revenue.
\end{example}
When considering relations between objects, there are often groups of indistinguishable objects that behave identically (or at least similarly).
Lifted representations such as \acp{pfg} exploit identical behaviour to enable scalable probabilistic inference with respect to domain sizes of \acp{lv}.
To illustrate the idea behind lifting, consider the following example.
\begin{example} \label{ex:eacp_lifting_idea}
	Consider the \ac{fg} depicted in \cref{fig:eacp_example_fg} and the query $P(Rev = \high)$.
	Then, it holds that
	\begingroup
	\allowdisplaybreaks
	\begin{align*}
		P&(Rev = \high) \\
		&= \sum_{a \in \range{SalA}} \sum_{b \in \range{salB}} P(a, b,\high) \\
		&= \frac{1}{Z} \sum_{a \in \range{SalA}} \sum_{b \in \range{salB}} \phi_1(a, \high) \phi_2(b, \high) \\
		&= \frac{1}{Z} \Big( \varphi_1 \varphi'_1 + \varphi_3 \varphi'_1 + \varphi_1 \varphi'_3 + \varphi_3 \varphi'_3 \Big).
	\end{align*}
	\endgroup
	If employees $A$ and $B$ are indistinguishable, that is, if it holds that $\varphi_i = \varphi'_i$ for all $i \in \{1, \ldots, 4\}$, we can simplify the computation and obtain
	\begingroup
	\allowdisplaybreaks
	\begin{align*}
		P&(Rev = \high) \\
		&= \frac{1}{Z} \sum_{a \in \range{SalA}} \phi_1(a, \high) \sum_{b \in \range{salB}} \phi_2(b, \high) \\
		&= \frac{1}{Z} \Bigg( \sum_{a \in \range{SalA}} \phi_1(a, \high) \Bigg)^2 \\
		&= \frac{1}{Z} \Big( \varphi_1 + \varphi_3 \Big)^2.
	\end{align*}
	\endgroup
\end{example}
\Cref{ex:eacp_lifting_idea} illustrates that in case $A$ and $B$ are indistinguishable, we can select one representative (e.g., $A$) and reduce the number of factors to consider for computations.
The idea of exploiting exponentiation can be generalised to groups of $k$ indistinguishable objects (e.g., employees) to significantly reduce the computational effort when answering queries.
Indistinguishable objects frequently occur in relational models and are relevant in many
real world domains.
For example, in an epidemic domain, each person influences the probability of an epidemic equally, i.e., the probability of an epidemic depends on the number of sick people and is independent of which individual people are sick.

As we have seen, to exploit indistinguishabilities, we need to find factors with identical potential tables.
Currently, the \ac{acp} algorithm is the state of the art to find factors with identical potential tables and group them together to obtain a lifted representation such as a \ac{pfg}.\footnote{A formal description and a detailed explanation of the \ac{acp} algorithm is provided in \cref{appendix:eacp_acp_in_detail}.}
In \cref{ex:eacp_lifting_idea}, we assume $\varphi_i = \varphi'_i$ for all $i \in \{1, \ldots, 4\}$, which is required by \ac{acp}.
However, in practice, we often face situations where estimates of potentials lead to deviations such that $\varphi_i = \varphi'_i \cdot (1 \pm \varepsilon)$ for a small $\varepsilon \in \mathbb{R}^+$.
The \ac{acp} algorithm does not group factors if they are not strictly equal and thus is hardly applicable in practice to identify factors that should be grouped.
To address this limitation, we next investigate how indistinguishabilities can be approximated when constructing a lifted representation.

\section{Approximation of Indistinguishabilities}
To control the trade-off between the exactness and compactness of the resulting lifted representation when grouping factors with approximately equivalent semantics, we now introduce a hyperparameter $\varepsilon \in \mathbb{R}^+$.
More specifically, we allow for a maximum relative deviation of factor $(1 \pm \varepsilon)$, i.e., two potentials $\varphi$ and $\varphi'$ are considered approximately equivalent if $\varphi \in [\varphi' \cdot (1 - \varepsilon), \varphi' \cdot (1 + \varepsilon)]$ and $\varphi' \in [\varphi \cdot (1 - \varepsilon), \varphi \cdot (1 + \varepsilon)]$.\footnote{Since potentials are arbitrary positive real numbers (and thus might differ in their order of magnitude), we allow for a relative deviation instead of using an absolute deviation.}
The notion of $\varepsilon$-equivalence formally captures the idea of approximately equivalent factors.
\begin{definition}[$\varepsilon$-Equivalent Factors] \label{def:eacp_epsilon_equivalence}
	Let $\varepsilon \in \mathbb{R}^+$ be a positive real number.
	Two potentials $\varphi_1 \in \mathbb{R}^+$ and $\varphi_2 \in \mathbb{R}^+$ are \emph{$\varepsilon$-equivalent}, denoted as $\varphi_1 \approxEquiv \varphi_2$, if $\varphi_1 \in [\varphi_2 \cdot (1 - \varepsilon), \varphi_2 \cdot (1 + \varepsilon)]$ and $\varphi_2 \in [\varphi_1 \cdot (1 - \varepsilon), \varphi_1 \cdot (1 + \varepsilon)]$.
	Further, two factors $\phi_1(R_1, \ldots, R_n)$ and $\phi_2(R'_1, \ldots, R'_n)$ are \emph{$\varepsilon$-equivalent}, denoted as $\phi_1 \approxEquiv \phi_2$, if there exists a permutation $\pi$ of $\{1, \ldots, n\}$ such that for all assignments $(r_1, \ldots, r_n) \in \times_{i=1}^n \range{R_i}$, where $\phi_1(r_1, \ldots, r_n) = \varphi_1$ and $\phi_2(r_{\pi(1)}, \ldots, r_{\pi(n)}) = \varphi_2$, it holds that $\varphi_1 \approxEquiv \varphi_2$.
\end{definition}
Note that the notion of $\varepsilon$-equivalence is symmetric and as a necessary condition to be $\varepsilon$-equivalent, $\phi_1$ and $\phi_2$ must be defined over the same function domain and hence must have the same number of arguments.
We further remark that indistinguishable objects are not guaranteed to be located at the same position in their respective factors, which is the reason we consider permutations of the arguments.
For example, in \cref{fig:eacp_example_fg}, $SalB$ could also be the second argument of $\phi_2$: Then, the potential table of $\phi_2$ would read $\varphi'_1$, $\varphi'_3$, $\varphi'_2$, $\varphi'_4$ from top to bottom (if we keep the order of the assignments), i.e., even if $\varphi_i = \varphi'_i$ for all $i \in \{1, \ldots, 4\}$, we would only be able to exploit this property if we permute the arguments of $\phi_2$ (or of $\phi_1$) such that $SalA$ and $SalB$ are located at the same positions in their respective argument lists.
A full example to showcase the role of permutations is given in \cref{appendix:eacp_permuted_arguments}.
For simplicity, we assume that $\pi$ is the identity function throughout this paper (however, all results also apply for arbitrary choices of $\pi$~\citep{Luttermann2024a}).
\begin{example} \label{ex:eacp_epsilon_equivalence}
	Let $\varphi = 0.49$, $\varphi' = 0.5$, and $\varepsilon = 0.1$.
	Then, it holds that $\varphi' = 0.5 \in [\varphi \cdot (1 - \varepsilon) = 0.441, \varphi \cdot (1 + \varepsilon) = 0.539]$ and $\varphi = 0.49 \in [\varphi' \cdot (1 - \varepsilon) = 0.45, \varphi' \cdot (1 + \varepsilon) = 0.55]$.
	In consequence, $\varphi$ and $\varphi'$ are $\varepsilon$-equivalent.
\end{example}
To group $\varepsilon$-equivalent factors such that we can use a representative and exploit exponentiation to reduce the number of factors to consider during computations, we need to find $\varepsilon$-equivalent factors and change their potentials in a way that their potential tables become identical.
We first address the issue of detecting $\varepsilon$-equivalent factors and then show how potentials are changed to minimise the approximation error.

\subsection{Finding and Grouping \texorpdfstring{$\boldsymbol \varepsilon$}{e}-Equivalent Factors}
A problem when searching for groups of $\varepsilon$-equivalent factors is that $\varepsilon$-equivalence is not transitive.
More specifically, it might happen that there are factors $\phi_1$, $\phi_2$, and $\phi_3$ such that $\phi_1 \approxEquiv \phi_2$ and $\phi_2 \approxEquiv \phi_3$ but $\phi_1 \not\approxEquiv \phi_3$.
\begin{table}
	\centering
	\begin{subtable}[t]{\linewidth}
		\centering
		\input{files/example_eps_equiv_mappings.tex}
		\caption{}
		\label{tab:example_eps_equiv_mappings}
	\end{subtable}

	\begin{subtable}[t]{\linewidth}
		\centering
		\input{files/example_eps_equiv_intervals.tex}
		\caption{}
		\label{tab:example_eps_equiv_intervals}
	\end{subtable}
	\caption{(a) The potential tables of exemplary factors $\phi_1(R_1^1,R_2^1)$, $\phi_2(R_1^2,R_2^2)$, and $\phi_3(R_1^3,R_2^3)$, where the \acp{rv} $R_1^i$ and $R_2^i$, $i \in \{1, 2, 3\}$, all have the same range $\{ \low, \high \}$, and (b) the intervals resulting from a deviation of factor $\varepsilon = 0.1$. We omit the arguments of the factors and their assignments for brevity (the order of the assignments is identical to the order in (a)).}
	\label{tab:example_eps_equiv}
\end{table}
\begin{example}
	Consider the factors $\phi_1(R_1^1,R_2^1)$, $\phi_2(R_1^2,R_2^2)$, and $\phi_3(R_1^3,R_2^3)$ and their potential tables depicted in \cref{tab:example_eps_equiv_mappings}.
	For the sake of this example, let $\varepsilon = 0.1$.
	The intervals allowing for a deviation of factor $(1 \pm \varepsilon)$ according to \cref{def:eacp_epsilon_equivalence} are shown in \cref{tab:example_eps_equiv_intervals}.
	Since all potentials of $\phi_1$ lie within the corresponding intervals of $\phi_2$ (and vice versa), it holds that $\phi_1 \approxEquiv \phi_2$.
	Analogously, it holds that $\phi_2 \approxEquiv \phi_3$.
	However, due to $0.75 \notin [0.756,0.924]$ (as well as $0.84 \notin [0.675,0.825]$), it holds that $\phi_1 \not\approxEquiv \phi_3$.
\end{example}
Due to the non-transitivity of $\varepsilon$-equivalence, we cannot simply group a factor $\phi$ with a group of $\varepsilon$-equivalent factors $\boldsymbol G = \{ \phi_1, \allowbreak \dots, \allowbreak \phi_k \}$ if $\phi$ is $\varepsilon$-equivalent to any $\phi_i \in \boldsymbol G$.
Doing so would give rise to the issue of cascading errors, that is, in the worst case, completely different factors could be grouped together (e.g., assuming $\varepsilon = 0.1$, the potential $1.0$ can be grouped with the potential $0.9$, which itself can be grouped with the potential $0.81$, and so on).
To avoid cascading errors, we thus ensure a factor $\phi$ is only added to a group of $\varepsilon$-equivalent factors $\boldsymbol G$ if $\phi$ is $\varepsilon$-equivalent to \emph{all} factors in $\boldsymbol G$.

Next, we need to solve the problem of changing the potentials for every group of pairwise $\varepsilon$-equivalent factors $\boldsymbol G = \{ \phi_1, \dots, \phi_k \}$.
To exploit exponentiation and thus avoid looking at every factor individually, the changes must ensure that all factors map to the same potentials.
At the same time, we aim to minimise the approximation error, that is, we want to apply the smallest possible change to the potentials.
Formally, the goal is to find $\phi^*$ such that
\begin{align}
	\phi^* = \argmin_{\phi_j} \sum_{\phi_i \in \boldsymbol G} Err(\phi_i, \phi_j), \label{eq:eacp_argmin_potential_change}
\end{align}
where $Err(\phi_i, \phi_j)$ is the sum of squared deviations between the potentials of $\phi_i$ and $\phi_j$:
\begin{align}
	Err(\phi_i, \phi_j) = \sum_{r_1, \ldots, r_n} \Big( \phi_i(r_1, \ldots, r_n) - \phi_j(r_1, \ldots, r_n) \Big)^2, \label{eq:eacp_error_between_factors}
\end{align}
with $r_1, \ldots, r_n$ denoting the possible assignments of the arguments of $\phi_i$ and $\phi_j$.\footnote{Recall that we assume $\pi$ from \cref{def:eacp_epsilon_equivalence} to be the identity function. In case $\pi$ is not the identity function, we end up with $Err(\phi_i, \phi_j) = \sum_{r_1, \ldots, r_n} ( \phi_i(r_1, \ldots, r_n) - \phi_j(r_{\pi(1)}, \ldots, r_{\pi(n)}) )^2$.}
To obtain identical potentials within a group $\boldsymbol G = \{ \phi_1, \dots, \phi_k \}$, our goal is to update the factors in $\boldsymbol G$ such that $\phi_1 = \phi^*, \dots, \phi_k = \phi^*$.

Thus, we now solve the problem of finding $\phi^*$.
In fact, it holds that for any set of numbers $\{ \varphi_1, \ldots, \varphi_k \}$, the arithmetic mean $\bar{\varphi} = \frac{1}{k} \sum_{i=1}^{k} \varphi_i$ minimises the sum of squared deviations $\sum_{i=1}^{k} (\varphi_i - \bar{\varphi})^2$, i.e., replacing $\bar{\varphi}$ by any other value would increase the sum of squared deviations.
\begin{restatable}{theorem}{eacpMeanMinimisesSumSquaredTheorem} \label{th:eacp_mean_minimal_sum_squared}
	Let $\varphi_1, \ldots, \varphi_k \in \mathbb{R}^+$.
	It holds that the arithmetic mean $\bar{\varphi} = \frac{1}{k} \sum_{i=1}^{k} \varphi_i$ is the optimal choice for $\varphi^* = \argmin_{\hat{\varphi}} \sum_{i=1}^{k} (\varphi_i - \hat{\varphi})^2$.
\end{restatable}
\Cref{th:eacp_mean_minimal_sum_squared} is a well-known property of the arithmetic mean (a proof is given in \cref{appendix:eacp_missing_proofs}).
As \cref{eq:eacp_argmin_potential_change} aims to minimise a sum over a sum of squared deviations $Err(\phi_i, \phi_j)$, the sum in \cref{eq:eacp_argmin_potential_change} becomes minimal if we minimise $Err(\phi_i, \phi_j)$, i.e., the right hand side of \cref{eq:eacp_error_between_factors}, according to \cref{th:eacp_mean_minimal_sum_squared}.
Therefore, for any group $\boldsymbol G = \{ \phi_1, \dots, \phi_k \}$ of pairwise $\varepsilon$-equivalent factors, we set $\phi_1 = \phi^*, \dots, \phi_k = \phi^*$ such that
\begin{align}
	\phi^*(\boldsymbol r) = \frac{1}{k} \sum_{i=1}^{k} \phi_i(\boldsymbol r) \label{eq:eacp_mean_potential}
\end{align}
for all possible assignments $\boldsymbol r = r_1, \ldots, r_n$ to ensure that all factors in $\boldsymbol G$ map to the same potentials while minimising the cumulative squared deviation of the group $\boldsymbol G$.

Next, we compile the insights on finding and grouping $\varepsilon$-equivalent factors into the \ac{eacp} algorithm, which paves the way to apply lifted model construction in practice.

\subsection{The \texorpdfstring{$\boldsymbol \varepsilon$}{e}-Advanced Colour Passing Algorithm} \label{sec:eacp_algorithm}

\begin{algorithm}[t]
	\caption{$\varepsilon$-Advanced Colour Passing}
	\label{alg:eacp_eacp}
	\alginput{An \ac{fg} $M = (\boldsymbol R \cup \boldsymbol \Phi, \boldsymbol E)$, a hyperparameter \\\hspace*{\algorithmicindent} $\varepsilon \in \mathbb{R}^+$, and a set of observed events (evidence) $\boldsymbol O = \\\hspace*{\algorithmicindent} \{ E_1 = e_1, \ldots, E_{\ell} = e_{\ell} \}$.} \\
	\algoutput{A lifted representation $M'$, encoded as a \ac{pfg}, \\\hspace*{\algorithmicindent} which is approximately equivalent to $M$.}
	\begin{algorithmic}[1]
		\LeftComment{Phase I: Find groups of pairwise $\varepsilon$-equivalent factors}\;
		\State $\boldsymbol G \gets \{ \{ \phi_1 \} \}$\;
		\ForEach{factor $\phi_i \in \boldsymbol \Phi \setminus \{ \phi_1 \}$}
			\State $\boldsymbol C \gets \emptyset$\; \label{line:eacp_initialise_candidates}
			\ForEach{group $\boldsymbol G_j \in \boldsymbol G$} \label{line:eacp_for_each_loop_candidates}
				\If{$\forall \phi_k \in \boldsymbol G_j \colon \phi_i =_{\varepsilon} \phi_k$} \label{line:eacp_check_eps_equiv}
					\State $\boldsymbol C \gets \boldsymbol C \cup \{ \boldsymbol G_j \}$\; \label{line:eacp_add_to_candidates}
				\EndIf
			\EndFor
			\If{$\boldsymbol C \neq \emptyset$}
				\State $\boldsymbol G_j \gets \argmin_{\boldsymbol C_i \in \boldsymbol C} \sum_{\phi_j \in \boldsymbol C_i} Err(\phi_i, \phi_j)$\; \label{line:eacp_argmin_group_candidates}
				\State $\boldsymbol G_j \gets \boldsymbol G_j \cup \{ \phi_i \}$\; \label{line:eacp_add_to_existing_group}
			\Else \label{line:eacp_if_create_new_group}
				\State $\boldsymbol G \gets \boldsymbol G \cup \{ \{ \phi_i \} \}$\; \label{line:eacp_add_to_new_group}
			\EndIf
		\EndFor
		\LeftComment{Phase II: Assign colours to factors and run \ac{acp}}\;
		\ForEach{group $\boldsymbol G_j \in \boldsymbol G$}
			\ForEach{factor $\phi_i \in \boldsymbol G_j$}
				\State $\phi_i.colour \gets j$\;
			\EndFor
		\EndFor
		\State $\boldsymbol G' \gets$ Call \ac{acp} on $M$ and $\boldsymbol O$ using the assigned colours\;\label{line:eacp_call_acp}
		\LeftComment{Phase III: Update potentials}\;
		\ForEach{group $\boldsymbol G_j \in \boldsymbol G'$} \label{line:eacp_for_each_loop_update_potentials}
			\State $\phi^*(\boldsymbol r) \gets \frac{1}{\abs{\boldsymbol G_j}} \sum_{\phi_i \in \boldsymbol G_j} \phi_i(\boldsymbol r)$ for all assignments $\boldsymbol r$\; \label{line:eacp_mean_potential}
			\ForEach{factor $\phi_i \in \boldsymbol G_j$} \label{line:eacp_for_each_loop_update_potentials_factors}
				\State $\phi_i \gets \phi^*$\; \label{line:eacp_update_potentials}
			\EndFor
		\EndFor
		\State $M' \gets$ construct \ac{pfg} from groupings of \ac{acp}\; \label{line:eacp_eacp_pfg_construction}
	\end{algorithmic}
\end{algorithm}

The \ac{eacp} algorithm consists of three phases and is described in \cref{alg:eacp_eacp}.
In the first phase, \ac{eacp} computes groups of factors that are pairwise $\varepsilon$-equivalent.
For every factor $\phi_i$ in the input \ac{fg}, \ac{eacp} checks whether it can be added to an existing group or if a new group has to be created.
As it is possible for $\phi_i$ to be $\varepsilon$-equivalent to all factors of multiple existing groups (e.g., in \cref{tab:example_eps_equiv}, $\phi_2$ could be grouped both with $\{ \phi_1 \}$ and $\{ \phi_3 \}$), \ac{eacp} computes all candidate groups $\boldsymbol C$ (\cref{line:eacp_initialise_candidates,line:eacp_for_each_loop_candidates,line:eacp_check_eps_equiv,line:eacp_add_to_candidates}) and then adds $\phi_i$ to the group that minimises the sum of squared deviations between $\phi_i$ and all factors in the group (\cref{line:eacp_argmin_group_candidates,line:eacp_add_to_existing_group}).
If $\phi_i$ cannot be added to an existing group, \ac{eacp} creates a new group for $\phi_i$ (\cref{line:eacp_add_to_new_group}).
Then, in the second phase, \ac{eacp} assigns to every factor a colour based on the group it belongs to, that is, all factors within the same group receive the same colour (and factors in different groups receive different colours).
Factors within the same group could potentially be grouped together in a lifted representation if their arguments are indistinguishable.
To ensure the factors' arguments are indistinguishable, \ac{eacp} runs the \ac{acp} algorithm using the previously assigned colours (instead of \ac{acp}'s original colour assignment).
By running \ac{acp} with the assigned colours, \ac{eacp} ensures that in addition to the potential tables, the surrounding graph structure of the factors is taken into account, thereby enforcing that the arguments of factors within a group are indistinguishable (more details on this are given in \cref{appendix:eacp_acp_in_detail}).
Finally, in phase three, \ac{eacp} updates the potentials of every group of factors computed by \ac{acp} according to \cref{eq:eacp_mean_potential} to ensure that all factors in a group have identical potential tables (\cref{line:eacp_for_each_loop_update_potentials,line:eacp_mean_potential,line:eacp_for_each_loop_update_potentials_factors,line:eacp_update_potentials}).
As potentials within a group are now strictly equal, the corresponding \ac{pfg} is constructed from the groups as in the original \ac{acp} algorithm.\footnote{For a detailed description of the \ac{pfg} construction in \cref{line:eacp_eacp_pfg_construction} of \cref{alg:eacp_eacp}, we refer the reader to \citet{Luttermann2024a}.}
Commonly used lifted inference algorithms, such as \acl{lve}, operate on \acp{pfg} and thus can directly be run on the output of \ac{eacp}.\footnote{We remark that $\varepsilon$-equivalence can also be applied to exploit approximate symmetries \emph{within} factors that map assignments of their arguments to identical potentials independent of the order of the assigned values (for more details, see \cref{appendix:eacp_approximate_symmetries_within_factors}).}
\begin{example} \label{ex:eacp_example_eacp}
	Take a look at the \ac{fg} given in \cref{fig:eacp_example_fg} and assume the potential tables of $\phi_1$ and $\phi_2$ are as given in \cref{tab:example_eps_equiv_mappings} (i.e., $\varphi_1 = 0.75$, $\varphi'_1 = 0.8$, and so on).
	Further, let $\varepsilon = 0.1$ and assume we do not have any evidence, i.e., $\boldsymbol O = \emptyset$.
	As $\phi_1$ and $\phi_2$ are $\varepsilon$-equivalent, \ac{eacp} puts them into the same group and after the first phase, \ac{eacp} ends up with $\boldsymbol G = \{ \{ \phi_1, \phi_2 \} \}$.
	Then, \ac{acp} is called with $\phi_1$ and $\phi_2$ having the same colour, and after passing the colours around, $\phi_1$ and $\phi_2$ remain in the same group because their surrounding graph structure is symmetric (and thus, their arguments are indistinguishable).
	After the third phase, the potential tables are updated by computing a row-wise arithmetic mean, that is, $\varphi_1 = \varphi'_1 = (0.75 + 0.8) \mathbin{/} 2 = 0.775$, $\varphi_2 = \varphi'_2 = 0.315$, $\varphi_3 = \varphi'_3 = 0.49$, and $\varphi_4 = \varphi'_4 = 0.21$.
\end{example}
The \ac{eacp} algorithm takes a fundamental step towards the practical applicability of lifted inference algorithms by generalising the \ac{acp} algorithm to account for inaccurate estimates of potentials, which are abundant in practice.
In particular, it holds that \ac{eacp} is identical to \ac{acp} when setting $\varepsilon$ to zero because $\varepsilon$-equivalence reduces to strict equivalence if $\varepsilon = 0$.
\begin{restatable}{corollary}{eacpGeneralisationCorollary}
	If $\varepsilon = 0$, \ac{eacp} returns the same \ac{pfg} as \ac{acp}.
\end{restatable}
So far, we have shown how $\varepsilon$-equivalent factors can be grouped and updated to enable lifted inference with a minimal approximation error.
As we show later, the approximation error is often even negligible in practice.
To get an initial idea about the extent of the approximation error, consider \cref{ex:eacp_example_eacp} and the query $P(SalA \mid Rev = \high)$.
In the original \ac{fg}, we obtain $P(SalA \mid Rev = \high) \approx \langle 0.6098, 0.3902 \rangle$ and after running \ac{eacp}, we have $P(SalA \mid Rev = \high) \approx \langle 0.6126, 0.3874 \rangle$.
An essential question now is how much query results can change in general when using the approximate lifted representation instead of the initial exact \ac{fg} for query answering.
We answer this question next.

\section{Bounding the Change in Query Results} \label{sec:eacp_theoretical}
We now bound the change in query results when modifying a given \ac{fg} by grouping and updating the potentials of $\varepsilon$-equivalent factors according to \cref{alg:eacp_eacp}.
For the sake of our analysis, let $M$ denote the input for \cref{alg:eacp_eacp} and $M'$ the output of \cref{alg:eacp_eacp} such that $M$ encodes the distribution $P_M$ and $M'$ encodes the distribution $P_{M'}$.
In our analysis, we use the following distance measure between two distributions $P_M$ and $P_{M'}$ introduced by \citet{Chan2005a}:
\begingroup
\allowdisplaybreaks
\begin{align}
	D(P_M, P_{M'}) &= \ln \max_{\boldsymbol r} \frac{P_{M'}(\boldsymbol r)}{P_M(\boldsymbol r)} - \ln \min_{\boldsymbol r} \frac{P_{M'}(\boldsymbol r)}{P_M(\boldsymbol r)} \label{eq:eacp_distance_measure} \\
	&= \ln \max_{\boldsymbol r} \frac{\frac{1}{Z'} \psi'(\boldsymbol r)}{\frac{1}{Z} \psi(\boldsymbol r)} - \ln \min_{\boldsymbol r} \frac{\frac{1}{Z'} \psi'(\boldsymbol r)}{\frac{1}{Z} \psi(\boldsymbol r)} \\
	&= \ln \max_{\boldsymbol r} \frac{\psi'(\boldsymbol r)}{\psi(\boldsymbol r)} - \ln \min_{\boldsymbol r} \frac{\psi'(\boldsymbol r)}{\psi(\boldsymbol r)}, \label{eq:eacp_distance_measure_only_joint_potential}
\end{align}
\endgroup
where we define $0 \mathbin{/} 0 := 1$ and $\infty \mathbin{/} \infty := 1$.
$D$ satisfies important properties of a distance measure (positiveness, symmetry, and the triangle inequality) and a major advantage of $D$ is that it allows us to bound the change in query results, which is not possible with other common distance measures such as the \acl{kld}~\citep{Chan2005a}.
In particular, if it holds that $D(P_M, P_{M'}) = d$, the change in a query result is bounded by
\begin{align} \label{eq:eacp_distance_bound_odds}
	e^{-d} \leq \frac{O_{M'}(r \mid \boldsymbol e)}{O_M(r \mid \boldsymbol e)} \leq e^d,
\end{align}
where $O_M(r \mid \boldsymbol e) = P_M(r \mid \boldsymbol e) \mathbin{/} (1 - P_M(r \mid \boldsymbol e))$ defines the odds of $r$ given $\boldsymbol e$.
We can also write \cref{eq:eacp_distance_bound_odds} in terms of probabilities instead of odds and obtain
\begin{align}
	\frac{p e^{-d}}{p(e^{-d} -1) + 1} \leq P_{M'}(r \mid \boldsymbol e) \leq \frac{p e^{d}}{p(e^{d} -1) + 1}, \label{eq:eacp_error_bound_prob}
\end{align}
where $p = P_M(r \mid \boldsymbol e)$ is the initial probability of $r$ given $\boldsymbol e$ in model $M$ and $P_{M'}(r \mid \boldsymbol e)$ is the probability of $r$ given $\boldsymbol e$ in the modified model $M'$~\citep{Chan2005a}.
The bounds given in \cref{eq:eacp_distance_bound_odds,eq:eacp_error_bound_prob} are sharp. 
To obtain a bound on the change in query results, we thus need to determine the value of $d = D(P_M, P_{M'})$ for a given choice of $\varepsilon$.
In general, the normalisation constant $Z$ changes when modifying the original model $M$.
Rewriting \cref{eq:eacp_distance_measure} as \cref{eq:eacp_distance_measure_only_joint_potential}, however, allows us to avoid dealing with the change from $Z$ to $Z'$ (a full derivation is given in \cref{appendix:eacp_missing_proofs}).

We next give a general bound on the distance $D(P_M, P_{M'})$ that applies to arbitrary \acp{fg} $M$ where updates of factors resulting in an \ac{fg} $M'$ ensure that all factors in $M'$ remain $\varepsilon$-equivalent to their original values after the update.
\begin{figure*}[t]
	\centering
	\begin{subfigure}[t]{0.3\linewidth}
		\centering
		\input{files/bound_plot_m=10.tex}
		\caption{}
		\label{fig:bound_plot_m=10}
	\end{subfigure}
	\begin{subfigure}[t]{0.3\linewidth}
		\centering
		\input{files/bound_plot_m=100.tex}
		\caption{}
		\label{fig:bound_plot_m=100}
	\end{subfigure}
	\begin{subfigure}[t]{0.3\linewidth}
		\centering
		\input{files/bound_plot_m=1000.tex}
		\caption{}
		\label{fig:bound_plot_m=1000}
	\end{subfigure}
	\caption{Plots of the bound given in \cref{eq:eacp_error_bound_prob} with $d = \ln\ (1 + \varepsilon)^{m} - \ln\ (1 - \varepsilon)^{m}$. Bounds are illustrated for (a) $m = 10$, (b) $m = 100$, and (c) $m = 1000$ where $\varepsilon = 0.01$ (dashed line) and $\varepsilon = 0.001$ (solid line), respectively. The x-axes depict the initial probability $p = P_M(r \mid \boldsymbol e)$ and the y-axes reflect the bound on the change in the query result.}
	\label{fig:bound_plot}
\end{figure*}
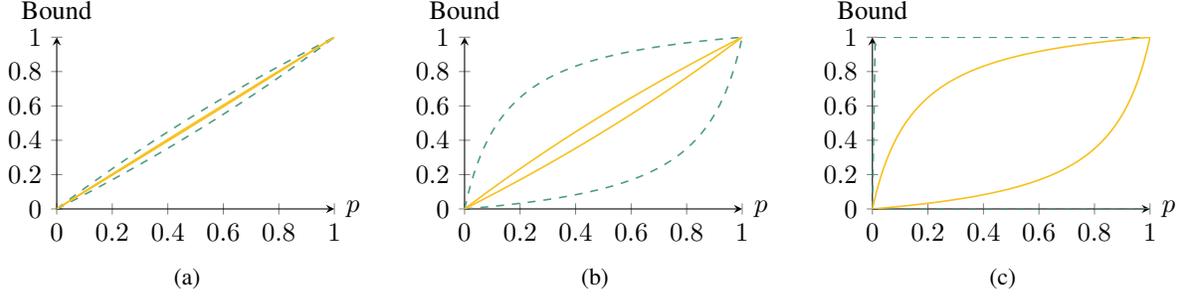
\begin{restatable}{theorem}{eacpErrorBoundTheorem} \label{th:eacp_error_bound_given_eps}
	Let $M = (\boldsymbol R \cup \boldsymbol \Phi, \boldsymbol E)$ be an \ac{fg} and let $M'$ be an \ac{fg} obtained by updating arbitrary potentials of factors in $M$ such that every updated potential remains $\varepsilon$-equivalent to its original value.
	Then, it holds that $D(P_M, P_{M'}) \leq \ln\ (1 + \varepsilon)^m - \ln\ (1 - \varepsilon)^m$, where $P_M$ and $P_{M'}$ are the underlying full joint probability distributions encoded by $M$ and $M'$, respectively, and $m = \abs{\boldsymbol \Phi}$.
\end{restatable}
\begin{proof}[Proof Sketch]
	By definition, every potential in $M'$ differs from its original value in $M$ by factor at most $(1 \pm \varepsilon)$.
	Adding a deviation by factor $(1 \pm \varepsilon)$ to every potential in $M'$ and entering this into \cref{eq:eacp_distance_measure_only_joint_potential} yields the desired result.
\end{proof}
\begin{corollary} \label{co:eacp_pairwise_odds_bound}
	Given the bound from \cref{th:eacp_error_bound_given_eps}, \cref{eq:eacp_distance_bound_odds} leads to
	\begin{align} \label{eq:eacp_distance_specific_estimate}
		\left( \frac{1 - \varepsilon}{1 + \varepsilon} \right)^m \leq \frac{O_{M'}(r \mid \boldsymbol e)}{O_M(r \mid \boldsymbol e)} \leq \left( \frac{1 + \varepsilon}{1 - \varepsilon} \right)^m.
	\end{align}
\end{corollary}
The next lemma shows that updating the potentials within a group of pairwise $\varepsilon$-equivalent factors according to \cref{eq:eacp_mean_potential} satisfies the premise of \cref{th:eacp_error_bound_given_eps} and hence, the bound given in \cref{th:eacp_error_bound_given_eps} holds if $M'$ is the output of \cref{alg:eacp_eacp} run on $M$.
\begin{restatable}{lemma}{pairwiseEquivalenceGroupMeanLemma} \label{lemma:eacp_mean_in_group_equivalence}
	Let $\boldsymbol G = \{ \phi_1, \dots, \phi_k \}$ denote a group of pairwise $\varepsilon$-equivalent factors and let $\phi^*(\boldsymbol r) = \frac{1}{k} \sum_{i=1}^{k} \phi_i(\boldsymbol r)$ for all assignments $\boldsymbol r$.
	Then, $\boldsymbol G^* = \{ \phi_1, \dots, \phi_k, \phi^* \}$ is a group of pairwise $\varepsilon$-equivalent factors.
\end{restatable}
\begin{proof}[Proof Sketch]
	As for the arithmetic mean $\phi^*(\boldsymbol r)$ it holds that $\min_{\phi_j \in \boldsymbol G} \phi_j(\boldsymbol r) \leq \phi^*(\boldsymbol r) \leq \max_{\phi_j \in \boldsymbol G} \phi_j(\boldsymbol r)$ and all $\phi_i, \phi_j \in \boldsymbol G$ are pairwise $\varepsilon$-equivalent, it follows that $\phi^*(\boldsymbol r) \in [\phi_i(\boldsymbol r) \cdot (1 - \varepsilon), \phi_i(\boldsymbol r) \cdot (1 + \varepsilon)]$ and $\phi_i(\boldsymbol r) \in [\phi^*(\boldsymbol r) \cdot (1 - \varepsilon), \phi^*(\boldsymbol r) \cdot (1 + \varepsilon)]$ for any assignment $\boldsymbol r$ and $\phi_i \in \boldsymbol G$.
\end{proof}
\Cref{lemma:eacp_mean_in_group_equivalence} implies that all updated potentials for every factor differ by factor at most $(1 \pm \varepsilon)$ from their original potential after running \cref{alg:eacp_eacp}.
To obtain a bound on the change in query results depending on the choice of $\varepsilon$, we enter the bound from \cref{th:eacp_error_bound_given_eps} into \cref{eq:eacp_error_bound_prob}. 
\Cref{fig:bound_plot} provides plots of the bound for different values of $\varepsilon$ and $m = \abs{\boldsymbol \Phi}$ to give a better idea on how the bound behaves.
Observe that $\varepsilon = 0.01$ yields a strong bound for $m = 10$, however, from $m = 100$ onward, the bound becomes weak (in particular, for $m = 1000$, the change in query results is essentially unbounded when choosing $\varepsilon = 0.01$).
When choosing $\varepsilon = 0.001$, the bound remains strong for $m = 100$, however, for $m = 1000$, the bound weakens as well.
Fortunately, the bound from \cref{th:eacp_error_bound_given_eps} is overly pessimistic for the output of \cref{alg:eacp_eacp}, as we show in the following.
\begin{lemma} \label{le:smallimprovment}
	For two $\varepsilon$-equivalent factors $\phi_1$ and $\phi_2$, it holds that $\phi_1 \in [\phi_2 \cdot \frac{1}{1 + \varepsilon}, \phi_2 \cdot (1 + \varepsilon)]$ and $\phi_2 \in [\phi_1 \cdot \frac{1}{1 + \varepsilon}, \phi_1 \cdot (1 + \varepsilon)]$.
\end{lemma}
\begin{proof}
	Due to the symmetric definition of $\varepsilon$-equivalence, we get $\phi_{2-i} \leq \phi_{i+1} \cdot (1 + \varepsilon)$ for $i \in \{ 0, 1 \}$, resulting in $\phi_{2-i} \cdot \frac{1}{1 + \varepsilon} \leq \phi_{i+1}$.
	Since $1 - \varepsilon \leq \frac{1}{1 + \varepsilon}$ holds for any $\varepsilon > 0$, $\phi_{2-i}$ is contained in the strict subset $[\phi_{i+1} \cdot \frac{1}{1 + \varepsilon}, \phi_{i+1} \cdot (1 + \varepsilon)] \subsetneq [\phi_{i+1} \cdot (1 - \varepsilon), \phi_{i+1} \cdot (1 + \varepsilon)]$.
\end{proof}
Using \cref{le:smallimprovment} and the properties of the arithmetic mean, we obtain the following stronger bound on $D(P_M, P_{M'})$.
\begin{restatable}{theorem}{lowerboundeacpErrorBoundTheorem} \label{th:lowerboundeacp_error_bound_given_eps}
	Let $M = (\boldsymbol R \cup \boldsymbol \Phi, \boldsymbol E)$ be an \ac{fg} and let $M'$ be the output of \cref{alg:eacp_eacp} when run on $M$. With $P_M$ and $P_{M'}$ being the underlying full joint probability distributions encoded by $M$ and $M'$, respectively, and $m = \abs{\boldsymbol \Phi}$, it holds that
	\begin{align}
		D(P_M, P_{M'})   &\leq 
		\ln\left(\frac {1+\frac{m-1}{m}\varepsilon}{\frac{1+\frac{1}{m}\varepsilon}{1+\varepsilon}}\right)^m\\
		&=\ln\left(\frac {\big( 1+\frac{m-1}{m}\varepsilon \big) \big( 1+\varepsilon \big)}{1+\frac{1}{m}\varepsilon}\right)^m\\
		&< \ln \big( 1 + \varepsilon \big)^{2m} \\
		&< \ln \left(\frac{1+\varepsilon}{1-\varepsilon}\right)^m.
	\end{align}
\end{restatable}
\begin{corollary} \label{co:smallinequalitywithimprovedbound}
	Given the bound from \cref{th:lowerboundeacp_error_bound_given_eps}, \cref{eq:eacp_distance_bound_odds} leads to
	\begin{align} \label{eq:eacp_distance_specific_estimate_moreprecise}
	\left(\frac{\frac{1+\frac{1}{m}\varepsilon}{1+\varepsilon} }{1+\frac{m-1}{m}\varepsilon}\right)^m \leq \frac{O_{M'}(r \mid \boldsymbol e)}{O_M(r \mid \boldsymbol e)} \leq \left(\frac {1+\frac{m-1}{m}\varepsilon}{\frac{1+\frac{1}{m}\varepsilon}{1+\varepsilon}}\right)^m.
	\end{align}
\end{corollary}
We give a proof of \cref{th:lowerboundeacp_error_bound_given_eps} in \cref{appendix:eacp_missing_proofs}.
The plot of the bound from \cref{th:lowerboundeacp_error_bound_given_eps} looks similar to the plot of \cref{th:eacp_error_bound_given_eps} (see \cref{fig:bound_plot}) and is optimal (i.e., it is the best bound we can find).
\begin{restatable}{theorem}{optimalBoundTheorem}
	The bound given in \cref{th:lowerboundeacp_error_bound_given_eps} is optimal.
\end{restatable}
\begin{proof}[Proof Sketch]
	We construct an \ac{fg} hitting the boundary from \cref{th:lowerboundeacp_error_bound_given_eps}.
	For the construction, see \cref{tab:example_worst_case} in \cref{appendix:eacp_missing_proofs}.
\end{proof}
\begin{figure*}[t]
	\centering
	\input{files/plot-times-avg.tex}
	\input{files/plot-quots-avg.tex}
	\caption{Average query times of \acl{lve} on the output of \ac{acp} and \ac{eacp} for every choice of $k$ (left), and a boxplot showing the distribution of the quotient $p' \mathbin{/} p$, where $p' = P_{M'}(r \mid \boldsymbol e)$ and $p = P_{M}(r \mid \boldsymbol e)$, for each choice of $k$ (right).}
	\label{fig:eacp_plot-avg}
\end{figure*}
\begin{table}
	\centering
	\input{files/table_mimic_results}
	\caption{Average query times and quotients of query results on parts of the MIMIC-IV dataset~\citep{Johnson2023a}.}
	\label{tab:eacp_mimic_results}
\end{table}
Fortunately, in practice, the change in query results is often close to zero (and thus well below the theoretical bound), as we will show in our experiments.
The reason for this is that the worst-case scenario is an extreme case and slightly deviating from it significantly improves the bounds.
For instance, if there are more factors in a group than rows in their potential tables, the worst-case scenario can no longer occur, resulting in notably smaller values for the distance measure $D$.
More details are given in the proof of \cref{th:lowerboundeacp_error_bound_given_eps} in \cref{appendix:eacp_missing_proofs}.

\section{Experiments} \label{sec:eacp_experiments}
We test the practicality of the \ac{eacp} algorithm in a series of experiments.
\Ac{eacp} is not only required to make \ac{acp} applicable in practice but also allows for more compression (and thus faster inference) if we are willing to trade the exactness of query results for additional speedup.
We thus report the run time gain and the resulting approximation error to get a better understanding of the trade-off between the exactness and the compactness of the lifted representation obtained by \ac{eacp}.
For our experiments, we generate a variety of \acp{fg} with different graph structures and graph sizes (i.e., numbers of \acp{rv} and factors).
More specifically, we generate \acp{fg} containing between $2k + 1$ and $2k + k \cdot \lfloor \log_2(k) \rfloor + 1$ Boolean \acp{rv} as well as between $2k$ and $k + k \cdot \lfloor \log_2(k) \rfloor + 1$ factors, where $k \in \{ 2, 4, 8, 16, 32, 64, 128 \}$ is the \emph{domain size}.
The domain size $k$ controls the number of objects in the models and thus the size of the \acp{fg}.
We provide all data set generators along with our source code in the supplementary material.

In every \ac{fg}, we modify a proportion of $x \in \{ 0.1, \allowbreak 0.3, \allowbreak 0.5, \allowbreak 0.7, \allowbreak 0.9, \allowbreak 1.0 \}$ of the factors such that their potential tables differ by at most factor $(1 \pm \varepsilon)$ from their original potential tables, where $\varepsilon \in \{ 0.001, 0.01, 0.1 \}$.
For each setting, we pose multiple queries to each \ac{fg}.
We report the average run time of \acl{lve} (the state-of-the-art lifted inference algorithm) on the output of \ac{acp} and \ac{eacp}, respectively, over all settings for each choice of $k$ in the left plot of \cref{fig:eacp_plot-avg} and show the distribution of $P_{M'}(r \mid \boldsymbol e) \mathbin{/} P_{M}(r \mid \boldsymbol e)$ over all queries for each choice of $k$ in the right plot of \cref{fig:eacp_plot-avg}.

Taking a look at the left plot in \cref{fig:eacp_plot-avg}, it becomes evident that \ac{eacp} yields a speedup of up to factor 100 compared to \ac{acp}.
The question now is at what cost \ac{eacp} achieves this speedup.
The right plot in \cref{fig:eacp_plot-avg} demonstrates that the price \ac{eacp} pays for the speedup is close to zero: Most of the quotients are nearly equal to one (i.e., most query results hardly differ from their original value).
As expected, the larger the domain size (and hence the size of the \ac{fg}), the larger quotients become.
However, even the outliers (denoted by the dots outside of the boxes) only deviate at the third decimal place from the optimal value one.
The experimental results highlight the practical effectiveness of \ac{eacp} as the approximation error is significantly smaller in practice than suggested by the theoretical bounds.
To give a better overview on how the approximation error behaves for specific choices of $x$ and $\varepsilon$, we provide additional results for individual choices of $x$ and $\varepsilon$ in \cref{appendix:eacp_further_experimental_results}.

In addition to the generated \acp{fg}, we learn an \ac{fg} from the MIMIC-IV dataset~\citep{Johnson2023a} and apply \ac{eacp} with $\varepsilon = 0.1$ to it.
MIMIC-IV contains real-world medical data and we consider a subset of $4000$ patients and their treatments from it.
The learned \ac{fg} contains $344$ \acp{rv} and factors, respectively, and we query each \ac{rv} in it.
We report average run times and average quotients over all queries in \cref{tab:eacp_mimic_results}.
While the speedup of \ac{eacp} is smaller than in \cref{fig:eacp_plot-avg}, the error quotients are also reduced by an order of magnitude, showing that the approximation error is again close to zero.

\section{Conclusion} \label{sec:eacp_conclusion}
Potentials learnt from data often slightly differ even for indistinguishable objects.
Therefore, we solve the problem of constructing a lifted representation from a given propositional representation taking inaccurate estimates of potentials into account, while previous approaches require exact matches.
We present the \ac{eacp} algorithm, which allows for a small deviation of potentials depending on a hyperparameter $\varepsilon$.
By not relying on strictly identical potentials, \ac{eacp} makes a fundamental step towards the practicality of obtaining a compact representation for lifted inference.
We further show that the approximation error of \ac{eacp} is strictly bounded and demonstrate that it is even close to zero in practice.

\section*{Acknowledgements}
This work was partially funded by the Ministry of Culture and Science of the German State of North Rhine-Westphalia.
The research of Malte Luttermann was funded by the BMBF project AnoMed 16KISA057.

\bibliographystyle{named}
\bibliography{references}

\clearpage
\appendix

\section{Missing Proofs} \label{appendix:eacp_missing_proofs}
\eacpMeanMinimisesSumSquaredTheorem*
\begin{proof}
	Let $\hat{\varphi}$ denote an arbitrary estimate for $\varphi^*$ and let $\bar{\varphi} = \frac{1}{k} \sum_{i=1}^{k} \varphi_i$ denote the arithmetic mean of $\varphi_1, \ldots, \varphi_k$.
	We now show that choosing $\hat{\varphi} = \bar{\varphi}$ minimises the expression $\sum_{i=1}^{k} (\varphi_i - \hat{\varphi})^2$.
	Since $\bar{\varphi} - \bar{\varphi} = 0$, we can add it without changing the result and then rewrite the expression:
	\begingroup
	\allowdisplaybreaks
	\begin{align}
		&\sum_{i=1}^{k} \big( \varphi_i - \hat{\varphi} \big)^2 \\
		= &\sum_{i=1}^{k} \Big( \varphi_i - \hat{\varphi} + (\bar{\varphi} - \bar{\varphi}) \Big)^2 \\
		= &\sum_{i=1}^{k} \Big( (\varphi_i - \bar{\varphi}) + (\bar{\varphi} - \hat{\varphi}) \Big)^2 \\
		= &\sum_{i=1}^{k} \Big( \big( \varphi_i - \bar{\varphi} \big)^2 + 2 \big( \varphi_i - \bar{\varphi} \big) \big( \bar{\varphi} - \hat{\varphi} \big) + \big( \bar{\varphi} - \hat{\varphi} \big)^2 \Big) \\
		= &\sum_{i=1}^{k} \big( \varphi_i - \bar{\varphi} \big)^2 + \sum_{i=1}^{k} 2 \big( \varphi_i - \bar{\varphi} \big) \big( \bar{\varphi} - \hat{\varphi} \big) + \sum_{i=1}^{k} \big( \bar{\varphi} - \hat{\varphi} \big)^2 \\
		= &\sum_{i=1}^{k} \big( \varphi_i - \bar{\varphi} \big)^2 + 2 \big( \bar{\varphi} - \hat{\varphi} \big) \sum_{i=1}^{k} \big( \varphi_i - \bar{\varphi} \big) + k \big( \bar{\varphi} - \hat{\varphi} \big)^2. \label{eq:eacp_sum_squared_rewritten}
	\end{align}
	\endgroup
	Due to $\bar{\varphi} = \frac{1}{k} \sum_{i=1}^{k} \varphi_i$ being the arithmetic mean of $\varphi_1, \allowbreak \ldots, \allowbreak \varphi_k$, it holds that $\sum_{i=1}^{k} (\varphi_i - \bar{\varphi}) = 0$:
	\begingroup
	\allowdisplaybreaks
	\begin{align}
		\sum_{i=1}^{k} \big( \varphi_i - \bar{\varphi} \big)
		&= \sum_{i=1}^{k} \varphi_i - k \cdot \bar{\varphi} \\
		&= \sum_{i=1}^{k} \varphi_i - k \cdot \frac{1}{k} \sum_{i=1}^{k} \varphi_i \\
		&= 0.
	\end{align}
	\endgroup
	By entering $\sum_{i=1}^{k} (\varphi_i - \bar{\varphi}) = 0$ into \cref{eq:eacp_sum_squared_rewritten}, we thus obtain
	\begingroup
	\allowdisplaybreaks
	\begin{align}
		&\sum_{i=1}^{k} \big( \varphi_i - \bar{\varphi} \big)^2 + 2 \big( \bar{\varphi} - \hat{\varphi} \big) \sum_{i=1}^{k} \big( \varphi_i - \bar{\varphi} \big) + k \big( \bar{\varphi} - \hat{\varphi} \big)^2 \\
		= &\sum_{i=1}^{k} \big( \varphi_i - \bar{\varphi} \big)^2 + k \big( \bar{\varphi} - \hat{\varphi} \big)^2. \label{eq:eacp_before_mean_enter}
	\end{align}
	\endgroup
	Now, if we set $\hat{\varphi} = \bar{\varphi}$, it holds that $\bar{\varphi} - \hat{\varphi} = 0$ and hence, \cref{eq:eacp_before_mean_enter} simplifies to
	\begingroup
	\allowdisplaybreaks
	\begin{align}
		&\sum_{i=1}^{k} \big( \varphi_i - \bar{\varphi} \big)^2 + k \big( \bar{\varphi} - \hat{\varphi} \big)^2 \\
		= &\sum_{i=1}^{k} \big( \varphi_i - \bar{\varphi} \big)^2.
	\end{align}
	\endgroup
	Choosing $\hat{\varphi}$ different from $\bar{\varphi}$ thus increases the value of the expression by $k (\bar{\varphi} - \hat{\varphi})^2$, which completes the proof.\footnote{\Cref{th:eacp_mean_minimal_sum_squared} is a well-known property of the arithmetic mean and there are different ways to prove this property. The proof given here is taken from \url{http://faculty.washington.edu/swithers/seestats/SeeingStatisticsFiles/seeing/center/meanproof/meanProof.html}.}
\end{proof}

\eacpGeneralisationCorollary*
\begin{proof}
	Let $\varepsilon = 0$.
	Recall that two potentials $\varphi \in \mathbb{R}^+$ and $\varphi' \in \mathbb{R}^+$ are $\varepsilon$-equivalent if $\varphi \in [\varphi' \cdot (1 - \varepsilon), \varphi' \cdot (1 + \varepsilon)]$ and $\varphi' \in [\varphi \cdot (1 - \varepsilon), \varphi \cdot (1 + \varepsilon)]$.
	Now, as $\varepsilon = 0$, $\varphi$ and $\varphi'$ are $\varepsilon$-equivalent if $\varphi = \varphi'$.
	In consequence, when running \ac{eacp} (\cref{alg:eacp_eacp}) on an arbitrary input \ac{fg} $M$, in the first phase all groups of $\varepsilon$-equivalent factors contain only factors whose potentials are strictly equivalent.
	Therefore, in phase two of \ac{eacp}, only factors with equivalent potentials receive the same colour and hence, the colour assignment is identical to the colour assignment of \ac{acp} (\cref{alg:eacp_acp}).
	Then, \ac{eacp} calls \ac{acp} and as the colour assignment of \ac{eacp} to factors is identical to the colour assignment of \ac{acp}, phase one of \ac{eacp} has no impact on the output of \ac{acp}.
	It remains to be shown that the update of potentials in phase three of \ac{eacp} does not alter the output of \ac{acp}.
	As all factors in any group have strictly equivalent potentials already and the arithmetic mean of a set of equal numbers $\varphi, \ldots, \varphi$ is $\varphi$ itself, the update in phase three does not alter the output of \ac{acp}.
	Hence, \ac{eacp} ends up with the same groups as \ac{acp} and therefore outputs the same \ac{pfg} as \ac{acp}, which completes the proof.
\end{proof}

\paragraph{Full derivation of writing \cref{eq:eacp_distance_measure} as \cref{eq:eacp_distance_measure_only_joint_potential}:}
\begingroup
\allowdisplaybreaks
\begin{align} \label{eq:eacp_distance_measure_without_z}
	&\phantom{=~} D(P_M, P_{M'}) \\
	&= \ln \max_{\boldsymbol r} \frac{P_{M'}(\boldsymbol r)}{P_M(\boldsymbol r)} - \ln \min_{\boldsymbol r} \frac{P_{M'}(\boldsymbol r)}{P_M(\boldsymbol r)} \\
	&= \ln \max_{\boldsymbol r} \frac{\frac{1}{Z'} \psi'(\boldsymbol r)}{\frac{1}{Z} \psi(\boldsymbol r)} - \ln \min_{\boldsymbol r} \frac{\frac{1}{Z'} \psi'(\boldsymbol r)}{\frac{1}{Z} \psi(\boldsymbol r)} \\
	&= \ln \left( \frac{\frac{1}{Z'}}{\frac{1}{Z}} \max_{\boldsymbol r} \frac{\psi'(\boldsymbol r)}{\psi(\boldsymbol r)} \right) - \ln \left( \frac{\frac{1}{Z'}}{\frac{1}{Z}} \min_{\boldsymbol r} \frac{\psi'(\boldsymbol r)}{\psi(\boldsymbol r)} \right) \\
	&= \ln \frac{\frac{1}{Z'}}{\frac{1}{Z}} + \ln \max_{\boldsymbol r} \frac{\psi'(\boldsymbol r)}{\psi(\boldsymbol r)} - \ln \frac{\frac{1}{Z'}}{\frac{1}{Z}} - \ln \min_{\boldsymbol r} \frac{\psi'(\boldsymbol r)}{\psi(\boldsymbol r)} \\
	&= \ln \max_{\boldsymbol r} \frac{\psi'(\boldsymbol r)}{\psi(\boldsymbol r)} - \ln \min_{\boldsymbol r} \frac{\psi'(\boldsymbol r)}{\psi(\boldsymbol r)}
\end{align}
\endgroup

\paragraph{Full derivation of writing \cref{eq:eacp_distance_bound_odds} as \cref{eq:eacp_error_bound_prob}.}
Remember that $O_M(r \mid \boldsymbol e) = P_M(r \mid \boldsymbol e) \mathbin{/} (1 - P_M(r \mid \boldsymbol e))$ defines the odds of $r$ given $\boldsymbol e$.
Entering the definition of the odds into \cref{eq:eacp_distance_bound_odds} then results in
\begin{align}
	e^{-d} \leq \frac{P_{M'}(r \mid \boldsymbol e) \mathbin{/} (1 - P_{M'}(r \mid \boldsymbol e))}{P_M(r \mid \boldsymbol e) \mathbin{/} (1 - P_M(r \mid \boldsymbol e))} \leq e^d,
\end{align}
which can be rewritten as
\begin{align}
	e^{-d} \leq \frac{P_{M'}(r \mid \boldsymbol e)}{1 - P_{M'}(r \mid \boldsymbol e)} \cdot \frac{1 - P_M(r \mid \boldsymbol e)}{P_M(r \mid \boldsymbol e)} \leq e^d.
\end{align}
With $p = P_M(r \mid \boldsymbol e)$ and $p' = P_{M'}(r \mid \boldsymbol e)$, we obtain
\begin{align}
	e^{-d} \leq \frac{p' (1 - p)}{p (1 - p')} \leq e^d.
\end{align}
Multiplying by $p (1 - p')$ (which is always positive) yields
\begin{align}
	e^{-d} p (1 - p') \leq p' (1 - p) \leq e^d p (1 - p').
\end{align}
Expanding the terms gives us
\begin{align}
	e^{-d} p - e^{-d} p p' \leq p' - p' p \leq e^d p - e^d p p'.
\end{align}
By rearranging the terms, we end up with
\begin{align}
	e^{-d} p &\leq p' - p' p + e^{-d} p p' \\
	&= p' (1 - p + e^{-d} p),~\text{and} \\
	e^d p &\geq p' - p' p + e^d p p' \\
	&= p' (1 - p + e^d p).
\end{align}
Dividing by $1 - p + e^{-d} p$ and $1 - p + e^d p$ (both terms are always positive), respectively, results in
\begin{align}
	\frac{e^{-d} p}{1 - p + e^{-d} p} &\leq p',~\text{and} \\
	\frac{e^d p}{1 - p + e^d p} &\geq p'.
\end{align}
By rearranging the terms once more, we obtain
\begin{align}
	\frac{p e^{-d}}{p (e^{-d} - 1) + 1} &\leq p' \leq \frac{p e^d}{p (e^d - 1) + 1}.
\end{align}

\eacpErrorBoundTheorem*
\begin{proof}
	Let $\boldsymbol \Phi = \{ \phi_1, \ldots, \phi_m \}$ denote the set of factors in $M$.
	By definition, it holds that every updated potential $\varphi'$ in $M'$ differs by factor at most $(1 \pm \varepsilon)$ from its original potential $\varphi$ in $M$, independent of the distribution of the groups (that is, it is irrelevant whether all $m$ factors are in the same group or whether groups are distributed otherwise).
	Therefore, as $\psi(\boldsymbol r) = \prod_{j=1}^{m} \phi_j(\boldsymbol r_j)$ for any assignment $\boldsymbol r$, we obtain
	\begingroup
	\allowdisplaybreaks
	\begin{align}
		\psi'(\boldsymbol r) &\geq \prod_{j=1}^{m} \phi_j(\boldsymbol r_j) \cdot (1 - \varepsilon),~\text{and} \\
		\psi'(\boldsymbol r) &\leq \prod_{j=1}^{m} \phi_j(\boldsymbol r_j) \cdot (1 + \varepsilon).
	\end{align}
	\endgroup
	In consequence, we get the following bounds:
	\begingroup
	\allowdisplaybreaks
	\begin{align}
		\min_{\boldsymbol r} \frac{\psi'(\boldsymbol r)}{\psi(\boldsymbol r)} &\geq \frac{\prod\limits_{j = 1}^m \big( \phi_j(\boldsymbol r_j) \cdot (1 - \varepsilon) \big)}{\prod\limits_{j = 1}^m \phi_j(\boldsymbol r_j)},~\text{and} \\
		\max_{\boldsymbol r} \frac{\psi'(\boldsymbol r)}{\psi(\boldsymbol r)} &\leq \frac{\prod\limits_{j = 1}^m \big( \phi_j(\boldsymbol r_j) \cdot (1 + \varepsilon) \big)}{\prod\limits_{j = 1}^m \phi_j(\boldsymbol r_j)},
	\end{align}
	\endgroup
	where $\boldsymbol r_j$ is a projection of \emph{any} assignment $\boldsymbol r$ to the argument list of $\phi_j$.
	Entering these bounds into \cref{eq:eacp_distance_measure_only_joint_potential} then yields
	\begingroup
	\allowdisplaybreaks
	\begin{align}
		\phantom{=~} &D(P_M, P_{M'}) \\
		= &\ln \max_{\boldsymbol r} \frac{\psi'(\boldsymbol r)}{\psi(\boldsymbol r)} - \ln \min_{\boldsymbol r} \frac{\psi'(\boldsymbol r)}{\psi(\boldsymbol r)} \\
		\leq &\ln \frac{\prod\limits_{j = 1}^m \big( \phi_j(\boldsymbol r_j) \cdot (1 + \varepsilon) \big)}{\prod\limits_{j = 1}^m \phi_j(\boldsymbol r_j)}
		- \ln \frac{\prod\limits_{j = 1}^m \big( \phi_j(\boldsymbol r_j) \cdot (1 - \varepsilon) \big)}{\prod\limits_{j = 1}^m \phi_j(\boldsymbol r_j)} \\
		= &\ln \frac{(1 + \varepsilon)^m \prod\limits_{j = 1}^m \phi_j(\boldsymbol r_j)}{\prod\limits_{j = 1}^m \phi_j(\boldsymbol r_j)}
		- \ln \frac{(1 - \varepsilon)^m \prod\limits_{j = 1}^m \phi_j(\boldsymbol r_j)}{\prod\limits_{j = 1}^m \phi_j(\boldsymbol r_j)} \\
		= &\ln\ (1 + \varepsilon)^m - \ln\ (1 - \varepsilon)^m.
	\end{align}
	\endgroup
\end{proof}

\pairwiseEquivalenceGroupMeanLemma*
\begin{proof}
	We show the claim in two directions by proving that $\phi^*(\boldsymbol r) \in [\phi_i(\boldsymbol r) \cdot (1 - \varepsilon), \phi_i(\boldsymbol r) \cdot (1 + \varepsilon)]$ and $\phi_i(\boldsymbol r) \in [\phi^*(\boldsymbol r) \cdot (1 - \varepsilon), \phi^*(\boldsymbol r) \cdot (1 + \varepsilon)]$ for any assignment $\boldsymbol r$ and $\phi_i \in \boldsymbol G$.

	For the first direction, let $\boldsymbol r$ be an arbitrary assignment and let $\phi_i \in \boldsymbol G$.
	As all factors in $\boldsymbol G$ are pairwise $\varepsilon$-equivalent, it holds that $\phi_i(\boldsymbol r) \cdot (1 - \varepsilon) \leq \min_{\phi_j \in \boldsymbol G} \phi_j(\boldsymbol r)$ and $\max_{\phi_j \in \boldsymbol G} \phi_j(\boldsymbol r) \leq \phi_i(\boldsymbol r) \cdot (1 + \varepsilon)$.
	Further, as $\phi^*(\boldsymbol r)$ is the arithmetic mean over all $\phi_j(\boldsymbol r) \in \boldsymbol G$, it holds that $\min_{\phi_j \in \boldsymbol G} \phi_j(\boldsymbol r) \leq \phi^*(\boldsymbol r) \leq \max_{\phi_j \in \boldsymbol G} \phi_j(\boldsymbol r)$ and thus $\phi^*(\boldsymbol r) \in [\phi_i(\boldsymbol r) \cdot (1 - \varepsilon), \phi_i(\boldsymbol r) \cdot (1 + \varepsilon)]$.

	For the second direction, it holds that for any assignment $\boldsymbol r$, every $\phi_i \in \boldsymbol G$ is contained in the interval $[\phi_j(\boldsymbol r) \cdot (1 - \varepsilon), \phi_j(\boldsymbol r) \cdot (1 + \varepsilon)]$ for any $j \in \{ 1, \ldots, k \}$, and thus also in the smallest possible composite interval $[\max_{\phi_j \in \boldsymbol G} \phi_j(\boldsymbol r) \cdot (1 - \varepsilon), \min_{\phi_j \in \boldsymbol G} \phi_j(\boldsymbol r) \cdot (1 + \varepsilon)]$. With the same argument as before, namely that for the arithmetic mean $\phi^*(\boldsymbol r)$ we have $\min_{\phi_j \in \boldsymbol G} \phi_j(\boldsymbol r) \leq \phi^*(\boldsymbol r) \leq \max_{\phi_j \in \boldsymbol G} \phi_j(\boldsymbol r)$, the composite interval $[\max_{\phi_j \in \boldsymbol G} \phi_j(\boldsymbol r) \cdot (1 - \varepsilon), \min_{\phi_j \in \boldsymbol G} \phi_j(\boldsymbol r) \cdot (1 + \varepsilon)]$ is a subset of the interval $[\phi^*(\boldsymbol r) \cdot (1 - \varepsilon), \phi^*(\boldsymbol r)\cdot (1 + \varepsilon)]$ and hence, $\phi_i(\boldsymbol r) \in [\phi^*(\boldsymbol r) \cdot (1 - \varepsilon), \phi^*(\boldsymbol r) \cdot (1 + \varepsilon)]$.
\end{proof}

\lowerboundeacpErrorBoundTheorem*
\begin{proof}
	Let $\boldsymbol \Phi = \{ \phi_1, \ldots, \phi_m \}$ denote the set of factors in $M$, representing the distribution of $M$ via $P_M(\boldsymbol r) = \frac{1}{Z} \psi(\boldsymbol r) = \frac{1}{Z} \prod_{j=1}^{m} \phi_j(\boldsymbol r_j)$ for an assignment $\boldsymbol r = (r_1, \ldots, r_n) \in \times_{i=1}^n \range{R_i}$ to the \acp{rv} in $\boldsymbol R$.
	On the level of potentials, this means that for a specific assignment $\boldsymbol r$, there exist $j_i$, $i \in \{ 1, \ldots, m \}$, such that $\psi(\boldsymbol r) = \prod_{i=1}^{m} \varphi_{j_i,i}$, where $\varphi_{j,i}$ represents the potential in the $j^{th}$ row in the potential table of factor $\phi_i$.
	For $\phi^*(\boldsymbol r) = \frac{1}{m} \sum_{i=1}^{m} \phi_i(\boldsymbol r)$, its potentials $\varphi^*_{j} = \frac{1}{m} \sum_{i=1}^{m} \varphi_{j,i}$ are given by the row-wise arithmetic mean over all factors and thus are independent of $i$.

	According to \cref{lemma:eacp_mean_in_group_equivalence}, $\phi^*$ is pairwise $\varepsilon$-equivalent to all $\phi_i$ of a group of pairwise $\varepsilon$-equivalent factors.
	We first prove the claim for the case where all factors $\{ \phi_1, \ldots, \phi_m \}$ belong to the same group of pairwise $\varepsilon$-equivalence factors and afterwards generalise the proof to arbitrary distributions of groups.

	By ordering the potentials of every row $j$, we adopt, without loss of generality, the notation
	\begin{align}
		\varphi_{j,1} \leq \ldots \leq \varphi_{j,k_j} \leq \varphi_j^* \leq \varphi_{j,k_j + 1} \leq \ldots \leq \varphi_{j,m}
	\end{align}
	with $k_j \in \{ 1, \ldots, m-1 \}$.
	Since $\varphi_{j,i}\geq \varphi_j^*$ holds already for $ i=k_j+1,\ldots,m$, we want to determine a minimal $\alpha_2(j) \in \mathbb{R}_{\geq 1}^+$ such that also $\varphi_{j,i} \cdot \alpha_2 \geq \varphi_j^*$ holds for all $ i=1,\ldots,m$. This means that $\varphi_j^* \leq \min_{i=1,\ldots,m} \varphi_{j,i}\cdot \alpha_2(j) =\varphi_{j,1}\cdot \alpha_2(j)$ has to hold. The minimal possible constant to fulfil this equation is $\alpha_2(j):= \frac{\varphi_j^*}{\varphi_{j,1}}$. In order to get the value of this constant, we have to assume the worst case scenario for the distribution of the $\varphi_{j,i}$'s. Therefore, we are looking for a small $\varphi_{j,1}$ and the largest $\varphi_j^*$, which is $\varphi_j^* = \frac{1}{m}\sum_{i=1}^{m} \varphi_{j,i}$ and $\varphi_{j,1}(1+\varepsilon) \geq \max_{i=1,\ldots,m}\varphi_{j,i}=\varphi_{j,m}$ according to the definition of $\varepsilon$-equivalence. Under these conditions, the maximal $\varphi_j^*$ is possible by the choice of $\varphi_{j,i}:= \varphi_j^*$ for $i=2,\ldots,k_j$ and $\varphi_{j,i}:= (1+\varepsilon)\varphi_{j,1}$ for $i=k_j+1,\ldots,m$. This results in the following mean
	\begingroup
	\allowdisplaybreaks
	\begin{align}
		\varphi_j^* &= \frac{1}{m}\left(\varphi_{j,1} + \sum_{i=2}^{k_j} \varphi_{j,i} + \sum_{i=k_j+1}^{m} \varphi_{j,1}(1+\varepsilon) \right)\\
		&=  \frac{1}{m}\Big(\varphi_{j,1} + (k_j-1)\varphi_j^* +(1+\varepsilon)(m-k_j)\varphi_{j,1}\Big),
	\end{align}
	\endgroup
	which is equivalent to
	\begingroup
	\allowdisplaybreaks
	\begin{align}
		\varphi_j^*& \left(1-\frac{k_j-1}{m} \right) = \frac{1}{m} \varphi_{j,1}\big(1+(m-k_j)(1+\varepsilon) \big)\\
		\Leftrightarrow \varphi_j^*& = \frac{1}{m-k_j+1} \varphi_{j,1}\big(1+(m-k_j)(1+\varepsilon) \big)\\
		&= \frac{m-k_j+1+(m-k_j)\varepsilon}{m-k_j+1} \varphi_{j,1}\\
		&= \left(1+\frac{m-k_j}{m-k_j+1}\varepsilon\right)\varphi_{j,1}.
	\end{align}
	\endgroup
	This results in $\alpha_2(j) = \left(1+\frac{m-k_j}{m-k_j+1}\varepsilon\right)>1$ for a $k_j\in \{1,\ldots,m-1\}$. Within this condition, $k_j=1$ leads to the largest $\varepsilon$-amount for $\alpha_2 = \left(1+\frac{m-1}{m}\varepsilon\right)$ in the worst case and therefore for any assignment independently of $j$.

	The estimation of a second constant $\alpha_1(j)$ as a lower constant works similar.
	Since $\varphi_{j,i}\leq \phi_j^*$ for $i=1,\ldots,k_j$ holds already, we determine a maximal $\alpha_1(j)\in \mathbb{R}^+_{\leq 1}$ such that $\varphi_{j,i}\leq \varphi_j^*$ holds for all $i=1,\ldots,m$. This means that $\varphi_j^*\geq \max_{i=1,\ldots,m} \varphi_{j,i}\cdot \alpha_1(j)=\varphi_{j,m}\cdot \alpha_1(j)$.
	The maximal possible constant is $\alpha_1(j):= \frac{\varphi_j^*}{\varphi_{j,m}}$.
	For the worst case scenario for the distribution of the $\varphi_{j,i}$'s, we get a large $\varphi_{j,m}$ and a low $\varphi_j^*$ and according to \cref{le:smallimprovment} this means $\varphi_{j,m} \cdot \frac{1}{1+\varepsilon}\leq \min_{i=1,\ldots,m}\varphi_{j,i}$. A minimal $\varphi_j^*$ can be reached by the choice of $\varphi_{j,i}:=\varphi_j^*$ for $i=k_j+1,\ldots,m$ and $\varphi_{j,i}:=\varphi_{j,m}\frac{1}{1+\varepsilon}$ for $i=1,\ldots,k_j$. This results in
	\begingroup
	\allowdisplaybreaks
	\begin{align}
		\varphi_j^* &= \frac{1}{m}\left(\sum_{i=1}^{k_j} \varphi_{j,m}\frac{1}{1+\varepsilon} + \sum_{i=k_j+1}^{m-1} \varphi_j^* +\varphi_{j,m} \right)\\
		&= \frac{1}{m} \left( \frac{k_j}{1 + \varepsilon} \varphi_{j,m} + (m - k_j - 1) \varphi_j^* + \varphi_{j,m} \right),
	\end{align}
	\endgroup
	which is equivalent to
	\begingroup
	\allowdisplaybreaks
	\begin{align}
		\varphi_j^*& \left(1-\frac{m-k_j-1}{m} \right) = \frac{1}{m}\left(\frac{k_j}{1+\varepsilon}+1\right)\varphi_{j,m}\\
		\Leftrightarrow \varphi_j^*& = \frac{1}{m} \frac{1}{1-\frac{m-k_j-1}{m}} \left(\frac{k_j}{1+\varepsilon}+1\right)\varphi_{j,m}\\
		&= \frac{1}{k_j+1}\frac{k_j+1+\varepsilon}{1+\varepsilon} \varphi_{j,m}\\
		&= \left(1+\frac{1}{k_j+1}\varepsilon\right)\frac{1}{1+\varepsilon} \varphi_{j,m}.
	\end{align}
	\endgroup
	This results in $1>\alpha_1(j)=\left(1+\frac{1}{k_j+1}\varepsilon\right)\frac{1}{1+\varepsilon}\geq \frac{1}{1+\varepsilon}$ for $k_j\in \{1,\ldots,m-1\}$, which is minimal for $k_j=m-1$, leading to $\alpha_1=\left(1+\frac{1}{m}\varepsilon\right)\frac{1}{1+\varepsilon}$ in the worst case for any assignment and independently of $j$.

	If we combine these calculations, we obtain
	\begingroup
	\allowdisplaybreaks
	\begin{align}
		\psi'(\boldsymbol r) &\geq \prod_{j=1}^{m} \phi_j(\boldsymbol r_j) \cdot \alpha_1,~\text{and} \\
		\psi'(\boldsymbol r) &\leq \prod_{j=1}^{m} \phi_j(\boldsymbol r_j) \cdot \alpha_2.
	\end{align}
	\endgroup
	In consequence, we get the following bounds:
	\begingroup
	\allowdisplaybreaks
	\begin{align}
		\min_{\boldsymbol r} \frac{\psi'(\boldsymbol r)}{\psi(\boldsymbol r)} &\geq \frac{\prod\limits_{j = 1}^m  \phi_j(\boldsymbol r_j) \cdot \alpha_1}{\prod\limits_{j = 1}^m \phi_j(\boldsymbol r_j)},~\text{and} \\
		\max_{\boldsymbol r} \frac{\psi'(\boldsymbol r)}{\psi(\boldsymbol r)} &\leq \frac{\prod\limits_{j = 1}^m \phi_j(\boldsymbol r_j) \cdot \alpha_2}{\prod\limits_{j = 1}^m \phi_j(\boldsymbol r_j)},
	\end{align}
	\endgroup
	where $\boldsymbol r_j$ is a projection of \emph{any} assignment $\boldsymbol r$ to the argument list of $\phi_j$.
	Entering these bounds into \cref{eq:eacp_distance_measure_only_joint_potential} then yields
	\begingroup
	\allowdisplaybreaks
	\begin{align}
		&\phantom{=} D(P_M, P_{M'}) \\
		&= \ln \max_{\boldsymbol r} \frac{\psi'(\boldsymbol r)}{\psi(\boldsymbol r)} - \ln \min_{\boldsymbol r} \frac{\psi'(\boldsymbol r)}{\psi(\boldsymbol r)} \\
		&\leq \ln \frac{\prod\limits_{j = 1}^m  \phi_j(\boldsymbol r_j) \cdot \alpha_2 }{\prod\limits_{j = 1}^m \phi_j(\boldsymbol r_j)}
		- \ln \frac{\prod\limits_{j = 1}^m \phi_j(\boldsymbol r_j) \cdot \alpha_1 }{\prod\limits_{j = 1}^m \phi_j(\boldsymbol r_j)} \\
		&= \ln \prod_{j=1}^{m} \alpha_2 - \ln \prod_{j=1}^{m} \alpha_1\\
		&= \ln \left(\frac{\alpha_2}{\alpha_1}\right)^m\\
		&= \ln\left(\frac {1+\frac{m-1}{m}\varepsilon}{\frac{1+\frac{1}{m}\varepsilon}{1+\varepsilon}}\right)^m\\
		&= \ln \left(1+\frac{m-1}{m}\varepsilon\right)^m\left(\frac{1+\varepsilon}{1+\frac{\varepsilon}{m}}\right)^m \\
		&=\ln\left(\frac {\big( 1+\frac{m-1}{m}\varepsilon \big) \big( 1+\varepsilon \big)}{1+\frac{1}{m}\varepsilon}\right)^m\\
		&< \ln\ \big( 1 + \varepsilon \big)^{2m} \\
		&< \ln \left(\frac{1+\varepsilon}{1-\varepsilon}\right)^m.
	\end{align}
	\endgroup
	Next, we consider the general case of having an arbitrary distribution of groups of pairwise $\varepsilon$-equivalent factors.
	More specifically, let the factors now be distributed into $k \geq 2$ groups of pairwise $\varepsilon$-equivalent factors, meaning there is a set of summands $\{m_1,\ldots,m_k\}\in \mathbb{N}^k$ with $\sum_{i=1}^{k} m_i= m$ and $\phi_1,\ldots,\phi_{m_1}$ being pairwise $\varepsilon$-equivalent, $\phi_{m_1+1},\ldots,\phi_{m_1+m_2}$ being pairwise $\varepsilon$-equivalent, and so on. Since the previous case holds for a group of fully pairwise $\varepsilon$-equivalent factors we can apply this case $k$ times to get the maximal boundary as follows.
	\begingroup
	\allowdisplaybreaks
	\begin{align}
		&\phantom{=} D(P_M,P_{M'}) \\
		&= \ln \max_{\boldsymbol r} \frac{\psi'(\boldsymbol r)}{\psi(\boldsymbol r)} - \ln \min_{\boldsymbol r} \frac{\psi'(\boldsymbol r)}{\psi(\boldsymbol r)}\\
		&\leq \ln \prod_{j=1}^{m_1} \frac {1+\frac{m_1-1}{m_1}\varepsilon}{\frac{1+\frac{1}{m_1}\varepsilon}{1+\varepsilon}}\cdot\ldots\cdot \prod_{j=1}^{m_k} \frac {1+\frac{m_k-1}{m_k}\varepsilon}{\frac{1+\frac{1}{m_k}\varepsilon}{1+\varepsilon}}\\
		&= \ln \left(\frac {1+\frac{m_1-1}{m_1}\varepsilon}{\frac{1+\frac{1}{m_1}\varepsilon}{1+\varepsilon}}\right)^{m_1} \cdot\ldots\cdot
		\left(\frac {1+\frac{m_k-1}{m_k}\varepsilon}{\frac{1+\frac{1}{m_k}\varepsilon}{1+\varepsilon}}\right)^{m_k}
	\end{align}
	\endgroup
	Since $1+\frac{m_i-1}{m_i}\varepsilon < 1+\frac{m-1}{m}\varepsilon$ holds for the numerator and $\frac{1+\frac{1}{m}\varepsilon}{1+\varepsilon}< \frac{1+\frac{1}{m_i}\varepsilon}{1+\varepsilon}$ holds for the denominator for every $i=1,\ldots,k$ due to $m_i<m$, we can bound each sub-product individually from above by
	\begingroup
	\allowdisplaybreaks
	\begin{align}
		\left(\frac {1+\frac{m_i-1}{m_i}\varepsilon}{\frac{1+\frac{1}{m_i}\varepsilon}{1+\varepsilon}}\right)^{m_i}< \left(\frac {1+\frac{m-1}{m}\varepsilon}{\frac{1+\frac{1}{m}\varepsilon}{1+\varepsilon}}\right)^{m_i}.
	\end{align}
	\endgroup
	Putting everything back together yields the desired result:
	\begingroup
	\allowdisplaybreaks
	\begin{align}
		&\phantom{=} D(P_M,P_{M'}) \\
		&\leq \ln \left(\frac {1+\frac{m_1-1}{m_1}\varepsilon}{\frac{1+\frac{1}{m_1}\varepsilon}{1+\varepsilon}}\right)^{m_1} \cdot\ldots\cdot
		\left(\frac {1+\frac{m_k-1}{m_k}\varepsilon}{\frac{1+\frac{1}{m_k}\varepsilon}{1+\varepsilon}}\right)^{m_k}\\
		& < \ln \prod_{i=1}^{k} 
		\left(\frac {1+\frac{m-1}{m}\varepsilon}{\frac{1+\frac{1}{m}\varepsilon}{1+\varepsilon}}\right)^{m_i}\\
		&=\ln \left(\frac {1+\frac{m-1}{m}\varepsilon}{\frac{1+\frac{1}{m}\varepsilon}{1+\varepsilon}}\right)^{\sum_{i=1}^{k} m_i}\\
		&=\ln \left(\frac {1+\frac{m-1}{m}\varepsilon}{\frac{1+\frac{1}{m}\varepsilon}{1+\varepsilon}}\right)^m,
	\end{align}
	\endgroup
	which implies due to the strict inequality, in particular, that the second case of multiple groups of pairwise $\varepsilon$-equivalent factors cannot reach the upper bound.
\end{proof}

\optimalBoundTheorem*
\begin{table*}[t]
	\centering
	\input{files/example_worst_case.tex}
	\caption{A construction of an exemplary \ac{fg} $M$ such that running \cref{alg:eacp_eacp} on $M$ to obtain $M'$ results in $D(P_M, P_{M'})$ becoming maximal and reaching the boundary given in \cref{th:lowerboundeacp_error_bound_given_eps}.}
	\label{tab:example_worst_case}
\end{table*}
\begin{proof}
	We show that the boundary of \cref{th:lowerboundeacp_error_bound_given_eps} can be reached by constructing an \ac{fg} $M$ such that running \cref{alg:eacp_eacp} on $M$ to obtain $M'$ results in $D(P_M, P_{M'})$ hitting the boundary.
	Consider the outcome of \cref{alg:eacp_eacp} for the \ac{fg} $M = (\boldsymbol R \cup \boldsymbol \Phi, \boldsymbol E)$ whose factors are depicted in \cref{tab:example_worst_case}.
	More specifically, the set of \acp{rv} is given by $\boldsymbol R = \{ R_1, \ldots, R_m \}$ with $\range{R_1} = \ldots = \range{R_m} = \{ r_1, \ldots, r_{2m} \}$, and the set of factors is defined as $\boldsymbol \Phi = \{ \phi_1(R_i), \ldots, \phi_m(R_m) \}$ (i.e., every factor has exactly one \ac{rv} as argument and the arguments of all factors are distinct).
	The set of edges is given by $\boldsymbol E = \{ \{R_1, \allowbreak \phi_1\}, \allowbreak \ldots, \allowbreak \{R_m, \allowbreak \phi_m\} \}$ but is not relevant for the proof.
	All factors in $\boldsymbol \Phi$ are pairwise $\varepsilon$-equivalent and hence end up in the same group after running \cref{alg:eacp_eacp}.

	Therefore, after updating the potentials, all factors in $\boldsymbol \Phi$ are replaced by the arithmetic mean $\phi^* = (\varphi^*_{1}, \ldots, \varphi^*_{2m})$ (shown in the rightmost column of \cref{tab:example_worst_case}) such that
	\begin{align}
		\varphi^*_{r} =
		\begin{cases}
			r \big( 1 + \frac{1}{m} \varepsilon \big) & \text{for } r = 1, \ldots, m \\
			r \big( 1 + \frac{m-1}{m} \varepsilon \big) & \text{for } r = m+1 ,\ldots, 2m.
		\end{cases}
	\end{align}
	We next compute $\min_{\boldsymbol r} \frac{\psi'(\boldsymbol r)}{\psi (\boldsymbol r)}$.
	By definition, it holds that $\psi'(\boldsymbol r) = \prod_{i=1}^{m} \phi^*(\boldsymbol r)$.
	To obtain the minimum quotient, the denominator thus must be maximal while at the same time keeping the arithmetic mean small, which is the case for the diagonal entries $\varphi_{i,i}$, $i = 1, \ldots, m$, where $\varphi_{j,i}$ refers to the potential in the $j^{th}$ row in the potential table of factor $\phi_i$.
	Choosing any other columns within the first $m$ rows would decrease the denominator (while leaving the nominator unchanged) and thus cannot lead to the minimum quotient.
	Furthermore, choosing any rows starting from row $m+1$ also increases the result of the quotient because
	\begin{align}
		\frac{i (1 + \frac{1}{m} \varepsilon)}{i (1 + \varepsilon)} \leq \frac{i (1 + \frac{m-1}{m} \varepsilon)}{i (1 + \varepsilon)},
	\end{align}
	with a minimal right hand side for row $m+1$ to $2m$. 
	In other words, the factor of the denominator for the minimum would remain $(1+\varepsilon)$, while the mean $\varphi^*_i$ for the numerator increases independently of $i$, resulting in an overall increased quotient compared to choosing rows $1$ to $m$.
	Consequently, it holds that $\boldsymbol r_{\min} = (r_1, \allowbreak \ldots, \allowbreak r_m)$ fulfils the equation $\min_{\boldsymbol r} \frac{\psi'(\boldsymbol r)}{\psi(\boldsymbol r)} = \frac{\psi'(\boldsymbol r_{\min})}{\psi(\boldsymbol r_{\min})}$, resulting in
	\begingroup
	\allowdisplaybreaks
	\begin{align}
		\min_{\boldsymbol r} \frac{\psi'(\boldsymbol r)}{\psi (\boldsymbol r)} &= \prod_{i=1}^{m} \frac{\varphi_i^*}{\varphi_{i,i}} \\
		&= \prod_{i=1}^{m} \frac{i\left( 1 + \frac{1}{m} \varepsilon \right)}{i \left( 1 + \varepsilon \right)}\\
		&= \left( \frac{1 + \frac{1}{m} \varepsilon}{1 + \varepsilon} \right)^m.
	\end{align}
	\endgroup
	Analogously, we compute the maximum quotient $\max_{\boldsymbol r}\frac{\psi'(\boldsymbol r)}{\psi (\boldsymbol r)}$ by choosing $\boldsymbol r_{\max} = (r_{m+1}, \allowbreak \ldots, \allowbreak r_{2m})$ in the following way:
	\begingroup
	\allowdisplaybreaks
	\begin{align}
		\max_{\boldsymbol r} \frac{\psi'(\boldsymbol r)}{\psi (\boldsymbol r)} &= \prod_{i=m+1}^{2m} \frac{\varphi_i^*}{\varphi_{i,i-m}} \\
		&= \prod_{i=m+1}^{2m} \frac{i \left( 1 + \frac{m-1}{m} \varepsilon \right)}{i} \\
		&= \prod_{i=1}^{m} \frac{i \left( 1 + \frac{m-1}{m} \varepsilon \right)}{i} \\
		&= \left( 1 + \frac{m-1}{m} \varepsilon \right)^m.
	\end{align}
	\endgroup
	Inserting the minimum and maximum quotients into the definition of the distance measure $D(P_M, P_{M'})$, we obtain
	\begingroup
	\allowdisplaybreaks
	\begin{align}
		D(P_M, P_{M'}) &= \ln \max_{\boldsymbol r} \frac{\psi'(\boldsymbol r)}{\psi (\boldsymbol r)} - \ln \min_{\boldsymbol r} \frac{\psi'(\boldsymbol r)}{\psi (\boldsymbol r)} \\
		&= \ln \left( \frac{1+\frac{m-1}{m} \varepsilon}{\frac{1 + \frac{1}{m} \varepsilon}{1 + \varepsilon}} \right)^m \\
		&= \ln \left( \frac{(1+\frac{m-1}{m} \varepsilon)(1 + \varepsilon)}{1 + \frac{1}{m} \varepsilon} \right)^m,
	\end{align}
	\endgroup
	which is strictly smaller than $\ln\ (1 + \varepsilon)^m - \ln\ (1 - \varepsilon)^m$ but also strictly larger than $\ln\ (1 + \varepsilon)^m$ and effectively hits the bound given in \cref{th:lowerboundeacp_error_bound_given_eps}.

	In practice, however, our bounds improve significantly as soon as we deviate from this extreme scenario.
	This is because even minor dependencies between the maximum and minimum quotients suffice to obtain a tighter estimate.
	For instance, if there are more factors in a group than rows in their potential tables or if multiple groups of pairwise $\varepsilon$-equivalent factors exist, we can immediately conclude that the worst-case condition no longer holds, resulting in smaller overall values of $D$.
	This effect occurs since even minimal deviations from the worst-case scenario result in dependencies between the maximum and minimum for any assignment $\boldsymbol r$.
\end{proof}

\section{The Advanced Colour Passing Algorithm} \label{appendix:eacp_acp_in_detail}
The \ac{acp} algorithm introduced by \citet{Luttermann2024a} builds on the \acl{cp} algorithm (originally named CompressFactorGraph)~\citep{Kersting2009a,Ahmadi2013a} and solves the problem of constructing a lifted representation, encoded as a \ac{pfg}, from a given propositional \ac{fg}.
The idea of \ac{acp} is to first find indistinguishabilities in a propositional \ac{fg} and then group together symmetric subgraphs.
In particular, \ac{acp} looks for indistinguishabilities based on potentials of factors, on ranges and evidence of \acp{rv}, as well as on the graph structure by passing around colours.
\Cref{alg:eacp_acp} provides a formal description of the \ac{acp} algorithm, which proceeds as follows.

\begin{algorithm}[t]
	\caption{Advanced Colour Passing}
	\label{alg:eacp_acp}
	\alginput{An \ac{fg} $M = (\boldsymbol R \cup \boldsymbol \Phi, \boldsymbol E)$ and a set of observed \\\hspace*{\algorithmicindent} events (evidence) $\boldsymbol O = \{ E_1 = e_1, \ldots, E_{\ell} = e_{\ell} \}$.} \\
	\algoutput{A lifted representation $M'$, encoded as a \ac{pfg}, \\\hspace*{\algorithmicindent} which entails equivalent semantics as $M$.}
	\begin{algorithmic}[1]
		\State Assign each $R_i$ a colour according to $\range{R_i}$ and $\boldsymbol O$\;
		\State Assign each $\phi_i$ a colour according to order-independent potentials and rearrange arguments accordingly\; \label{line:eacp_acp_init_factor_colours}
		\Repeat
			\For{each factor $\phi \in \boldsymbol \Phi$}
				\State $signature_{\phi} \gets [\,]$\;
				\For{each \ac{rv} $R \in neighbours(M, \phi)$}
					\State\Comment{In order of appearance in $\phi$}\;
					\State $append(signature_{\phi}, R.colour)$\;
				\EndFor
				\State $append(signature_{\phi}, \phi.colour)$\;
			\EndFor
			\State Group together all $\phi$s with the same signature\;
			\State Assign each such cluster a unique colour\;
			\State Set $\phi.colour$ correspondingly for all $\phi$s\;
			\For{each \ac{rv} $R \in \boldsymbol R$}
				\State $signature_{R} \gets [\,]$\;
				\For{each factor $\phi \in neighbours(M, R)$}
					\If{$\phi$ is commutative w.r.t.\ $\boldsymbol S$ and $R \in \boldsymbol S$}
						\State $append(signature_{R}, (\phi.colour, 0))$\;
					\Else
						\State $append(signature_{R}, (\phi.colour, p(R, \phi)))$\;
					\EndIf
				\EndFor
				\State Sort $signature_{R}$ according to colour\;
				\State $append(signature_{R}, R.colour)$\;
			\EndFor
			\State Group together all $R$s with the same signature\;
			\State Assign each such cluster a unique colour\;
			\State Set $R.colour$ correspondingly for all $R$s\;
		\Until{grouping does not change}
		\State $M' \gets$ construct \acs{pfg} from groupings\; \label{line:eacp_acp_construct_pfg}
	\end{algorithmic}
\end{algorithm}

\Ac{acp} begins with the colour assignment to variable nodes, meaning that all \acp{rv} that have the same range and observed event are assigned the same colour.
Thereafter, \ac{acp} assigns a colour to every factor node such that factors representing equivalent potentials are assigned the same colour.
After the initial colour assignments, \ac{acp} begins to pass the colours around.
\Ac{acp} first passes the colours from every variable node to its neighbouring factor nodes and afterwards, every factor node $\phi$ sends both its colour and the position $p(R, \phi)$ of $R$ in $\phi$'s argument list to all of its neighbouring variable nodes $R$.
Factors that have symmetries within themselves, formally denoted as being commutative with respect to a subset $\boldsymbol S$ of their arguments, omit the position when sending their colour to a neighbouring variable node $R \in \boldsymbol S$.
We provide more details on commutative factors in \cref{appendix:eacp_approximate_symmetries_within_factors}.
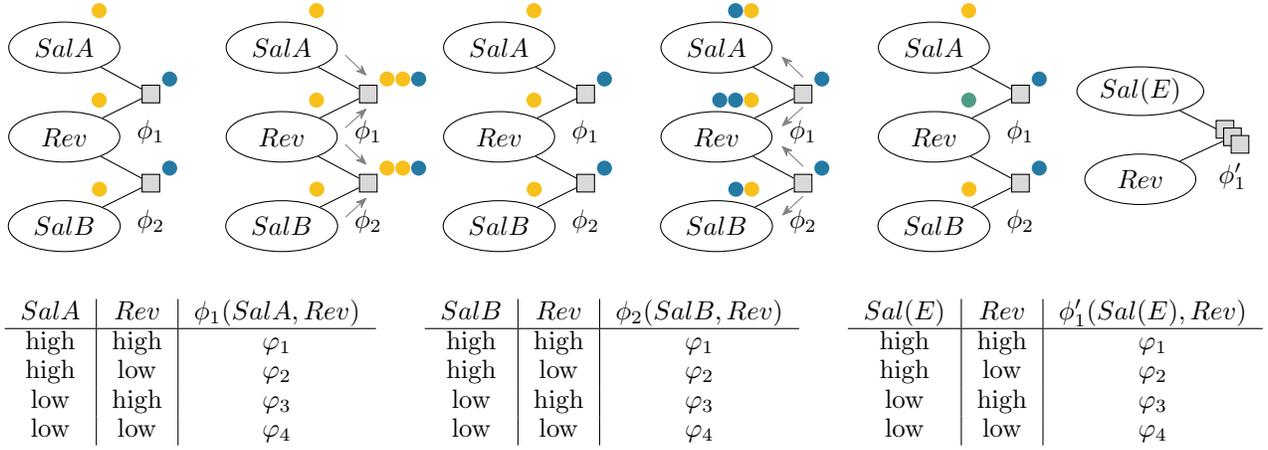
\begin{figure*}[t]
	\centering
	\input{files/acp_example.tex}
	\caption{A visualisation of the steps undertaken by the \ac{acp} algorithm (\cref{alg:eacp_acp}) on the input \ac{fg} from \cref{fig:eacp_example_fg} (left). All \acp{rv} have the same range ($\{ \low, \high \}$) and there are no observed events (evidence) available. Colours are first passed from variable nodes to factor nodes, followed by a recolouring, and then passed back from factor nodes to variable nodes, again followed by a recolouring. The colour passing procedure is then iterated until convergence (here, the colour assignments remain the same in the next iteration) and the resulting \ac{pfg} is depicted on the right. In the resulting \ac{pfg}, $Sal(E)$ is a so-called \acl{prv} that represents a group of \acp{rv} ($SalA$ and $SalB$) by using a \acl{lv} $E$ with domain $\domain{E} = \{ SalA, SalB \}$. Analogously, $\phi'_1$ now represents both $\phi_1$ and $\phi_2$ simultaneously.}
	\label{fig:acp_example}
\end{figure*}
\begin{example}
	\Cref{fig:acp_example} illustrates the course of the \ac{acp} algorithm on the \ac{fg} from \cref{fig:eacp_example_fg} with the modification that $\varphi_i = \varphi'_i$ for all $i \in \{ 1, \ldots, 4 \}$.
	Recall that $SalA$, $SalB$, and $Rev$ all have the range $\{ \low, \high \}$.
	We further assume that there are no observed events (evidence) and thus, $SalA$, $SalB$, and $Rev$ receive the same colour (e.g., $\mathrm{yellow}$).
	As the potentials of $\phi_1$ and $\phi_2$ are identical, $\phi_1$ and $\phi_2$ are assigned the same colour as well (e.g., $\mathrm{blue}$).
	The colour passing then starts from variable nodes to factor nodes, that is, $SalA$ and $Rev$ send their colour ($\mathrm{yellow}$) to $\phi_1$, and $Rev$ and $SalB$ send their colour ($\mathrm{yellow}$) to $\phi_2$.
	$\phi_1$ and $\phi_2$ are then recoloured according to the colours they received from their neighbours to reduce the communication overhead.
	Since $\phi_1$ and $\phi_2$ received identical colours (two times the colour $\mathrm{yellow}$), they are assigned the same colour during recolouring.
	Afterwards, the colours are passed from factor nodes to variable nodes and this time not only the colours but also the position of the \acp{rv} in the argument list of the corresponding factor are shared (because none of the factors is commutative with respect to a subset of its arguments having size at least two---see \cref{appendix:eacp_approximate_symmetries_within_factors} for more details).
	Consequently, $\phi_1$ sends a tuple $(\mathrm{blue}, 1)$ to $SalA$ and a tuple $(\mathrm{blue}, 2)$ to $Rev$, and $\phi_2$ sends a tuple $(\mathrm{blue}, 2)$ to $Rev$ and a tuple $(\mathrm{blue}, 1)$ to $SalB$ (positions are not shown in \cref{fig:acp_example}).
	As $SalA$ and $SalB$ are both at position one in the argument list of their respective neighbouring factor, they receive identical messages and are recoloured with the same colour.
	$Rev$ is assigned a different colour during recolouring than $SalA$ and $SalB$ because $Rev$ received different messages than $SalA$ and $SalB$.
	The groupings do not change in further iterations and hence the algorithm terminates.
	The output is the \ac{pfg} shown on the right in \cref{fig:acp_example}, where both $SalA$ and $SalB$ as well as $\phi_1$ and $\phi_2$ are grouped.
\end{example}
For more details on the colour passing procedure and the construction of the \ac{pfg} in the final step (\cref{line:eacp_acp_construct_pfg}), we refer the reader to \citet{Luttermann2024a}.
Note that when \ac{eacp} (\cref{alg:eacp_eacp}) calls \ac{acp} in \cref{line:eacp_call_acp}, the colour assignment of \ac{acp} to factors (\cref{line:eacp_acp_init_factor_colours} in \cref{alg:eacp_acp}) is replaced by the colour assignment of \ac{eacp} prior to calling \ac{acp}.
We also remark that \ac{eacp} does not use the \ac{pfg} output by \ac{acp} but instead takes the groups computed by \ac{acp}, updates them in phase three, and then applies the final step to construct the \ac{pfg}.

\section{Permutations of Factors' Arguments} \label{appendix:eacp_permuted_arguments}
In general, we cannot assume that the tables of equivalent factors (i.e., factors that encode equivalent underlying functions) read identical values from top to bottom.
More specifically, it is not always the case that indistinguishable \acp{rv} are located at the same position in their respective factors' argument lists.
The following example illustrates this point.
\begin{figure}
	\centering
	\input{files/example_fg_permuted.tex}
	\caption{Another \ac{fg} modeling the interplay between the revenue of a company ($Rev$) and the salaries of its employees ($SalA$ and $SalB$). In comparison to \cref{fig:eacp_example_fg}, the order of $\phi_2$'s arguments has now changed. Nevertheless, the semantics (i.e., the underlying full joint probability distribution) of the \ac{fg} remains the same as in \cref{fig:eacp_example_fg}.}
	\label{fig:eacp_example_fg_permuted}
\end{figure}
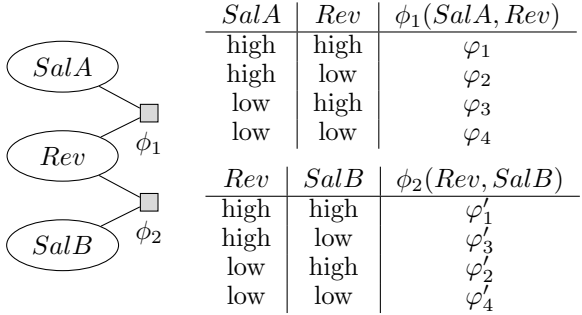
\begin{example}
	Consider the \ac{fg} $M_1$ shown in \cref{fig:eacp_example_fg_permuted}.
	In fact, $M_1$ entails equivalent semantics as the \ac{fg} $M_2$ depicted in \cref{fig:eacp_example_fg}.
	Observe that the function definition of $\phi_1$ is identical in $M_1$ and $M_2$ (i.e., its potential table is exactly the same).
	If we now take a look at $\phi_2$, we see that in $M_1$, the order of its arguments differs from the order of its arguments in $M_2$.
	In particular, $Rev$ is now located at position one and $SalB$ at position two in $M_1$ (opposed to $SalB$ at position one and $Rev$ at position two in $M_2$).
	The semantics of $\phi_2$, however, remains unchanged as all assignments are mapped to the same potential: $Rev = \high$, $SalB = \high$ is mapped to $\varphi'_1$, $Rev = \high, SalB = \low$ is mapped to $\varphi'_3$, $Rev = \low$, $SalB = \high$ is mapped to $\varphi'_2$, and $Rev = \low$, $SalB = \low$ is mapped to $\varphi'_4$.
	In consequence, $\phi_2$ entails the same semantics in $M_1$ as in $M_2$.
\end{example}
To construct a lifted representation such as a \ac{pfg}, indistinguishable \acp{rv} like $SalA$ and $SalB$ are required to be located at the same position in the argument list of their respective factors (otherwise, they cannot be grouped).
Therefore, to account for permutations of factors' arguments when looking for ($\varepsilon$-)equivalent factors, we first have to check whether there exists a rearrangement of the arguments such that the potential tables read identical ($\varepsilon$-equivalent, respectively) values from top to bottom.
If such a rearrangement exists, arguments are rearranged accordingly to ensure that indistinguishable \acp{rv} are actually located at the same position in their respective factors' argument lists.
After the rearrangement, it is guaranteed that the potential tables of ($\varepsilon$-)equivalent factors can be compared row-wise and the \ac{acp} algorithm (\cref{alg:eacp_acp}) assigns identical colours to them.
More details on how such rearrangements of arguments can be computed efficiently are given in \citep{Luttermann2024d}.

\section{Approximate Symmetries Within Factors} \label{appendix:eacp_approximate_symmetries_within_factors}
Symmetries within factors arise when arguments of the same factor are indistinguishable.
A factor that contains symmetries within itself is referred to as a \emph{commutative factor} in the literature.
The next definition formally introduces the notion of a commutative factor.
\begin{definition}[Commutative Factor, \citealp{Luttermann2024a}] \label{def:eacp_commutative_factor}
	Let $\phi(R_1, \ldots, R_n)$ denote a factor.
	We say that $\phi$ is \emph{commutative with respect to} $\boldsymbol S \subseteq \{R_1, \ldots, R_n\}$ if for all events $r_1, \ldots, r_n \in \times_{i=1}^n \range{R_i}$ it holds that $\phi(r_1, \ldots, r_n) = \phi(r_{\pi(1)}, \ldots, r_{\pi(n)})$ for all permutations $\pi$ of $\{1, \ldots, n\}$ with $\pi(i) = i$ for all $R_i \notin \boldsymbol S$.
	If $\phi$ is commutative with respect to $\boldsymbol S$, all arguments in $\boldsymbol S$ are called \emph{commutative arguments}.
\end{definition}
\begin{figure*}
	\centering
	\begin{subfigure}[t]{0.49\linewidth}
		\centering
		\input{files/example_fg_crv.tex}
		\caption{}
		\label{fig:eacp_example_fg_crv}
	\end{subfigure}
	\begin{subfigure}[t]{0.49\linewidth}
		\centering
		\input{files/example_pfg_crv.tex}
		\caption{}
		\label{fig:eacp_example_pfg_crv}
	\end{subfigure}
	\caption{(a) An \ac{fg} modelling the interplay between the revenue $Rev$ of a company and the competences $ComA$ and $ComB$ of its employees, and (b) a lifted representation, encoded as a \ac{pfg}, of the \ac{fg} shown in (a). For brevity, we omit the argument lists of $\phi_1$ and $\phi'_1$ and write $\phi_1$ instead of $\phi_1(Rev, ComA, ComB)$ and $\phi'_1$ instead of $\phi'_1(Rev, \#_E[Com(E)])$.}
	\label{fig:eacp_example_crv}
\end{figure*}
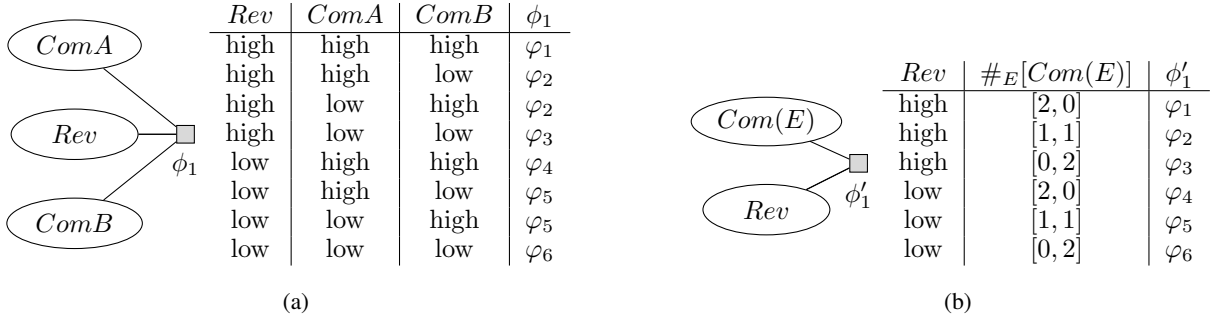
\begin{example}
	Consider the \ac{fg} shown in \cref{fig:eacp_example_fg_crv}, which models the interplay between a company's revenue $Rev$ and the competences $ComA$ and $ComB$ of its employees.
	It holds that $\range{ComA} = \allowbreak \range{ComB} = \allowbreak \range{Rev} = \allowbreak \{ \low, \allowbreak \high \}$.
	Here, $\phi_1$ is commutative with respect to $\boldsymbol S = \{ ComA, ComB \}$ because $\phi_1(Rev = \high, \allowbreak ComA = \high, \allowbreak ComB = \low) = \allowbreak \phi_1(Rev = \high, \allowbreak ComA = \low, \allowbreak ComB = \high) = \varphi_2$ and $\phi_1(Rev = \low, \allowbreak ComA = \high, \allowbreak ComB = \low) = \allowbreak \phi_1(Rev = \low, \allowbreak ComA = \low, \allowbreak ComB = \high) = \varphi_5$.
	In other words, the order of $ComA$ and $ComB$ in $\phi_1$ does not matter, i.e., it is only relevant how many employees have a high competence and how many have a low competence but not which specific employees have a high or low competence.
	In consequence, $\phi_1$ can be represented more compactly without losing any information.
	\Cref{fig:eacp_example_pfg_crv} shows a \ac{pfg} encoding equivalent semantics as the \ac{fg} from \cref{fig:eacp_example_fg_crv} by using a so-called \acl{crv} $\#_E[Com(E)]$ that counts over the competences of all employees.
	More specifically, instead of listing the competence of every employee separately, the \acl{crv} now specifies counts for the ranges values of $ComA$ and $ComB$: $[2,0]$ represents the case that there are two employees with a high competence and none with a low competence, $[1,1]$ represents the case that there is one employee with a high competence and one with a low competence, and $[0,2]$ represents the case that there are two employees with a low competence and none with a high competence.
	Note that the potential table is now smaller than in the initial \ac{fg} from \cref{fig:eacp_example_fg_crv} but still encodes exactly the same semantics.
\end{example}
The previous example gives a glimpse of how \emph{exact} symmetries within a factor can be exploited to obtain a more compact lifted representation.
We remark that it is also possible to allow for approximate symmetries within factors by replacing the exact equality in \cref{def:eacp_commutative_factor} with the notion of $\varepsilon$-equivalence.
To enable the exploitation of approximate symmetries within factors in the \ac{eacp} algorithm, the call of \ac{acp} is adjusted such that \ac{acp} computes commutative subsets of arguments using $\varepsilon$-equivalence instead of exact equality.
When inserting a \acl{crv} in the final \ac{pfg}, the potential table is again constructed by using the arithmetic mean over the original potentials.
By doing so, every updated potential still differs by factor at most $(1 \pm \varepsilon)$ from its original potential and hence, the bounds on the change in query results from \cref{sec:eacp_theoretical} continue to hold.

\section{Further Experimental Results} \label{appendix:eacp_further_experimental_results}
In addition to the results given in \cref{sec:eacp_experiments}, we give separate plots illustrating the distributions of quotients of query results in the modified model and query results in the original model for specific choices of $x$ (proportions of factors whose potential tables differ by factor at most $(1 \pm \varepsilon)$ from their original potential table) and $\varepsilon$.
We also investigate the overhead introduced by \ac{eacp} compared to \ac{acp} in terms of the number of queries needed to amortise the additional offline effort of computing groups of pairwise $\varepsilon$-equivalent factors.

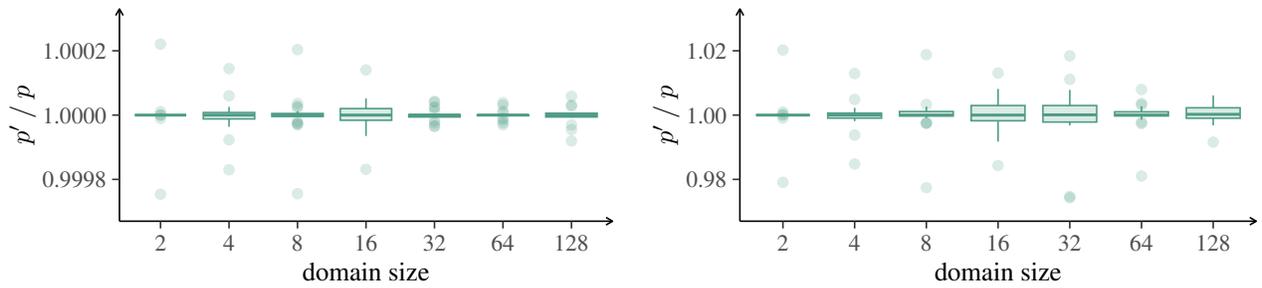
\begin{figure*}
	\centering
	\input{files/plot-quots-p=0.1-eps=0.001.tex}
	\input{files/plot-quots-p=0.1-eps=0.1.tex}
	\caption{Boxplots showing the distribution of the quotient $p' \mathbin{/} p$ with $p' = P_{M'}(r \mid \boldsymbol e)$ and $p = P_{M}(r \mid \boldsymbol e)$ for a proportion of $x = 0.1$ of factors whose potential tables deviate by factor at most $(1 \pm \varepsilon)$ from the original potential tables, where $\varepsilon = 0.001$ (left) and $\varepsilon = 0.1$ (right).}
	\label{fig:eacp_plot-quots-p=0.1}
\end{figure*}
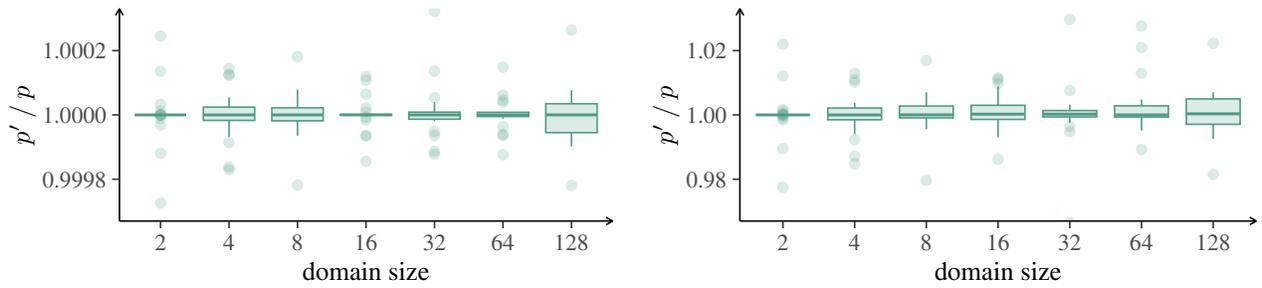
\begin{figure*}
	\centering
	\input{files/plot-quots-p=0.3-eps=0.001.tex}
	\input{files/plot-quots-p=0.3-eps=0.1.tex}
	\caption{Boxplots showing the distribution of the quotient $p' \mathbin{/} p$ with $p' = P_{M'}(r \mid \boldsymbol e)$ and $p = P_{M}(r \mid \boldsymbol e)$ for a proportion of $x = 0.3$ of factors whose potential tables deviate by factor at most $(1 \pm \varepsilon)$ from the original potential tables, where $\varepsilon = 0.001$ (left) and $\varepsilon = 0.1$ (right).}
	\label{fig:eacp_plot-quots-p=0.3}
\end{figure*}
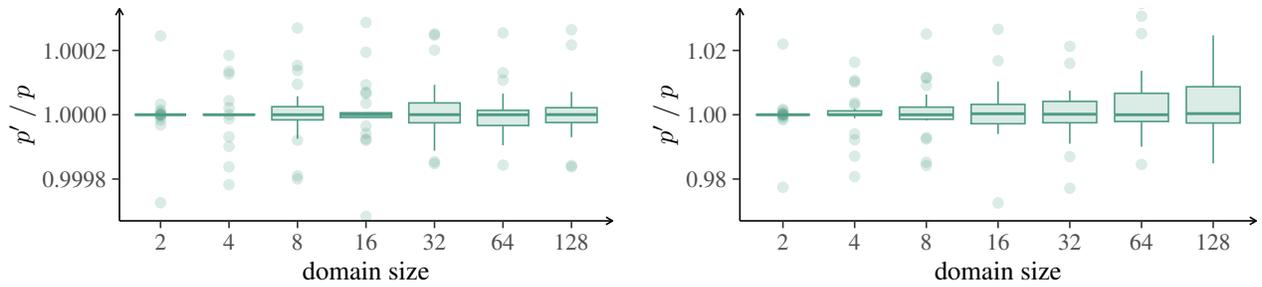
\begin{figure*}
	\centering
	\input{files/plot-quots-p=0.5-eps=0.001.tex}
	\input{files/plot-quots-p=0.5-eps=0.1.tex}
	\caption{Boxplots showing the distribution of the quotient $p' \mathbin{/} p$ with $p' = P_{M'}(r \mid \boldsymbol e)$ and $p = P_{M}(r \mid \boldsymbol e)$ for a proportion of $x = 0.5$ of factors whose potential tables deviate by factor at most $(1 \pm \varepsilon)$ from the original potential tables, where $\varepsilon = 0.001$ (left) and $\varepsilon = 0.1$ (right).}
	\label{fig:eacp_plot-quots-p=0.5}
\end{figure*}
\begin{figure*}
	\centering
	\input{files/plot-quots-p=0.7-eps=0.001.tex}
	\input{files/plot-quots-p=0.7-eps=0.1.tex}
	\caption{Boxplots showing the distribution of the quotient $p' \mathbin{/} p$ with $p' = P_{M'}(r \mid \boldsymbol e)$ and $p = P_{M}(r \mid \boldsymbol e)$ for a proportion of $x = 0.7$ of factors whose potential tables deviate by factor at most $(1 \pm \varepsilon)$ from the original potential tables, where $\varepsilon = 0.001$ (left) and $\varepsilon = 0.1$ (right).}
	\label{fig:eacp_plot-quots-p=0.7}
\end{figure*}
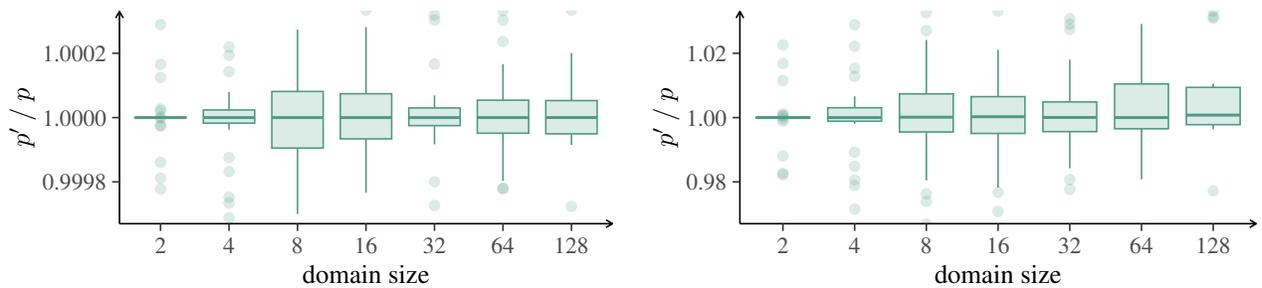
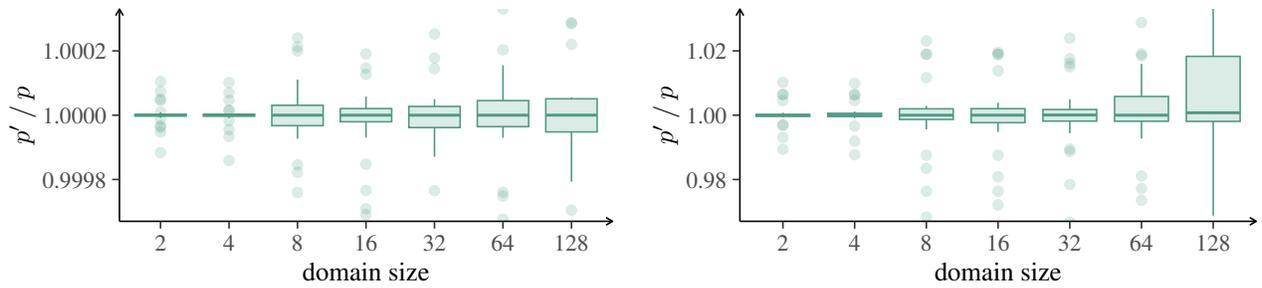
\begin{figure*}
	\centering
	\input{files/plot-quots-p=0.9-eps=0.001.tex}
	\input{files/plot-quots-p=0.9-eps=0.1.tex}
	\caption{Boxplots showing the distribution of the quotient $p' \mathbin{/} p$ with $p' = P_{M'}(r \mid \boldsymbol e)$ and $p = P_{M}(r \mid \boldsymbol e)$ for a proportion of $x = 0.9$ of factors whose potential tables deviate by factor at most $(1 \pm \varepsilon)$ from the original potential tables, where $\varepsilon = 0.001$ (left) and $\varepsilon = 0.1$ (right).}
	\label{fig:eacp_plot-quots-p=0.9}
\end{figure*}
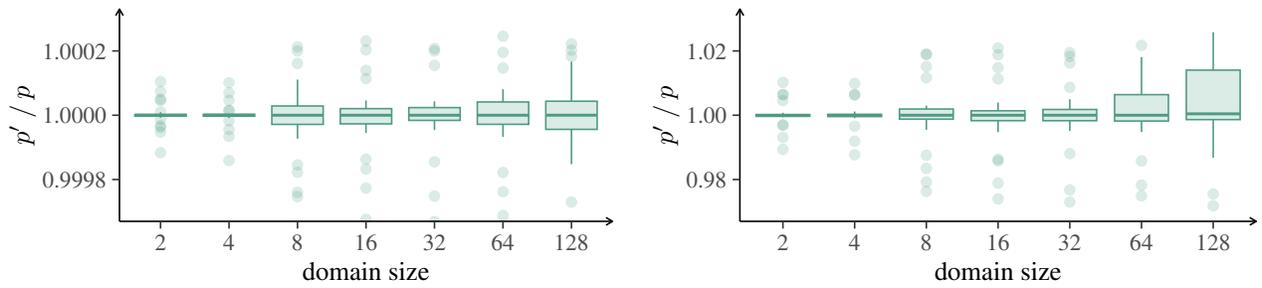
\begin{figure*}
	\centering
	\input{files/plot-quots-p=1-eps=0.001.tex}
	\input{files/plot-quots-p=1-eps=0.1.tex}
	\caption{Boxplots showing the distribution of the quotient $p' \mathbin{/} p$ with $p' = P_{M'}(r \mid \boldsymbol e)$ and $p = P_{M}(r \mid \boldsymbol e)$ for a proportion of $x = 1.0$ of factors whose potential tables deviate by factor at most $(1 \pm \varepsilon)$ from the original potential tables, where $\varepsilon = 0.001$ (left) and $\varepsilon = 0.1$ (right).}
	\label{fig:eacp_plot-quots-p=1.0}
\end{figure*}

\Cref{fig:eacp_plot-quots-p=0.1,fig:eacp_plot-quots-p=0.3,fig:eacp_plot-quots-p=0.5,fig:eacp_plot-quots-p=0.7,fig:eacp_plot-quots-p=0.9,fig:eacp_plot-quots-p=1.0} depict the distributions of quotients of query results in the modified model and query results in the original model for various choices of $x$.
In each of the figures, the left plot shows distributions of quotients for each domain size $k \in \{ 2, 4, 8, 16, 32, 64, 128 \}$ for $\varepsilon = 0.001$ while the right plot shows distributions of quotients for each domain size $k$ for $\varepsilon = 0.1$.
\Cref{fig:eacp_plot-quots-p=0.1} shows distributions of quotients for a proportion of $x = 0.1$ of factors that are manipulated such that their potential tables deviate by factor at most $(1 \pm \varepsilon)$, \cref{fig:eacp_plot-quots-p=0.3} shows distributions of quotients for $x = 0.3$, \cref{fig:eacp_plot-quots-p=0.5} for $x = 0.5$, \cref{fig:eacp_plot-quots-p=0.7} for $x = 0.7$, \cref{fig:eacp_plot-quots-p=0.9} for $x = 0.9$, and \cref{fig:eacp_plot-quots-p=1.0} for $x = 1.0$.
As expected, the quotients are generally larger for $\varepsilon = 0.1$ than for $\varepsilon = 0.001$ independent of the choice of $x$.
All quotients are again close to one---in particular, \cref{fig:eacp_plot-quots-p=0.1,fig:eacp_plot-quots-p=0.3,fig:eacp_plot-quots-p=0.5,fig:eacp_plot-quots-p=0.7,fig:eacp_plot-quots-p=0.9,fig:eacp_plot-quots-p=1.0} exhibit similar patterns as the right plot in \cref{fig:eacp_plot-avg} (again, even outliers remain very close to one).
With increasing value of $x$, the distributions of quotients become less concentrated at value one (that is, the boxes span a wider range when going from \cref{fig:eacp_plot-quots-p=0.1} to \cref{fig:eacp_plot-quots-p=1.0}).
In other words, there are more queries for which the quotient is further away from one (however, \enquote{further away} still refers to numbers extremely close to one).
To summarise, even if a large proportion of potential tables is modified by adding noise controlled by a factor $(1 \pm \varepsilon)$, the approximation error remains very small---even for the choice of $\varepsilon = 0.1$.
Next, we take a look at the overhead introduced by \ac{eacp} to compute groups of pairwise $\varepsilon$-equivalent factors by comparing the offline run times of \ac{eacp} and \ac{acp}.

\Cref{fig:eacp_plot-offline-p=0.1-3,fig:eacp_plot-offline-p=0.5-7,fig:eacp_plot-offline-p=0.9-1} report the average number $\alpha$ of queries after which the additional offline effort of \ac{eacp} compared to \ac{acp} amortises.
More specifically, $\alpha$ is defined as $\alpha = \Delta_o \mathbin{/} \Delta_g$ with $\Delta_o = t_{\ac{eacp}} - t_{\ac{acp}}$ being the difference between the offline run time of \ac{eacp} and \ac{acp} and $\Delta_g = t_{\acs{lve}-\ac{acp}} - t_{\acs{lve}-\ac{eacp}}$ being the difference of the run time of \acl{lve} on the output of \ac{acp} and \ac{eacp} to answer a query.
In other words, after $\alpha$ queries, the additional time needed by \ac{eacp} to compute groups of pairwise $\varepsilon$-equivalent factors is saved by the faster inference times to answer queries on the output of \ac{eacp}.

\begin{figure*}
	\centering
	\input{files/plot-offline-p=0.1.tex}
	\input{files/plot-offline-p=0.3.tex}
	\caption{Boxplots illustrating the distributions of the number $\alpha$ of queries after which the additional offline effort of \ac{eacp} amortises for input \acp{fg} containing a proportion of $x = 0.1$ (left) and $x = 0.3$ (right) of factors that are modified by factor at most $(1 \pm \varepsilon)$.}
	\label{fig:eacp_plot-offline-p=0.1-3}
\end{figure*}
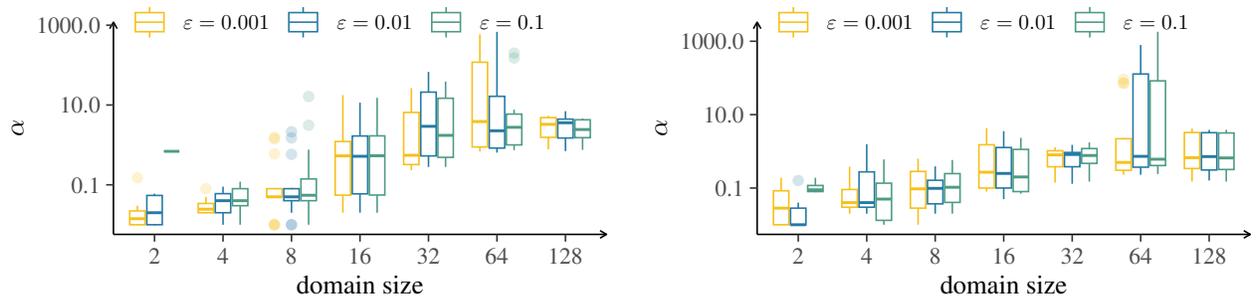
\begin{figure*}
	\centering
	\input{files/plot-offline-p=0.5.tex}
	\input{files/plot-offline-p=0.7.tex}
	\caption{Boxplots illustrating the distributions of the number $\alpha$ of queries after which the additional offline effort of \ac{eacp} amortises for input \acp{fg} containing a proportion of $x = 0.5$ (left) and $x = 0.7$ (right) of factors that are modified by factor at most $(1 \pm \varepsilon)$.}
	\label{fig:eacp_plot-offline-p=0.5-7}
\end{figure*}
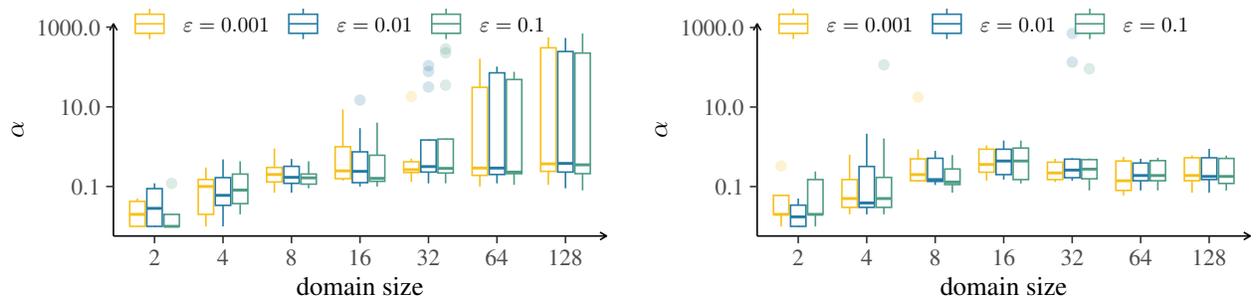
\begin{figure*}
	\centering
	\input{files/plot-offline-p=0.9.tex}
	\input{files/plot-offline-p=1.tex}
	\caption{Boxplots illustrating the distributions of the number $\alpha$ of queries after which the additional offline effort of \ac{eacp} amortises for input \acp{fg} containing a proportion of $x = 0.9$ (left) and $x = 1.0$ (right) of factors that are modified by factor at most $(1 \pm \varepsilon)$.}
	\label{fig:eacp_plot-offline-p=0.9-1}
\end{figure*}
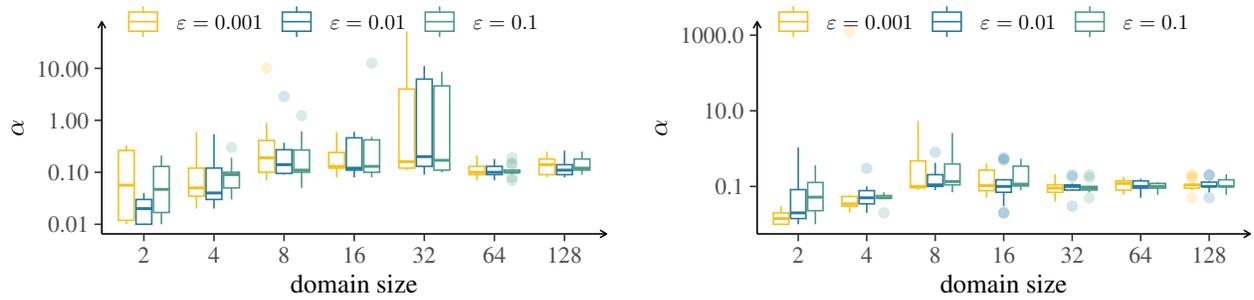

The boxplots depicted in \cref{fig:eacp_plot-offline-p=0.1-3,fig:eacp_plot-offline-p=0.5-7,fig:eacp_plot-offline-p=0.9-1} show that the median value for $\alpha$ is always smaller than ten and most of the time even smaller than one.
Thus, after less than ten queries, the offline overhead of \ac{eacp} amortises in most scenarios.
There are some outliers where $\alpha > 1000$, however, the overhead introduced by \ac{eacp} is in the scale of milliseconds to a few seconds in these cases (in fact, the overhead is always smaller than ten seconds, that is, \ac{eacp} takes at most ten seconds longer than \ac{acp} in every scenario).
We can therefore conclude that the offline overhead is not only small but also amortises after just a few queries in general.
Overall, there is a tendency that the larger $x$, the smaller is $\alpha$ (apart from a few outliers), which is expected as the advantage of \ac{eacp} increases with larger values of $x$.
However, when comparing the different domain sizes $k \in \{ 2, 4, 8, 16, 32, 64, 128 \}$, there is no clear pattern on how $\alpha$ behaves.
In the left plot of \cref{fig:eacp_plot-offline-p=0.5-7}, for example, $\alpha$ has a tendency to become larger for larger domain sizes.
This pattern, however, cannot be observed in general in the other plots.
\end{document}

%% file: defs.tex
\acrodef{acp}[ACP]{Advanced Colour Passing}
\acrodef{bn}[BN]{Bayesian network}
\acrodef{cp}[CP]{Colour Passing}
\acrodef{crv}[CRV]{counting randvar}
\acrodef{decor}[DECOR]{detection of commutative factors}
\acrodef{deft}[DEFT]{detection of exchangeable factors}
\acrodef{eacp}[$\varepsilon$-ACP]{$\varepsilon$-Advanced Colour Passing}
\acrodef{er}[ER]{entity-relationship}
\acrodef{fg}[FG]{factor graph}
\acrodef{kld}[KLD]{Kullback-Leibler divergence}
\acrodef{ljt}[LJT]{lifted junction tree}
\acrodef{lv}[logvar]{logical variable}
\acrodef{lve}[LVE]{lifted variable elimination}
\acrodef{mln}[MLN]{Markov logic network}
\acrodef{mn}[MN]{Markov network}
\acrodef{mpdag}[MPDAG]{maximally oriented partially directed acyclic graph}
\acrodef{pcfg}[PCFG]{parametric causal factor graph}
\acrodef{pcrv}[PCRV]{parameterised CRV}
\acrodef{ppcfg}[PPCFG]{partially directed parametric causal factor graph}
\acrodef{pf}[parfactor]{parametric factor}
\acrodef{pfg}[PFG]{parametric factor graph}
\acrodef{prv}[PRV]{parameterised randvar}
\acrodef{rv}[randvar]{random variable}
\acrodef{ve}[VE]{variable elimination}
\acrodef{wl}[WL]{Weisfeiler-Leman}

\crefname{algocf}{Alg.}{Algs.}
\Crefname{algocf}{Algorithm}{Algorithms}
\crefname{algorithm}{Alg.}{Algs.}
\Crefname{algorithm}{Algorithm}{Algorithms}
\crefname{corollary}{Cor.}{Cors.}
\Crefname{corollary}{Corollary}{Corollaries}
\crefname{definition}{Def.}{Defs.}
\Crefname{definition}{Definition}{Definitions}
\crefname{example}{Ex.}{Exs.}
\Crefname{example}{Example}{Examples}
\crefname{line}{Line}{Lines}
\Crefname{line}{Line}{Lines}
\crefname{proposition}{Prop.}{Props.}
\Crefname{proposition}{Proposition}{Propositions}
\crefname{section}{Sec.}{Secs.}
\Crefname{section}{Section}{Sections}
\crefname{theorem}{Thm.}{Thms.}
\Crefname{theorem}{Theorem}{Theorems}

\DeclareMathOperator*{\argmin}{arg\,min}

\newcommand{\alginput}[1]{\hspace*{\algorithmicindent} \textbf{Input:} #1}
\newcommand{\algoutput}[1]{\hspace*{\algorithmicindent} \textbf{Output:} #1}
\newcommand{\abs}[1]{\lvert #1 \rvert}
\newcommand{\approxEquiv}{\ensuremath{=_{\varepsilon}}}

\newcommand{\domain}[1]{\ensuremath{\mathrm{dom}(#1)}}

\newcommand{\range}[1]{\ensuremath{\mathrm{range}(#1)}}

\newcommand{\low}{\ensuremath{\mathrm{low}}}

\newcommand{\high}{\ensuremath{\mathrm{high}}}

\definecolor{myyellow}{RGB}{247,192,26}
\definecolor{myblue}{RGB}{37,122,164}
\definecolor{mygreen}{RGB}{78,155,133}
\definecolor{mypurple}{RGB}{86,51,94}

\definecolor{newblue}{RGB}{50,113,173}
\definecolor{newred}{RGB}{222,32,36}
\definecolor{newgreen}{RGB}{70,165,69}
\definecolor{newpurple}{RGB}{140,69,152}

\definecolor{cborange}{RGB}{230,159,0}
\definecolor{cbblue}{RGB}{30,136,229}
\definecolor{cbbluedark}{RGB}{46,37,133}
\definecolor{cbpurple}{RGB}{170,68,153}
\definecolor{cbgreen}{RGB}{0,77,64}
\definecolor{cbgreenlight}{RGB}{93,168,153}
\definecolor{cbbrown}{RGB}{126,41,84}

\pgfdeclarelayer{bg}
\pgfsetlayers{bg,main}

\tikzset{
	rv/.style={draw, ellipse},
	pf/.style={draw, rectangle, fill = gray!30},
	arc/.style = {->, >={[round,sep]Stealth}},
}

\newcommand\factor[6]{
	\node[pf, #1=#3 of #2, label={#4:{#5}}](#6) {};
}

\newcommand\nodecolorshift[5]{
	\node[circle, fill=#1, above right=0.1cm of #2, inner sep=0pt, minimum size=2mm, xshift=#4, yshift=#5](#3) {};
}

\newcommand\factorcolor[3]{
	\node[circle, fill=#1, above right=0cm and 0.05cm of #2, inner sep=0pt, minimum size=2mm](#3) {};
}
\newcommand\factorcolorshift[4]{
	\node[circle, fill=#1, above right=0cm and 0.05cm of #2, inner sep=0pt, minimum size=2mm, xshift=#4](#3) {};
}

\newcommand\pfs[8]{
	\node[pf, #1=#3 of #2, xshift=-1mm, yshift=1mm](#6) {};
	\node[pf, #1=#3 of #2, label={[label distance=1mm]#4:{#5}}](#7) {};
	\node[pf, #1=#3 of #2, xshift=1mm, yshift=-1mm](#8) {};
}

%% file: files/example_fg.tex
\begin{tikzpicture}
	\node[ellipse, draw, minimum width = 4.2em] (A) {$SalA$};
	\node[ellipse, draw, minimum width = 4.2em] (B) [below = 0.5cm of A] {$Rev$};
	\node[ellipse, draw, minimum width = 4.2em] (C) [below = 0.5cm of B] {$SalB$};
	\factor{below right}{A}{0.25cm and 0.5cm}{270}{$\phi_1$}{f1}
	\factor{below right}{B}{0.25cm and 0.5cm}{270}{$\phi_2$}{f2}

	\node[right = 0.5cm of f1, yshift=5.0mm] (tab_f1) {
		\begin{tabular}{c|c|c}
			$SalA$ & $Rev$ & $\phi_1(SalA,Rev)$ \\ \hline
			\high  & \high & $\varphi_1$ \\
			\high  & \low  & $\varphi_2$ \\
			\low   & \high & $\varphi_3$ \\
			\low   & \low  & $\varphi_4$ \\
		\end{tabular}
	};

	\node[right = 0.5cm of f2, yshift=-5.0mm] (tab_f2) {
		\begin{tabular}{c|c|c}
			$SalB$ & $Rev$ & $\phi_2(SalB,Rev)$ \\ \hline
			\high  & \high & $\varphi'_1$ \\
			\high  & \low  & $\varphi'_2$ \\
			\low   & \high & $\varphi'_3$ \\
			\low   & \low  & $\varphi'_4$ \\
		\end{tabular}
	};

	\draw (A) -- (f1);
	\draw (B) -- (f1);
	\draw (B) -- (f2);
	\draw (C) -- (f2);
\end{tikzpicture}

%% file: files/example_eps_equiv_mappings.tex
\begin{tikzpicture}
	\node (t1) {
		\setlength{\tabcolsep}{5pt}
		\begin{tabular}{ccccc}
			\toprule
			$R_1^i$ & $R_2^i$ & $\phi_1(R_1^1,R_2^1)$ & $\phi_2(R_1^2,R_2^2)$ & $\phi_3(R_1^3,R_2^3)$ \\ \midrule
			\high   & \high   & $0.75$                & $0.8$                 & $0.84$ \\
			\high   & \low    & $0.33$                & $0.3$                 & $0.31$ \\
			\low    & \high   & $0.48$                & $0.5$                 & $0.51$ \\
			\low    & \low    & $0.22$                & $0.2$                 & $0.22$ \\ \bottomrule
		\end{tabular}
	};
\end{tikzpicture}

%% file: files/example_eps_equiv_intervals.tex
\begin{tikzpicture}
	\node (t1) {
		\begin{tabular}{ccc}
			\toprule
			$\phi_1 \cdot (1 \mp \varepsilon)$ & $\phi_2 \cdot (1 \mp \varepsilon)$ & $\phi_3 \cdot (1 \mp \varepsilon)$ \\ \midrule
			$[0.675,0.825]$                    & $[0.72,0.88]$                      & $[0.756,0.924]$ \\
			$[0.297,0.363]$                    & $[0.27,0.33]$                      & $[0.279,0.341]$ \\
			$[0.432,0.528]$                    & $[0.45,0.55]$                      & $[0.459,0.561]$ \\
			$[0.198,0.242]$                    & $[0.18,0.22]$                      & $[0.198,0.242]$ \\ \bottomrule
		\end{tabular}
	};
\end{tikzpicture}

%% file: files/bound_plot_m=10.tex
\begin{tikzpicture}
	\pgfmathdeclarefunction{d}{2}{%
		\pgfmathparse{ln((1 + #1)^#2) - ln((1 - #1)^#2)}%
	}
	\pgfmathdeclarefunction{lb}{2}{%
		\pgfmathparse{(#1 * exp(-#2)) / (#1 * (exp(-#2) - 1) + 1)}%
	}
	\pgfmathdeclarefunction{ub}{2}{%
		\pgfmathparse{(#1 * exp(#2)) / (#1 * (exp(#2) - 1) + 1)}%
	}

	\begin{axis}[
		axis lines = left,
		xlabel = \(p\),
		x label style = {
			at = {(axis description cs:1.01,0)},
			anchor = west,
		},
		ylabel = {Bound},
		y label style = {
			at = {(axis description cs:0,1.05)},
			anchor = south,
			rotate = -90,
		},
		width = 15em,
		height = 11em,
		legend style = {
			draw = none,
			anchor = north west,
			at = {(0.02, 1.00)},
		},
	]

	\addplot[
		domain = 0:1, 
		samples = 100, 
		color = mygreen,
		semithick,
		dashed,
	]
	(x, {lb(x, d(0.01, 10))});

	\addplot[
		domain = 0:1, 
		samples = 100, 
		color = mygreen,
		semithick,
		dashed,
	]
	(x, {ub(x, d(0.01, 10))});

	\addplot[
		domain = 0:1, 
		samples = 100, 
		color = myyellow,
		semithick,
	]
	(x, {lb(x, d(0.001, 10))});

	\addplot[
		domain = 0:1, 
		samples = 100, 
		color = myyellow,
		semithick,
	]
	(x, {ub(x, d(0.001, 10))});
	\end{axis}
\end{tikzpicture}

%% file: files/bound_plot_m=100.tex
\begin{tikzpicture}
	\pgfmathdeclarefunction{d}{2}{%
		\pgfmathparse{ln((1 + #1)^#2) - ln((1 - #1)^#2)}%
	}
	\pgfmathdeclarefunction{lb}{2}{%
		\pgfmathparse{(#1 * exp(-#2)) / (#1 * (exp(-#2) - 1) + 1)}%
	}
	\pgfmathdeclarefunction{ub}{2}{%
		\pgfmathparse{(#1 * exp(#2)) / (#1 * (exp(#2) - 1) + 1)}%
	}

	\begin{axis}[
		axis lines = left,
		xlabel = \(p\),
		x label style = {
			at = {(axis description cs:1.01,0)},
			anchor = west,
		},
		ylabel = {Bound},
		y label style = {
			at = {(axis description cs:0,1.05)},
			anchor = south,
			rotate = -90,
		},
		width = 15em,
		height = 11em,
		legend style = {
			draw = none,
			anchor = north west,
			at = {(0.02, 1.00)},
		},
	]

	\addplot[
		domain = 0:1, 
		samples = 100, 
		color = mygreen,
		semithick,
		dashed,
	]
	(x, {lb(x, d(0.01, 100))});

	\addplot[
		domain = 0:1, 
		samples = 100, 
		color = mygreen,
		semithick,
		dashed,
	]
	(x, {ub(x, d(0.01, 100))});

	\addplot[
		domain = 0:1, 
		samples = 100, 
		color = myyellow,
		semithick,
	]
	(x, {lb(x, d(0.001, 100))});

	\addplot[
		domain = 0:1, 
		samples = 100, 
		color = myyellow,
		semithick,
	]
	(x, {ub(x, d(0.001, 100))});
	\end{axis}
\end{tikzpicture}

%% file: files/bound_plot_m=1000.tex
\begin{tikzpicture}
	\pgfmathdeclarefunction{d}{2}{%
		\pgfmathparse{ln((1 + #1)^#2) - ln((1 - #1)^#2)}%
	}
	\pgfmathdeclarefunction{lb}{2}{%
		\pgfmathparse{(#1 * exp(-#2)) / (#1 * (exp(-#2) - 1) + 1)}%
	}
	\pgfmathdeclarefunction{ub}{2}{%
		\pgfmathparse{(#1 * exp(#2)) / (#1 * (exp(#2) - 1) + 1)}%
	}

	\begin{axis}[
		axis lines = left,
		xlabel = \(p\),
		x label style = {
			at = {(axis description cs:1.01,0)},
			anchor = west,
		},
		ylabel = {Bound},
		y label style = {
			at = {(axis description cs:0,1.05)},
			anchor = south,
			rotate = -90,
		},
		width = 15em,
		height = 11em,
		legend style = {
			draw = none,
			anchor = north west,
			at = {(0.02, 1.00)},
		},
	]

	\addplot[
		domain = 0:1, 
		samples = 100, 
		color = mygreen,
		semithick,
		dashed,
	]
	(x, {lb(x, d(0.01, 1000))});

	\addplot[
		domain = 0:1, 
		samples = 100, 
		color = mygreen,
		semithick,
		dashed,
	]
	(x, {ub(x, d(0.01, 1000))});

	\addplot[
		domain = 0:1, 
		samples = 100, 
		color = myyellow,
		semithick,
	]
	(x, {lb(x, d(0.001, 1000))});

	\addplot[
		domain = 0:1, 
		samples = 100, 
		color = myyellow,
		semithick,
	]
	(x, {ub(x, d(0.001, 1000))});
	\end{axis}
\end{tikzpicture}

%% file: files/plot-times-avg.tex
\begin{tikzpicture}[x=1pt,y=1pt]
\definecolor{fillColor}{RGB}{255,255,255}
\path[use as bounding box,fill=fillColor,fill opacity=0.00] (0,0) rectangle (238.49,115.63);
\begin{scope}
\path[clip] (  0.00,  0.00) rectangle (238.49,115.63);
\definecolor{drawColor}{RGB}{255,255,255}
\definecolor{fillColor}{RGB}{255,255,255}

\path[draw=drawColor,line width= 0.6pt,line join=round,line cap=round,fill=fillColor] (  0.00,  0.00) rectangle (238.49,115.63);
\end{scope}
\begin{scope}
\path[clip] ( 44.03, 29.80) rectangle (232.99,110.13);
\definecolor{fillColor}{RGB}{255,255,255}

\path[fill=fillColor] ( 44.03, 29.80) rectangle (232.99,110.13);
\definecolor{drawColor}{RGB}{247,192,26}

\path[draw=drawColor,line width= 0.6pt,line join=round] ( 52.62, 33.45) --
	( 55.34, 36.78) --
	( 60.80, 40.03) --
	( 71.70, 43.09) --
	( 93.52, 44.12) --
	(137.15, 45.70) --
	(224.40, 46.52);
\definecolor{drawColor}{RGB}{78,155,133}

\path[draw=drawColor,line width= 0.6pt,dash pattern=on 2pt off 2pt ,line join=round] ( 52.62, 33.54) --
	( 55.34, 38.89) --
	( 60.80, 45.46) --
	( 71.70, 52.90) --
	( 93.52, 65.14) --
	(137.15, 85.93) --
	(224.40,106.48);
\definecolor{fillColor}{RGB}{78,155,133}

\path[fill=fillColor] ( 91.55, 63.18) --
	( 95.48, 63.18) --
	( 95.48, 67.10) --
	( 91.55, 67.10) --
	cycle;

\path[fill=fillColor] ( 53.38, 36.93) --
	( 57.30, 36.93) --
	( 57.30, 40.86) --
	( 53.38, 40.86) --
	cycle;

\path[fill=fillColor] ( 69.74, 50.94) --
	( 73.66, 50.94) --
	( 73.66, 54.86) --
	( 69.74, 54.86) --
	cycle;

\path[fill=fillColor] (135.18, 83.97) --
	(139.11, 83.97) --
	(139.11, 87.90) --
	(135.18, 87.90) --
	cycle;

\path[fill=fillColor] ( 50.65, 31.57) --
	( 54.58, 31.57) --
	( 54.58, 35.50) --
	( 50.65, 35.50) --
	cycle;

\path[fill=fillColor] ( 58.83, 43.49) --
	( 62.76, 43.49) --
	( 62.76, 47.42) --
	( 58.83, 47.42) --
	cycle;

\path[fill=fillColor] (222.44,104.52) --
	(226.36,104.52) --
	(226.36,108.44) --
	(222.44,108.44) --
	cycle;
\definecolor{drawColor}{RGB}{247,192,26}
\definecolor{fillColor}{RGB}{247,192,26}

\path[draw=drawColor,line width= 0.4pt,line join=round,line cap=round,fill=fillColor] ( 93.52, 44.12) circle (  1.96);

\path[draw=drawColor,line width= 0.4pt,line join=round,line cap=round,fill=fillColor] ( 55.34, 36.78) circle (  1.96);

\path[draw=drawColor,line width= 0.4pt,line join=round,line cap=round,fill=fillColor] ( 71.70, 43.09) circle (  1.96);

\path[draw=drawColor,line width= 0.4pt,line join=round,line cap=round,fill=fillColor] (137.15, 45.70) circle (  1.96);

\path[draw=drawColor,line width= 0.4pt,line join=round,line cap=round,fill=fillColor] ( 52.62, 33.45) circle (  1.96);

\path[draw=drawColor,line width= 0.4pt,line join=round,line cap=round,fill=fillColor] ( 60.80, 40.03) circle (  1.96);

\path[draw=drawColor,line width= 0.4pt,line join=round,line cap=round,fill=fillColor] (224.40, 46.52) circle (  1.96);
\end{scope}
\begin{scope}
\path[clip] (  0.00,  0.00) rectangle (238.49,115.63);
\definecolor{drawColor}{RGB}{0,0,0}

\path[draw=drawColor,line width= 0.6pt,line join=round] ( 44.03, 29.80) --
	( 44.03,110.13);

\path[draw=drawColor,line width= 0.6pt,line join=round] ( 45.45,107.67) --
	( 44.03,110.13) --
	( 42.60,107.67);
\end{scope}
\begin{scope}
\path[clip] (  0.00,  0.00) rectangle (238.49,115.63);
\definecolor{drawColor}{gray}{0.30}

\node[text=drawColor,anchor=base east,inner sep=0pt, outer sep=0pt, scale=  0.88] at ( 39.08, 50.78) {100};

\node[text=drawColor,anchor=base east,inner sep=0pt, outer sep=0pt, scale=  0.88] at ( 39.08, 77.60) {1000};

\node[text=drawColor,anchor=base east,inner sep=0pt, outer sep=0pt, scale=  0.88] at ( 39.08,104.42) {10000};
\end{scope}
\begin{scope}
\path[clip] (  0.00,  0.00) rectangle (238.49,115.63);
\definecolor{drawColor}{gray}{0.20}

\path[draw=drawColor,line width= 0.6pt,line join=round] ( 41.28, 53.81) --
	( 44.03, 53.81);

\path[draw=drawColor,line width= 0.6pt,line join=round] ( 41.28, 80.63) --
	( 44.03, 80.63);

\path[draw=drawColor,line width= 0.6pt,line join=round] ( 41.28,107.45) --
	( 44.03,107.45);
\end{scope}
\begin{scope}
\path[clip] (  0.00,  0.00) rectangle (238.49,115.63);
\definecolor{drawColor}{RGB}{0,0,0}

\path[draw=drawColor,line width= 0.6pt,line join=round] ( 44.03, 29.80) --
	(232.99, 29.80);

\path[draw=drawColor,line width= 0.6pt,line join=round] (230.53, 28.38) --
	(232.99, 29.80) --
	(230.53, 31.23);
\end{scope}
\begin{scope}
\path[clip] (  0.00,  0.00) rectangle (238.49,115.63);
\definecolor{drawColor}{gray}{0.20}

\path[draw=drawColor,line width= 0.6pt,line join=round] ( 49.89, 27.05) --
	( 49.89, 29.80);

\path[draw=drawColor,line width= 0.6pt,line join=round] (118.06, 27.05) --
	(118.06, 29.80);

\path[draw=drawColor,line width= 0.6pt,line join=round] (186.23, 27.05) --
	(186.23, 29.80);
\end{scope}
\begin{scope}
\path[clip] (  0.00,  0.00) rectangle (238.49,115.63);
\definecolor{drawColor}{gray}{0.30}

\node[text=drawColor,anchor=base,inner sep=0pt, outer sep=0pt, scale=  0.88] at ( 49.89, 18.79) {0};

\node[text=drawColor,anchor=base,inner sep=0pt, outer sep=0pt, scale=  0.88] at (118.06, 18.79) {50};

\node[text=drawColor,anchor=base,inner sep=0pt, outer sep=0pt, scale=  0.88] at (186.23, 18.79) {100};
\end{scope}
\begin{scope}
\path[clip] (  0.00,  0.00) rectangle (238.49,115.63);
\definecolor{drawColor}{RGB}{0,0,0}

\node[text=drawColor,anchor=base,inner sep=0pt, outer sep=0pt, scale=  1.00] at (138.51,  7.44) {domain size};
\end{scope}
\begin{scope}
\path[clip] (  0.00,  0.00) rectangle (238.49,115.63);
\definecolor{drawColor}{RGB}{0,0,0}

\node[text=drawColor,rotate= 90.00,anchor=base,inner sep=0pt, outer sep=0pt, scale=  1.00] at ( 12.39, 69.97) {time (ms)};
\end{scope}
\begin{scope}
\path[clip] (  0.00,  0.00) rectangle (238.49,115.63);

\path[] ( 41.42, 78.13) rectangle ( 95.76,118.04);
\end{scope}
\begin{scope}
\path[clip] (  0.00,  0.00) rectangle (238.49,115.63);
\definecolor{drawColor}{RGB}{247,192,26}

\path[draw=drawColor,line width= 0.6pt,line join=round] ( 48.36,105.31) -- ( 59.93,105.31);
\end{scope}
\begin{scope}
\path[clip] (  0.00,  0.00) rectangle (238.49,115.63);
\definecolor{drawColor}{RGB}{247,192,26}
\definecolor{fillColor}{RGB}{247,192,26}

\path[draw=drawColor,line width= 0.4pt,line join=round,line cap=round,fill=fillColor] ( 54.15,105.31) circle (  1.96);
\end{scope}
\begin{scope}
\path[clip] (  0.00,  0.00) rectangle (238.49,115.63);
\definecolor{drawColor}{RGB}{78,155,133}

\path[draw=drawColor,line width= 0.6pt,dash pattern=on 2pt off 2pt ,line join=round] ( 48.36, 90.86) -- ( 59.93, 90.86);
\end{scope}
\begin{scope}
\path[clip] (  0.00,  0.00) rectangle (238.49,115.63);
\definecolor{fillColor}{RGB}{78,155,133}

\path[fill=fillColor] ( 52.18, 88.89) --
	( 56.11, 88.89) --
	( 56.11, 92.82) --
	( 52.18, 92.82) --
	cycle;
\end{scope}
\begin{scope}
\path[clip] (  0.00,  0.00) rectangle (238.49,115.63);
\definecolor{drawColor}{RGB}{0,0,0}

\node[text=drawColor,anchor=base west,inner sep=0pt, outer sep=0pt, scale=  0.80] at ( 66.87,102.55) {$\varepsilon$-ACP};
\end{scope}
\begin{scope}
\path[clip] (  0.00,  0.00) rectangle (238.49,115.63);
\definecolor{drawColor}{RGB}{0,0,0}

\node[text=drawColor,anchor=base west,inner sep=0pt, outer sep=0pt, scale=  0.80] at ( 66.87, 88.10) {ACP};
\end{scope}
\end{tikzpicture}

%% file: files/table_mimic_results.tex
\begin{tabular}{rcc}
	\toprule
	Algorithm  & Avg. Run Time        & Avg. $p' \mathbin{/} p$ \\ \midrule
	\acs{acp}  & $183$ ms ($\pm\ 21$) & $1.0$ ($\pm\ 0.0$)      \\
	\acs{eacp} & $105$ ms ($\pm\ 9$)  & $1.0001$ ($\pm\ 0.01$)  \\ \bottomrule
\end{tabular}

%% file: files/example_worst_case.tex
\begin{tikzpicture}
	\node (t1) {
		\setlength{\tabcolsep}{5pt}
		\begin{tabular}{c|ccccc|c}
			\toprule
			$R_i$ & 
            $\phi_1(R_1)$ & $\phi_2(R_2)$ & $\cdots$  & $\phi_{m-1}(R_{m-1})$ & $\phi_m(R_m)$ & $\phi^*$\\ \midrule
			$r_1$ & $1 (1+\varepsilon)$                & $1$                 & $\cdots$ & $\cdots$ & $1$ & $1\left(1+\frac{1}{m}\varepsilon\right)$\\
			$r_2$    & $2$ & $2 (1+\varepsilon)$ &  $2$  &$\cdots$& $2$      &$2\left(1+\frac{1}{m}\varepsilon\right)$          \\
			$\vdots$   &    $\vdots$  & $\cdots$     &  $\ddots$      &             & $\vdots$ & $\vdots$\\
            $\vdots$   &    $\vdots$  &           &  &  $\ddots$           & $\vdots$ &$\vdots$\\
			  $r_m$ & $m$ & $\cdots$  & $\cdots$ &$m$  & $m (1+\varepsilon)$ &$m\left(1+\frac{1}{m}\varepsilon\right)$\\
              $r_{m+1}$ & $(m+1)$                & $(m+1)   (1+\varepsilon) $              & $\cdots$ & $\cdots$ & $(m+1) (1+\varepsilon)$ & $(m+1)\left(1+\frac{m-1}{m}\varepsilon\right) $\\
			$r_{m+2}$    & $(m+2)(1+\varepsilon)$ & $m+2 $ &  $(m+2)(1+\varepsilon)$  &$\cdots$& $(m+2) (1+\varepsilon)$   & $(m+2)\left(1+\frac{m-1}{m}\varepsilon\right) $            \\
            $\vdots$   &    $\vdots$  & $\cdots$     &  $\ddots$      &             & $\vdots$ & $\vdots$\\
            $\vdots$   &    $\vdots$  &           &  &  $\ddots$           & $\vdots$ & $\vdots$\\
			  $r_{2m}$ & $2m(1+\varepsilon)$ & $\cdots$  &$\cdots$ &$2m(1+\varepsilon)$  & $2m$ & $2m\left(1+\frac{m-1}{m}\varepsilon\right) $\\\bottomrule  
		\end{tabular}
	};
\end{tikzpicture}

%% file: files/acp_example.tex
\begin{tikzpicture}[label distance=1mm]
	\node[ellipse, draw, minimum width = 4.2em] (A) {$SalA$};
	\node[ellipse, draw, minimum width = 4.2em] (B) [below = 0.5cm of A] {$Rev$};
	\node[ellipse, draw, minimum width = 4.2em] (C) [below = 0.5cm of B] {$SalB$};
	\factor{below right}{A}{0.25cm and 0.5cm}{270}{$\phi_1$}{f1}
	\factor{below right}{B}{0.25cm and 0.5cm}{270}{$\phi_2$}{f2}

	\nodecolorshift{myyellow}{A}{Acol}{-2.1mm}{1mm}
	\nodecolorshift{myyellow}{B}{Bcol}{-2.1mm}{1mm}
	\nodecolorshift{myyellow}{C}{Ccol}{-2.1mm}{1mm}

	\factorcolor{myblue}{f1}{f1col}
	\factorcolor{myblue}{f2}{f2col}

	\draw (A) -- (f1);
	\draw (B) -- (f1);
	\draw (B) -- (f2);
	\draw (C) -- (f2);

	\node[ellipse, draw, minimum width = 4.2em, right = 1.4cm of A] (A1) {$SalA$};
	\node[ellipse, draw, minimum width = 4.2em, below = 0.5cm of A1] (B1) {$Rev$};
	\node[ellipse, draw, minimum width = 4.2em, below = 0.5cm of B1] (C1) {$SalB$};
	\factor{below right}{A1}{0.25cm and 0.5cm}{270}{$\phi_1$}{f1_1}
	\factor{below right}{B1}{0.25cm and 0.5cm}{270}{$\phi_2$}{f2_1}

	\nodecolorshift{myyellow}{A1}{A1col}{-2.1mm}{1mm}
	\nodecolorshift{myyellow}{B1}{B1col}{-2.1mm}{1mm}
	\nodecolorshift{myyellow}{C1}{C1col}{-2.1mm}{1mm}

	\factorcolor{myyellow}{f1_1}{f1_1col1}
	\factorcolorshift{myyellow}{f1_1}{f1_1col2}{2.1mm}
	\factorcolorshift{myblue}{f1_1}{f1_1col3}{4.2mm}
	\factorcolor{myyellow}{f2_1}{f2_1col1}
	\factorcolorshift{myyellow}{f2_1}{f2_1col2}{2.1mm}
	\factorcolorshift{myblue}{f2_1}{f2_1col3}{4.2mm}

	\coordinate[right=0.1cm of A1, yshift=-0.1cm] (CA1);
	\coordinate[above=0.2cm of f1_1, yshift=-0.1cm] (Cf1_1);
	\coordinate[right=0.1cm of B1, yshift=0.12cm] (CB1);
	\coordinate[right=0.1cm of B1, yshift=-0.1cm] (CB1_1);
	\coordinate[below=0.2cm of f1_1, yshift=0.15cm] (Cf1_1b);
	\coordinate[above=0.2cm of f2_1, yshift=-0.1cm] (Cf2_1);
	\coordinate[right=0.1cm of C1, yshift=0.12cm] (CC1);
	\coordinate[below=0.2cm of f2_1, yshift=0.15cm] (Cf2_1b);

	\begin{pgfonlayer}{bg}
		\draw (A1) -- (f1_1);
		\draw [arc, gray] (CA1) -- (Cf1_1);
		\draw (B1) -- (f1_1);
		\draw [arc, gray] (CB1) -- (Cf1_1b);
		\draw (B1) -- (f2_1);
		\draw [arc, gray] (CB1_1) -- (Cf2_1);
		\draw (C1) -- (f2_1);
		\draw [arc, gray] (CC1) -- (Cf2_1b);
	\end{pgfonlayer}

	\node[ellipse, draw, minimum width = 4.2em, right = 1.4cm of A1] (A2) {$SalA$};
	\node[ellipse, draw, minimum width = 4.2em, below = 0.5cm of A2] (B2) {$Rev$};
	\node[ellipse, draw, minimum width = 4.2em, below = 0.5cm of B2] (C2) {$SalB$};
	\factor{below right}{A2}{0.25cm and 0.5cm}{270}{$\phi_1$}{f1_2}
	\factor{below right}{B2}{0.25cm and 0.5cm}{270}{$\phi_2$}{f2_2}

	\nodecolorshift{myyellow}{A2}{A2col}{-2.1mm}{1mm}
	\nodecolorshift{myyellow}{B2}{B2col}{-2.1mm}{1mm}
	\nodecolorshift{myyellow}{C2}{C2col}{-2.1mm}{1mm}

	\factorcolor{myblue}{f1_2}{f1_2col1}
	\factorcolor{myblue}{f2_2}{f2_2col1}

	\draw (A2) -- (f1_2);
	\draw (B2) -- (f1_2);
	\draw (B2) -- (f2_2);
	\draw (C2) -- (f2_2);

	\node[ellipse, draw, minimum width = 4.2em, right = 1.4cm of A2] (A3) {$SalA$};
	\node[ellipse, draw, minimum width = 4.2em, below = 0.5cm of A3] (B3) {$Rev$};
	\node[ellipse, draw, minimum width = 4.2em, below = 0.5cm of B3] (C3) {$SalB$};
	\factor{below right}{A3}{0.25cm and 0.5cm}{270}{$\phi_1$}{f1_3}
	\factor{below right}{B3}{0.25cm and 0.5cm}{270}{$\phi_2$}{f2_3}

	\nodecolorshift{myblue}{A3}{A3col1}{-4.2mm}{1mm}
	\nodecolorshift{myyellow}{A3}{A3col2}{-2.1mm}{1mm}
	\nodecolorshift{myblue}{B3}{B3col1}{-6.3mm}{1mm}
	\nodecolorshift{myblue}{B3}{B3col2}{-4.2mm}{1mm}
	\nodecolorshift{myyellow}{B3}{B3col3}{-2.1mm}{1mm}
	\nodecolorshift{myblue}{C3}{C3col1}{-4.2mm}{1mm}
	\nodecolorshift{myyellow}{C3}{C3col2}{-2.1mm}{1mm}

	\factorcolor{myblue}{f1_3}{f1_3col1}
	\factorcolor{myblue}{f2_3}{f2_3col1}

	\coordinate[right=0.1cm of A3, yshift=-0.1cm] (CA3);
	\coordinate[above=0.2cm of f1_3, yshift=-0.1cm] (Cf1_3);
	\coordinate[right=0.1cm of B3, yshift=0.12cm] (CB3);
	\coordinate[right=0.1cm of B3, yshift=-0.1cm] (CB1_3);
	\coordinate[below=0.2cm of f1_3, yshift=0.15cm] (Cf1_3b);
	\coordinate[above=0.2cm of f2_3, yshift=-0.1cm] (Cf2_3);
	\coordinate[right=0.1cm of C3, yshift=0.12cm] (CC3);
	\coordinate[below=0.2cm of f2_3, yshift=0.15cm] (Cf2_3b);

	\begin{pgfonlayer}{bg}
		\draw (A3) -- (f1_3);
		\draw [arc, gray] (Cf1_3) -- (CA3);
		\draw (B3) -- (f1_3);
		\draw [arc, gray] (Cf1_3b) -- (CB3);
		\draw (B3) -- (f2_3);
		\draw [arc, gray] (Cf2_3) -- (CB1_3);
		\draw (C3) -- (f2_3);
		\draw [arc, gray] (Cf2_3b) -- (CC3);
	\end{pgfonlayer}

	\node[ellipse, draw, minimum width = 4.2em, right = 1.4cm of A3] (A4) {$SalA$};
	\node[ellipse, draw, minimum width = 4.2em, below = 0.5cm of A4] (B4) {$Rev$};
	\node[ellipse, draw, minimum width = 4.2em, below = 0.5cm of B4] (C4) {$SalB$};
	\factor{below right}{A4}{0.25cm and 0.5cm}{270}{$\phi_1$}{f1_4}
	\factor{below right}{B4}{0.25cm and 0.5cm}{270}{$\phi_2$}{f2_4}

	\nodecolorshift{myyellow}{A4}{A4col}{-2.1mm}{1mm}
	\nodecolorshift{mygreen}{B4}{B4col}{-2.1mm}{1mm}
	\nodecolorshift{myyellow}{C4}{C4col}{-2.1mm}{1mm}

	\factorcolor{myblue}{f1_4}{f1_4col1}
	\factorcolor{myblue}{f2_4}{f2_4col1}

	\draw (A4) -- (f1_4);
	\draw (B4) -- (f1_4);
	\draw (B4) -- (f2_4);
	\draw (C4) -- (f2_4);

	\pfs{right}{B4}{3.1cm}{270}{$\phi'_1$}{f12a}{f12}{f12b}

	\node[ellipse, draw, minimum width = 4.2em, inner sep = 1.5pt, above left = 0.25cm and 0.5cm of f12] (AC) {$Sal(E)$};
	\node[ellipse, draw, minimum width = 4.2em] (B) [below = 0.5cm of AC] {$Rev$};

	\begin{pgfonlayer}{bg}
		\draw (AC) -- (f12);
		\draw (B) -- (f12);
	\end{pgfonlayer}

	\node[below = 0.5cm of C2, xshift=1.5cm] (tab_f2) {
		\begin{tabular}{c|c|c}
			$SalB$ & $Rev$ & $\phi_2(SalB,Rev)$ \\ \hline
			\high  & \high & $\varphi_1$ \\
			\high  & \low  & $\varphi_2$ \\
			\low   & \high & $\varphi_3$ \\
			\low   & \low  & $\varphi_4$ \\
		\end{tabular}
	};

	\node[left = 0.4cm of tab_f2] (tab_f1) {
		\begin{tabular}{c|c|c}
			$SalA$ & $Rev$ & $\phi_1(SalA,Rev)$ \\ \hline
			\high  & \high & $\varphi_1$ \\
			\high  & \low  & $\varphi_2$ \\
			\low   & \high & $\varphi_3$ \\
			\low   & \low  & $\varphi_4$ \\
		\end{tabular}
	};

	\node[right = 0.4cm of tab_f2] (tab_f12) {
		\begin{tabular}{c|c|c}
			$Sal(E)$ & $Rev$ & $\phi'_1(Sal(E),Rev)$ \\ \hline
			\high    & \high & $\varphi_1$ \\
			\high    & \low  & $\varphi_2$ \\
			\low     & \high & $\varphi_3$ \\
			\low     & \low  & $\varphi_4$ \\
		\end{tabular}
	};
\end{tikzpicture}

%% file: files/example_fg_permuted.tex
\begin{tikzpicture}
	\node[ellipse, draw, minimum width = 4.2em] (A) {$SalA$};
	\node[ellipse, draw, minimum width = 4.2em] (B) [below = 0.5cm of A] {$Rev$};
	\node[ellipse, draw, minimum width = 4.2em] (C) [below = 0.5cm of B] {$SalB$};
	\factor{below right}{A}{0.25cm and 0.5cm}{270}{$\phi_1$}{f1}
	\factor{below right}{B}{0.25cm and 0.5cm}{270}{$\phi_2$}{f2}

	\node[right = 0.5cm of f1, yshift=5.0mm] (tab_f1) {
		\begin{tabular}{c|c|c}
			$SalA$ & $Rev$ & $\phi_1(SalA,Rev)$ \\ \hline
			\high  & \high & $\varphi_1$ \\
			\high  & \low  & $\varphi_2$ \\
			\low   & \high & $\varphi_3$ \\
			\low   & \low  & $\varphi_4$ \\
		\end{tabular}
	};

	\node[right = 0.5cm of f2, yshift=-5.0mm] (tab_f2) {
		\begin{tabular}{c|c|c}
			$Rev$ & $SalB$ & $\phi_2(Rev,SalB)$ \\ \hline
			\high  & \high & $\varphi'_1$ \\
			\high  & \low  & $\varphi'_3$ \\
			\low   & \high & $\varphi'_2$ \\
			\low   & \low  & $\varphi'_4$ \\
		\end{tabular}
	};

	\draw (A) -- (f1);
	\draw (B) -- (f1);
	\draw (B) -- (f2);
	\draw (C) -- (f2);
\end{tikzpicture}

%% file: files/example_fg_crv.tex
\begin{tikzpicture}
	\node[ellipse, draw, minimum width = 4.8em] (A) {$ComA$};
	\node[ellipse, draw, minimum width = 4.8em] (B) [below = 0.5cm of A] {$Rev$};
	\node[ellipse, draw, minimum width = 4.8em] (C) [below = 0.5cm of B] {$ComB$};
	\factor{right}{B}{0.25cm and 0.5cm}{270}{$\phi_1$}{f1}

	\node[right = 0.2em of f1] (tab_f1) {
		\begin{tabular}{c|c|c|c}
			$Rev$ & $ComA$ & $ComB$ & $\phi_1$    \\ \hline
			\high & \high  & \high  & $\varphi_1$ \\
			\high & \high  & \low   & $\varphi_2$ \\
			\high & \low   & \high  & $\varphi_2$ \\
			\high & \low   & \low   & $\varphi_3$ \\
			\low  & \high  & \high  & $\varphi_4$ \\
			\low  & \high  & \low   & $\varphi_5$ \\
			\low  & \low   & \high  & $\varphi_5$ \\
			\low  & \low   & \low   & $\varphi_6$ \\
		\end{tabular}
	};

	\draw (A) -- (f1);
	\draw (B) -- (f1);
	\draw (B) -- (f1);
	\draw (C) -- (f1);
\end{tikzpicture}

%% file: files/example_pfg_crv.tex
\begin{tikzpicture}
	\node[ellipse, draw, minimum width = 4.2em, inner sep = 1.5pt] (AC) {$Com(E)$};
	\node[ellipse, draw, minimum width = 4.8em] (B) [below = 0.5cm of AC] {$Rev$};
	\factor{above right}{B}{0.25cm and 0.5cm}{270}{$\phi'_1$}{f1}

	\node[right = 0.2em of f1] (tab_f1_crv) {
		\begin{tabular}{c|c|c}
			$Rev$ & $\#_E[Com(E)]$ & $\phi'_1$   \\ \hline
			\high & $[2,0]$        & $\varphi_1$ \\
			\high & $[1,1]$        & $\varphi_2$ \\
			\high & $[0,2]$        & $\varphi_3$ \\
			\low  & $[2,0]$        & $\varphi_4$ \\
			\low  & $[1,1]$        & $\varphi_5$ \\
			\low  & $[0,2]$        & $\varphi_6$ \\
		\end{tabular}
	};

	\draw (A) -- (f1);
	\draw (B) -- (f1);
	\draw (B) -- (f1);
\end{tikzpicture}

%% file: files/plot-quots-p=0.1-eps=0.001.tex
\begin{tikzpicture}[x=1pt,y=1pt]
\definecolor{fillColor}{RGB}{255,255,255}
\path[use as bounding box,fill=fillColor,fill opacity=0.00] (0,0) rectangle (238.49,115.63);
\begin{scope}
\path[clip] (  0.00,  0.00) rectangle (238.49,115.63);
\definecolor{drawColor}{RGB}{255,255,255}
\definecolor{fillColor}{RGB}{255,255,255}

\path[draw=drawColor,line width= 0.6pt,line join=round,line cap=round,fill=fillColor] (  0.00,  0.00) rectangle (238.49,115.63);
\end{scope}
\begin{scope}
\path[clip] ( 46.47, 29.80) rectangle (232.99,110.13);
\definecolor{fillColor}{RGB}{255,255,255}

\path[fill=fillColor] ( 46.47, 29.80) rectangle (232.99,110.13);
\definecolor{drawColor}{RGB}{78,155,133}
\definecolor{fillColor}{RGB}{78,155,133}

\path[draw=drawColor,draw opacity=0.20,line width= 0.4pt,line join=round,line cap=round,fill=fillColor,fill opacity=0.20] ( 62.01, 69.97) circle (  1.96);

\path[draw=drawColor,draw opacity=0.20,line width= 0.4pt,line join=round,line cap=round,fill=fillColor,fill opacity=0.20] ( 62.01, 69.97) circle (  1.96);

\path[draw=drawColor,draw opacity=0.20,line width= 0.4pt,line join=round,line cap=round,fill=fillColor,fill opacity=0.20] ( 62.01, 40.00) circle (  1.96);

\path[draw=drawColor,draw opacity=0.20,line width= 0.4pt,line join=round,line cap=round,fill=fillColor,fill opacity=0.20] ( 62.01, 96.82) circle (  1.96);

\path[draw=drawColor,draw opacity=0.20,line width= 0.4pt,line join=round,line cap=round,fill=fillColor,fill opacity=0.20] ( 62.01, 68.69) circle (  1.96);

\path[draw=drawColor,draw opacity=0.20,line width= 0.4pt,line join=round,line cap=round,fill=fillColor,fill opacity=0.20] ( 62.01, 71.29) circle (  1.96);
\definecolor{drawColor}{RGB}{78,155,133}

\path[draw=drawColor,line width= 0.6pt,line join=round] ( 62.01, 69.97) -- ( 62.01, 69.97);

\path[draw=drawColor,line width= 0.6pt,line join=round] ( 62.01, 69.97) -- ( 62.01, 69.97);

\path[draw=drawColor,line width= 0.6pt,fill=fillColor,fill opacity=0.20] ( 52.30, 69.97) --
	( 52.30, 69.97) --
	( 71.73, 69.97) --
	( 71.73, 69.97) --
	( 52.30, 69.97) --
	cycle;

\path[draw=drawColor,line width= 1.1pt] ( 52.30, 69.97) -- ( 71.73, 69.97);
\definecolor{drawColor}{RGB}{78,155,133}

\path[draw=drawColor,draw opacity=0.20,line width= 0.4pt,line join=round,line cap=round,fill=fillColor,fill opacity=0.20] ( 87.92,  0.90) circle (  1.96);

\path[draw=drawColor,draw opacity=0.20,line width= 0.4pt,line join=round,line cap=round,fill=fillColor,fill opacity=0.20] ( 87.92,119.49) circle (  1.96);

\path[draw=drawColor,draw opacity=0.20,line width= 0.4pt,line join=round,line cap=round,fill=fillColor,fill opacity=0.20] ( 87.92, 87.55) circle (  1.96);

\path[draw=drawColor,draw opacity=0.20,line width= 0.4pt,line join=round,line cap=round,fill=fillColor,fill opacity=0.20] ( 87.92, 49.27) circle (  1.96);

\path[draw=drawColor,draw opacity=0.20,line width= 0.4pt,line join=round,line cap=round,fill=fillColor,fill opacity=0.20] ( 87.92, 77.27) circle (  1.96);

\path[draw=drawColor,draw opacity=0.20,line width= 0.4pt,line join=round,line cap=round,fill=fillColor,fill opacity=0.20] ( 87.92, 60.61) circle (  1.96);
\definecolor{drawColor}{RGB}{78,155,133}

\path[draw=drawColor,line width= 0.6pt,line join=round] ( 87.92, 70.95) -- ( 87.92, 73.07);

\path[draw=drawColor,line width= 0.6pt,line join=round] ( 87.92, 68.51) -- ( 87.92, 65.55);

\path[draw=drawColor,line width= 0.6pt,fill=fillColor,fill opacity=0.20] ( 78.20, 70.95) --
	( 78.20, 68.51) --
	( 97.63, 68.51) --
	( 97.63, 70.95) --
	( 78.20, 70.95) --
	cycle;

\path[draw=drawColor,line width= 1.1pt] ( 78.20, 69.97) -- ( 97.63, 69.97);
\definecolor{drawColor}{RGB}{78,155,133}

\path[draw=drawColor,draw opacity=0.20,line width= 0.4pt,line join=round,line cap=round,fill=fillColor,fill opacity=0.20] (113.82, 67.55) circle (  1.96);

\path[draw=drawColor,draw opacity=0.20,line width= 0.4pt,line join=round,line cap=round,fill=fillColor,fill opacity=0.20] (113.82,-29.81) circle (  1.96);

\path[draw=drawColor,draw opacity=0.20,line width= 0.4pt,line join=round,line cap=round,fill=fillColor,fill opacity=0.20] (113.82,133.40) circle (  1.96);

\path[draw=drawColor,draw opacity=0.20,line width= 0.4pt,line join=round,line cap=round,fill=fillColor,fill opacity=0.20] (113.82, 66.76) circle (  1.96);

\path[draw=drawColor,draw opacity=0.20,line width= 0.4pt,line join=round,line cap=round,fill=fillColor,fill opacity=0.20] (113.82, 73.25) circle (  1.96);

\path[draw=drawColor,draw opacity=0.20,line width= 0.4pt,line join=round,line cap=round,fill=fillColor,fill opacity=0.20] (113.82, 94.75) circle (  1.96);

\path[draw=drawColor,draw opacity=0.20,line width= 0.4pt,line join=round,line cap=round,fill=fillColor,fill opacity=0.20] (113.82, 40.20) circle (  1.96);

\path[draw=drawColor,draw opacity=0.20,line width= 0.4pt,line join=round,line cap=round,fill=fillColor,fill opacity=0.20] (113.82, 66.78) circle (  1.96);

\path[draw=drawColor,draw opacity=0.20,line width= 0.4pt,line join=round,line cap=round,fill=fillColor,fill opacity=0.20] (113.82, 74.36) circle (  1.96);

\path[draw=drawColor,draw opacity=0.20,line width= 0.4pt,line join=round,line cap=round,fill=fillColor,fill opacity=0.20] (113.82, 72.84) circle (  1.96);

\path[draw=drawColor,draw opacity=0.20,line width= 0.4pt,line join=round,line cap=round,fill=fillColor,fill opacity=0.20] (113.82, 66.21) circle (  1.96);
\definecolor{drawColor}{RGB}{78,155,133}

\path[draw=drawColor,line width= 0.6pt,line join=round] (113.82, 70.57) -- (113.82, 71.47);

\path[draw=drawColor,line width= 0.6pt,line join=round] (113.82, 69.44) -- (113.82, 69.43);

\path[draw=drawColor,line width= 0.6pt,fill=fillColor,fill opacity=0.20] (104.11, 70.57) --
	(104.11, 69.44) --
	(123.54, 69.44) --
	(123.54, 70.57) --
	(104.11, 70.57) --
	cycle;

\path[draw=drawColor,line width= 1.1pt] (104.11, 69.97) -- (123.54, 69.97);
\definecolor{drawColor}{RGB}{78,155,133}

\path[draw=drawColor,draw opacity=0.20,line width= 0.4pt,line join=round,line cap=round,fill=fillColor,fill opacity=0.20] (139.73,-38.68) circle (  1.96);

\path[draw=drawColor,draw opacity=0.20,line width= 0.4pt,line join=round,line cap=round,fill=fillColor,fill opacity=0.20] (139.73,138.07) circle (  1.96);

\path[draw=drawColor,draw opacity=0.20,line width= 0.4pt,line join=round,line cap=round,fill=fillColor,fill opacity=0.20] (139.73, 87.05) circle (  1.96);

\path[draw=drawColor,draw opacity=0.20,line width= 0.4pt,line join=round,line cap=round,fill=fillColor,fill opacity=0.20] (139.73, 49.45) circle (  1.96);
\definecolor{drawColor}{RGB}{78,155,133}

\path[draw=drawColor,line width= 0.6pt,line join=round] (139.73, 72.43) -- (139.73, 76.29);

\path[draw=drawColor,line width= 0.6pt,line join=round] (139.73, 68.01) -- (139.73, 62.06);

\path[draw=drawColor,line width= 0.6pt,fill=fillColor,fill opacity=0.20] (130.02, 72.43) --
	(130.02, 68.01) --
	(149.45, 68.01) --
	(149.45, 72.43) --
	(130.02, 72.43) --
	cycle;

\path[draw=drawColor,line width= 1.1pt] (130.02, 69.97) -- (149.45, 69.97);
\definecolor{drawColor}{RGB}{78,155,133}

\path[draw=drawColor,draw opacity=0.20,line width= 0.4pt,line join=round,line cap=round,fill=fillColor,fill opacity=0.20] (165.64, 71.50) circle (  1.96);

\path[draw=drawColor,draw opacity=0.20,line width= 0.4pt,line join=round,line cap=round,fill=fillColor,fill opacity=0.20] (165.64,-35.96) circle (  1.96);

\path[draw=drawColor,draw opacity=0.20,line width= 0.4pt,line join=round,line cap=round,fill=fillColor,fill opacity=0.20] (165.64,136.37) circle (  1.96);

\path[draw=drawColor,draw opacity=0.20,line width= 0.4pt,line join=round,line cap=round,fill=fillColor,fill opacity=0.20] (165.64, 67.00) circle (  1.96);

\path[draw=drawColor,draw opacity=0.20,line width= 0.4pt,line join=round,line cap=round,fill=fillColor,fill opacity=0.20] (165.64, 73.01) circle (  1.96);

\path[draw=drawColor,draw opacity=0.20,line width= 0.4pt,line join=round,line cap=round,fill=fillColor,fill opacity=0.20] (165.64, 65.71) circle (  1.96);

\path[draw=drawColor,draw opacity=0.20,line width= 0.4pt,line join=round,line cap=round,fill=fillColor,fill opacity=0.20] (165.64, 75.08) circle (  1.96);

\path[draw=drawColor,draw opacity=0.20,line width= 0.4pt,line join=round,line cap=round,fill=fillColor,fill opacity=0.20] (165.64, 67.89) circle (  1.96);

\path[draw=drawColor,draw opacity=0.20,line width= 0.4pt,line join=round,line cap=round,fill=fillColor,fill opacity=0.20] (165.64, 72.83) circle (  1.96);

\path[draw=drawColor,draw opacity=0.20,line width= 0.4pt,line join=round,line cap=round,fill=fillColor,fill opacity=0.20] (165.64, 75.00) circle (  1.96);

\path[draw=drawColor,draw opacity=0.20,line width= 0.4pt,line join=round,line cap=round,fill=fillColor,fill opacity=0.20] (165.64, 65.77) circle (  1.96);
\definecolor{drawColor}{RGB}{78,155,133}

\path[draw=drawColor,line width= 0.6pt,line join=round] (165.64, 70.13) -- (165.64, 70.13);

\path[draw=drawColor,line width= 0.6pt,line join=round] (165.64, 69.25) -- (165.64, 69.02);

\path[draw=drawColor,line width= 0.6pt,fill=fillColor,fill opacity=0.20] (155.92, 70.13) --
	(155.92, 69.25) --
	(175.35, 69.25) --
	(175.35, 70.13) --
	(155.92, 70.13) --
	cycle;

\path[draw=drawColor,line width= 1.1pt] (155.92, 69.97) -- (175.35, 69.97);
\definecolor{drawColor}{RGB}{78,155,133}

\path[draw=drawColor,draw opacity=0.20,line width= 0.4pt,line join=round,line cap=round,fill=fillColor,fill opacity=0.20] (191.54, 74.69) circle (  1.96);

\path[draw=drawColor,draw opacity=0.20,line width= 0.4pt,line join=round,line cap=round,fill=fillColor,fill opacity=0.20] (191.54, 67.06) circle (  1.96);

\path[draw=drawColor,draw opacity=0.20,line width= 0.4pt,line join=round,line cap=round,fill=fillColor,fill opacity=0.20] (191.54,-33.30) circle (  1.96);

\path[draw=drawColor,draw opacity=0.20,line width= 0.4pt,line join=round,line cap=round,fill=fillColor,fill opacity=0.20] (191.54,134.70) circle (  1.96);

\path[draw=drawColor,draw opacity=0.20,line width= 0.4pt,line join=round,line cap=round,fill=fillColor,fill opacity=0.20] (191.54, 66.27) circle (  1.96);

\path[draw=drawColor,draw opacity=0.20,line width= 0.4pt,line join=round,line cap=round,fill=fillColor,fill opacity=0.20] (191.54, 73.75) circle (  1.96);

\path[draw=drawColor,draw opacity=0.20,line width= 0.4pt,line join=round,line cap=round,fill=fillColor,fill opacity=0.20] (191.54, 71.49) circle (  1.96);

\path[draw=drawColor,draw opacity=0.20,line width= 0.4pt,line join=round,line cap=round,fill=fillColor,fill opacity=0.20] (191.54, 68.14) circle (  1.96);

\path[draw=drawColor,draw opacity=0.20,line width= 0.4pt,line join=round,line cap=round,fill=fillColor,fill opacity=0.20] (191.54, 70.81) circle (  1.96);

\path[draw=drawColor,draw opacity=0.20,line width= 0.4pt,line join=round,line cap=round,fill=fillColor,fill opacity=0.20] (191.54, 68.81) circle (  1.96);
\definecolor{drawColor}{RGB}{78,155,133}

\path[draw=drawColor,line width= 0.6pt,line join=round] (191.54, 70.05) -- (191.54, 70.36);

\path[draw=drawColor,line width= 0.6pt,line join=round] (191.54, 69.75) -- (191.54, 69.64);

\path[draw=drawColor,line width= 0.6pt,fill=fillColor,fill opacity=0.20] (181.83, 70.05) --
	(181.83, 69.75) --
	(201.26, 69.75) --
	(201.26, 70.05) --
	(181.83, 70.05) --
	cycle;

\path[draw=drawColor,line width= 1.1pt] (181.83, 69.97) -- (201.26, 69.97);
\definecolor{drawColor}{RGB}{78,155,133}

\path[draw=drawColor,draw opacity=0.20,line width= 0.4pt,line join=round,line cap=round,fill=fillColor,fill opacity=0.20] (217.45, 64.49) circle (  1.96);

\path[draw=drawColor,draw opacity=0.20,line width= 0.4pt,line join=round,line cap=round,fill=fillColor,fill opacity=0.20] (217.45, 73.34) circle (  1.96);

\path[draw=drawColor,draw opacity=0.20,line width= 0.4pt,line join=round,line cap=round,fill=fillColor,fill opacity=0.20] (217.45,-39.85) circle (  1.96);

\path[draw=drawColor,draw opacity=0.20,line width= 0.4pt,line join=round,line cap=round,fill=fillColor,fill opacity=0.20] (217.45,138.80) circle (  1.96);

\path[draw=drawColor,draw opacity=0.20,line width= 0.4pt,line join=round,line cap=round,fill=fillColor,fill opacity=0.20] (217.45, 66.25) circle (  1.96);

\path[draw=drawColor,draw opacity=0.20,line width= 0.4pt,line join=round,line cap=round,fill=fillColor,fill opacity=0.20] (217.45, 73.77) circle (  1.96);

\path[draw=drawColor,draw opacity=0.20,line width= 0.4pt,line join=round,line cap=round,fill=fillColor,fill opacity=0.20] (217.45, 77.08) circle (  1.96);

\path[draw=drawColor,draw opacity=0.20,line width= 0.4pt,line join=round,line cap=round,fill=fillColor,fill opacity=0.20] (217.45, 60.18) circle (  1.96);
\definecolor{drawColor}{RGB}{78,155,133}

\path[draw=drawColor,line width= 0.6pt,line join=round] (217.45, 70.68) -- (217.45, 70.73);

\path[draw=drawColor,line width= 0.6pt,line join=round] (217.45, 69.24) -- (217.45, 69.06);

\path[draw=drawColor,line width= 0.6pt,fill=fillColor,fill opacity=0.20] (207.73, 70.68) --
	(207.73, 69.24) --
	(227.16, 69.24) --
	(227.16, 70.68) --
	(207.73, 70.68) --
	cycle;

\path[draw=drawColor,line width= 1.1pt] (207.73, 69.98) -- (227.16, 69.98);
\end{scope}
\begin{scope}
\path[clip] (  0.00,  0.00) rectangle (238.49,115.63);
\definecolor{drawColor}{RGB}{0,0,0}

\path[draw=drawColor,line width= 0.6pt,line join=round] ( 46.47, 29.80) --
	( 46.47,110.13);

\path[draw=drawColor,line width= 0.6pt,line join=round] ( 47.89,107.67) --
	( 46.47,110.13) --
	( 45.05,107.67);
\end{scope}
\begin{scope}
\path[clip] (  0.00,  0.00) rectangle (238.49,115.63);
\definecolor{drawColor}{gray}{0.30}

\node[text=drawColor,anchor=base east,inner sep=0pt, outer sep=0pt, scale=  0.88] at ( 41.52, 42.59) {0.9998};

\node[text=drawColor,anchor=base east,inner sep=0pt, outer sep=0pt, scale=  0.88] at ( 41.52, 66.94) {1.0000};

\node[text=drawColor,anchor=base east,inner sep=0pt, outer sep=0pt, scale=  0.88] at ( 41.52, 91.28) {1.0002};
\end{scope}
\begin{scope}
\path[clip] (  0.00,  0.00) rectangle (238.49,115.63);
\definecolor{drawColor}{gray}{0.20}

\path[draw=drawColor,line width= 0.6pt,line join=round] ( 43.72, 45.63) --
	( 46.47, 45.63);

\path[draw=drawColor,line width= 0.6pt,line join=round] ( 43.72, 69.97) --
	( 46.47, 69.97);

\path[draw=drawColor,line width= 0.6pt,line join=round] ( 43.72, 94.31) --
	( 46.47, 94.31);
\end{scope}
\begin{scope}
\path[clip] (  0.00,  0.00) rectangle (238.49,115.63);
\definecolor{drawColor}{RGB}{0,0,0}

\path[draw=drawColor,line width= 0.6pt,line join=round] ( 46.47, 29.80) --
	(232.99, 29.80);

\path[draw=drawColor,line width= 0.6pt,line join=round] (230.53, 28.38) --
	(232.99, 29.80) --
	(230.53, 31.23);
\end{scope}
\begin{scope}
\path[clip] (  0.00,  0.00) rectangle (238.49,115.63);
\definecolor{drawColor}{gray}{0.20}

\path[draw=drawColor,line width= 0.6pt,line join=round] ( 62.01, 27.05) --
	( 62.01, 29.80);

\path[draw=drawColor,line width= 0.6pt,line join=round] ( 87.92, 27.05) --
	( 87.92, 29.80);

\path[draw=drawColor,line width= 0.6pt,line join=round] (113.82, 27.05) --
	(113.82, 29.80);

\path[draw=drawColor,line width= 0.6pt,line join=round] (139.73, 27.05) --
	(139.73, 29.80);

\path[draw=drawColor,line width= 0.6pt,line join=round] (165.64, 27.05) --
	(165.64, 29.80);

\path[draw=drawColor,line width= 0.6pt,line join=round] (191.54, 27.05) --
	(191.54, 29.80);

\path[draw=drawColor,line width= 0.6pt,line join=round] (217.45, 27.05) --
	(217.45, 29.80);
\end{scope}
\begin{scope}
\path[clip] (  0.00,  0.00) rectangle (238.49,115.63);
\definecolor{drawColor}{gray}{0.30}

\node[text=drawColor,anchor=base,inner sep=0pt, outer sep=0pt, scale=  0.88] at ( 62.01, 18.79) {2};

\node[text=drawColor,anchor=base,inner sep=0pt, outer sep=0pt, scale=  0.88] at ( 87.92, 18.79) {4};

\node[text=drawColor,anchor=base,inner sep=0pt, outer sep=0pt, scale=  0.88] at (113.82, 18.79) {8};

\node[text=drawColor,anchor=base,inner sep=0pt, outer sep=0pt, scale=  0.88] at (139.73, 18.79) {16};

\node[text=drawColor,anchor=base,inner sep=0pt, outer sep=0pt, scale=  0.88] at (165.64, 18.79) {32};

\node[text=drawColor,anchor=base,inner sep=0pt, outer sep=0pt, scale=  0.88] at (191.54, 18.79) {64};

\node[text=drawColor,anchor=base,inner sep=0pt, outer sep=0pt, scale=  0.88] at (217.45, 18.79) {128};
\end{scope}
\begin{scope}
\path[clip] (  0.00,  0.00) rectangle (238.49,115.63);
\definecolor{drawColor}{RGB}{0,0,0}

\node[text=drawColor,anchor=base,inner sep=0pt, outer sep=0pt, scale=  1.00] at (139.73,  7.44) {domain size};
\end{scope}
\begin{scope}
\path[clip] (  0.00,  0.00) rectangle (238.49,115.63);
\definecolor{drawColor}{RGB}{0,0,0}

\node[text=drawColor,rotate= 90.00,anchor=base,inner sep=0pt, outer sep=0pt, scale=  1.00] at ( 12.39, 69.97) {$p' \mathbin{/} p$};
\end{scope}
\end{tikzpicture}

%% file: files/plot-quots-p=0.1-eps=0.1.tex
\begin{tikzpicture}[x=1pt,y=1pt]
\definecolor{fillColor}{RGB}{255,255,255}
\path[use as bounding box,fill=fillColor,fill opacity=0.00] (0,0) rectangle (238.49,115.63);
\begin{scope}
\path[clip] (  0.00,  0.00) rectangle (238.49,115.63);
\definecolor{drawColor}{RGB}{255,255,255}
\definecolor{fillColor}{RGB}{255,255,255}

\path[draw=drawColor,line width= 0.6pt,line join=round,line cap=round,fill=fillColor] (  0.00,  0.00) rectangle (238.49,115.63);
\end{scope}
\begin{scope}
\path[clip] ( 37.67, 29.80) rectangle (232.99,110.13);
\definecolor{fillColor}{RGB}{255,255,255}

\path[fill=fillColor] ( 37.67, 29.80) rectangle (232.99,110.13);
\definecolor{drawColor}{RGB}{78,155,133}
\definecolor{fillColor}{RGB}{78,155,133}

\path[draw=drawColor,draw opacity=0.20,line width= 0.4pt,line join=round,line cap=round,fill=fillColor,fill opacity=0.20] ( 53.95, 69.90) circle (  1.96);

\path[draw=drawColor,draw opacity=0.20,line width= 0.4pt,line join=round,line cap=round,fill=fillColor,fill opacity=0.20] ( 53.95, 70.19) circle (  1.96);

\path[draw=drawColor,draw opacity=0.20,line width= 0.4pt,line join=round,line cap=round,fill=fillColor,fill opacity=0.20] ( 53.95, 44.52) circle (  1.96);

\path[draw=drawColor,draw opacity=0.20,line width= 0.4pt,line join=round,line cap=round,fill=fillColor,fill opacity=0.20] ( 53.95, 94.59) circle (  1.96);

\path[draw=drawColor,draw opacity=0.20,line width= 0.4pt,line join=round,line cap=round,fill=fillColor,fill opacity=0.20] ( 53.95, 68.84) circle (  1.96);

\path[draw=drawColor,draw opacity=0.20,line width= 0.4pt,line join=round,line cap=round,fill=fillColor,fill opacity=0.20] ( 53.95, 71.14) circle (  1.96);
\definecolor{drawColor}{RGB}{78,155,133}

\path[draw=drawColor,line width= 0.6pt,line join=round] ( 53.95, 69.97) -- ( 53.95, 69.97);

\path[draw=drawColor,line width= 0.6pt,line join=round] ( 53.95, 69.97) -- ( 53.95, 69.97);

\path[draw=drawColor,line width= 0.6pt,fill=fillColor,fill opacity=0.20] ( 43.78, 69.97) --
	( 43.78, 69.97) --
	( 64.12, 69.97) --
	( 64.12, 69.97) --
	( 43.78, 69.97) --
	cycle;

\path[draw=drawColor,line width= 1.1pt] ( 43.78, 69.97) -- ( 64.12, 69.97);
\definecolor{drawColor}{RGB}{78,155,133}

\path[draw=drawColor,draw opacity=0.20,line width= 0.4pt,line join=round,line cap=round,fill=fillColor,fill opacity=0.20] ( 81.08, 12.30) circle (  1.96);

\path[draw=drawColor,draw opacity=0.20,line width= 0.4pt,line join=round,line cap=round,fill=fillColor,fill opacity=0.20] ( 81.08,116.17) circle (  1.96);

\path[draw=drawColor,draw opacity=0.20,line width= 0.4pt,line join=round,line cap=round,fill=fillColor,fill opacity=0.20] ( 81.08, 85.69) circle (  1.96);

\path[draw=drawColor,draw opacity=0.20,line width= 0.4pt,line join=round,line cap=round,fill=fillColor,fill opacity=0.20] ( 81.08, 51.47) circle (  1.96);

\path[draw=drawColor,draw opacity=0.20,line width= 0.4pt,line join=round,line cap=round,fill=fillColor,fill opacity=0.20] ( 81.08, 75.89) circle (  1.96);

\path[draw=drawColor,draw opacity=0.20,line width= 0.4pt,line join=round,line cap=round,fill=fillColor,fill opacity=0.20] ( 81.08, 62.39) circle (  1.96);
\definecolor{drawColor}{RGB}{78,155,133}

\path[draw=drawColor,line width= 0.6pt,line join=round] ( 81.08, 70.72) -- ( 81.08, 72.71);

\path[draw=drawColor,line width= 0.6pt,line join=round] ( 81.08, 68.82) -- ( 81.08, 67.65);

\path[draw=drawColor,line width= 0.6pt,fill=fillColor,fill opacity=0.20] ( 70.90, 70.72) --
	( 70.90, 68.82) --
	( 91.25, 68.82) --
	( 91.25, 70.72) --
	( 70.90, 70.72) --
	cycle;

\path[draw=drawColor,line width= 1.1pt] ( 70.90, 69.97) -- ( 91.25, 69.97);
\definecolor{drawColor}{RGB}{78,155,133}

\path[draw=drawColor,draw opacity=0.20,line width= 0.4pt,line join=round,line cap=round,fill=fillColor,fill opacity=0.20] (108.20,-11.69) circle (  1.96);

\path[draw=drawColor,draw opacity=0.20,line width= 0.4pt,line join=round,line cap=round,fill=fillColor,fill opacity=0.20] (108.20,129.14) circle (  1.96);

\path[draw=drawColor,draw opacity=0.20,line width= 0.4pt,line join=round,line cap=round,fill=fillColor,fill opacity=0.20] (108.20, 66.84) circle (  1.96);

\path[draw=drawColor,draw opacity=0.20,line width= 0.4pt,line join=round,line cap=round,fill=fillColor,fill opacity=0.20] (108.20, 92.85) circle (  1.96);

\path[draw=drawColor,draw opacity=0.20,line width= 0.4pt,line join=round,line cap=round,fill=fillColor,fill opacity=0.20] (108.20, 42.49) circle (  1.96);

\path[draw=drawColor,draw opacity=0.20,line width= 0.4pt,line join=round,line cap=round,fill=fillColor,fill opacity=0.20] (108.20, 67.02) circle (  1.96);

\path[draw=drawColor,draw opacity=0.20,line width= 0.4pt,line join=round,line cap=round,fill=fillColor,fill opacity=0.20] (108.20, 74.02) circle (  1.96);

\path[draw=drawColor,draw opacity=0.20,line width= 0.4pt,line join=round,line cap=round,fill=fillColor,fill opacity=0.20] (108.20, 66.96) circle (  1.96);
\definecolor{drawColor}{RGB}{78,155,133}

\path[draw=drawColor,line width= 0.6pt,line join=round] (108.20, 71.32) -- (108.20, 73.19);

\path[draw=drawColor,line width= 0.6pt,line join=round] (108.20, 69.63) -- (108.20, 68.88);

\path[draw=drawColor,line width= 0.6pt,fill=fillColor,fill opacity=0.20] ( 98.03, 71.32) --
	( 98.03, 69.63) --
	(118.38, 69.63) --
	(118.38, 71.32) --
	( 98.03, 71.32) --
	cycle;

\path[draw=drawColor,line width= 1.1pt] ( 98.03, 69.97) -- (118.38, 69.97);
\definecolor{drawColor}{RGB}{78,155,133}

\path[draw=drawColor,draw opacity=0.20,line width= 0.4pt,line join=round,line cap=round,fill=fillColor,fill opacity=0.20] (135.33,-18.94) circle (  1.96);

\path[draw=drawColor,draw opacity=0.20,line width= 0.4pt,line join=round,line cap=round,fill=fillColor,fill opacity=0.20] (135.33,133.65) circle (  1.96);

\path[draw=drawColor,draw opacity=0.20,line width= 0.4pt,line join=round,line cap=round,fill=fillColor,fill opacity=0.20] (135.33, 85.86) circle (  1.96);

\path[draw=drawColor,draw opacity=0.20,line width= 0.4pt,line join=round,line cap=round,fill=fillColor,fill opacity=0.20] (135.33, 50.87) circle (  1.96);
\definecolor{drawColor}{RGB}{78,155,133}

\path[draw=drawColor,line width= 0.6pt,line join=round] (135.33, 73.59) -- (135.33, 79.85);

\path[draw=drawColor,line width= 0.6pt,line join=round] (135.33, 67.82) -- (135.33, 59.95);

\path[draw=drawColor,line width= 0.6pt,fill=fillColor,fill opacity=0.20] (125.16, 73.59) --
	(125.16, 67.82) --
	(145.50, 67.82) --
	(145.50, 73.59) --
	(125.16, 73.59) --
	cycle;

\path[draw=drawColor,line width= 1.1pt] (125.16, 69.97) -- (145.50, 69.97);
\definecolor{drawColor}{RGB}{78,155,133}

\path[draw=drawColor,draw opacity=0.20,line width= 0.4pt,line join=round,line cap=round,fill=fillColor,fill opacity=0.20] (162.46,-17.00) circle (  1.96);

\path[draw=drawColor,draw opacity=0.20,line width= 0.4pt,line join=round,line cap=round,fill=fillColor,fill opacity=0.20] (162.46,132.25) circle (  1.96);

\path[draw=drawColor,draw opacity=0.20,line width= 0.4pt,line join=round,line cap=round,fill=fillColor,fill opacity=0.20] (162.46, 83.48) circle (  1.96);

\path[draw=drawColor,draw opacity=0.20,line width= 0.4pt,line join=round,line cap=round,fill=fillColor,fill opacity=0.20] (162.46, 92.39) circle (  1.96);

\path[draw=drawColor,draw opacity=0.20,line width= 0.4pt,line join=round,line cap=round,fill=fillColor,fill opacity=0.20] (162.46, 39.16) circle (  1.96);

\path[draw=drawColor,draw opacity=0.20,line width= 0.4pt,line join=round,line cap=round,fill=fillColor,fill opacity=0.20] (162.46, 38.64) circle (  1.96);
\definecolor{drawColor}{RGB}{78,155,133}

\path[draw=drawColor,line width= 0.6pt,line join=round] (162.46, 73.59) -- (162.46, 79.50);

\path[draw=drawColor,line width= 0.6pt,line join=round] (162.46, 67.33) -- (162.46, 66.04);

\path[draw=drawColor,line width= 0.6pt,fill=fillColor,fill opacity=0.20] (152.29, 73.59) --
	(152.29, 67.33) --
	(172.63, 67.33) --
	(172.63, 73.59) --
	(152.29, 73.59) --
	cycle;

\path[draw=drawColor,line width= 1.1pt] (152.29, 70.06) -- (172.63, 70.06);
\definecolor{drawColor}{RGB}{78,155,133}

\path[draw=drawColor,draw opacity=0.20,line width= 0.4pt,line join=round,line cap=round,fill=fillColor,fill opacity=0.20] (189.59, 73.81) circle (  1.96);

\path[draw=drawColor,draw opacity=0.20,line width= 0.4pt,line join=round,line cap=round,fill=fillColor,fill opacity=0.20] (189.59, 74.39) circle (  1.96);

\path[draw=drawColor,draw opacity=0.20,line width= 0.4pt,line join=round,line cap=round,fill=fillColor,fill opacity=0.20] (189.59, 67.24) circle (  1.96);

\path[draw=drawColor,draw opacity=0.20,line width= 0.4pt,line join=round,line cap=round,fill=fillColor,fill opacity=0.20] (189.59,-14.87) circle (  1.96);

\path[draw=drawColor,draw opacity=0.20,line width= 0.4pt,line join=round,line cap=round,fill=fillColor,fill opacity=0.20] (189.59,130.72) circle (  1.96);

\path[draw=drawColor,draw opacity=0.20,line width= 0.4pt,line join=round,line cap=round,fill=fillColor,fill opacity=0.20] (189.59, 66.74) circle (  1.96);

\path[draw=drawColor,draw opacity=0.20,line width= 0.4pt,line join=round,line cap=round,fill=fillColor,fill opacity=0.20] (189.59, 79.65) circle (  1.96);

\path[draw=drawColor,draw opacity=0.20,line width= 0.4pt,line join=round,line cap=round,fill=fillColor,fill opacity=0.20] (189.59, 46.89) circle (  1.96);
\definecolor{drawColor}{RGB}{78,155,133}

\path[draw=drawColor,line width= 0.6pt,line join=round] (189.59, 71.21) -- (189.59, 73.29);

\path[draw=drawColor,line width= 0.6pt,line join=round] (189.59, 69.67) -- (189.59, 68.23);

\path[draw=drawColor,line width= 0.6pt,fill=fillColor,fill opacity=0.20] (179.41, 71.21) --
	(179.41, 69.67) --
	(199.76, 69.67) --
	(199.76, 71.21) --
	(179.41, 71.21) --
	cycle;

\path[draw=drawColor,line width= 1.1pt] (179.41, 69.97) -- (199.76, 69.97);
\definecolor{drawColor}{RGB}{78,155,133}

\path[draw=drawColor,draw opacity=0.20,line width= 0.4pt,line join=round,line cap=round,fill=fillColor,fill opacity=0.20] (216.71, -9.87) circle (  1.96);

\path[draw=drawColor,draw opacity=0.20,line width= 0.4pt,line join=round,line cap=round,fill=fillColor,fill opacity=0.20] (216.71,127.15) circle (  1.96);

\path[draw=drawColor,draw opacity=0.20,line width= 0.4pt,line join=round,line cap=round,fill=fillColor,fill opacity=0.20] (216.71,120.12) circle (  1.96);

\path[draw=drawColor,draw opacity=0.20,line width= 0.4pt,line join=round,line cap=round,fill=fillColor,fill opacity=0.20] (216.71, 59.71) circle (  1.96);
\definecolor{drawColor}{RGB}{78,155,133}

\path[draw=drawColor,line width= 0.6pt,line join=round] (216.71, 72.73) -- (216.71, 77.41);

\path[draw=drawColor,line width= 0.6pt,line join=round] (216.71, 68.78) -- (216.71, 66.09);

\path[draw=drawColor,line width= 0.6pt,fill=fillColor,fill opacity=0.20] (206.54, 72.73) --
	(206.54, 68.78) --
	(226.89, 68.78) --
	(226.89, 72.73) --
	(206.54, 72.73) --
	cycle;

\path[draw=drawColor,line width= 1.1pt] (206.54, 70.26) -- (226.89, 70.26);
\end{scope}
\begin{scope}
\path[clip] (  0.00,  0.00) rectangle (238.49,115.63);
\definecolor{drawColor}{RGB}{0,0,0}

\path[draw=drawColor,line width= 0.6pt,line join=round] ( 37.67, 29.80) --
	( 37.67,110.13);

\path[draw=drawColor,line width= 0.6pt,line join=round] ( 39.09,107.67) --
	( 37.67,110.13) --
	( 36.25,107.67);
\end{scope}
\begin{scope}
\path[clip] (  0.00,  0.00) rectangle (238.49,115.63);
\definecolor{drawColor}{gray}{0.30}

\node[text=drawColor,anchor=base east,inner sep=0pt, outer sep=0pt, scale=  0.88] at ( 32.72, 42.59) {0.98};

\node[text=drawColor,anchor=base east,inner sep=0pt, outer sep=0pt, scale=  0.88] at ( 32.72, 66.94) {1.00};

\node[text=drawColor,anchor=base east,inner sep=0pt, outer sep=0pt, scale=  0.88] at ( 32.72, 91.28) {1.02};
\end{scope}
\begin{scope}
\path[clip] (  0.00,  0.00) rectangle (238.49,115.63);
\definecolor{drawColor}{gray}{0.20}

\path[draw=drawColor,line width= 0.6pt,line join=round] ( 34.92, 45.63) --
	( 37.67, 45.63);

\path[draw=drawColor,line width= 0.6pt,line join=round] ( 34.92, 69.97) --
	( 37.67, 69.97);

\path[draw=drawColor,line width= 0.6pt,line join=round] ( 34.92, 94.31) --
	( 37.67, 94.31);
\end{scope}
\begin{scope}
\path[clip] (  0.00,  0.00) rectangle (238.49,115.63);
\definecolor{drawColor}{RGB}{0,0,0}

\path[draw=drawColor,line width= 0.6pt,line join=round] ( 37.67, 29.80) --
	(232.99, 29.80);

\path[draw=drawColor,line width= 0.6pt,line join=round] (230.53, 28.38) --
	(232.99, 29.80) --
	(230.53, 31.23);
\end{scope}
\begin{scope}
\path[clip] (  0.00,  0.00) rectangle (238.49,115.63);
\definecolor{drawColor}{gray}{0.20}

\path[draw=drawColor,line width= 0.6pt,line join=round] ( 53.95, 27.05) --
	( 53.95, 29.80);

\path[draw=drawColor,line width= 0.6pt,line join=round] ( 81.08, 27.05) --
	( 81.08, 29.80);

\path[draw=drawColor,line width= 0.6pt,line join=round] (108.20, 27.05) --
	(108.20, 29.80);

\path[draw=drawColor,line width= 0.6pt,line join=round] (135.33, 27.05) --
	(135.33, 29.80);

\path[draw=drawColor,line width= 0.6pt,line join=round] (162.46, 27.05) --
	(162.46, 29.80);

\path[draw=drawColor,line width= 0.6pt,line join=round] (189.59, 27.05) --
	(189.59, 29.80);

\path[draw=drawColor,line width= 0.6pt,line join=round] (216.71, 27.05) --
	(216.71, 29.80);
\end{scope}
\begin{scope}
\path[clip] (  0.00,  0.00) rectangle (238.49,115.63);
\definecolor{drawColor}{gray}{0.30}

\node[text=drawColor,anchor=base,inner sep=0pt, outer sep=0pt, scale=  0.88] at ( 53.95, 18.79) {2};

\node[text=drawColor,anchor=base,inner sep=0pt, outer sep=0pt, scale=  0.88] at ( 81.08, 18.79) {4};

\node[text=drawColor,anchor=base,inner sep=0pt, outer sep=0pt, scale=  0.88] at (108.20, 18.79) {8};

\node[text=drawColor,anchor=base,inner sep=0pt, outer sep=0pt, scale=  0.88] at (135.33, 18.79) {16};

\node[text=drawColor,anchor=base,inner sep=0pt, outer sep=0pt, scale=  0.88] at (162.46, 18.79) {32};

\node[text=drawColor,anchor=base,inner sep=0pt, outer sep=0pt, scale=  0.88] at (189.59, 18.79) {64};

\node[text=drawColor,anchor=base,inner sep=0pt, outer sep=0pt, scale=  0.88] at (216.71, 18.79) {128};
\end{scope}
\begin{scope}
\path[clip] (  0.00,  0.00) rectangle (238.49,115.63);
\definecolor{drawColor}{RGB}{0,0,0}

\node[text=drawColor,anchor=base,inner sep=0pt, outer sep=0pt, scale=  1.00] at (135.33,  7.44) {domain size};
\end{scope}
\begin{scope}
\path[clip] (  0.00,  0.00) rectangle (238.49,115.63);
\definecolor{drawColor}{RGB}{0,0,0}

\node[text=drawColor,rotate= 90.00,anchor=base,inner sep=0pt, outer sep=0pt, scale=  1.00] at ( 12.39, 69.97) {$p' \mathbin{/} p$};
\end{scope}
\end{tikzpicture}

%% file: files/plot-quots-p=0.3-eps=0.001.tex
\begin{tikzpicture}[x=1pt,y=1pt]
\definecolor{fillColor}{RGB}{255,255,255}
\path[use as bounding box,fill=fillColor,fill opacity=0.00] (0,0) rectangle (238.49,115.63);
\begin{scope}
\path[clip] (  0.00,  0.00) rectangle (238.49,115.63);
\definecolor{drawColor}{RGB}{255,255,255}
\definecolor{fillColor}{RGB}{255,255,255}

\path[draw=drawColor,line width= 0.6pt,line join=round,line cap=round,fill=fillColor] (  0.00,  0.00) rectangle (238.49,115.63);
\end{scope}
\begin{scope}
\path[clip] ( 46.47, 29.80) rectangle (232.99,110.13);
\definecolor{fillColor}{RGB}{255,255,255}

\path[fill=fillColor] ( 46.47, 29.80) rectangle (232.99,110.13);
\definecolor{drawColor}{RGB}{78,155,133}
\definecolor{fillColor}{RGB}{78,155,133}

\path[draw=drawColor,draw opacity=0.20,line width= 0.4pt,line join=round,line cap=round,fill=fillColor,fill opacity=0.20] ( 62.01, 69.97) circle (  1.96);

\path[draw=drawColor,draw opacity=0.20,line width= 0.4pt,line join=round,line cap=round,fill=fillColor,fill opacity=0.20] ( 62.01, 69.97) circle (  1.96);

\path[draw=drawColor,draw opacity=0.20,line width= 0.4pt,line join=round,line cap=round,fill=fillColor,fill opacity=0.20] ( 62.01, 86.51) circle (  1.96);

\path[draw=drawColor,draw opacity=0.20,line width= 0.4pt,line join=round,line cap=round,fill=fillColor,fill opacity=0.20] ( 62.01, 55.44) circle (  1.96);

\path[draw=drawColor,draw opacity=0.20,line width= 0.4pt,line join=round,line cap=round,fill=fillColor,fill opacity=0.20] ( 62.01, 68.67) circle (  1.96);

\path[draw=drawColor,draw opacity=0.20,line width= 0.4pt,line join=round,line cap=round,fill=fillColor,fill opacity=0.20] ( 62.01, 71.47) circle (  1.96);

\path[draw=drawColor,draw opacity=0.20,line width= 0.4pt,line join=round,line cap=round,fill=fillColor,fill opacity=0.20] ( 62.01, 69.97) circle (  1.96);

\path[draw=drawColor,draw opacity=0.20,line width= 0.4pt,line join=round,line cap=round,fill=fillColor,fill opacity=0.20] ( 62.01, 69.97) circle (  1.96);

\path[draw=drawColor,draw opacity=0.20,line width= 0.4pt,line join=round,line cap=round,fill=fillColor,fill opacity=0.20] ( 62.01, 36.65) circle (  1.96);

\path[draw=drawColor,draw opacity=0.20,line width= 0.4pt,line join=round,line cap=round,fill=fillColor,fill opacity=0.20] ( 62.01, 99.82) circle (  1.96);

\path[draw=drawColor,draw opacity=0.20,line width= 0.4pt,line join=round,line cap=round,fill=fillColor,fill opacity=0.20] ( 62.01, 66.11) circle (  1.96);

\path[draw=drawColor,draw opacity=0.20,line width= 0.4pt,line join=round,line cap=round,fill=fillColor,fill opacity=0.20] ( 62.01, 73.98) circle (  1.96);
\definecolor{drawColor}{RGB}{78,155,133}

\path[draw=drawColor,line width= 0.6pt,line join=round] ( 62.01, 69.97) -- ( 62.01, 69.97);

\path[draw=drawColor,line width= 0.6pt,line join=round] ( 62.01, 69.97) -- ( 62.01, 69.97);

\path[draw=drawColor,line width= 0.6pt,fill=fillColor,fill opacity=0.20] ( 52.30, 69.97) --
	( 52.30, 69.97) --
	( 71.73, 69.97) --
	( 71.73, 69.97) --
	( 52.30, 69.97) --
	cycle;

\path[draw=drawColor,line width= 1.1pt] ( 52.30, 69.97) -- ( 71.73, 69.97);
\definecolor{drawColor}{RGB}{78,155,133}

\path[draw=drawColor,draw opacity=0.20,line width= 0.4pt,line join=round,line cap=round,fill=fillColor,fill opacity=0.20] ( 87.92, 84.95) circle (  1.96);

\path[draw=drawColor,draw opacity=0.20,line width= 0.4pt,line join=round,line cap=round,fill=fillColor,fill opacity=0.20] ( 87.92, 59.45) circle (  1.96);

\path[draw=drawColor,draw opacity=0.20,line width= 0.4pt,line join=round,line cap=round,fill=fillColor,fill opacity=0.20] ( 87.92,  5.13) circle (  1.96);

\path[draw=drawColor,draw opacity=0.20,line width= 0.4pt,line join=round,line cap=round,fill=fillColor,fill opacity=0.20] ( 87.92,116.46) circle (  1.96);

\path[draw=drawColor,draw opacity=0.20,line width= 0.4pt,line join=round,line cap=round,fill=fillColor,fill opacity=0.20] ( 87.92, 87.55) circle (  1.96);

\path[draw=drawColor,draw opacity=0.20,line width= 0.4pt,line join=round,line cap=round,fill=fillColor,fill opacity=0.20] ( 87.92, 49.27) circle (  1.96);

\path[draw=drawColor,draw opacity=0.20,line width= 0.4pt,line join=round,line cap=round,fill=fillColor,fill opacity=0.20] ( 87.92, 85.39) circle (  1.96);

\path[draw=drawColor,draw opacity=0.20,line width= 0.4pt,line join=round,line cap=round,fill=fillColor,fill opacity=0.20] ( 87.92, 50.21) circle (  1.96);
\definecolor{drawColor}{RGB}{78,155,133}

\path[draw=drawColor,line width= 0.6pt,line join=round] ( 87.92, 72.93) -- ( 87.92, 76.59);

\path[draw=drawColor,line width= 0.6pt,line join=round] ( 87.92, 67.90) -- ( 87.92, 61.61);

\path[draw=drawColor,line width= 0.6pt,fill=fillColor,fill opacity=0.20] ( 78.20, 72.93) --
	( 78.20, 67.90) --
	( 97.63, 67.90) --
	( 97.63, 72.93) --
	( 78.20, 72.93) --
	cycle;

\path[draw=drawColor,line width= 1.1pt] ( 78.20, 69.97) -- ( 97.63, 69.97);
\definecolor{drawColor}{RGB}{78,155,133}

\path[draw=drawColor,draw opacity=0.20,line width= 0.4pt,line join=round,line cap=round,fill=fillColor,fill opacity=0.20] (113.82,-28.31) circle (  1.96);

\path[draw=drawColor,draw opacity=0.20,line width= 0.4pt,line join=round,line cap=round,fill=fillColor,fill opacity=0.20] (113.82,132.46) circle (  1.96);

\path[draw=drawColor,draw opacity=0.20,line width= 0.4pt,line join=round,line cap=round,fill=fillColor,fill opacity=0.20] (113.82, 92.09) circle (  1.96);

\path[draw=drawColor,draw opacity=0.20,line width= 0.4pt,line join=round,line cap=round,fill=fillColor,fill opacity=0.20] (113.82, 43.39) circle (  1.96);
\definecolor{drawColor}{RGB}{78,155,133}

\path[draw=drawColor,line width= 0.6pt,line join=round] (113.82, 72.67) -- (113.82, 79.57);

\path[draw=drawColor,line width= 0.6pt,line join=round] (113.82, 67.75) -- (113.82, 62.09);

\path[draw=drawColor,line width= 0.6pt,fill=fillColor,fill opacity=0.20] (104.11, 72.67) --
	(104.11, 67.75) --
	(123.54, 67.75) --
	(123.54, 72.67) --
	(104.11, 72.67) --
	cycle;

\path[draw=drawColor,line width= 1.1pt] (104.11, 69.97) -- (123.54, 69.97);
\definecolor{drawColor}{RGB}{78,155,133}

\path[draw=drawColor,draw opacity=0.20,line width= 0.4pt,line join=round,line cap=round,fill=fillColor,fill opacity=0.20] (139.73, 83.10) circle (  1.96);

\path[draw=drawColor,draw opacity=0.20,line width= 0.4pt,line join=round,line cap=round,fill=fillColor,fill opacity=0.20] (139.73, 61.89) circle (  1.96);

\path[draw=drawColor,draw opacity=0.20,line width= 0.4pt,line join=round,line cap=round,fill=fillColor,fill opacity=0.20] (139.73, 69.08) circle (  1.96);

\path[draw=drawColor,draw opacity=0.20,line width= 0.4pt,line join=round,line cap=round,fill=fillColor,fill opacity=0.20] (139.73, 71.03) circle (  1.96);

\path[draw=drawColor,draw opacity=0.20,line width= 0.4pt,line join=round,line cap=round,fill=fillColor,fill opacity=0.20] (139.73,-27.01) circle (  1.96);

\path[draw=drawColor,draw opacity=0.20,line width= 0.4pt,line join=round,line cap=round,fill=fillColor,fill opacity=0.20] (139.73,130.77) circle (  1.96);

\path[draw=drawColor,draw opacity=0.20,line width= 0.4pt,line join=round,line cap=round,fill=fillColor,fill opacity=0.20] (139.73, 62.30) circle (  1.96);

\path[draw=drawColor,draw opacity=0.20,line width= 0.4pt,line join=round,line cap=round,fill=fillColor,fill opacity=0.20] (139.73, 77.81) circle (  1.96);

\path[draw=drawColor,draw opacity=0.20,line width= 0.4pt,line join=round,line cap=round,fill=fillColor,fill opacity=0.20] (139.73, 84.57) circle (  1.96);

\path[draw=drawColor,draw opacity=0.20,line width= 0.4pt,line join=round,line cap=round,fill=fillColor,fill opacity=0.20] (139.73, 52.42) circle (  1.96);

\path[draw=drawColor,draw opacity=0.20,line width= 0.4pt,line join=round,line cap=round,fill=fillColor,fill opacity=0.20] (139.73, 68.09) circle (  1.96);

\path[draw=drawColor,draw opacity=0.20,line width= 0.4pt,line join=round,line cap=round,fill=fillColor,fill opacity=0.20] (139.73, 72.56) circle (  1.96);
\definecolor{drawColor}{RGB}{78,155,133}

\path[draw=drawColor,line width= 0.6pt,line join=round] (139.73, 70.15) -- (139.73, 70.17);

\path[draw=drawColor,line width= 0.6pt,line join=round] (139.73, 69.92) -- (139.73, 69.92);

\path[draw=drawColor,line width= 0.6pt,fill=fillColor,fill opacity=0.20] (130.02, 70.15) --
	(130.02, 69.92) --
	(149.45, 69.92) --
	(149.45, 70.15) --
	(130.02, 70.15) --
	cycle;

\path[draw=drawColor,line width= 1.1pt] (130.02, 69.97) -- (149.45, 69.97);
\definecolor{drawColor}{RGB}{78,155,133}

\path[draw=drawColor,draw opacity=0.20,line width= 0.4pt,line join=round,line cap=round,fill=fillColor,fill opacity=0.20] (165.64, 62.14) circle (  1.96);

\path[draw=drawColor,draw opacity=0.20,line width= 0.4pt,line join=round,line cap=round,fill=fillColor,fill opacity=0.20] (165.64,  7.55) circle (  1.96);

\path[draw=drawColor,draw opacity=0.20,line width= 0.4pt,line join=round,line cap=round,fill=fillColor,fill opacity=0.20] (165.64,109.10) circle (  1.96);

\path[draw=drawColor,draw opacity=0.20,line width= 0.4pt,line join=round,line cap=round,fill=fillColor,fill opacity=0.20] (165.64, 63.60) circle (  1.96);

\path[draw=drawColor,draw opacity=0.20,line width= 0.4pt,line join=round,line cap=round,fill=fillColor,fill opacity=0.20] (165.64, 76.48) circle (  1.96);

\path[draw=drawColor,draw opacity=0.20,line width= 0.4pt,line join=round,line cap=round,fill=fillColor,fill opacity=0.20] (165.64, 56.19) circle (  1.96);

\path[draw=drawColor,draw opacity=0.20,line width= 0.4pt,line join=round,line cap=round,fill=fillColor,fill opacity=0.20] (165.64, 86.52) circle (  1.96);

\path[draw=drawColor,draw opacity=0.20,line width= 0.4pt,line join=round,line cap=round,fill=fillColor,fill opacity=0.20] (165.64, 55.10) circle (  1.96);
\definecolor{drawColor}{RGB}{78,155,133}

\path[draw=drawColor,line width= 0.6pt,line join=round] (165.64, 71.02) -- (165.64, 74.78);

\path[draw=drawColor,line width= 0.6pt,line join=round] (165.64, 68.35) -- (165.64, 67.53);

\path[draw=drawColor,line width= 0.6pt,fill=fillColor,fill opacity=0.20] (155.92, 71.02) --
	(155.92, 68.35) --
	(175.35, 68.35) --
	(175.35, 71.02) --
	(155.92, 71.02) --
	cycle;

\path[draw=drawColor,line width= 1.1pt] (155.92, 69.97) -- (175.35, 69.97);
\definecolor{drawColor}{RGB}{78,155,133}

\path[draw=drawColor,draw opacity=0.20,line width= 0.4pt,line join=round,line cap=round,fill=fillColor,fill opacity=0.20] (191.54, 62.11) circle (  1.96);

\path[draw=drawColor,draw opacity=0.20,line width= 0.4pt,line join=round,line cap=round,fill=fillColor,fill opacity=0.20] (191.54, 74.80) circle (  1.96);

\path[draw=drawColor,draw opacity=0.20,line width= 0.4pt,line join=round,line cap=round,fill=fillColor,fill opacity=0.20] (191.54,-12.28) circle (  1.96);

\path[draw=drawColor,draw opacity=0.20,line width= 0.4pt,line join=round,line cap=round,fill=fillColor,fill opacity=0.20] (191.54,121.53) circle (  1.96);

\path[draw=drawColor,draw opacity=0.20,line width= 0.4pt,line join=round,line cap=round,fill=fillColor,fill opacity=0.20] (191.54, 62.63) circle (  1.96);

\path[draw=drawColor,draw opacity=0.20,line width= 0.4pt,line join=round,line cap=round,fill=fillColor,fill opacity=0.20] (191.54, 77.47) circle (  1.96);

\path[draw=drawColor,draw opacity=0.20,line width= 0.4pt,line join=round,line cap=round,fill=fillColor,fill opacity=0.20] (191.54, 54.91) circle (  1.96);

\path[draw=drawColor,draw opacity=0.20,line width= 0.4pt,line join=round,line cap=round,fill=fillColor,fill opacity=0.20] (191.54, 88.05) circle (  1.96);

\path[draw=drawColor,draw opacity=0.20,line width= 0.4pt,line join=round,line cap=round,fill=fillColor,fill opacity=0.20] (191.54,228.03) circle (  1.96);

\path[draw=drawColor,draw opacity=0.20,line width= 0.4pt,line join=round,line cap=round,fill=fillColor,fill opacity=0.20] (191.54, 75.58) circle (  1.96);

\path[draw=drawColor,draw opacity=0.20,line width= 0.4pt,line join=round,line cap=round,fill=fillColor,fill opacity=0.20] (191.54, 65.60) circle (  1.96);
\definecolor{drawColor}{RGB}{78,155,133}

\path[draw=drawColor,line width= 0.6pt,line join=round] (191.54, 70.97) -- (191.54, 71.18);

\path[draw=drawColor,line width= 0.6pt,line join=round] (191.54, 69.36) -- (191.54, 68.51);

\path[draw=drawColor,line width= 0.6pt,fill=fillColor,fill opacity=0.20] (181.83, 70.97) --
	(181.83, 69.36) --
	(201.26, 69.36) --
	(201.26, 70.97) --
	(181.83, 70.97) --
	cycle;

\path[draw=drawColor,line width= 1.1pt] (181.83, 69.97) -- (201.26, 69.97);
\definecolor{drawColor}{RGB}{78,155,133}

\path[draw=drawColor,draw opacity=0.20,line width= 0.4pt,line join=round,line cap=round,fill=fillColor,fill opacity=0.20] (217.45,-36.40) circle (  1.96);

\path[draw=drawColor,draw opacity=0.20,line width= 0.4pt,line join=round,line cap=round,fill=fillColor,fill opacity=0.20] (217.45,136.65) circle (  1.96);

\path[draw=drawColor,draw opacity=0.20,line width= 0.4pt,line join=round,line cap=round,fill=fillColor,fill opacity=0.20] (217.45, 43.28) circle (  1.96);

\path[draw=drawColor,draw opacity=0.20,line width= 0.4pt,line join=round,line cap=round,fill=fillColor,fill opacity=0.20] (217.45,102.03) circle (  1.96);
\definecolor{drawColor}{RGB}{78,155,133}

\path[draw=drawColor,line width= 0.6pt,line join=round] (217.45, 74.20) -- (217.45, 79.28);

\path[draw=drawColor,line width= 0.6pt,line join=round] (217.45, 63.27) -- (217.45, 58.06);

\path[draw=drawColor,line width= 0.6pt,fill=fillColor,fill opacity=0.20] (207.73, 74.20) --
	(207.73, 63.27) --
	(227.16, 63.27) --
	(227.16, 74.20) --
	(207.73, 74.20) --
	cycle;

\path[draw=drawColor,line width= 1.1pt] (207.73, 69.98) -- (227.16, 69.98);
\end{scope}
\begin{scope}
\path[clip] (  0.00,  0.00) rectangle (238.49,115.63);
\definecolor{drawColor}{RGB}{0,0,0}

\path[draw=drawColor,line width= 0.6pt,line join=round] ( 46.47, 29.80) --
	( 46.47,110.13);

\path[draw=drawColor,line width= 0.6pt,line join=round] ( 47.89,107.67) --
	( 46.47,110.13) --
	( 45.05,107.67);
\end{scope}
\begin{scope}
\path[clip] (  0.00,  0.00) rectangle (238.49,115.63);
\definecolor{drawColor}{gray}{0.30}

\node[text=drawColor,anchor=base east,inner sep=0pt, outer sep=0pt, scale=  0.88] at ( 41.52, 42.59) {0.9998};

\node[text=drawColor,anchor=base east,inner sep=0pt, outer sep=0pt, scale=  0.88] at ( 41.52, 66.94) {1.0000};

\node[text=drawColor,anchor=base east,inner sep=0pt, outer sep=0pt, scale=  0.88] at ( 41.52, 91.28) {1.0002};
\end{scope}
\begin{scope}
\path[clip] (  0.00,  0.00) rectangle (238.49,115.63);
\definecolor{drawColor}{gray}{0.20}

\path[draw=drawColor,line width= 0.6pt,line join=round] ( 43.72, 45.63) --
	( 46.47, 45.63);

\path[draw=drawColor,line width= 0.6pt,line join=round] ( 43.72, 69.97) --
	( 46.47, 69.97);

\path[draw=drawColor,line width= 0.6pt,line join=round] ( 43.72, 94.31) --
	( 46.47, 94.31);
\end{scope}
\begin{scope}
\path[clip] (  0.00,  0.00) rectangle (238.49,115.63);
\definecolor{drawColor}{RGB}{0,0,0}

\path[draw=drawColor,line width= 0.6pt,line join=round] ( 46.47, 29.80) --
	(232.99, 29.80);

\path[draw=drawColor,line width= 0.6pt,line join=round] (230.53, 28.38) --
	(232.99, 29.80) --
	(230.53, 31.23);
\end{scope}
\begin{scope}
\path[clip] (  0.00,  0.00) rectangle (238.49,115.63);
\definecolor{drawColor}{gray}{0.20}

\path[draw=drawColor,line width= 0.6pt,line join=round] ( 62.01, 27.05) --
	( 62.01, 29.80);

\path[draw=drawColor,line width= 0.6pt,line join=round] ( 87.92, 27.05) --
	( 87.92, 29.80);

\path[draw=drawColor,line width= 0.6pt,line join=round] (113.82, 27.05) --
	(113.82, 29.80);

\path[draw=drawColor,line width= 0.6pt,line join=round] (139.73, 27.05) --
	(139.73, 29.80);

\path[draw=drawColor,line width= 0.6pt,line join=round] (165.64, 27.05) --
	(165.64, 29.80);

\path[draw=drawColor,line width= 0.6pt,line join=round] (191.54, 27.05) --
	(191.54, 29.80);

\path[draw=drawColor,line width= 0.6pt,line join=round] (217.45, 27.05) --
	(217.45, 29.80);
\end{scope}
\begin{scope}
\path[clip] (  0.00,  0.00) rectangle (238.49,115.63);
\definecolor{drawColor}{gray}{0.30}

\node[text=drawColor,anchor=base,inner sep=0pt, outer sep=0pt, scale=  0.88] at ( 62.01, 18.79) {2};

\node[text=drawColor,anchor=base,inner sep=0pt, outer sep=0pt, scale=  0.88] at ( 87.92, 18.79) {4};

\node[text=drawColor,anchor=base,inner sep=0pt, outer sep=0pt, scale=  0.88] at (113.82, 18.79) {8};

\node[text=drawColor,anchor=base,inner sep=0pt, outer sep=0pt, scale=  0.88] at (139.73, 18.79) {16};

\node[text=drawColor,anchor=base,inner sep=0pt, outer sep=0pt, scale=  0.88] at (165.64, 18.79) {32};

\node[text=drawColor,anchor=base,inner sep=0pt, outer sep=0pt, scale=  0.88] at (191.54, 18.79) {64};

\node[text=drawColor,anchor=base,inner sep=0pt, outer sep=0pt, scale=  0.88] at (217.45, 18.79) {128};
\end{scope}
\begin{scope}
\path[clip] (  0.00,  0.00) rectangle (238.49,115.63);
\definecolor{drawColor}{RGB}{0,0,0}

\node[text=drawColor,anchor=base,inner sep=0pt, outer sep=0pt, scale=  1.00] at (139.73,  7.44) {domain size};
\end{scope}
\begin{scope}
\path[clip] (  0.00,  0.00) rectangle (238.49,115.63);
\definecolor{drawColor}{RGB}{0,0,0}

\node[text=drawColor,rotate= 90.00,anchor=base,inner sep=0pt, outer sep=0pt, scale=  1.00] at ( 12.39, 69.97) {$p' \mathbin{/} p$};
\end{scope}
\end{tikzpicture}

%% file: files/plot-quots-p=0.3-eps=0.1.tex
\begin{tikzpicture}[x=1pt,y=1pt]
\definecolor{fillColor}{RGB}{255,255,255}
\path[use as bounding box,fill=fillColor,fill opacity=0.00] (0,0) rectangle (238.49,115.63);
\begin{scope}
\path[clip] (  0.00,  0.00) rectangle (238.49,115.63);
\definecolor{drawColor}{RGB}{255,255,255}
\definecolor{fillColor}{RGB}{255,255,255}

\path[draw=drawColor,line width= 0.6pt,line join=round,line cap=round,fill=fillColor] (  0.00,  0.00) rectangle (238.49,115.63);
\end{scope}
\begin{scope}
\path[clip] ( 37.67, 29.80) rectangle (232.99,110.13);
\definecolor{fillColor}{RGB}{255,255,255}

\path[fill=fillColor] ( 37.67, 29.80) rectangle (232.99,110.13);
\definecolor{drawColor}{RGB}{78,155,133}
\definecolor{fillColor}{RGB}{78,155,133}

\path[draw=drawColor,draw opacity=0.20,line width= 0.4pt,line join=round,line cap=round,fill=fillColor,fill opacity=0.20] ( 53.95, 69.90) circle (  1.96);

\path[draw=drawColor,draw opacity=0.20,line width= 0.4pt,line join=round,line cap=round,fill=fillColor,fill opacity=0.20] ( 53.95, 70.18) circle (  1.96);

\path[draw=drawColor,draw opacity=0.20,line width= 0.4pt,line join=round,line cap=round,fill=fillColor,fill opacity=0.20] ( 53.95, 84.62) circle (  1.96);

\path[draw=drawColor,draw opacity=0.20,line width= 0.4pt,line join=round,line cap=round,fill=fillColor,fill opacity=0.20] ( 53.95, 57.26) circle (  1.96);

\path[draw=drawColor,draw opacity=0.20,line width= 0.4pt,line join=round,line cap=round,fill=fillColor,fill opacity=0.20] ( 53.95, 68.83) circle (  1.96);

\path[draw=drawColor,draw opacity=0.20,line width= 0.4pt,line join=round,line cap=round,fill=fillColor,fill opacity=0.20] ( 53.95, 71.29) circle (  1.96);

\path[draw=drawColor,draw opacity=0.20,line width= 0.4pt,line join=round,line cap=round,fill=fillColor,fill opacity=0.20] ( 53.95, 69.84) circle (  1.96);

\path[draw=drawColor,draw opacity=0.20,line width= 0.4pt,line join=round,line cap=round,fill=fillColor,fill opacity=0.20] ( 53.95, 70.38) circle (  1.96);

\path[draw=drawColor,draw opacity=0.20,line width= 0.4pt,line join=round,line cap=round,fill=fillColor,fill opacity=0.20] ( 53.95, 42.52) circle (  1.96);

\path[draw=drawColor,draw opacity=0.20,line width= 0.4pt,line join=round,line cap=round,fill=fillColor,fill opacity=0.20] ( 53.95, 96.72) circle (  1.96);

\path[draw=drawColor,draw opacity=0.20,line width= 0.4pt,line join=round,line cap=round,fill=fillColor,fill opacity=0.20] ( 53.95, 68.02) circle (  1.96);

\path[draw=drawColor,draw opacity=0.20,line width= 0.4pt,line join=round,line cap=round,fill=fillColor,fill opacity=0.20] ( 53.95, 72.00) circle (  1.96);
\definecolor{drawColor}{RGB}{78,155,133}

\path[draw=drawColor,line width= 0.6pt,line join=round] ( 53.95, 69.97) -- ( 53.95, 69.97);

\path[draw=drawColor,line width= 0.6pt,line join=round] ( 53.95, 69.97) -- ( 53.95, 69.97);

\path[draw=drawColor,line width= 0.6pt,fill=fillColor,fill opacity=0.20] ( 43.78, 69.97) --
	( 43.78, 69.97) --
	( 64.12, 69.97) --
	( 64.12, 69.97) --
	( 43.78, 69.97) --
	cycle;

\path[draw=drawColor,line width= 1.1pt] ( 43.78, 69.97) -- ( 64.12, 69.97);
\definecolor{drawColor}{RGB}{78,155,133}

\path[draw=drawColor,draw opacity=0.20,line width= 0.4pt,line join=round,line cap=round,fill=fillColor,fill opacity=0.20] ( 81.08, 83.37) circle (  1.96);

\path[draw=drawColor,draw opacity=0.20,line width= 0.4pt,line join=round,line cap=round,fill=fillColor,fill opacity=0.20] ( 81.08, 60.61) circle (  1.96);

\path[draw=drawColor,draw opacity=0.20,line width= 0.4pt,line join=round,line cap=round,fill=fillColor,fill opacity=0.20] ( 81.08, 16.84) circle (  1.96);

\path[draw=drawColor,draw opacity=0.20,line width= 0.4pt,line join=round,line cap=round,fill=fillColor,fill opacity=0.20] ( 81.08,112.79) circle (  1.96);

\path[draw=drawColor,draw opacity=0.20,line width= 0.4pt,line join=round,line cap=round,fill=fillColor,fill opacity=0.20] ( 81.08, 85.69) circle (  1.96);

\path[draw=drawColor,draw opacity=0.20,line width= 0.4pt,line join=round,line cap=round,fill=fillColor,fill opacity=0.20] ( 81.08, 51.47) circle (  1.96);

\path[draw=drawColor,draw opacity=0.20,line width= 0.4pt,line join=round,line cap=round,fill=fillColor,fill opacity=0.20] ( 81.08, 82.29) circle (  1.96);

\path[draw=drawColor,draw opacity=0.20,line width= 0.4pt,line join=round,line cap=round,fill=fillColor,fill opacity=0.20] ( 81.08, 54.27) circle (  1.96);
\definecolor{drawColor}{RGB}{78,155,133}

\path[draw=drawColor,line width= 0.6pt,line join=round] ( 81.08, 72.58) -- ( 81.08, 74.52);

\path[draw=drawColor,line width= 0.6pt,line join=round] ( 81.08, 68.13) -- ( 81.08, 62.70);

\path[draw=drawColor,line width= 0.6pt,fill=fillColor,fill opacity=0.20] ( 70.90, 72.58) --
	( 70.90, 68.13) --
	( 91.25, 68.13) --
	( 91.25, 72.58) --
	( 70.90, 72.58) --
	cycle;

\path[draw=drawColor,line width= 1.1pt] ( 70.90, 69.95) -- ( 91.25, 69.95);
\definecolor{drawColor}{RGB}{78,155,133}

\path[draw=drawColor,draw opacity=0.20,line width= 0.4pt,line join=round,line cap=round,fill=fillColor,fill opacity=0.20] (108.20, -9.30) circle (  1.96);

\path[draw=drawColor,draw opacity=0.20,line width= 0.4pt,line join=round,line cap=round,fill=fillColor,fill opacity=0.20] (108.20,127.83) circle (  1.96);

\path[draw=drawColor,draw opacity=0.20,line width= 0.4pt,line join=round,line cap=round,fill=fillColor,fill opacity=0.20] (108.20, 90.61) circle (  1.96);

\path[draw=drawColor,draw opacity=0.20,line width= 0.4pt,line join=round,line cap=round,fill=fillColor,fill opacity=0.20] (108.20, 45.18) circle (  1.96);
\definecolor{drawColor}{RGB}{78,155,133}

\path[draw=drawColor,line width= 0.6pt,line join=round] (108.20, 73.33) -- (108.20, 78.53);

\path[draw=drawColor,line width= 0.6pt,line join=round] (108.20, 68.84) -- (108.20, 64.63);

\path[draw=drawColor,line width= 0.6pt,fill=fillColor,fill opacity=0.20] ( 98.03, 73.33) --
	( 98.03, 68.84) --
	(118.38, 68.84) --
	(118.38, 73.33) --
	( 98.03, 73.33) --
	cycle;

\path[draw=drawColor,line width= 1.1pt] ( 98.03, 69.99) -- (118.38, 69.99);
\definecolor{drawColor}{RGB}{78,155,133}

\path[draw=drawColor,draw opacity=0.20,line width= 0.4pt,line join=round,line cap=round,fill=fillColor,fill opacity=0.20] (135.33, -8.18) circle (  1.96);

\path[draw=drawColor,draw opacity=0.20,line width= 0.4pt,line join=round,line cap=round,fill=fillColor,fill opacity=0.20] (135.33,126.36) circle (  1.96);

\path[draw=drawColor,draw opacity=0.20,line width= 0.4pt,line join=round,line cap=round,fill=fillColor,fill opacity=0.20] (135.33, 81.82) circle (  1.96);

\path[draw=drawColor,draw opacity=0.20,line width= 0.4pt,line join=round,line cap=round,fill=fillColor,fill opacity=0.20] (135.33, 83.99) circle (  1.96);

\path[draw=drawColor,draw opacity=0.20,line width= 0.4pt,line join=round,line cap=round,fill=fillColor,fill opacity=0.20] (135.33, 53.13) circle (  1.96);

\path[draw=drawColor,draw opacity=0.20,line width= 0.4pt,line join=round,line cap=round,fill=fillColor,fill opacity=0.20] (135.33, 83.36) circle (  1.96);
\definecolor{drawColor}{RGB}{78,155,133}

\path[draw=drawColor,line width= 0.6pt,line join=round] (135.33, 73.61) -- (135.33, 80.82);

\path[draw=drawColor,line width= 0.6pt,line join=round] (135.33, 68.26) -- (135.33, 61.52);

\path[draw=drawColor,line width= 0.6pt,fill=fillColor,fill opacity=0.20] (125.16, 73.61) --
	(125.16, 68.26) --
	(145.50, 68.26) --
	(145.50, 73.61) --
	(125.16, 73.61) --
	cycle;

\path[draw=drawColor,line width= 1.1pt] (125.16, 70.26) -- (145.50, 70.26);
\definecolor{drawColor}{RGB}{78,155,133}

\path[draw=drawColor,draw opacity=0.20,line width= 0.4pt,line join=round,line cap=round,fill=fillColor,fill opacity=0.20] (162.46, 63.64) circle (  1.96);

\path[draw=drawColor,draw opacity=0.20,line width= 0.4pt,line join=round,line cap=round,fill=fillColor,fill opacity=0.20] (162.46, 19.86) circle (  1.96);

\path[draw=drawColor,draw opacity=0.20,line width= 0.4pt,line join=round,line cap=round,fill=fillColor,fill opacity=0.20] (162.46,106.12) circle (  1.96);

\path[draw=drawColor,draw opacity=0.20,line width= 0.4pt,line join=round,line cap=round,fill=fillColor,fill opacity=0.20] (162.46, 79.23) circle (  1.96);

\path[draw=drawColor,draw opacity=0.20,line width= 0.4pt,line join=round,line cap=round,fill=fillColor,fill opacity=0.20] (162.46, 65.58) circle (  1.96);

\path[draw=drawColor,draw opacity=0.20,line width= 0.4pt,line join=round,line cap=round,fill=fillColor,fill opacity=0.20] (162.46, 29.09) circle (  1.96);

\path[draw=drawColor,draw opacity=0.20,line width= 0.4pt,line join=round,line cap=round,fill=fillColor,fill opacity=0.20] (162.46, 17.50) circle (  1.96);
\definecolor{drawColor}{RGB}{78,155,133}

\path[draw=drawColor,line width= 0.6pt,line join=round] (162.46, 71.63) -- (162.46, 73.86);

\path[draw=drawColor,line width= 0.6pt,line join=round] (162.46, 69.23) -- (162.46, 66.95);

\path[draw=drawColor,line width= 0.6pt,fill=fillColor,fill opacity=0.20] (152.29, 71.63) --
	(152.29, 69.23) --
	(172.63, 69.23) --
	(172.63, 71.63) --
	(152.29, 71.63) --
	cycle;

\path[draw=drawColor,line width= 1.1pt] (152.29, 70.26) -- (172.63, 70.26);
\definecolor{drawColor}{RGB}{78,155,133}

\path[draw=drawColor,draw opacity=0.20,line width= 0.4pt,line join=round,line cap=round,fill=fillColor,fill opacity=0.20] (189.59, 23.39) circle (  1.96);

\path[draw=drawColor,draw opacity=0.20,line width= 0.4pt,line join=round,line cap=round,fill=fillColor,fill opacity=0.20] (189.59,103.57) circle (  1.96);

\path[draw=drawColor,draw opacity=0.20,line width= 0.4pt,line join=round,line cap=round,fill=fillColor,fill opacity=0.20] (189.59, 95.45) circle (  1.96);

\path[draw=drawColor,draw opacity=0.20,line width= 0.4pt,line join=round,line cap=round,fill=fillColor,fill opacity=0.20] (189.59, 56.86) circle (  1.96);

\path[draw=drawColor,draw opacity=0.20,line width= 0.4pt,line join=round,line cap=round,fill=fillColor,fill opacity=0.20] (189.59, 85.72) circle (  1.96);

\path[draw=drawColor,draw opacity=0.20,line width= 0.4pt,line join=round,line cap=round,fill=fillColor,fill opacity=0.20] (189.59,  6.76) circle (  1.96);

\path[draw=drawColor,draw opacity=0.20,line width= 0.4pt,line join=round,line cap=round,fill=fillColor,fill opacity=0.20] (189.59,208.99) circle (  1.96);
\definecolor{drawColor}{RGB}{78,155,133}

\path[draw=drawColor,line width= 0.6pt,line join=round] (189.59, 73.40) -- (189.59, 75.76);

\path[draw=drawColor,line width= 0.6pt,line join=round] (189.59, 69.17) -- (189.59, 64.08);

\path[draw=drawColor,line width= 0.6pt,fill=fillColor,fill opacity=0.20] (179.41, 73.40) --
	(179.41, 69.17) --
	(199.76, 69.17) --
	(199.76, 73.40) --
	(179.41, 73.40) --
	cycle;

\path[draw=drawColor,line width= 1.1pt] (179.41, 69.97) -- (199.76, 69.97);
\definecolor{drawColor}{RGB}{78,155,133}

\path[draw=drawColor,draw opacity=0.20,line width= 0.4pt,line join=round,line cap=round,fill=fillColor,fill opacity=0.20] (216.71,-11.55) circle (  1.96);

\path[draw=drawColor,draw opacity=0.20,line width= 0.4pt,line join=round,line cap=round,fill=fillColor,fill opacity=0.20] (216.71,128.78) circle (  1.96);

\path[draw=drawColor,draw opacity=0.20,line width= 0.4pt,line join=round,line cap=round,fill=fillColor,fill opacity=0.20] (216.71,143.39) circle (  1.96);

\path[draw=drawColor,draw opacity=0.20,line width= 0.4pt,line join=round,line cap=round,fill=fillColor,fill opacity=0.20] (216.71, 47.42) circle (  1.96);

\path[draw=drawColor,draw opacity=0.20,line width= 0.4pt,line join=round,line cap=round,fill=fillColor,fill opacity=0.20] (216.71, 97.05) circle (  1.96);
\definecolor{drawColor}{RGB}{78,155,133}

\path[draw=drawColor,line width= 0.6pt,line join=round] (216.71, 76.03) -- (216.71, 78.57);

\path[draw=drawColor,line width= 0.6pt,line join=round] (216.71, 66.41) -- (216.71, 60.92);

\path[draw=drawColor,line width= 0.6pt,fill=fillColor,fill opacity=0.20] (206.54, 76.03) --
	(206.54, 66.41) --
	(226.89, 66.41) --
	(226.89, 76.03) --
	(206.54, 76.03) --
	cycle;

\path[draw=drawColor,line width= 1.1pt] (206.54, 70.39) -- (226.89, 70.39);
\end{scope}
\begin{scope}
\path[clip] (  0.00,  0.00) rectangle (238.49,115.63);
\definecolor{drawColor}{RGB}{0,0,0}

\path[draw=drawColor,line width= 0.6pt,line join=round] ( 37.67, 29.80) --
	( 37.67,110.13);

\path[draw=drawColor,line width= 0.6pt,line join=round] ( 39.09,107.67) --
	( 37.67,110.13) --
	( 36.25,107.67);
\end{scope}
\begin{scope}
\path[clip] (  0.00,  0.00) rectangle (238.49,115.63);
\definecolor{drawColor}{gray}{0.30}

\node[text=drawColor,anchor=base east,inner sep=0pt, outer sep=0pt, scale=  0.88] at ( 32.72, 42.59) {0.98};

\node[text=drawColor,anchor=base east,inner sep=0pt, outer sep=0pt, scale=  0.88] at ( 32.72, 66.94) {1.00};

\node[text=drawColor,anchor=base east,inner sep=0pt, outer sep=0pt, scale=  0.88] at ( 32.72, 91.28) {1.02};
\end{scope}
\begin{scope}
\path[clip] (  0.00,  0.00) rectangle (238.49,115.63);
\definecolor{drawColor}{gray}{0.20}

\path[draw=drawColor,line width= 0.6pt,line join=round] ( 34.92, 45.63) --
	( 37.67, 45.63);

\path[draw=drawColor,line width= 0.6pt,line join=round] ( 34.92, 69.97) --
	( 37.67, 69.97);

\path[draw=drawColor,line width= 0.6pt,line join=round] ( 34.92, 94.31) --
	( 37.67, 94.31);
\end{scope}
\begin{scope}
\path[clip] (  0.00,  0.00) rectangle (238.49,115.63);
\definecolor{drawColor}{RGB}{0,0,0}

\path[draw=drawColor,line width= 0.6pt,line join=round] ( 37.67, 29.80) --
	(232.99, 29.80);

\path[draw=drawColor,line width= 0.6pt,line join=round] (230.53, 28.38) --
	(232.99, 29.80) --
	(230.53, 31.23);
\end{scope}
\begin{scope}
\path[clip] (  0.00,  0.00) rectangle (238.49,115.63);
\definecolor{drawColor}{gray}{0.20}

\path[draw=drawColor,line width= 0.6pt,line join=round] ( 53.95, 27.05) --
	( 53.95, 29.80);

\path[draw=drawColor,line width= 0.6pt,line join=round] ( 81.08, 27.05) --
	( 81.08, 29.80);

\path[draw=drawColor,line width= 0.6pt,line join=round] (108.20, 27.05) --
	(108.20, 29.80);

\path[draw=drawColor,line width= 0.6pt,line join=round] (135.33, 27.05) --
	(135.33, 29.80);

\path[draw=drawColor,line width= 0.6pt,line join=round] (162.46, 27.05) --
	(162.46, 29.80);

\path[draw=drawColor,line width= 0.6pt,line join=round] (189.59, 27.05) --
	(189.59, 29.80);

\path[draw=drawColor,line width= 0.6pt,line join=round] (216.71, 27.05) --
	(216.71, 29.80);
\end{scope}
\begin{scope}
\path[clip] (  0.00,  0.00) rectangle (238.49,115.63);
\definecolor{drawColor}{gray}{0.30}

\node[text=drawColor,anchor=base,inner sep=0pt, outer sep=0pt, scale=  0.88] at ( 53.95, 18.79) {2};

\node[text=drawColor,anchor=base,inner sep=0pt, outer sep=0pt, scale=  0.88] at ( 81.08, 18.79) {4};

\node[text=drawColor,anchor=base,inner sep=0pt, outer sep=0pt, scale=  0.88] at (108.20, 18.79) {8};

\node[text=drawColor,anchor=base,inner sep=0pt, outer sep=0pt, scale=  0.88] at (135.33, 18.79) {16};

\node[text=drawColor,anchor=base,inner sep=0pt, outer sep=0pt, scale=  0.88] at (162.46, 18.79) {32};

\node[text=drawColor,anchor=base,inner sep=0pt, outer sep=0pt, scale=  0.88] at (189.59, 18.79) {64};

\node[text=drawColor,anchor=base,inner sep=0pt, outer sep=0pt, scale=  0.88] at (216.71, 18.79) {128};
\end{scope}
\begin{scope}
\path[clip] (  0.00,  0.00) rectangle (238.49,115.63);
\definecolor{drawColor}{RGB}{0,0,0}

\node[text=drawColor,anchor=base,inner sep=0pt, outer sep=0pt, scale=  1.00] at (135.33,  7.44) {domain size};
\end{scope}
\begin{scope}
\path[clip] (  0.00,  0.00) rectangle (238.49,115.63);
\definecolor{drawColor}{RGB}{0,0,0}

\node[text=drawColor,rotate= 90.00,anchor=base,inner sep=0pt, outer sep=0pt, scale=  1.00] at ( 12.39, 69.97) {$p' \mathbin{/} p$};
\end{scope}
\end{tikzpicture}

%% file: files/plot-quots-p=0.5-eps=0.001.tex
\begin{tikzpicture}[x=1pt,y=1pt]
\definecolor{fillColor}{RGB}{255,255,255}
\path[use as bounding box,fill=fillColor,fill opacity=0.00] (0,0) rectangle (238.49,115.63);
\begin{scope}
\path[clip] (  0.00,  0.00) rectangle (238.49,115.63);
\definecolor{drawColor}{RGB}{255,255,255}
\definecolor{fillColor}{RGB}{255,255,255}

\path[draw=drawColor,line width= 0.6pt,line join=round,line cap=round,fill=fillColor] (  0.00,  0.00) rectangle (238.49,115.63);
\end{scope}
\begin{scope}
\path[clip] ( 46.47, 29.80) rectangle (232.99,110.13);
\definecolor{fillColor}{RGB}{255,255,255}

\path[fill=fillColor] ( 46.47, 29.80) rectangle (232.99,110.13);
\definecolor{drawColor}{RGB}{78,155,133}
\definecolor{fillColor}{RGB}{78,155,133}

\path[draw=drawColor,draw opacity=0.20,line width= 0.4pt,line join=round,line cap=round,fill=fillColor,fill opacity=0.20] ( 62.01, 69.97) circle (  1.96);

\path[draw=drawColor,draw opacity=0.20,line width= 0.4pt,line join=round,line cap=round,fill=fillColor,fill opacity=0.20] ( 62.01, 69.97) circle (  1.96);

\path[draw=drawColor,draw opacity=0.20,line width= 0.4pt,line join=round,line cap=round,fill=fillColor,fill opacity=0.20] ( 62.01, 70.93) circle (  1.96);

\path[draw=drawColor,draw opacity=0.20,line width= 0.4pt,line join=round,line cap=round,fill=fillColor,fill opacity=0.20] ( 62.01, 69.12) circle (  1.96);

\path[draw=drawColor,draw opacity=0.20,line width= 0.4pt,line join=round,line cap=round,fill=fillColor,fill opacity=0.20] ( 62.01, 71.68) circle (  1.96);

\path[draw=drawColor,draw opacity=0.20,line width= 0.4pt,line join=round,line cap=round,fill=fillColor,fill opacity=0.20] ( 62.01, 67.98) circle (  1.96);

\path[draw=drawColor,draw opacity=0.20,line width= 0.4pt,line join=round,line cap=round,fill=fillColor,fill opacity=0.20] ( 62.01, 69.97) circle (  1.96);

\path[draw=drawColor,draw opacity=0.20,line width= 0.4pt,line join=round,line cap=round,fill=fillColor,fill opacity=0.20] ( 62.01, 69.97) circle (  1.96);

\path[draw=drawColor,draw opacity=0.20,line width= 0.4pt,line join=round,line cap=round,fill=fillColor,fill opacity=0.20] ( 62.01, 36.65) circle (  1.96);

\path[draw=drawColor,draw opacity=0.20,line width= 0.4pt,line join=round,line cap=round,fill=fillColor,fill opacity=0.20] ( 62.01, 99.82) circle (  1.96);

\path[draw=drawColor,draw opacity=0.20,line width= 0.4pt,line join=round,line cap=round,fill=fillColor,fill opacity=0.20] ( 62.01, 66.11) circle (  1.96);

\path[draw=drawColor,draw opacity=0.20,line width= 0.4pt,line join=round,line cap=round,fill=fillColor,fill opacity=0.20] ( 62.01, 73.98) circle (  1.96);
\definecolor{drawColor}{RGB}{78,155,133}

\path[draw=drawColor,line width= 0.6pt,line join=round] ( 62.01, 69.97) -- ( 62.01, 69.97);

\path[draw=drawColor,line width= 0.6pt,line join=round] ( 62.01, 69.97) -- ( 62.01, 69.97);

\path[draw=drawColor,line width= 0.6pt,fill=fillColor,fill opacity=0.20] ( 52.30, 69.97) --
	( 52.30, 69.97) --
	( 71.73, 69.97) --
	( 71.73, 69.97) --
	( 52.30, 69.97) --
	cycle;

\path[draw=drawColor,line width= 1.1pt] ( 52.30, 69.97) -- ( 71.73, 69.97);
\definecolor{drawColor}{RGB}{78,155,133}

\path[draw=drawColor,draw opacity=0.20,line width= 0.4pt,line join=round,line cap=round,fill=fillColor,fill opacity=0.20] ( 87.92, 70.01) circle (  1.96);

\path[draw=drawColor,draw opacity=0.20,line width= 0.4pt,line join=round,line cap=round,fill=fillColor,fill opacity=0.20] ( 87.92, 92.44) circle (  1.96);

\path[draw=drawColor,draw opacity=0.20,line width= 0.4pt,line join=round,line cap=round,fill=fillColor,fill opacity=0.20] ( 87.92, 43.50) circle (  1.96);

\path[draw=drawColor,draw opacity=0.20,line width= 0.4pt,line join=round,line cap=round,fill=fillColor,fill opacity=0.20] ( 87.92, 57.97) circle (  1.96);

\path[draw=drawColor,draw opacity=0.20,line width= 0.4pt,line join=round,line cap=round,fill=fillColor,fill opacity=0.20] ( 87.92, 86.38) circle (  1.96);

\path[draw=drawColor,draw opacity=0.20,line width= 0.4pt,line join=round,line cap=round,fill=fillColor,fill opacity=0.20] ( 87.92, 68.43) circle (  1.96);

\path[draw=drawColor,draw opacity=0.20,line width= 0.4pt,line join=round,line cap=round,fill=fillColor,fill opacity=0.20] ( 87.92, 72.58) circle (  1.96);

\path[draw=drawColor,draw opacity=0.20,line width= 0.4pt,line join=round,line cap=round,fill=fillColor,fill opacity=0.20] ( 87.92, 61.61) circle (  1.96);

\path[draw=drawColor,draw opacity=0.20,line width= 0.4pt,line join=round,line cap=round,fill=fillColor,fill opacity=0.20] ( 87.92, 75.31) circle (  1.96);

\path[draw=drawColor,draw opacity=0.20,line width= 0.4pt,line join=round,line cap=round,fill=fillColor,fill opacity=0.20] ( 87.92, 85.39) circle (  1.96);

\path[draw=drawColor,draw opacity=0.20,line width= 0.4pt,line join=round,line cap=round,fill=fillColor,fill opacity=0.20] ( 87.92, 50.21) circle (  1.96);
\definecolor{drawColor}{RGB}{78,155,133}

\path[draw=drawColor,line width= 0.6pt,line join=round] ( 87.92, 69.99) -- ( 87.92, 69.99);

\path[draw=drawColor,line width= 0.6pt,line join=round] ( 87.92, 69.97) -- ( 87.92, 69.96);

\path[draw=drawColor,line width= 0.6pt,fill=fillColor,fill opacity=0.20] ( 78.20, 69.99) --
	( 78.20, 69.97) --
	( 97.63, 69.97) --
	( 97.63, 69.99) --
	( 78.20, 69.99) --
	cycle;

\path[draw=drawColor,line width= 1.1pt] ( 78.20, 69.97) -- ( 97.63, 69.97);
\definecolor{drawColor}{RGB}{78,155,133}

\path[draw=drawColor,draw opacity=0.20,line width= 0.4pt,line join=round,line cap=round,fill=fillColor,fill opacity=0.20] (113.82, 18.36) circle (  1.96);

\path[draw=drawColor,draw opacity=0.20,line width= 0.4pt,line join=round,line cap=round,fill=fillColor,fill opacity=0.20] (113.82,102.78) circle (  1.96);

\path[draw=drawColor,draw opacity=0.20,line width= 0.4pt,line join=round,line cap=round,fill=fillColor,fill opacity=0.20] (113.82, 60.36) circle (  1.96);

\path[draw=drawColor,draw opacity=0.20,line width= 0.4pt,line join=round,line cap=round,fill=fillColor,fill opacity=0.20] (113.82, 81.51) circle (  1.96);

\path[draw=drawColor,draw opacity=0.20,line width= 0.4pt,line join=round,line cap=round,fill=fillColor,fill opacity=0.20] (113.82, 86.74) circle (  1.96);

\path[draw=drawColor,draw opacity=0.20,line width= 0.4pt,line join=round,line cap=round,fill=fillColor,fill opacity=0.20] (113.82, 46.87) circle (  1.96);

\path[draw=drawColor,draw opacity=0.20,line width= 0.4pt,line join=round,line cap=round,fill=fillColor,fill opacity=0.20] (113.82, 88.60) circle (  1.96);

\path[draw=drawColor,draw opacity=0.20,line width= 0.4pt,line join=round,line cap=round,fill=fillColor,fill opacity=0.20] (113.82, 45.66) circle (  1.96);
\definecolor{drawColor}{RGB}{78,155,133}

\path[draw=drawColor,line width= 0.6pt,line join=round] (113.82, 73.02) -- (113.82, 76.96);

\path[draw=drawColor,line width= 0.6pt,line join=round] (113.82, 68.02) -- (113.82, 60.86);

\path[draw=drawColor,line width= 0.6pt,fill=fillColor,fill opacity=0.20] (104.11, 73.02) --
	(104.11, 68.02) --
	(123.54, 68.02) --
	(123.54, 73.02) --
	(104.11, 73.02) --
	cycle;

\path[draw=drawColor,line width= 1.1pt] (104.11, 69.97) -- (123.54, 69.97);
\definecolor{drawColor}{RGB}{78,155,133}

\path[draw=drawColor,draw opacity=0.20,line width= 0.4pt,line join=round,line cap=round,fill=fillColor,fill opacity=0.20] (139.73, 31.53) circle (  1.96);

\path[draw=drawColor,draw opacity=0.20,line width= 0.4pt,line join=round,line cap=round,fill=fillColor,fill opacity=0.20] (139.73, 93.62) circle (  1.96);

\path[draw=drawColor,draw opacity=0.20,line width= 0.4pt,line join=round,line cap=round,fill=fillColor,fill opacity=0.20] (139.73, 78.00) circle (  1.96);

\path[draw=drawColor,draw opacity=0.20,line width= 0.4pt,line join=round,line cap=round,fill=fillColor,fill opacity=0.20] (139.73, 60.32) circle (  1.96);

\path[draw=drawColor,draw opacity=0.20,line width= 0.4pt,line join=round,line cap=round,fill=fillColor,fill opacity=0.20] (139.73, 14.30) circle (  1.96);

\path[draw=drawColor,draw opacity=0.20,line width= 0.4pt,line join=round,line cap=round,fill=fillColor,fill opacity=0.20] (139.73,104.87) circle (  1.96);

\path[draw=drawColor,draw opacity=0.20,line width= 0.4pt,line join=round,line cap=round,fill=fillColor,fill opacity=0.20] (139.73, 65.67) circle (  1.96);

\path[draw=drawColor,draw opacity=0.20,line width= 0.4pt,line join=round,line cap=round,fill=fillColor,fill opacity=0.20] (139.73, 74.36) circle (  1.96);

\path[draw=drawColor,draw opacity=0.20,line width= 0.4pt,line join=round,line cap=round,fill=fillColor,fill opacity=0.20] (139.73, 62.87) circle (  1.96);

\path[draw=drawColor,draw opacity=0.20,line width= 0.4pt,line join=round,line cap=round,fill=fillColor,fill opacity=0.20] (139.73, 78.49) circle (  1.96);

\path[draw=drawColor,draw opacity=0.20,line width= 0.4pt,line join=round,line cap=round,fill=fillColor,fill opacity=0.20] (139.73, 60.77) circle (  1.96);

\path[draw=drawColor,draw opacity=0.20,line width= 0.4pt,line join=round,line cap=round,fill=fillColor,fill opacity=0.20] (139.73, 81.34) circle (  1.96);
\definecolor{drawColor}{RGB}{78,155,133}

\path[draw=drawColor,line width= 0.6pt,line join=round] (139.73, 70.71) -- (139.73, 70.88);

\path[draw=drawColor,line width= 0.6pt,line join=round] (139.73, 68.95) -- (139.73, 68.71);

\path[draw=drawColor,line width= 0.6pt,fill=fillColor,fill opacity=0.20] (130.02, 70.71) --
	(130.02, 68.95) --
	(149.45, 68.95) --
	(149.45, 70.71) --
	(130.02, 70.71) --
	cycle;

\path[draw=drawColor,line width= 1.1pt] (130.02, 69.98) -- (149.45, 69.98);
\definecolor{drawColor}{RGB}{78,155,133}

\path[draw=drawColor,draw opacity=0.20,line width= 0.4pt,line join=round,line cap=round,fill=fillColor,fill opacity=0.20] (165.64,100.10) circle (  1.96);

\path[draw=drawColor,draw opacity=0.20,line width= 0.4pt,line join=round,line cap=round,fill=fillColor,fill opacity=0.20] (165.64, 51.43) circle (  1.96);

\path[draw=drawColor,draw opacity=0.20,line width= 0.4pt,line join=round,line cap=round,fill=fillColor,fill opacity=0.20] (165.64, -3.90) circle (  1.96);

\path[draw=drawColor,draw opacity=0.20,line width= 0.4pt,line join=round,line cap=round,fill=fillColor,fill opacity=0.20] (165.64,116.27) circle (  1.96);

\path[draw=drawColor,draw opacity=0.20,line width= 0.4pt,line join=round,line cap=round,fill=fillColor,fill opacity=0.20] (165.64, 52.19) circle (  1.96);

\path[draw=drawColor,draw opacity=0.20,line width= 0.4pt,line join=round,line cap=round,fill=fillColor,fill opacity=0.20] (165.64, 94.45) circle (  1.96);

\path[draw=drawColor,draw opacity=0.20,line width= 0.4pt,line join=round,line cap=round,fill=fillColor,fill opacity=0.20] (165.64,100.55) circle (  1.96);
\definecolor{drawColor}{RGB}{78,155,133}

\path[draw=drawColor,line width= 0.6pt,line join=round] (165.64, 74.45) -- (165.64, 81.31);

\path[draw=drawColor,line width= 0.6pt,line join=round] (165.64, 66.92) -- (165.64, 56.35);

\path[draw=drawColor,line width= 0.6pt,fill=fillColor,fill opacity=0.20] (155.92, 74.45) --
	(155.92, 66.92) --
	(175.35, 66.92) --
	(175.35, 74.45) --
	(155.92, 74.45) --
	cycle;

\path[draw=drawColor,line width= 1.1pt] (155.92, 70.00) -- (175.35, 70.00);
\definecolor{drawColor}{RGB}{78,155,133}

\path[draw=drawColor,draw opacity=0.20,line width= 0.4pt,line join=round,line cap=round,fill=fillColor,fill opacity=0.20] (191.54,100.95) circle (  1.96);

\path[draw=drawColor,draw opacity=0.20,line width= 0.4pt,line join=round,line cap=round,fill=fillColor,fill opacity=0.20] (191.54, 50.91) circle (  1.96);

\path[draw=drawColor,draw opacity=0.20,line width= 0.4pt,line join=round,line cap=round,fill=fillColor,fill opacity=0.20] (191.54, -6.73) circle (  1.96);

\path[draw=drawColor,draw opacity=0.20,line width= 0.4pt,line join=round,line cap=round,fill=fillColor,fill opacity=0.20] (191.54,118.05) circle (  1.96);

\path[draw=drawColor,draw opacity=0.20,line width= 0.4pt,line join=round,line cap=round,fill=fillColor,fill opacity=0.20] (191.54, 83.03) circle (  1.96);

\path[draw=drawColor,draw opacity=0.20,line width= 0.4pt,line join=round,line cap=round,fill=fillColor,fill opacity=0.20] (191.54, 85.85) circle (  1.96);

\path[draw=drawColor,draw opacity=0.20,line width= 0.4pt,line join=round,line cap=round,fill=fillColor,fill opacity=0.20] (191.54,-80.52) circle (  1.96);
\definecolor{drawColor}{RGB}{78,155,133}

\path[draw=drawColor,line width= 0.6pt,line join=round] (191.54, 71.63) -- (191.54, 77.97);

\path[draw=drawColor,line width= 0.6pt,line join=round] (191.54, 65.90) -- (191.54, 58.43);

\path[draw=drawColor,line width= 0.6pt,fill=fillColor,fill opacity=0.20] (181.83, 71.63) --
	(181.83, 65.90) --
	(201.26, 65.90) --
	(201.26, 71.63) --
	(181.83, 71.63) --
	cycle;

\path[draw=drawColor,line width= 1.1pt] (181.83, 69.97) -- (201.26, 69.97);
\definecolor{drawColor}{RGB}{78,155,133}

\path[draw=drawColor,draw opacity=0.20,line width= 0.4pt,line join=round,line cap=round,fill=fillColor,fill opacity=0.20] (217.45,102.15) circle (  1.96);

\path[draw=drawColor,draw opacity=0.20,line width= 0.4pt,line join=round,line cap=round,fill=fillColor,fill opacity=0.20] (217.45, 50.17) circle (  1.96);

\path[draw=drawColor,draw opacity=0.20,line width= 0.4pt,line join=round,line cap=round,fill=fillColor,fill opacity=0.20] (217.45,-24.55) circle (  1.96);

\path[draw=drawColor,draw opacity=0.20,line width= 0.4pt,line join=round,line cap=round,fill=fillColor,fill opacity=0.20] (217.45,129.22) circle (  1.96);

\path[draw=drawColor,draw opacity=0.20,line width= 0.4pt,line join=round,line cap=round,fill=fillColor,fill opacity=0.20] (217.45, 50.77) circle (  1.96);

\path[draw=drawColor,draw opacity=0.20,line width= 0.4pt,line join=round,line cap=round,fill=fillColor,fill opacity=0.20] (217.45, 96.40) circle (  1.96);
\definecolor{drawColor}{RGB}{78,155,133}

\path[draw=drawColor,line width= 0.6pt,line join=round] (217.45, 72.64) -- (217.45, 78.63);

\path[draw=drawColor,line width= 0.6pt,line join=round] (217.45, 67.02) -- (217.45, 61.50);

\path[draw=drawColor,line width= 0.6pt,fill=fillColor,fill opacity=0.20] (207.73, 72.64) --
	(207.73, 67.02) --
	(227.16, 67.02) --
	(227.16, 72.64) --
	(207.73, 72.64) --
	cycle;

\path[draw=drawColor,line width= 1.1pt] (207.73, 70.01) -- (227.16, 70.01);
\end{scope}
\begin{scope}
\path[clip] (  0.00,  0.00) rectangle (238.49,115.63);
\definecolor{drawColor}{RGB}{0,0,0}

\path[draw=drawColor,line width= 0.6pt,line join=round] ( 46.47, 29.80) --
	( 46.47,110.13);

\path[draw=drawColor,line width= 0.6pt,line join=round] ( 47.89,107.67) --
	( 46.47,110.13) --
	( 45.05,107.67);
\end{scope}
\begin{scope}
\path[clip] (  0.00,  0.00) rectangle (238.49,115.63);
\definecolor{drawColor}{gray}{0.30}

\node[text=drawColor,anchor=base east,inner sep=0pt, outer sep=0pt, scale=  0.88] at ( 41.52, 42.59) {0.9998};

\node[text=drawColor,anchor=base east,inner sep=0pt, outer sep=0pt, scale=  0.88] at ( 41.52, 66.94) {1.0000};

\node[text=drawColor,anchor=base east,inner sep=0pt, outer sep=0pt, scale=  0.88] at ( 41.52, 91.28) {1.0002};
\end{scope}
\begin{scope}
\path[clip] (  0.00,  0.00) rectangle (238.49,115.63);
\definecolor{drawColor}{gray}{0.20}

\path[draw=drawColor,line width= 0.6pt,line join=round] ( 43.72, 45.63) --
	( 46.47, 45.63);

\path[draw=drawColor,line width= 0.6pt,line join=round] ( 43.72, 69.97) --
	( 46.47, 69.97);

\path[draw=drawColor,line width= 0.6pt,line join=round] ( 43.72, 94.31) --
	( 46.47, 94.31);
\end{scope}
\begin{scope}
\path[clip] (  0.00,  0.00) rectangle (238.49,115.63);
\definecolor{drawColor}{RGB}{0,0,0}

\path[draw=drawColor,line width= 0.6pt,line join=round] ( 46.47, 29.80) --
	(232.99, 29.80);

\path[draw=drawColor,line width= 0.6pt,line join=round] (230.53, 28.38) --
	(232.99, 29.80) --
	(230.53, 31.23);
\end{scope}
\begin{scope}
\path[clip] (  0.00,  0.00) rectangle (238.49,115.63);
\definecolor{drawColor}{gray}{0.20}

\path[draw=drawColor,line width= 0.6pt,line join=round] ( 62.01, 27.05) --
	( 62.01, 29.80);

\path[draw=drawColor,line width= 0.6pt,line join=round] ( 87.92, 27.05) --
	( 87.92, 29.80);

\path[draw=drawColor,line width= 0.6pt,line join=round] (113.82, 27.05) --
	(113.82, 29.80);

\path[draw=drawColor,line width= 0.6pt,line join=round] (139.73, 27.05) --
	(139.73, 29.80);

\path[draw=drawColor,line width= 0.6pt,line join=round] (165.64, 27.05) --
	(165.64, 29.80);

\path[draw=drawColor,line width= 0.6pt,line join=round] (191.54, 27.05) --
	(191.54, 29.80);

\path[draw=drawColor,line width= 0.6pt,line join=round] (217.45, 27.05) --
	(217.45, 29.80);
\end{scope}
\begin{scope}
\path[clip] (  0.00,  0.00) rectangle (238.49,115.63);
\definecolor{drawColor}{gray}{0.30}

\node[text=drawColor,anchor=base,inner sep=0pt, outer sep=0pt, scale=  0.88] at ( 62.01, 18.79) {2};

\node[text=drawColor,anchor=base,inner sep=0pt, outer sep=0pt, scale=  0.88] at ( 87.92, 18.79) {4};

\node[text=drawColor,anchor=base,inner sep=0pt, outer sep=0pt, scale=  0.88] at (113.82, 18.79) {8};

\node[text=drawColor,anchor=base,inner sep=0pt, outer sep=0pt, scale=  0.88] at (139.73, 18.79) {16};

\node[text=drawColor,anchor=base,inner sep=0pt, outer sep=0pt, scale=  0.88] at (165.64, 18.79) {32};

\node[text=drawColor,anchor=base,inner sep=0pt, outer sep=0pt, scale=  0.88] at (191.54, 18.79) {64};

\node[text=drawColor,anchor=base,inner sep=0pt, outer sep=0pt, scale=  0.88] at (217.45, 18.79) {128};
\end{scope}
\begin{scope}
\path[clip] (  0.00,  0.00) rectangle (238.49,115.63);
\definecolor{drawColor}{RGB}{0,0,0}

\node[text=drawColor,anchor=base,inner sep=0pt, outer sep=0pt, scale=  1.00] at (139.73,  7.44) {domain size};
\end{scope}
\begin{scope}
\path[clip] (  0.00,  0.00) rectangle (238.49,115.63);
\definecolor{drawColor}{RGB}{0,0,0}

\node[text=drawColor,rotate= 90.00,anchor=base,inner sep=0pt, outer sep=0pt, scale=  1.00] at ( 12.39, 69.97) {$p' \mathbin{/} p$};
\end{scope}
\end{tikzpicture}

%% file: files/plot-quots-p=0.5-eps=0.1.tex
\begin{tikzpicture}[x=1pt,y=1pt]
\definecolor{fillColor}{RGB}{255,255,255}
\path[use as bounding box,fill=fillColor,fill opacity=0.00] (0,0) rectangle (238.49,115.63);
\begin{scope}
\path[clip] (  0.00,  0.00) rectangle (238.49,115.63);
\definecolor{drawColor}{RGB}{255,255,255}
\definecolor{fillColor}{RGB}{255,255,255}

\path[draw=drawColor,line width= 0.6pt,line join=round,line cap=round,fill=fillColor] (  0.00,  0.00) rectangle (238.49,115.63);
\end{scope}
\begin{scope}
\path[clip] ( 37.67, 29.80) rectangle (232.99,110.13);
\definecolor{fillColor}{RGB}{255,255,255}

\path[fill=fillColor] ( 37.67, 29.80) rectangle (232.99,110.13);
\definecolor{drawColor}{RGB}{78,155,133}
\definecolor{fillColor}{RGB}{78,155,133}

\path[draw=drawColor,draw opacity=0.20,line width= 0.4pt,line join=round,line cap=round,fill=fillColor,fill opacity=0.20] ( 53.95, 69.97) circle (  1.96);

\path[draw=drawColor,draw opacity=0.20,line width= 0.4pt,line join=round,line cap=round,fill=fillColor,fill opacity=0.20] ( 53.95, 69.97) circle (  1.96);

\path[draw=drawColor,draw opacity=0.20,line width= 0.4pt,line join=round,line cap=round,fill=fillColor,fill opacity=0.20] ( 53.95, 71.48) circle (  1.96);

\path[draw=drawColor,draw opacity=0.20,line width= 0.4pt,line join=round,line cap=round,fill=fillColor,fill opacity=0.20] ( 53.95, 68.65) circle (  1.96);

\path[draw=drawColor,draw opacity=0.20,line width= 0.4pt,line join=round,line cap=round,fill=fillColor,fill opacity=0.20] ( 53.95, 69.44) circle (  1.96);

\path[draw=drawColor,draw opacity=0.20,line width= 0.4pt,line join=round,line cap=round,fill=fillColor,fill opacity=0.20] ( 53.95, 70.58) circle (  1.96);

\path[draw=drawColor,draw opacity=0.20,line width= 0.4pt,line join=round,line cap=round,fill=fillColor,fill opacity=0.20] ( 53.95, 69.84) circle (  1.96);

\path[draw=drawColor,draw opacity=0.20,line width= 0.4pt,line join=round,line cap=round,fill=fillColor,fill opacity=0.20] ( 53.95, 70.38) circle (  1.96);

\path[draw=drawColor,draw opacity=0.20,line width= 0.4pt,line join=round,line cap=round,fill=fillColor,fill opacity=0.20] ( 53.95, 42.52) circle (  1.96);

\path[draw=drawColor,draw opacity=0.20,line width= 0.4pt,line join=round,line cap=round,fill=fillColor,fill opacity=0.20] ( 53.95, 96.72) circle (  1.96);

\path[draw=drawColor,draw opacity=0.20,line width= 0.4pt,line join=round,line cap=round,fill=fillColor,fill opacity=0.20] ( 53.95, 68.02) circle (  1.96);

\path[draw=drawColor,draw opacity=0.20,line width= 0.4pt,line join=round,line cap=round,fill=fillColor,fill opacity=0.20] ( 53.95, 72.00) circle (  1.96);
\definecolor{drawColor}{RGB}{78,155,133}

\path[draw=drawColor,line width= 0.6pt,line join=round] ( 53.95, 69.97) -- ( 53.95, 69.97);

\path[draw=drawColor,line width= 0.6pt,line join=round] ( 53.95, 69.97) -- ( 53.95, 69.97);

\path[draw=drawColor,line width= 0.6pt,fill=fillColor,fill opacity=0.20] ( 43.78, 69.97) --
	( 43.78, 69.97) --
	( 64.12, 69.97) --
	( 64.12, 69.97) --
	( 43.78, 69.97) --
	cycle;

\path[draw=drawColor,line width= 1.1pt] ( 43.78, 69.97) -- ( 64.12, 69.97);
\definecolor{drawColor}{RGB}{78,155,133}

\path[draw=drawColor,draw opacity=0.20,line width= 0.4pt,line join=round,line cap=round,fill=fillColor,fill opacity=0.20] ( 81.08, 73.69) circle (  1.96);

\path[draw=drawColor,draw opacity=0.20,line width= 0.4pt,line join=round,line cap=round,fill=fillColor,fill opacity=0.20] ( 81.08, 89.86) circle (  1.96);

\path[draw=drawColor,draw opacity=0.20,line width= 0.4pt,line join=round,line cap=round,fill=fillColor,fill opacity=0.20] ( 81.08, 46.57) circle (  1.96);

\path[draw=drawColor,draw opacity=0.20,line width= 0.4pt,line join=round,line cap=round,fill=fillColor,fill opacity=0.20] ( 81.08, 60.49) circle (  1.96);

\path[draw=drawColor,draw opacity=0.20,line width= 0.4pt,line join=round,line cap=round,fill=fillColor,fill opacity=0.20] ( 81.08, 82.93) circle (  1.96);

\path[draw=drawColor,draw opacity=0.20,line width= 0.4pt,line join=round,line cap=round,fill=fillColor,fill opacity=0.20] ( 81.08, 62.70) circle (  1.96);

\path[draw=drawColor,draw opacity=0.20,line width= 0.4pt,line join=round,line cap=round,fill=fillColor,fill opacity=0.20] ( 81.08, 74.52) circle (  1.96);

\path[draw=drawColor,draw opacity=0.20,line width= 0.4pt,line join=round,line cap=round,fill=fillColor,fill opacity=0.20] ( 81.08, 82.29) circle (  1.96);

\path[draw=drawColor,draw opacity=0.20,line width= 0.4pt,line join=round,line cap=round,fill=fillColor,fill opacity=0.20] ( 81.08, 54.27) circle (  1.96);
\definecolor{drawColor}{RGB}{78,155,133}

\path[draw=drawColor,line width= 0.6pt,line join=round] ( 81.08, 71.43) -- ( 81.08, 72.19);

\path[draw=drawColor,line width= 0.6pt,line join=round] ( 81.08, 69.93) -- ( 81.08, 68.65);

\path[draw=drawColor,line width= 0.6pt,fill=fillColor,fill opacity=0.20] ( 70.90, 71.43) --
	( 70.90, 69.93) --
	( 91.25, 69.93) --
	( 91.25, 71.43) --
	( 70.90, 71.43) --
	cycle;

\path[draw=drawColor,line width= 1.1pt] ( 70.90, 69.97) -- ( 91.25, 69.97);
\definecolor{drawColor}{RGB}{78,155,133}

\path[draw=drawColor,draw opacity=0.20,line width= 0.4pt,line join=round,line cap=round,fill=fillColor,fill opacity=0.20] (108.20, 28.05) circle (  1.96);

\path[draw=drawColor,draw opacity=0.20,line width= 0.4pt,line join=round,line cap=round,fill=fillColor,fill opacity=0.20] (108.20,100.54) circle (  1.96);

\path[draw=drawColor,draw opacity=0.20,line width= 0.4pt,line join=round,line cap=round,fill=fillColor,fill opacity=0.20] (108.20, 60.73) circle (  1.96);

\path[draw=drawColor,draw opacity=0.20,line width= 0.4pt,line join=round,line cap=round,fill=fillColor,fill opacity=0.20] (108.20, 81.06) circle (  1.96);

\path[draw=drawColor,draw opacity=0.20,line width= 0.4pt,line join=round,line cap=round,fill=fillColor,fill opacity=0.20] (108.20, 83.92) circle (  1.96);

\path[draw=drawColor,draw opacity=0.20,line width= 0.4pt,line join=round,line cap=round,fill=fillColor,fill opacity=0.20] (108.20, 50.75) circle (  1.96);

\path[draw=drawColor,draw opacity=0.20,line width= 0.4pt,line join=round,line cap=round,fill=fillColor,fill opacity=0.20] (108.20, 61.32) circle (  1.96);

\path[draw=drawColor,draw opacity=0.20,line width= 0.4pt,line join=round,line cap=round,fill=fillColor,fill opacity=0.20] (108.20, 84.08) circle (  1.96);

\path[draw=drawColor,draw opacity=0.20,line width= 0.4pt,line join=round,line cap=round,fill=fillColor,fill opacity=0.20] (108.20, 51.97) circle (  1.96);
\definecolor{drawColor}{RGB}{78,155,133}

\path[draw=drawColor,line width= 0.6pt,line join=round] (108.20, 72.86) -- (108.20, 77.61);

\path[draw=drawColor,line width= 0.6pt,line join=round] (108.20, 68.26) -- (108.20, 67.77);

\path[draw=drawColor,line width= 0.6pt,fill=fillColor,fill opacity=0.20] ( 98.03, 72.86) --
	( 98.03, 68.26) --
	(118.38, 68.26) --
	(118.38, 72.86) --
	( 98.03, 72.86) --
	cycle;

\path[draw=drawColor,line width= 1.1pt] ( 98.03, 69.97) -- (118.38, 69.97);
\definecolor{drawColor}{RGB}{78,155,133}

\path[draw=drawColor,draw opacity=0.20,line width= 0.4pt,line join=round,line cap=round,fill=fillColor,fill opacity=0.20] (135.33, 36.57) circle (  1.96);

\path[draw=drawColor,draw opacity=0.20,line width= 0.4pt,line join=round,line cap=round,fill=fillColor,fill opacity=0.20] (135.33, 90.41) circle (  1.96);

\path[draw=drawColor,draw opacity=0.20,line width= 0.4pt,line join=round,line cap=round,fill=fillColor,fill opacity=0.20] (135.33, 25.16) circle (  1.96);

\path[draw=drawColor,draw opacity=0.20,line width= 0.4pt,line join=round,line cap=round,fill=fillColor,fill opacity=0.20] (135.33,102.30) circle (  1.96);
\definecolor{drawColor}{RGB}{78,155,133}

\path[draw=drawColor,line width= 0.6pt,line join=round] (135.33, 73.89) -- (135.33, 82.58);

\path[draw=drawColor,line width= 0.6pt,line join=round] (135.33, 66.56) -- (135.33, 62.68);

\path[draw=drawColor,line width= 0.6pt,fill=fillColor,fill opacity=0.20] (125.16, 73.89) --
	(125.16, 66.56) --
	(145.50, 66.56) --
	(145.50, 73.89) --
	(125.16, 73.89) --
	cycle;

\path[draw=drawColor,line width= 1.1pt] (125.16, 70.35) -- (145.50, 70.35);
\definecolor{drawColor}{RGB}{78,155,133}

\path[draw=drawColor,draw opacity=0.20,line width= 0.4pt,line join=round,line cap=round,fill=fillColor,fill opacity=0.20] (162.46, 95.86) circle (  1.96);

\path[draw=drawColor,draw opacity=0.20,line width= 0.4pt,line join=round,line cap=round,fill=fillColor,fill opacity=0.20] (162.46, 54.12) circle (  1.96);

\path[draw=drawColor,draw opacity=0.20,line width= 0.4pt,line join=round,line cap=round,fill=fillColor,fill opacity=0.20] (162.46, 10.81) circle (  1.96);

\path[draw=drawColor,draw opacity=0.20,line width= 0.4pt,line join=round,line cap=round,fill=fillColor,fill opacity=0.20] (162.46,112.65) circle (  1.96);

\path[draw=drawColor,draw opacity=0.20,line width= 0.4pt,line join=round,line cap=round,fill=fillColor,fill opacity=0.20] (162.46, 89.41) circle (  1.96);

\path[draw=drawColor,draw opacity=0.20,line width= 0.4pt,line join=round,line cap=round,fill=fillColor,fill opacity=0.20] (162.46, 14.66) circle (  1.96);

\path[draw=drawColor,draw opacity=0.20,line width= 0.4pt,line join=round,line cap=round,fill=fillColor,fill opacity=0.20] (162.46, 42.13) circle (  1.96);
\definecolor{drawColor}{RGB}{78,155,133}

\path[draw=drawColor,line width= 0.6pt,line join=round] (162.46, 75.02) -- (162.46, 79.11);

\path[draw=drawColor,line width= 0.6pt,line join=round] (162.46, 66.93) -- (162.46, 58.99);

\path[draw=drawColor,line width= 0.6pt,fill=fillColor,fill opacity=0.20] (152.29, 75.02) --
	(152.29, 66.93) --
	(172.63, 66.93) --
	(172.63, 75.02) --
	(152.29, 75.02) --
	cycle;

\path[draw=drawColor,line width= 1.1pt] (152.29, 70.17) -- (172.63, 70.17);
\definecolor{drawColor}{RGB}{78,155,133}

\path[draw=drawColor,draw opacity=0.20,line width= 0.4pt,line join=round,line cap=round,fill=fillColor,fill opacity=0.20] (189.59,100.70) circle (  1.96);

\path[draw=drawColor,draw opacity=0.20,line width= 0.4pt,line join=round,line cap=round,fill=fillColor,fill opacity=0.20] (189.59, 51.16) circle (  1.96);

\path[draw=drawColor,draw opacity=0.20,line width= 0.4pt,line join=round,line cap=round,fill=fillColor,fill opacity=0.20] (189.59, 11.64) circle (  1.96);

\path[draw=drawColor,draw opacity=0.20,line width= 0.4pt,line join=round,line cap=round,fill=fillColor,fill opacity=0.20] (189.59,112.05) circle (  1.96);

\path[draw=drawColor,draw opacity=0.20,line width= 0.4pt,line join=round,line cap=round,fill=fillColor,fill opacity=0.20] (189.59,114.74) circle (  1.96);

\path[draw=drawColor,draw opacity=0.20,line width= 0.4pt,line join=round,line cap=round,fill=fillColor,fill opacity=0.20] (189.59,107.25) circle (  1.96);

\path[draw=drawColor,draw opacity=0.20,line width= 0.4pt,line join=round,line cap=round,fill=fillColor,fill opacity=0.20] (189.59, 25.19) circle (  1.96);

\path[draw=drawColor,draw opacity=0.20,line width= 0.4pt,line join=round,line cap=round,fill=fillColor,fill opacity=0.20] (189.59,-59.58) circle (  1.96);

\path[draw=drawColor,draw opacity=0.20,line width= 0.4pt,line join=round,line cap=round,fill=fillColor,fill opacity=0.20] (189.59,-58.74) circle (  1.96);
\definecolor{drawColor}{RGB}{78,155,133}

\path[draw=drawColor,line width= 0.6pt,line join=round] (189.59, 78.08) -- (189.59, 86.61);

\path[draw=drawColor,line width= 0.6pt,line join=round] (189.59, 67.41) -- (189.59, 57.88);

\path[draw=drawColor,line width= 0.6pt,fill=fillColor,fill opacity=0.20] (179.41, 78.08) --
	(179.41, 67.41) --
	(199.76, 67.41) --
	(199.76, 78.08) --
	(179.41, 78.08) --
	cycle;

\path[draw=drawColor,line width= 1.1pt] (179.41, 69.97) -- (199.76, 69.97);
\definecolor{drawColor}{RGB}{78,155,133}

\path[draw=drawColor,draw opacity=0.20,line width= 0.4pt,line join=round,line cap=round,fill=fillColor,fill opacity=0.20] (216.71, -3.64) circle (  1.96);

\path[draw=drawColor,draw opacity=0.20,line width= 0.4pt,line join=round,line cap=round,fill=fillColor,fill opacity=0.20] (216.71,123.08) circle (  1.96);

\path[draw=drawColor,draw opacity=0.20,line width= 0.4pt,line join=round,line cap=round,fill=fillColor,fill opacity=0.20] (216.71,167.21) circle (  1.96);
\definecolor{drawColor}{RGB}{78,155,133}

\path[draw=drawColor,line width= 0.6pt,line join=round] (216.71, 80.58) -- (216.71,100.01);

\path[draw=drawColor,line width= 0.6pt,line join=round] (216.71, 66.84) -- (216.71, 51.58);

\path[draw=drawColor,line width= 0.6pt,fill=fillColor,fill opacity=0.20] (206.54, 80.58) --
	(206.54, 66.84) --
	(226.89, 66.84) --
	(226.89, 80.58) --
	(206.54, 80.58) --
	cycle;

\path[draw=drawColor,line width= 1.1pt] (206.54, 70.40) -- (226.89, 70.40);
\end{scope}
\begin{scope}
\path[clip] (  0.00,  0.00) rectangle (238.49,115.63);
\definecolor{drawColor}{RGB}{0,0,0}

\path[draw=drawColor,line width= 0.6pt,line join=round] ( 37.67, 29.80) --
	( 37.67,110.13);

\path[draw=drawColor,line width= 0.6pt,line join=round] ( 39.09,107.67) --
	( 37.67,110.13) --
	( 36.25,107.67);
\end{scope}
\begin{scope}
\path[clip] (  0.00,  0.00) rectangle (238.49,115.63);
\definecolor{drawColor}{gray}{0.30}

\node[text=drawColor,anchor=base east,inner sep=0pt, outer sep=0pt, scale=  0.88] at ( 32.72, 42.59) {0.98};

\node[text=drawColor,anchor=base east,inner sep=0pt, outer sep=0pt, scale=  0.88] at ( 32.72, 66.94) {1.00};

\node[text=drawColor,anchor=base east,inner sep=0pt, outer sep=0pt, scale=  0.88] at ( 32.72, 91.28) {1.02};
\end{scope}
\begin{scope}
\path[clip] (  0.00,  0.00) rectangle (238.49,115.63);
\definecolor{drawColor}{gray}{0.20}

\path[draw=drawColor,line width= 0.6pt,line join=round] ( 34.92, 45.63) --
	( 37.67, 45.63);

\path[draw=drawColor,line width= 0.6pt,line join=round] ( 34.92, 69.97) --
	( 37.67, 69.97);

\path[draw=drawColor,line width= 0.6pt,line join=round] ( 34.92, 94.31) --
	( 37.67, 94.31);
\end{scope}
\begin{scope}
\path[clip] (  0.00,  0.00) rectangle (238.49,115.63);
\definecolor{drawColor}{RGB}{0,0,0}

\path[draw=drawColor,line width= 0.6pt,line join=round] ( 37.67, 29.80) --
	(232.99, 29.80);

\path[draw=drawColor,line width= 0.6pt,line join=round] (230.53, 28.38) --
	(232.99, 29.80) --
	(230.53, 31.23);
\end{scope}
\begin{scope}
\path[clip] (  0.00,  0.00) rectangle (238.49,115.63);
\definecolor{drawColor}{gray}{0.20}

\path[draw=drawColor,line width= 0.6pt,line join=round] ( 53.95, 27.05) --
	( 53.95, 29.80);

\path[draw=drawColor,line width= 0.6pt,line join=round] ( 81.08, 27.05) --
	( 81.08, 29.80);

\path[draw=drawColor,line width= 0.6pt,line join=round] (108.20, 27.05) --
	(108.20, 29.80);

\path[draw=drawColor,line width= 0.6pt,line join=round] (135.33, 27.05) --
	(135.33, 29.80);

\path[draw=drawColor,line width= 0.6pt,line join=round] (162.46, 27.05) --
	(162.46, 29.80);

\path[draw=drawColor,line width= 0.6pt,line join=round] (189.59, 27.05) --
	(189.59, 29.80);

\path[draw=drawColor,line width= 0.6pt,line join=round] (216.71, 27.05) --
	(216.71, 29.80);
\end{scope}
\begin{scope}
\path[clip] (  0.00,  0.00) rectangle (238.49,115.63);
\definecolor{drawColor}{gray}{0.30}

\node[text=drawColor,anchor=base,inner sep=0pt, outer sep=0pt, scale=  0.88] at ( 53.95, 18.79) {2};

\node[text=drawColor,anchor=base,inner sep=0pt, outer sep=0pt, scale=  0.88] at ( 81.08, 18.79) {4};

\node[text=drawColor,anchor=base,inner sep=0pt, outer sep=0pt, scale=  0.88] at (108.20, 18.79) {8};

\node[text=drawColor,anchor=base,inner sep=0pt, outer sep=0pt, scale=  0.88] at (135.33, 18.79) {16};

\node[text=drawColor,anchor=base,inner sep=0pt, outer sep=0pt, scale=  0.88] at (162.46, 18.79) {32};

\node[text=drawColor,anchor=base,inner sep=0pt, outer sep=0pt, scale=  0.88] at (189.59, 18.79) {64};

\node[text=drawColor,anchor=base,inner sep=0pt, outer sep=0pt, scale=  0.88] at (216.71, 18.79) {128};
\end{scope}
\begin{scope}
\path[clip] (  0.00,  0.00) rectangle (238.49,115.63);
\definecolor{drawColor}{RGB}{0,0,0}

\node[text=drawColor,anchor=base,inner sep=0pt, outer sep=0pt, scale=  1.00] at (135.33,  7.44) {domain size};
\end{scope}
\begin{scope}
\path[clip] (  0.00,  0.00) rectangle (238.49,115.63);
\definecolor{drawColor}{RGB}{0,0,0}

\node[text=drawColor,rotate= 90.00,anchor=base,inner sep=0pt, outer sep=0pt, scale=  1.00] at ( 12.39, 69.97) {$p' \mathbin{/} p$};
\end{scope}
\end{tikzpicture}

%% file: files/plot-quots-p=0.7-eps=0.001.tex
\begin{tikzpicture}[x=1pt,y=1pt]
\definecolor{fillColor}{RGB}{255,255,255}
\path[use as bounding box,fill=fillColor,fill opacity=0.00] (0,0) rectangle (238.49,115.63);
\begin{scope}
\path[clip] (  0.00,  0.00) rectangle (238.49,115.63);
\definecolor{drawColor}{RGB}{255,255,255}
\definecolor{fillColor}{RGB}{255,255,255}

\path[draw=drawColor,line width= 0.6pt,line join=round,line cap=round,fill=fillColor] (  0.00,  0.00) rectangle (238.49,115.63);
\end{scope}
\begin{scope}
\path[clip] ( 46.47, 29.80) rectangle (232.99,110.13);
\definecolor{fillColor}{RGB}{255,255,255}

\path[fill=fillColor] ( 46.47, 29.80) rectangle (232.99,110.13);
\definecolor{drawColor}{RGB}{78,155,133}
\definecolor{fillColor}{RGB}{78,155,133}

\path[draw=drawColor,draw opacity=0.20,line width= 0.4pt,line join=round,line cap=round,fill=fillColor,fill opacity=0.20] ( 62.01, 47.04) circle (  1.96);

\path[draw=drawColor,draw opacity=0.20,line width= 0.4pt,line join=round,line cap=round,fill=fillColor,fill opacity=0.20] ( 62.01, 90.12) circle (  1.96);

\path[draw=drawColor,draw opacity=0.20,line width= 0.4pt,line join=round,line cap=round,fill=fillColor,fill opacity=0.20] ( 62.01, 72.55) circle (  1.96);

\path[draw=drawColor,draw opacity=0.20,line width= 0.4pt,line join=round,line cap=round,fill=fillColor,fill opacity=0.20] ( 62.01, 66.96) circle (  1.96);

\path[draw=drawColor,draw opacity=0.20,line width= 0.4pt,line join=round,line cap=round,fill=fillColor,fill opacity=0.20] ( 62.01, 53.02) circle (  1.96);

\path[draw=drawColor,draw opacity=0.20,line width= 0.4pt,line join=round,line cap=round,fill=fillColor,fill opacity=0.20] ( 62.01, 85.15) circle (  1.96);

\path[draw=drawColor,draw opacity=0.20,line width= 0.4pt,line join=round,line cap=round,fill=fillColor,fill opacity=0.20] ( 62.01, 66.81) circle (  1.96);

\path[draw=drawColor,draw opacity=0.20,line width= 0.4pt,line join=round,line cap=round,fill=fillColor,fill opacity=0.20] ( 62.01, 73.25) circle (  1.96);

\path[draw=drawColor,draw opacity=0.20,line width= 0.4pt,line join=round,line cap=round,fill=fillColor,fill opacity=0.20] ( 62.01, 70.29) circle (  1.96);

\path[draw=drawColor,draw opacity=0.20,line width= 0.4pt,line join=round,line cap=round,fill=fillColor,fill opacity=0.20] ( 62.01, 69.64) circle (  1.96);

\path[draw=drawColor,draw opacity=0.20,line width= 0.4pt,line join=round,line cap=round,fill=fillColor,fill opacity=0.20] ( 62.01, 42.91) circle (  1.96);

\path[draw=drawColor,draw opacity=0.20,line width= 0.4pt,line join=round,line cap=round,fill=fillColor,fill opacity=0.20] ( 62.01,105.12) circle (  1.96);
\definecolor{drawColor}{RGB}{78,155,133}

\path[draw=drawColor,line width= 0.6pt,line join=round] ( 62.01, 69.98) -- ( 62.01, 69.98);

\path[draw=drawColor,line width= 0.6pt,line join=round] ( 62.01, 69.96) -- ( 62.01, 69.96);

\path[draw=drawColor,line width= 0.6pt,fill=fillColor,fill opacity=0.20] ( 52.30, 69.98) --
	( 52.30, 69.96) --
	( 71.73, 69.96) --
	( 71.73, 69.98) --
	( 52.30, 69.98) --
	cycle;

\path[draw=drawColor,line width= 1.1pt] ( 52.30, 69.97) -- ( 71.73, 69.97);
\definecolor{drawColor}{RGB}{78,155,133}

\path[draw=drawColor,draw opacity=0.20,line width= 0.4pt,line join=round,line cap=round,fill=fillColor,fill opacity=0.20] ( 87.92, 32.00) circle (  1.96);

\path[draw=drawColor,draw opacity=0.20,line width= 0.4pt,line join=round,line cap=round,fill=fillColor,fill opacity=0.20] ( 87.92, 96.67) circle (  1.96);

\path[draw=drawColor,draw opacity=0.20,line width= 0.4pt,line join=round,line cap=round,fill=fillColor,fill opacity=0.20] ( 87.92, 87.34) circle (  1.96);

\path[draw=drawColor,draw opacity=0.20,line width= 0.4pt,line join=round,line cap=round,fill=fillColor,fill opacity=0.20] ( 87.92, 49.52) circle (  1.96);

\path[draw=drawColor,draw opacity=0.20,line width= 0.4pt,line join=round,line cap=round,fill=fillColor,fill opacity=0.20] ( 87.92, 37.59) circle (  1.96);

\path[draw=drawColor,draw opacity=0.20,line width= 0.4pt,line join=round,line cap=round,fill=fillColor,fill opacity=0.20] ( 87.92,114.27) circle (  1.96);

\path[draw=drawColor,draw opacity=0.20,line width= 0.4pt,line join=round,line cap=round,fill=fillColor,fill opacity=0.20] ( 87.92, 54.83) circle (  1.96);

\path[draw=drawColor,draw opacity=0.20,line width= 0.4pt,line join=round,line cap=round,fill=fillColor,fill opacity=0.20] ( 87.92, 93.51) circle (  1.96);

\path[draw=drawColor,draw opacity=0.20,line width= 0.4pt,line join=round,line cap=round,fill=fillColor,fill opacity=0.20] ( 87.92, 39.80) circle (  1.96);
\definecolor{drawColor}{RGB}{78,155,133}

\path[draw=drawColor,line width= 0.6pt,line join=round] ( 87.92, 72.85) -- ( 87.92, 79.64);

\path[draw=drawColor,line width= 0.6pt,line join=round] ( 87.92, 67.85) -- ( 87.92, 65.43);

\path[draw=drawColor,line width= 0.6pt,fill=fillColor,fill opacity=0.20] ( 78.20, 72.85) --
	( 78.20, 67.85) --
	( 97.63, 67.85) --
	( 97.63, 72.85) --
	( 78.20, 72.85) --
	cycle;

\path[draw=drawColor,line width= 1.1pt] ( 78.20, 69.97) -- ( 97.63, 69.97);
\definecolor{drawColor}{RGB}{78,155,133}

\path[draw=drawColor,draw opacity=0.20,line width= 0.4pt,line join=round,line cap=round,fill=fillColor,fill opacity=0.20] (113.82, 16.65) circle (  1.96);

\path[draw=drawColor,draw opacity=0.20,line width= 0.4pt,line join=round,line cap=round,fill=fillColor,fill opacity=0.20] (113.82, 22.57) circle (  1.96);

\path[draw=drawColor,draw opacity=0.20,line width= 0.4pt,line join=round,line cap=round,fill=fillColor,fill opacity=0.20] (113.82,120.22) circle (  1.96);
\definecolor{drawColor}{RGB}{78,155,133}

\path[draw=drawColor,line width= 0.6pt,line join=round] (113.82, 79.85) -- (113.82,103.27);

\path[draw=drawColor,line width= 0.6pt,line join=round] (113.82, 58.42) -- (113.82, 33.47);

\path[draw=drawColor,line width= 0.6pt,fill=fillColor,fill opacity=0.20] (104.11, 79.85) --
	(104.11, 58.42) --
	(123.54, 58.42) --
	(123.54, 79.85) --
	(104.11, 79.85) --
	cycle;

\path[draw=drawColor,line width= 1.1pt] (104.11, 69.97) -- (123.54, 69.97);
\definecolor{drawColor}{RGB}{78,155,133}

\path[draw=drawColor,draw opacity=0.20,line width= 0.4pt,line join=round,line cap=round,fill=fillColor,fill opacity=0.20] (139.73,  3.99) circle (  1.96);

\path[draw=drawColor,draw opacity=0.20,line width= 0.4pt,line join=round,line cap=round,fill=fillColor,fill opacity=0.20] (139.73,110.62) circle (  1.96);

\path[draw=drawColor,draw opacity=0.20,line width= 0.4pt,line join=round,line cap=round,fill=fillColor,fill opacity=0.20] (139.73, 15.30) circle (  1.96);
\definecolor{drawColor}{RGB}{78,155,133}

\path[draw=drawColor,line width= 0.6pt,line join=round] (139.73, 79.00) -- (139.73,104.24);

\path[draw=drawColor,line width= 0.6pt,line join=round] (139.73, 61.91) -- (139.73, 41.46);

\path[draw=drawColor,line width= 0.6pt,fill=fillColor,fill opacity=0.20] (130.02, 79.00) --
	(130.02, 61.91) --
	(149.45, 61.91) --
	(149.45, 79.00) --
	(130.02, 79.00) --
	cycle;

\path[draw=drawColor,line width= 1.1pt] (130.02, 69.97) -- (149.45, 69.97);
\definecolor{drawColor}{RGB}{78,155,133}

\path[draw=drawColor,draw opacity=0.20,line width= 0.4pt,line join=round,line cap=round,fill=fillColor,fill opacity=0.20] (165.64,  7.06) circle (  1.96);

\path[draw=drawColor,draw opacity=0.20,line width= 0.4pt,line join=round,line cap=round,fill=fillColor,fill opacity=0.20] (165.64,108.72) circle (  1.96);

\path[draw=drawColor,draw opacity=0.20,line width= 0.4pt,line join=round,line cap=round,fill=fillColor,fill opacity=0.20] (165.64, 11.10) circle (  1.96);

\path[draw=drawColor,draw opacity=0.20,line width= 0.4pt,line join=round,line cap=round,fill=fillColor,fill opacity=0.20] (165.64,106.87) circle (  1.96);

\path[draw=drawColor,draw opacity=0.20,line width= 0.4pt,line join=round,line cap=round,fill=fillColor,fill opacity=0.20] (165.64, 90.24) circle (  1.96);

\path[draw=drawColor,draw opacity=0.20,line width= 0.4pt,line join=round,line cap=round,fill=fillColor,fill opacity=0.20] (165.64, 45.62) circle (  1.96);

\path[draw=drawColor,draw opacity=0.20,line width= 0.4pt,line join=round,line cap=round,fill=fillColor,fill opacity=0.20] (165.64, 36.71) circle (  1.96);

\path[draw=drawColor,draw opacity=0.20,line width= 0.4pt,line join=round,line cap=round,fill=fillColor,fill opacity=0.20] (165.64,115.76) circle (  1.96);
\definecolor{drawColor}{RGB}{78,155,133}

\path[draw=drawColor,line width= 0.6pt,line join=round] (165.64, 73.61) -- (165.64, 78.39);

\path[draw=drawColor,line width= 0.6pt,line join=round] (165.64, 66.93) -- (165.64, 59.85);

\path[draw=drawColor,line width= 0.6pt,fill=fillColor,fill opacity=0.20] (155.92, 73.61) --
	(155.92, 66.93) --
	(175.35, 66.93) --
	(175.35, 73.61) --
	(155.92, 73.61) --
	cycle;

\path[draw=drawColor,line width= 1.1pt] (155.92, 69.97) -- (175.35, 69.97);
\definecolor{drawColor}{RGB}{78,155,133}

\path[draw=drawColor,draw opacity=0.20,line width= 0.4pt,line join=round,line cap=round,fill=fillColor,fill opacity=0.20] (191.54, 43.11) circle (  1.96);

\path[draw=drawColor,draw opacity=0.20,line width= 0.4pt,line join=round,line cap=round,fill=fillColor,fill opacity=0.20] (191.54,  5.40) circle (  1.96);

\path[draw=drawColor,draw opacity=0.20,line width= 0.4pt,line join=round,line cap=round,fill=fillColor,fill opacity=0.20] (191.54,110.44) circle (  1.96);

\path[draw=drawColor,draw opacity=0.20,line width= 0.4pt,line join=round,line cap=round,fill=fillColor,fill opacity=0.20] (191.54, 98.76) circle (  1.96);

\path[draw=drawColor,draw opacity=0.20,line width= 0.4pt,line join=round,line cap=round,fill=fillColor,fill opacity=0.20] (191.54, 43.17) circle (  1.96);

\path[draw=drawColor,draw opacity=0.20,line width= 0.4pt,line join=round,line cap=round,fill=fillColor,fill opacity=0.20] (191.54,106.86) circle (  1.96);
\definecolor{drawColor}{RGB}{78,155,133}

\path[draw=drawColor,line width= 0.6pt,line join=round] (191.54, 76.54) -- (191.54, 90.23);

\path[draw=drawColor,line width= 0.6pt,line join=round] (191.54, 64.05) -- (191.54, 46.00);

\path[draw=drawColor,line width= 0.6pt,fill=fillColor,fill opacity=0.20] (181.83, 76.54) --
	(181.83, 64.05) --
	(201.26, 64.05) --
	(201.26, 76.54) --
	(181.83, 76.54) --
	cycle;

\path[draw=drawColor,line width= 1.1pt] (181.83, 69.97) -- (201.26, 69.97);
\definecolor{drawColor}{RGB}{78,155,133}

\path[draw=drawColor,draw opacity=0.20,line width= 0.4pt,line join=round,line cap=round,fill=fillColor,fill opacity=0.20] (217.45,  3.99) circle (  1.96);

\path[draw=drawColor,draw opacity=0.20,line width= 0.4pt,line join=round,line cap=round,fill=fillColor,fill opacity=0.20] (217.45,110.62) circle (  1.96);

\path[draw=drawColor,draw opacity=0.20,line width= 0.4pt,line join=round,line cap=round,fill=fillColor,fill opacity=0.20] (217.45,  0.14) circle (  1.96);

\path[draw=drawColor,draw opacity=0.20,line width= 0.4pt,line join=round,line cap=round,fill=fillColor,fill opacity=0.20] (217.45,113.74) circle (  1.96);

\path[draw=drawColor,draw opacity=0.20,line width= 0.4pt,line join=round,line cap=round,fill=fillColor,fill opacity=0.20] (217.45, 36.32) circle (  1.96);
\definecolor{drawColor}{RGB}{78,155,133}

\path[draw=drawColor,line width= 0.6pt,line join=round] (217.45, 76.40) -- (217.45, 94.40);

\path[draw=drawColor,line width= 0.6pt,line join=round] (217.45, 63.81) -- (217.45, 59.62);

\path[draw=drawColor,line width= 0.6pt,fill=fillColor,fill opacity=0.20] (207.73, 76.40) --
	(207.73, 63.81) --
	(227.16, 63.81) --
	(227.16, 76.40) --
	(207.73, 76.40) --
	cycle;

\path[draw=drawColor,line width= 1.1pt] (207.73, 69.99) -- (227.16, 69.99);
\end{scope}
\begin{scope}
\path[clip] (  0.00,  0.00) rectangle (238.49,115.63);
\definecolor{drawColor}{RGB}{0,0,0}

\path[draw=drawColor,line width= 0.6pt,line join=round] ( 46.47, 29.80) --
	( 46.47,110.13);

\path[draw=drawColor,line width= 0.6pt,line join=round] ( 47.89,107.67) --
	( 46.47,110.13) --
	( 45.05,107.67);
\end{scope}
\begin{scope}
\path[clip] (  0.00,  0.00) rectangle (238.49,115.63);
\definecolor{drawColor}{gray}{0.30}

\node[text=drawColor,anchor=base east,inner sep=0pt, outer sep=0pt, scale=  0.88] at ( 41.52, 42.59) {0.9998};

\node[text=drawColor,anchor=base east,inner sep=0pt, outer sep=0pt, scale=  0.88] at ( 41.52, 66.94) {1.0000};

\node[text=drawColor,anchor=base east,inner sep=0pt, outer sep=0pt, scale=  0.88] at ( 41.52, 91.28) {1.0002};
\end{scope}
\begin{scope}
\path[clip] (  0.00,  0.00) rectangle (238.49,115.63);
\definecolor{drawColor}{gray}{0.20}

\path[draw=drawColor,line width= 0.6pt,line join=round] ( 43.72, 45.63) --
	( 46.47, 45.63);

\path[draw=drawColor,line width= 0.6pt,line join=round] ( 43.72, 69.97) --
	( 46.47, 69.97);

\path[draw=drawColor,line width= 0.6pt,line join=round] ( 43.72, 94.31) --
	( 46.47, 94.31);
\end{scope}
\begin{scope}
\path[clip] (  0.00,  0.00) rectangle (238.49,115.63);
\definecolor{drawColor}{RGB}{0,0,0}

\path[draw=drawColor,line width= 0.6pt,line join=round] ( 46.47, 29.80) --
	(232.99, 29.80);

\path[draw=drawColor,line width= 0.6pt,line join=round] (230.53, 28.38) --
	(232.99, 29.80) --
	(230.53, 31.23);
\end{scope}
\begin{scope}
\path[clip] (  0.00,  0.00) rectangle (238.49,115.63);
\definecolor{drawColor}{gray}{0.20}

\path[draw=drawColor,line width= 0.6pt,line join=round] ( 62.01, 27.05) --
	( 62.01, 29.80);

\path[draw=drawColor,line width= 0.6pt,line join=round] ( 87.92, 27.05) --
	( 87.92, 29.80);

\path[draw=drawColor,line width= 0.6pt,line join=round] (113.82, 27.05) --
	(113.82, 29.80);

\path[draw=drawColor,line width= 0.6pt,line join=round] (139.73, 27.05) --
	(139.73, 29.80);

\path[draw=drawColor,line width= 0.6pt,line join=round] (165.64, 27.05) --
	(165.64, 29.80);

\path[draw=drawColor,line width= 0.6pt,line join=round] (191.54, 27.05) --
	(191.54, 29.80);

\path[draw=drawColor,line width= 0.6pt,line join=round] (217.45, 27.05) --
	(217.45, 29.80);
\end{scope}
\begin{scope}
\path[clip] (  0.00,  0.00) rectangle (238.49,115.63);
\definecolor{drawColor}{gray}{0.30}

\node[text=drawColor,anchor=base,inner sep=0pt, outer sep=0pt, scale=  0.88] at ( 62.01, 18.79) {2};

\node[text=drawColor,anchor=base,inner sep=0pt, outer sep=0pt, scale=  0.88] at ( 87.92, 18.79) {4};

\node[text=drawColor,anchor=base,inner sep=0pt, outer sep=0pt, scale=  0.88] at (113.82, 18.79) {8};

\node[text=drawColor,anchor=base,inner sep=0pt, outer sep=0pt, scale=  0.88] at (139.73, 18.79) {16};

\node[text=drawColor,anchor=base,inner sep=0pt, outer sep=0pt, scale=  0.88] at (165.64, 18.79) {32};

\node[text=drawColor,anchor=base,inner sep=0pt, outer sep=0pt, scale=  0.88] at (191.54, 18.79) {64};

\node[text=drawColor,anchor=base,inner sep=0pt, outer sep=0pt, scale=  0.88] at (217.45, 18.79) {128};
\end{scope}
\begin{scope}
\path[clip] (  0.00,  0.00) rectangle (238.49,115.63);
\definecolor{drawColor}{RGB}{0,0,0}

\node[text=drawColor,anchor=base,inner sep=0pt, outer sep=0pt, scale=  1.00] at (139.73,  7.44) {domain size};
\end{scope}
\begin{scope}
\path[clip] (  0.00,  0.00) rectangle (238.49,115.63);
\definecolor{drawColor}{RGB}{0,0,0}

\node[text=drawColor,rotate= 90.00,anchor=base,inner sep=0pt, outer sep=0pt, scale=  1.00] at ( 12.39, 69.97) {$p' \mathbin{/} p$};
\end{scope}
\end{tikzpicture}

%% file: files/plot-quots-p=0.7-eps=0.1.tex
\begin{tikzpicture}[x=1pt,y=1pt]
\definecolor{fillColor}{RGB}{255,255,255}
\path[use as bounding box,fill=fillColor,fill opacity=0.00] (0,0) rectangle (238.49,115.63);
\begin{scope}
\path[clip] (  0.00,  0.00) rectangle (238.49,115.63);
\definecolor{drawColor}{RGB}{255,255,255}
\definecolor{fillColor}{RGB}{255,255,255}

\path[draw=drawColor,line width= 0.6pt,line join=round,line cap=round,fill=fillColor] (  0.00,  0.00) rectangle (238.49,115.63);
\end{scope}
\begin{scope}
\path[clip] ( 37.67, 29.80) rectangle (232.99,110.13);
\definecolor{fillColor}{RGB}{255,255,255}

\path[fill=fillColor] ( 37.67, 29.80) rectangle (232.99,110.13);
\definecolor{drawColor}{RGB}{78,155,133}
\definecolor{fillColor}{RGB}{78,155,133}

\path[draw=drawColor,draw opacity=0.20,line width= 0.4pt,line join=round,line cap=round,fill=fillColor,fill opacity=0.20] ( 53.95, 48.33) circle (  1.96);

\path[draw=drawColor,draw opacity=0.20,line width= 0.4pt,line join=round,line cap=round,fill=fillColor,fill opacity=0.20] ( 53.95, 90.42) circle (  1.96);

\path[draw=drawColor,draw opacity=0.20,line width= 0.4pt,line join=round,line cap=round,fill=fillColor,fill opacity=0.20] ( 53.95, 70.43) circle (  1.96);

\path[draw=drawColor,draw opacity=0.20,line width= 0.4pt,line join=round,line cap=round,fill=fillColor,fill opacity=0.20] ( 53.95, 69.43) circle (  1.96);

\path[draw=drawColor,draw opacity=0.20,line width= 0.4pt,line join=round,line cap=round,fill=fillColor,fill opacity=0.20] ( 53.95, 55.46) circle (  1.96);

\path[draw=drawColor,draw opacity=0.20,line width= 0.4pt,line join=round,line cap=round,fill=fillColor,fill opacity=0.20] ( 53.95, 83.95) circle (  1.96);

\path[draw=drawColor,draw opacity=0.20,line width= 0.4pt,line join=round,line cap=round,fill=fillColor,fill opacity=0.20] ( 53.95, 68.60) circle (  1.96);

\path[draw=drawColor,draw opacity=0.20,line width= 0.4pt,line join=round,line cap=round,fill=fillColor,fill opacity=0.20] ( 53.95, 71.40) circle (  1.96);

\path[draw=drawColor,draw opacity=0.20,line width= 0.4pt,line join=round,line cap=round,fill=fillColor,fill opacity=0.20] ( 53.95, 70.82) circle (  1.96);

\path[draw=drawColor,draw opacity=0.20,line width= 0.4pt,line join=round,line cap=round,fill=fillColor,fill opacity=0.20] ( 53.95, 48.85) circle (  1.96);

\path[draw=drawColor,draw opacity=0.20,line width= 0.4pt,line join=round,line cap=round,fill=fillColor,fill opacity=0.20] ( 53.95, 97.47) circle (  1.96);
\definecolor{drawColor}{RGB}{78,155,133}

\path[draw=drawColor,line width= 0.6pt,line join=round] ( 53.95, 70.06) -- ( 53.95, 70.08);

\path[draw=drawColor,line width= 0.6pt,line join=round] ( 53.95, 69.83) -- ( 53.95, 69.60);

\path[draw=drawColor,line width= 0.6pt,fill=fillColor,fill opacity=0.20] ( 43.78, 70.06) --
	( 43.78, 69.83) --
	( 64.12, 69.83) --
	( 64.12, 70.06) --
	( 43.78, 70.06) --
	cycle;

\path[draw=drawColor,line width= 1.1pt] ( 43.78, 69.97) -- ( 64.12, 69.97);
\definecolor{drawColor}{RGB}{78,155,133}

\path[draw=drawColor,draw opacity=0.20,line width= 0.4pt,line join=round,line cap=round,fill=fillColor,fill opacity=0.20] ( 81.08, 35.37) circle (  1.96);

\path[draw=drawColor,draw opacity=0.20,line width= 0.4pt,line join=round,line cap=round,fill=fillColor,fill opacity=0.20] ( 81.08, 96.99) circle (  1.96);

\path[draw=drawColor,draw opacity=0.20,line width= 0.4pt,line join=round,line cap=round,fill=fillColor,fill opacity=0.20] ( 81.08, 85.60) circle (  1.96);

\path[draw=drawColor,draw opacity=0.20,line width= 0.4pt,line join=round,line cap=round,fill=fillColor,fill opacity=0.20] ( 81.08, 51.55) circle (  1.96);

\path[draw=drawColor,draw opacity=0.20,line width= 0.4pt,line join=round,line cap=round,fill=fillColor,fill opacity=0.20] ( 81.08, 44.27) circle (  1.96);

\path[draw=drawColor,draw opacity=0.20,line width= 0.4pt,line join=round,line cap=round,fill=fillColor,fill opacity=0.20] ( 81.08,105.13) circle (  1.96);

\path[draw=drawColor,draw opacity=0.20,line width= 0.4pt,line join=round,line cap=round,fill=fillColor,fill opacity=0.20] ( 81.08, 56.89) circle (  1.96);

\path[draw=drawColor,draw opacity=0.20,line width= 0.4pt,line join=round,line cap=round,fill=fillColor,fill opacity=0.20] ( 81.08, 88.61) circle (  1.96);

\path[draw=drawColor,draw opacity=0.20,line width= 0.4pt,line join=round,line cap=round,fill=fillColor,fill opacity=0.20] ( 81.08, 46.37) circle (  1.96);
\definecolor{drawColor}{RGB}{78,155,133}

\path[draw=drawColor,line width= 0.6pt,line join=round] ( 81.08, 73.69) -- ( 81.08, 78.01);

\path[draw=drawColor,line width= 0.6pt,line join=round] ( 81.08, 68.58) -- ( 81.08, 67.60);

\path[draw=drawColor,line width= 0.6pt,fill=fillColor,fill opacity=0.20] ( 70.90, 73.69) --
	( 70.90, 68.58) --
	( 91.25, 68.58) --
	( 91.25, 73.69) --
	( 70.90, 73.69) --
	cycle;

\path[draw=drawColor,line width= 1.1pt] ( 70.90, 69.97) -- ( 91.25, 69.97);
\definecolor{drawColor}{RGB}{78,155,133}

\path[draw=drawColor,draw opacity=0.20,line width= 0.4pt,line join=round,line cap=round,fill=fillColor,fill opacity=0.20] (108.20, 23.42) circle (  1.96);

\path[draw=drawColor,draw opacity=0.20,line width= 0.4pt,line join=round,line cap=round,fill=fillColor,fill opacity=0.20] (108.20,102.93) circle (  1.96);

\path[draw=drawColor,draw opacity=0.20,line width= 0.4pt,line join=round,line cap=round,fill=fillColor,fill opacity=0.20] (108.20, 29.63) circle (  1.96);

\path[draw=drawColor,draw opacity=0.20,line width= 0.4pt,line join=round,line cap=round,fill=fillColor,fill opacity=0.20] (108.20, 38.22) circle (  1.96);

\path[draw=drawColor,draw opacity=0.20,line width= 0.4pt,line join=round,line cap=round,fill=fillColor,fill opacity=0.20] (108.20, 41.09) circle (  1.96);

\path[draw=drawColor,draw opacity=0.20,line width= 0.4pt,line join=round,line cap=round,fill=fillColor,fill opacity=0.20] (108.20,109.73) circle (  1.96);
\definecolor{drawColor}{RGB}{78,155,133}

\path[draw=drawColor,line width= 0.6pt,line join=round] (108.20, 78.93) -- (108.20, 99.38);

\path[draw=drawColor,line width= 0.6pt,line join=round] (108.20, 64.52) -- (108.20, 46.19);

\path[draw=drawColor,line width= 0.6pt,fill=fillColor,fill opacity=0.20] ( 98.03, 78.93) --
	( 98.03, 64.52) --
	(118.38, 64.52) --
	(118.38, 78.93) --
	( 98.03, 78.93) --
	cycle;

\path[draw=drawColor,line width= 1.1pt] ( 98.03, 70.12) -- (118.38, 70.12);
\definecolor{drawColor}{RGB}{78,155,133}

\path[draw=drawColor,draw opacity=0.20,line width= 0.4pt,line join=round,line cap=round,fill=fillColor,fill opacity=0.20] (135.33, 12.33) circle (  1.96);

\path[draw=drawColor,draw opacity=0.20,line width= 0.4pt,line join=round,line cap=round,fill=fillColor,fill opacity=0.20] (135.33,110.34) circle (  1.96);

\path[draw=drawColor,draw opacity=0.20,line width= 0.4pt,line join=round,line cap=round,fill=fillColor,fill opacity=0.20] (135.33, 34.41) circle (  1.96);

\path[draw=drawColor,draw opacity=0.20,line width= 0.4pt,line join=round,line cap=round,fill=fillColor,fill opacity=0.20] (135.33, 41.68) circle (  1.96);
\definecolor{drawColor}{RGB}{78,155,133}

\path[draw=drawColor,line width= 0.6pt,line join=round] (135.33, 77.89) -- (135.33, 95.62);

\path[draw=drawColor,line width= 0.6pt,line join=round] (135.33, 63.96) -- (135.33, 43.46);

\path[draw=drawColor,line width= 0.6pt,fill=fillColor,fill opacity=0.20] (125.16, 77.89) --
	(125.16, 63.96) --
	(145.50, 63.96) --
	(145.50, 77.89) --
	(125.16, 77.89) --
	cycle;

\path[draw=drawColor,line width= 1.1pt] (125.16, 70.29) -- (145.50, 70.29);
\definecolor{drawColor}{RGB}{78,155,133}

\path[draw=drawColor,draw opacity=0.20,line width= 0.4pt,line join=round,line cap=round,fill=fillColor,fill opacity=0.20] (162.46, 22.35) circle (  1.96);

\path[draw=drawColor,draw opacity=0.20,line width= 0.4pt,line join=round,line cap=round,fill=fillColor,fill opacity=0.20] (162.46,103.32) circle (  1.96);

\path[draw=drawColor,draw opacity=0.20,line width= 0.4pt,line join=round,line cap=round,fill=fillColor,fill opacity=0.20] (162.46, 20.99) circle (  1.96);

\path[draw=drawColor,draw opacity=0.20,line width= 0.4pt,line join=round,line cap=round,fill=fillColor,fill opacity=0.20] (162.46,105.30) circle (  1.96);

\path[draw=drawColor,draw opacity=0.20,line width= 0.4pt,line join=round,line cap=round,fill=fillColor,fill opacity=0.20] (162.46, 46.50) circle (  1.96);

\path[draw=drawColor,draw opacity=0.20,line width= 0.4pt,line join=round,line cap=round,fill=fillColor,fill opacity=0.20] (162.46, 42.79) circle (  1.96);

\path[draw=drawColor,draw opacity=0.20,line width= 0.4pt,line join=round,line cap=round,fill=fillColor,fill opacity=0.20] (162.46,107.39) circle (  1.96);
\definecolor{drawColor}{RGB}{78,155,133}

\path[draw=drawColor,line width= 0.6pt,line join=round] (162.46, 75.89) -- (162.46, 91.89);

\path[draw=drawColor,line width= 0.6pt,line join=round] (162.46, 64.65) -- (162.46, 50.77);

\path[draw=drawColor,line width= 0.6pt,fill=fillColor,fill opacity=0.20] (152.29, 75.89) --
	(152.29, 64.65) --
	(172.63, 64.65) --
	(172.63, 75.89) --
	(152.29, 75.89) --
	cycle;

\path[draw=drawColor,line width= 1.1pt] (152.29, 70.01) -- (172.63, 70.01);
\definecolor{drawColor}{RGB}{78,155,133}

\path[draw=drawColor,draw opacity=0.20,line width= 0.4pt,line join=round,line cap=round,fill=fillColor,fill opacity=0.20] (189.59, 20.85) circle (  1.96);

\path[draw=drawColor,draw opacity=0.20,line width= 0.4pt,line join=round,line cap=round,fill=fillColor,fill opacity=0.20] (189.59, 19.88) circle (  1.96);
\definecolor{drawColor}{RGB}{78,155,133}

\path[draw=drawColor,line width= 0.6pt,line join=round] (189.59, 82.71) -- (189.59,105.41);

\path[draw=drawColor,line width= 0.6pt,line join=round] (189.59, 65.74) -- (189.59, 46.61);

\path[draw=drawColor,line width= 0.6pt,fill=fillColor,fill opacity=0.20] (179.41, 82.71) --
	(179.41, 65.74) --
	(199.76, 65.74) --
	(199.76, 82.71) --
	(179.41, 82.71) --
	cycle;

\path[draw=drawColor,line width= 1.1pt] (179.41, 69.97) -- (199.76, 69.97);
\definecolor{drawColor}{RGB}{78,155,133}

\path[draw=drawColor,draw opacity=0.20,line width= 0.4pt,line join=round,line cap=round,fill=fillColor,fill opacity=0.20] (216.71, 16.34) circle (  1.96);

\path[draw=drawColor,draw opacity=0.20,line width= 0.4pt,line join=round,line cap=round,fill=fillColor,fill opacity=0.20] (216.71,107.53) circle (  1.96);

\path[draw=drawColor,draw opacity=0.20,line width= 0.4pt,line join=round,line cap=round,fill=fillColor,fill opacity=0.20] (216.71, 14.02) circle (  1.96);

\path[draw=drawColor,draw opacity=0.20,line width= 0.4pt,line join=round,line cap=round,fill=fillColor,fill opacity=0.20] (216.71,110.34) circle (  1.96);

\path[draw=drawColor,draw opacity=0.20,line width= 0.4pt,line join=round,line cap=round,fill=fillColor,fill opacity=0.20] (216.71,144.06) circle (  1.96);

\path[draw=drawColor,draw opacity=0.20,line width= 0.4pt,line join=round,line cap=round,fill=fillColor,fill opacity=0.20] (216.71, 42.22) circle (  1.96);

\path[draw=drawColor,draw opacity=0.20,line width= 0.4pt,line join=round,line cap=round,fill=fillColor,fill opacity=0.20] (216.71,108.17) circle (  1.96);
\definecolor{drawColor}{RGB}{78,155,133}

\path[draw=drawColor,line width= 0.6pt,line join=round] (216.71, 81.40) -- (216.71, 82.78);

\path[draw=drawColor,line width= 0.6pt,line join=round] (216.71, 67.29) -- (216.71, 65.53);

\path[draw=drawColor,line width= 0.6pt,fill=fillColor,fill opacity=0.20] (206.54, 81.40) --
	(206.54, 67.29) --
	(226.89, 67.29) --
	(226.89, 81.40) --
	(206.54, 81.40) --
	cycle;

\path[draw=drawColor,line width= 1.1pt] (206.54, 70.91) -- (226.89, 70.91);
\end{scope}
\begin{scope}
\path[clip] (  0.00,  0.00) rectangle (238.49,115.63);
\definecolor{drawColor}{RGB}{0,0,0}

\path[draw=drawColor,line width= 0.6pt,line join=round] ( 37.67, 29.80) --
	( 37.67,110.13);

\path[draw=drawColor,line width= 0.6pt,line join=round] ( 39.09,107.67) --
	( 37.67,110.13) --
	( 36.25,107.67);
\end{scope}
\begin{scope}
\path[clip] (  0.00,  0.00) rectangle (238.49,115.63);
\definecolor{drawColor}{gray}{0.30}

\node[text=drawColor,anchor=base east,inner sep=0pt, outer sep=0pt, scale=  0.88] at ( 32.72, 42.59) {0.98};

\node[text=drawColor,anchor=base east,inner sep=0pt, outer sep=0pt, scale=  0.88] at ( 32.72, 66.94) {1.00};

\node[text=drawColor,anchor=base east,inner sep=0pt, outer sep=0pt, scale=  0.88] at ( 32.72, 91.28) {1.02};
\end{scope}
\begin{scope}
\path[clip] (  0.00,  0.00) rectangle (238.49,115.63);
\definecolor{drawColor}{gray}{0.20}

\path[draw=drawColor,line width= 0.6pt,line join=round] ( 34.92, 45.63) --
	( 37.67, 45.63);

\path[draw=drawColor,line width= 0.6pt,line join=round] ( 34.92, 69.97) --
	( 37.67, 69.97);

\path[draw=drawColor,line width= 0.6pt,line join=round] ( 34.92, 94.31) --
	( 37.67, 94.31);
\end{scope}
\begin{scope}
\path[clip] (  0.00,  0.00) rectangle (238.49,115.63);
\definecolor{drawColor}{RGB}{0,0,0}

\path[draw=drawColor,line width= 0.6pt,line join=round] ( 37.67, 29.80) --
	(232.99, 29.80);

\path[draw=drawColor,line width= 0.6pt,line join=round] (230.53, 28.38) --
	(232.99, 29.80) --
	(230.53, 31.23);
\end{scope}
\begin{scope}
\path[clip] (  0.00,  0.00) rectangle (238.49,115.63);
\definecolor{drawColor}{gray}{0.20}

\path[draw=drawColor,line width= 0.6pt,line join=round] ( 53.95, 27.05) --
	( 53.95, 29.80);

\path[draw=drawColor,line width= 0.6pt,line join=round] ( 81.08, 27.05) --
	( 81.08, 29.80);

\path[draw=drawColor,line width= 0.6pt,line join=round] (108.20, 27.05) --
	(108.20, 29.80);

\path[draw=drawColor,line width= 0.6pt,line join=round] (135.33, 27.05) --
	(135.33, 29.80);

\path[draw=drawColor,line width= 0.6pt,line join=round] (162.46, 27.05) --
	(162.46, 29.80);

\path[draw=drawColor,line width= 0.6pt,line join=round] (189.59, 27.05) --
	(189.59, 29.80);

\path[draw=drawColor,line width= 0.6pt,line join=round] (216.71, 27.05) --
	(216.71, 29.80);
\end{scope}
\begin{scope}
\path[clip] (  0.00,  0.00) rectangle (238.49,115.63);
\definecolor{drawColor}{gray}{0.30}

\node[text=drawColor,anchor=base,inner sep=0pt, outer sep=0pt, scale=  0.88] at ( 53.95, 18.79) {2};

\node[text=drawColor,anchor=base,inner sep=0pt, outer sep=0pt, scale=  0.88] at ( 81.08, 18.79) {4};

\node[text=drawColor,anchor=base,inner sep=0pt, outer sep=0pt, scale=  0.88] at (108.20, 18.79) {8};

\node[text=drawColor,anchor=base,inner sep=0pt, outer sep=0pt, scale=  0.88] at (135.33, 18.79) {16};

\node[text=drawColor,anchor=base,inner sep=0pt, outer sep=0pt, scale=  0.88] at (162.46, 18.79) {32};

\node[text=drawColor,anchor=base,inner sep=0pt, outer sep=0pt, scale=  0.88] at (189.59, 18.79) {64};

\node[text=drawColor,anchor=base,inner sep=0pt, outer sep=0pt, scale=  0.88] at (216.71, 18.79) {128};
\end{scope}
\begin{scope}
\path[clip] (  0.00,  0.00) rectangle (238.49,115.63);
\definecolor{drawColor}{RGB}{0,0,0}

\node[text=drawColor,anchor=base,inner sep=0pt, outer sep=0pt, scale=  1.00] at (135.33,  7.44) {domain size};
\end{scope}
\begin{scope}
\path[clip] (  0.00,  0.00) rectangle (238.49,115.63);
\definecolor{drawColor}{RGB}{0,0,0}

\node[text=drawColor,rotate= 90.00,anchor=base,inner sep=0pt, outer sep=0pt, scale=  1.00] at ( 12.39, 69.97) {$p' \mathbin{/} p$};
\end{scope}
\end{tikzpicture}

%% file: files/plot-quots-p=0.9-eps=0.001.tex
\begin{tikzpicture}[x=1pt,y=1pt]
\definecolor{fillColor}{RGB}{255,255,255}
\path[use as bounding box,fill=fillColor,fill opacity=0.00] (0,0) rectangle (238.49,115.63);
\begin{scope}
\path[clip] (  0.00,  0.00) rectangle (238.49,115.63);
\definecolor{drawColor}{RGB}{255,255,255}
\definecolor{fillColor}{RGB}{255,255,255}

\path[draw=drawColor,line width= 0.6pt,line join=round,line cap=round,fill=fillColor] (  0.00,  0.00) rectangle (238.49,115.63);
\end{scope}
\begin{scope}
\path[clip] ( 46.47, 29.80) rectangle (232.99,110.13);
\definecolor{fillColor}{RGB}{255,255,255}

\path[fill=fillColor] ( 46.47, 29.80) rectangle (232.99,110.13);
\definecolor{drawColor}{RGB}{78,155,133}
\definecolor{fillColor}{RGB}{78,155,133}

\path[draw=drawColor,draw opacity=0.20,line width= 0.4pt,line join=round,line cap=round,fill=fillColor,fill opacity=0.20] ( 62.01, 63.57) circle (  1.96);

\path[draw=drawColor,draw opacity=0.20,line width= 0.4pt,line join=round,line cap=round,fill=fillColor,fill opacity=0.20] ( 62.01, 75.59) circle (  1.96);

\path[draw=drawColor,draw opacity=0.20,line width= 0.4pt,line join=round,line cap=round,fill=fillColor,fill opacity=0.20] ( 62.01, 71.26) circle (  1.96);

\path[draw=drawColor,draw opacity=0.20,line width= 0.4pt,line join=round,line cap=round,fill=fillColor,fill opacity=0.20] ( 62.01, 68.46) circle (  1.96);

\path[draw=drawColor,draw opacity=0.20,line width= 0.4pt,line join=round,line cap=round,fill=fillColor,fill opacity=0.20] ( 62.01, 55.77) circle (  1.96);

\path[draw=drawColor,draw opacity=0.20,line width= 0.4pt,line join=round,line cap=round,fill=fillColor,fill opacity=0.20] ( 62.01, 82.69) circle (  1.96);

\path[draw=drawColor,draw opacity=0.20,line width= 0.4pt,line join=round,line cap=round,fill=fillColor,fill opacity=0.20] ( 62.01, 66.01) circle (  1.96);

\path[draw=drawColor,draw opacity=0.20,line width= 0.4pt,line join=round,line cap=round,fill=fillColor,fill opacity=0.20] ( 62.01, 76.05) circle (  1.96);

\path[draw=drawColor,draw opacity=0.20,line width= 0.4pt,line join=round,line cap=round,fill=fillColor,fill opacity=0.20] ( 62.01, 78.96) circle (  1.96);

\path[draw=drawColor,draw opacity=0.20,line width= 0.4pt,line join=round,line cap=round,fill=fillColor,fill opacity=0.20] ( 62.01, 65.35) circle (  1.96);
\definecolor{drawColor}{RGB}{78,155,133}

\path[draw=drawColor,line width= 0.6pt,line join=round] ( 62.01, 70.28) -- ( 62.01, 71.09);

\path[draw=drawColor,line width= 0.6pt,line join=round] ( 62.01, 69.67) -- ( 62.01, 69.03);

\path[draw=drawColor,line width= 0.6pt,fill=fillColor,fill opacity=0.20] ( 52.30, 70.28) --
	( 52.30, 69.67) --
	( 71.73, 69.67) --
	( 71.73, 70.28) --
	( 52.30, 70.28) --
	cycle;

\path[draw=drawColor,line width= 1.1pt] ( 52.30, 69.97) -- ( 71.73, 69.97);
\definecolor{drawColor}{RGB}{78,155,133}

\path[draw=drawColor,draw opacity=0.20,line width= 0.4pt,line join=round,line cap=round,fill=fillColor,fill opacity=0.20] ( 87.92, 61.93) circle (  1.96);

\path[draw=drawColor,draw opacity=0.20,line width= 0.4pt,line join=round,line cap=round,fill=fillColor,fill opacity=0.20] ( 87.92, 75.62) circle (  1.96);

\path[draw=drawColor,draw opacity=0.20,line width= 0.4pt,line join=round,line cap=round,fill=fillColor,fill opacity=0.20] ( 87.92, 71.74) circle (  1.96);

\path[draw=drawColor,draw opacity=0.20,line width= 0.4pt,line join=round,line cap=round,fill=fillColor,fill opacity=0.20] ( 87.92, 67.86) circle (  1.96);

\path[draw=drawColor,draw opacity=0.20,line width= 0.4pt,line join=round,line cap=round,fill=fillColor,fill opacity=0.20] ( 87.92, 52.82) circle (  1.96);

\path[draw=drawColor,draw opacity=0.20,line width= 0.4pt,line join=round,line cap=round,fill=fillColor,fill opacity=0.20] ( 87.92, 82.27) circle (  1.96);

\path[draw=drawColor,draw opacity=0.20,line width= 0.4pt,line join=round,line cap=round,fill=fillColor,fill opacity=0.20] ( 87.92, 71.77) circle (  1.96);

\path[draw=drawColor,draw opacity=0.20,line width= 0.4pt,line join=round,line cap=round,fill=fillColor,fill opacity=0.20] ( 87.92, 78.46) circle (  1.96);

\path[draw=drawColor,draw opacity=0.20,line width= 0.4pt,line join=round,line cap=round,fill=fillColor,fill opacity=0.20] ( 87.92, 64.55) circle (  1.96);
\definecolor{drawColor}{RGB}{78,155,133}

\path[draw=drawColor,line width= 0.6pt,line join=round] ( 87.92, 70.34) -- ( 87.92, 70.47);

\path[draw=drawColor,line width= 0.6pt,line join=round] ( 87.92, 69.62) -- ( 87.92, 68.90);

\path[draw=drawColor,line width= 0.6pt,fill=fillColor,fill opacity=0.20] ( 78.20, 70.34) --
	( 78.20, 69.62) --
	( 97.63, 69.62) --
	( 97.63, 70.34) --
	( 78.20, 70.34) --
	cycle;

\path[draw=drawColor,line width= 1.1pt] ( 78.20, 69.97) -- ( 97.63, 69.97);
\definecolor{drawColor}{RGB}{78,155,133}

\path[draw=drawColor,draw opacity=0.20,line width= 0.4pt,line join=round,line cap=round,fill=fillColor,fill opacity=0.20] (113.82, 48.35) circle (  1.96);

\path[draw=drawColor,draw opacity=0.20,line width= 0.4pt,line join=round,line cap=round,fill=fillColor,fill opacity=0.20] (113.82, 23.97) circle (  1.96);

\path[draw=drawColor,draw opacity=0.20,line width= 0.4pt,line join=round,line cap=round,fill=fillColor,fill opacity=0.20] (113.82, 99.21) circle (  1.96);

\path[draw=drawColor,draw opacity=0.20,line width= 0.4pt,line join=round,line cap=round,fill=fillColor,fill opacity=0.20] (113.82, 94.28) circle (  1.96);

\path[draw=drawColor,draw opacity=0.20,line width= 0.4pt,line join=round,line cap=round,fill=fillColor,fill opacity=0.20] (113.82, 40.76) circle (  1.96);

\path[draw=drawColor,draw opacity=0.20,line width= 0.4pt,line join=round,line cap=round,fill=fillColor,fill opacity=0.20] (113.82, 51.18) circle (  1.96);

\path[draw=drawColor,draw opacity=0.20,line width= 0.4pt,line join=round,line cap=round,fill=fillColor,fill opacity=0.20] (113.82, 95.86) circle (  1.96);
\definecolor{drawColor}{RGB}{78,155,133}

\path[draw=drawColor,line width= 0.6pt,line join=round] (113.82, 73.72) -- (113.82, 83.47);

\path[draw=drawColor,line width= 0.6pt,line join=round] (113.82, 66.01) -- (113.82, 61.11);

\path[draw=drawColor,line width= 0.6pt,fill=fillColor,fill opacity=0.20] (104.11, 73.72) --
	(104.11, 66.01) --
	(123.54, 66.01) --
	(123.54, 73.72) --
	(104.11, 73.72) --
	cycle;

\path[draw=drawColor,line width= 1.1pt] (104.11, 69.97) -- (123.54, 69.97);
\definecolor{drawColor}{RGB}{78,155,133}

\path[draw=drawColor,draw opacity=0.20,line width= 0.4pt,line join=round,line cap=round,fill=fillColor,fill opacity=0.20] (139.73, 32.32) circle (  1.96);

\path[draw=drawColor,draw opacity=0.20,line width= 0.4pt,line join=round,line cap=round,fill=fillColor,fill opacity=0.20] (139.73, 93.17) circle (  1.96);

\path[draw=drawColor,draw opacity=0.20,line width= 0.4pt,line join=round,line cap=round,fill=fillColor,fill opacity=0.20] (139.73, 41.48) circle (  1.96);

\path[draw=drawColor,draw opacity=0.20,line width= 0.4pt,line join=round,line cap=round,fill=fillColor,fill opacity=0.20] (139.73, 87.83) circle (  1.96);

\path[draw=drawColor,draw opacity=0.20,line width= 0.4pt,line join=round,line cap=round,fill=fillColor,fill opacity=0.20] (139.73, 85.42) circle (  1.96);

\path[draw=drawColor,draw opacity=0.20,line width= 0.4pt,line join=round,line cap=round,fill=fillColor,fill opacity=0.20] (139.73, 51.40) circle (  1.96);

\path[draw=drawColor,draw opacity=0.20,line width= 0.4pt,line join=round,line cap=round,fill=fillColor,fill opacity=0.20] (139.73, 34.55) circle (  1.96);

\path[draw=drawColor,draw opacity=0.20,line width= 0.4pt,line join=round,line cap=round,fill=fillColor,fill opacity=0.20] (139.73,118.81) circle (  1.96);
\definecolor{drawColor}{RGB}{78,155,133}

\path[draw=drawColor,line width= 0.6pt,line join=round] (139.73, 72.47) -- (139.73, 77.02);

\path[draw=drawColor,line width= 0.6pt,line join=round] (139.73, 67.47) -- (139.73, 61.50);

\path[draw=drawColor,line width= 0.6pt,fill=fillColor,fill opacity=0.20] (130.02, 72.47) --
	(130.02, 67.47) --
	(149.45, 67.47) --
	(149.45, 72.47) --
	(130.02, 72.47) --
	cycle;

\path[draw=drawColor,line width= 1.1pt] (130.02, 69.97) -- (149.45, 69.97);
\definecolor{drawColor}{RGB}{78,155,133}

\path[draw=drawColor,draw opacity=0.20,line width= 0.4pt,line join=round,line cap=round,fill=fillColor,fill opacity=0.20] (165.64, 41.39) circle (  1.96);

\path[draw=drawColor,draw opacity=0.20,line width= 0.4pt,line join=round,line cap=round,fill=fillColor,fill opacity=0.20] (165.64, 87.58) circle (  1.96);

\path[draw=drawColor,draw opacity=0.20,line width= 0.4pt,line join=round,line cap=round,fill=fillColor,fill opacity=0.20] (165.64, 20.94) circle (  1.96);

\path[draw=drawColor,draw opacity=0.20,line width= 0.4pt,line join=round,line cap=round,fill=fillColor,fill opacity=0.20] (165.64,100.70) circle (  1.96);

\path[draw=drawColor,draw opacity=0.20,line width= 0.4pt,line join=round,line cap=round,fill=fillColor,fill opacity=0.20] (165.64, 91.59) circle (  1.96);
\definecolor{drawColor}{RGB}{78,155,133}

\path[draw=drawColor,line width= 0.6pt,line join=round] (165.64, 73.29) -- (165.64, 76.06);

\path[draw=drawColor,line width= 0.6pt,line join=round] (165.64, 65.28) -- (165.64, 54.28);

\path[draw=drawColor,line width= 0.6pt,fill=fillColor,fill opacity=0.20] (155.92, 73.29) --
	(155.92, 65.28) --
	(175.35, 65.28) --
	(175.35, 73.29) --
	(155.92, 73.29) --
	cycle;

\path[draw=drawColor,line width= 1.1pt] (155.92, 69.97) -- (175.35, 69.97);
\definecolor{drawColor}{RGB}{78,155,133}

\path[draw=drawColor,draw opacity=0.20,line width= 0.4pt,line join=round,line cap=round,fill=fillColor,fill opacity=0.20] (191.54, 39.39) circle (  1.96);

\path[draw=drawColor,draw opacity=0.20,line width= 0.4pt,line join=round,line cap=round,fill=fillColor,fill opacity=0.20] (191.54, 30.53) circle (  1.96);

\path[draw=drawColor,draw opacity=0.20,line width= 0.4pt,line join=round,line cap=round,fill=fillColor,fill opacity=0.20] (191.54, 94.69) circle (  1.96);

\path[draw=drawColor,draw opacity=0.20,line width= 0.4pt,line join=round,line cap=round,fill=fillColor,fill opacity=0.20] (191.54, 40.80) circle (  1.96);

\path[draw=drawColor,draw opacity=0.20,line width= 0.4pt,line join=round,line cap=round,fill=fillColor,fill opacity=0.20] (191.54,110.18) circle (  1.96);

\path[draw=drawColor,draw opacity=0.20,line width= 0.4pt,line join=round,line cap=round,fill=fillColor,fill opacity=0.20] (191.54,140.68) circle (  1.96);
\definecolor{drawColor}{RGB}{78,155,133}

\path[draw=drawColor,line width= 0.6pt,line join=round] (191.54, 75.49) -- (191.54, 88.81);

\path[draw=drawColor,line width= 0.6pt,line join=round] (191.54, 65.63) -- (191.54, 61.42);

\path[draw=drawColor,line width= 0.6pt,fill=fillColor,fill opacity=0.20] (181.83, 75.49) --
	(181.83, 65.63) --
	(201.26, 65.63) --
	(201.26, 75.49) --
	(181.83, 75.49) --
	cycle;

\path[draw=drawColor,line width= 1.1pt] (181.83, 69.97) -- (201.26, 69.97);
\definecolor{drawColor}{RGB}{78,155,133}

\path[draw=drawColor,draw opacity=0.20,line width= 0.4pt,line join=round,line cap=round,fill=fillColor,fill opacity=0.20] (217.45, 26.50) circle (  1.96);

\path[draw=drawColor,draw opacity=0.20,line width= 0.4pt,line join=round,line cap=round,fill=fillColor,fill opacity=0.20] (217.45, 96.75) circle (  1.96);

\path[draw=drawColor,draw opacity=0.20,line width= 0.4pt,line join=round,line cap=round,fill=fillColor,fill opacity=0.20] (217.45, 14.11) circle (  1.96);

\path[draw=drawColor,draw opacity=0.20,line width= 0.4pt,line join=round,line cap=round,fill=fillColor,fill opacity=0.20] (217.45,104.99) circle (  1.96);

\path[draw=drawColor,draw opacity=0.20,line width= 0.4pt,line join=round,line cap=round,fill=fillColor,fill opacity=0.20] (217.45, 34.01) circle (  1.96);

\path[draw=drawColor,draw opacity=0.20,line width= 0.4pt,line join=round,line cap=round,fill=fillColor,fill opacity=0.20] (217.45,113.18) circle (  1.96);

\path[draw=drawColor,draw opacity=0.20,line width= 0.4pt,line join=round,line cap=round,fill=fillColor,fill opacity=0.20] (217.45,104.62) circle (  1.96);
\definecolor{drawColor}{RGB}{78,155,133}

\path[draw=drawColor,line width= 0.6pt,line join=round] (217.45, 76.18) -- (217.45, 76.69);

\path[draw=drawColor,line width= 0.6pt,line join=round] (217.45, 63.63) -- (217.45, 44.84);

\path[draw=drawColor,line width= 0.6pt,fill=fillColor,fill opacity=0.20] (207.73, 76.18) --
	(207.73, 63.63) --
	(227.16, 63.63) --
	(227.16, 76.18) --
	(207.73, 76.18) --
	cycle;

\path[draw=drawColor,line width= 1.1pt] (207.73, 69.99) -- (227.16, 69.99);
\end{scope}
\begin{scope}
\path[clip] (  0.00,  0.00) rectangle (238.49,115.63);
\definecolor{drawColor}{RGB}{0,0,0}

\path[draw=drawColor,line width= 0.6pt,line join=round] ( 46.47, 29.80) --
	( 46.47,110.13);

\path[draw=drawColor,line width= 0.6pt,line join=round] ( 47.89,107.67) --
	( 46.47,110.13) --
	( 45.05,107.67);
\end{scope}
\begin{scope}
\path[clip] (  0.00,  0.00) rectangle (238.49,115.63);
\definecolor{drawColor}{gray}{0.30}

\node[text=drawColor,anchor=base east,inner sep=0pt, outer sep=0pt, scale=  0.88] at ( 41.52, 42.59) {0.9998};

\node[text=drawColor,anchor=base east,inner sep=0pt, outer sep=0pt, scale=  0.88] at ( 41.52, 66.94) {1.0000};

\node[text=drawColor,anchor=base east,inner sep=0pt, outer sep=0pt, scale=  0.88] at ( 41.52, 91.28) {1.0002};
\end{scope}
\begin{scope}
\path[clip] (  0.00,  0.00) rectangle (238.49,115.63);
\definecolor{drawColor}{gray}{0.20}

\path[draw=drawColor,line width= 0.6pt,line join=round] ( 43.72, 45.63) --
	( 46.47, 45.63);

\path[draw=drawColor,line width= 0.6pt,line join=round] ( 43.72, 69.97) --
	( 46.47, 69.97);

\path[draw=drawColor,line width= 0.6pt,line join=round] ( 43.72, 94.31) --
	( 46.47, 94.31);
\end{scope}
\begin{scope}
\path[clip] (  0.00,  0.00) rectangle (238.49,115.63);
\definecolor{drawColor}{RGB}{0,0,0}

\path[draw=drawColor,line width= 0.6pt,line join=round] ( 46.47, 29.80) --
	(232.99, 29.80);

\path[draw=drawColor,line width= 0.6pt,line join=round] (230.53, 28.38) --
	(232.99, 29.80) --
	(230.53, 31.23);
\end{scope}
\begin{scope}
\path[clip] (  0.00,  0.00) rectangle (238.49,115.63);
\definecolor{drawColor}{gray}{0.20}

\path[draw=drawColor,line width= 0.6pt,line join=round] ( 62.01, 27.05) --
	( 62.01, 29.80);

\path[draw=drawColor,line width= 0.6pt,line join=round] ( 87.92, 27.05) --
	( 87.92, 29.80);

\path[draw=drawColor,line width= 0.6pt,line join=round] (113.82, 27.05) --
	(113.82, 29.80);

\path[draw=drawColor,line width= 0.6pt,line join=round] (139.73, 27.05) --
	(139.73, 29.80);

\path[draw=drawColor,line width= 0.6pt,line join=round] (165.64, 27.05) --
	(165.64, 29.80);

\path[draw=drawColor,line width= 0.6pt,line join=round] (191.54, 27.05) --
	(191.54, 29.80);

\path[draw=drawColor,line width= 0.6pt,line join=round] (217.45, 27.05) --
	(217.45, 29.80);
\end{scope}
\begin{scope}
\path[clip] (  0.00,  0.00) rectangle (238.49,115.63);
\definecolor{drawColor}{gray}{0.30}

\node[text=drawColor,anchor=base,inner sep=0pt, outer sep=0pt, scale=  0.88] at ( 62.01, 18.79) {2};

\node[text=drawColor,anchor=base,inner sep=0pt, outer sep=0pt, scale=  0.88] at ( 87.92, 18.79) {4};

\node[text=drawColor,anchor=base,inner sep=0pt, outer sep=0pt, scale=  0.88] at (113.82, 18.79) {8};

\node[text=drawColor,anchor=base,inner sep=0pt, outer sep=0pt, scale=  0.88] at (139.73, 18.79) {16};

\node[text=drawColor,anchor=base,inner sep=0pt, outer sep=0pt, scale=  0.88] at (165.64, 18.79) {32};

\node[text=drawColor,anchor=base,inner sep=0pt, outer sep=0pt, scale=  0.88] at (191.54, 18.79) {64};

\node[text=drawColor,anchor=base,inner sep=0pt, outer sep=0pt, scale=  0.88] at (217.45, 18.79) {128};
\end{scope}
\begin{scope}
\path[clip] (  0.00,  0.00) rectangle (238.49,115.63);
\definecolor{drawColor}{RGB}{0,0,0}

\node[text=drawColor,anchor=base,inner sep=0pt, outer sep=0pt, scale=  1.00] at (139.73,  7.44) {domain size};
\end{scope}
\begin{scope}
\path[clip] (  0.00,  0.00) rectangle (238.49,115.63);
\definecolor{drawColor}{RGB}{0,0,0}

\node[text=drawColor,rotate= 90.00,anchor=base,inner sep=0pt, outer sep=0pt, scale=  1.00] at ( 12.39, 69.97) {$p' \mathbin{/} p$};
\end{scope}
\end{tikzpicture}

%% file: files/plot-quots-p=0.9-eps=0.1.tex
\begin{tikzpicture}[x=1pt,y=1pt]
\definecolor{fillColor}{RGB}{255,255,255}
\path[use as bounding box,fill=fillColor,fill opacity=0.00] (0,0) rectangle (238.49,115.63);
\begin{scope}
\path[clip] (  0.00,  0.00) rectangle (238.49,115.63);
\definecolor{drawColor}{RGB}{255,255,255}
\definecolor{fillColor}{RGB}{255,255,255}

\path[draw=drawColor,line width= 0.6pt,line join=round,line cap=round,fill=fillColor] (  0.00,  0.00) rectangle (238.49,115.63);
\end{scope}
\begin{scope}
\path[clip] ( 37.67, 29.80) rectangle (232.99,110.13);
\definecolor{fillColor}{RGB}{255,255,255}

\path[fill=fillColor] ( 37.67, 29.80) rectangle (232.99,110.13);
\definecolor{drawColor}{RGB}{78,155,133}
\definecolor{fillColor}{RGB}{78,155,133}

\path[draw=drawColor,draw opacity=0.20,line width= 0.4pt,line join=round,line cap=round,fill=fillColor,fill opacity=0.20] ( 53.95, 61.59) circle (  1.96);

\path[draw=drawColor,draw opacity=0.20,line width= 0.4pt,line join=round,line cap=round,fill=fillColor,fill opacity=0.20] ( 53.95, 77.80) circle (  1.96);

\path[draw=drawColor,draw opacity=0.20,line width= 0.4pt,line join=round,line cap=round,fill=fillColor,fill opacity=0.20] ( 53.95, 57.08) circle (  1.96);

\path[draw=drawColor,draw opacity=0.20,line width= 0.4pt,line join=round,line cap=round,fill=fillColor,fill opacity=0.20] ( 53.95, 82.39) circle (  1.96);

\path[draw=drawColor,draw opacity=0.20,line width= 0.4pt,line join=round,line cap=round,fill=fillColor,fill opacity=0.20] ( 53.95, 66.21) circle (  1.96);

\path[draw=drawColor,draw opacity=0.20,line width= 0.4pt,line join=round,line cap=round,fill=fillColor,fill opacity=0.20] ( 53.95, 75.54) circle (  1.96);

\path[draw=drawColor,draw opacity=0.20,line width= 0.4pt,line join=round,line cap=round,fill=fillColor,fill opacity=0.20] ( 53.95, 77.79) circle (  1.96);

\path[draw=drawColor,draw opacity=0.20,line width= 0.4pt,line join=round,line cap=round,fill=fillColor,fill opacity=0.20] ( 53.95, 66.28) circle (  1.96);
\definecolor{drawColor}{RGB}{78,155,133}

\path[draw=drawColor,line width= 0.6pt,line join=round] ( 53.95, 70.19) -- ( 53.95, 70.91);

\path[draw=drawColor,line width= 0.6pt,line join=round] ( 53.95, 69.54) -- ( 53.95, 69.21);

\path[draw=drawColor,line width= 0.6pt,fill=fillColor,fill opacity=0.20] ( 43.78, 70.19) --
	( 43.78, 69.54) --
	( 64.12, 69.54) --
	( 64.12, 70.19) --
	( 43.78, 70.19) --
	cycle;

\path[draw=drawColor,line width= 1.1pt] ( 43.78, 69.97) -- ( 64.12, 69.97);
\definecolor{drawColor}{RGB}{78,155,133}

\path[draw=drawColor,draw opacity=0.20,line width= 0.4pt,line join=round,line cap=round,fill=fillColor,fill opacity=0.20] ( 81.08, 60.14) circle (  1.96);

\path[draw=drawColor,draw opacity=0.20,line width= 0.4pt,line join=round,line cap=round,fill=fillColor,fill opacity=0.20] ( 81.08, 77.62) circle (  1.96);

\path[draw=drawColor,draw opacity=0.20,line width= 0.4pt,line join=round,line cap=round,fill=fillColor,fill opacity=0.20] ( 81.08, 55.02) circle (  1.96);

\path[draw=drawColor,draw opacity=0.20,line width= 0.4pt,line join=round,line cap=round,fill=fillColor,fill opacity=0.20] ( 81.08, 81.98) circle (  1.96);

\path[draw=drawColor,draw opacity=0.20,line width= 0.4pt,line join=round,line cap=round,fill=fillColor,fill opacity=0.20] ( 81.08, 75.52) circle (  1.96);

\path[draw=drawColor,draw opacity=0.20,line width= 0.4pt,line join=round,line cap=round,fill=fillColor,fill opacity=0.20] ( 81.08, 77.81) circle (  1.96);

\path[draw=drawColor,draw opacity=0.20,line width= 0.4pt,line join=round,line cap=round,fill=fillColor,fill opacity=0.20] ( 81.08, 65.86) circle (  1.96);
\definecolor{drawColor}{RGB}{78,155,133}

\path[draw=drawColor,line width= 0.6pt,line join=round] ( 81.08, 70.68) -- ( 81.08, 71.38);

\path[draw=drawColor,line width= 0.6pt,line join=round] ( 81.08, 69.47) -- ( 81.08, 68.86);

\path[draw=drawColor,line width= 0.6pt,fill=fillColor,fill opacity=0.20] ( 70.90, 70.68) --
	( 70.90, 69.47) --
	( 91.25, 69.47) --
	( 91.25, 70.68) --
	( 70.90, 70.68) --
	cycle;

\path[draw=drawColor,line width= 1.1pt] ( 70.90, 69.97) -- ( 91.25, 69.97);
\definecolor{drawColor}{RGB}{78,155,133}

\path[draw=drawColor,draw opacity=0.20,line width= 0.4pt,line join=round,line cap=round,fill=fillColor,fill opacity=0.20] (108.20, 49.88) circle (  1.96);

\path[draw=drawColor,draw opacity=0.20,line width= 0.4pt,line join=round,line cap=round,fill=fillColor,fill opacity=0.20] (108.20, 84.18) circle (  1.96);

\path[draw=drawColor,draw opacity=0.20,line width= 0.4pt,line join=round,line cap=round,fill=fillColor,fill opacity=0.20] (108.20, 31.39) circle (  1.96);

\path[draw=drawColor,draw opacity=0.20,line width= 0.4pt,line join=round,line cap=round,fill=fillColor,fill opacity=0.20] (108.20, 98.10) circle (  1.96);

\path[draw=drawColor,draw opacity=0.20,line width= 0.4pt,line join=round,line cap=round,fill=fillColor,fill opacity=0.20] (108.20, 92.92) circle (  1.96);

\path[draw=drawColor,draw opacity=0.20,line width= 0.4pt,line join=round,line cap=round,fill=fillColor,fill opacity=0.20] (108.20, 41.15) circle (  1.96);

\path[draw=drawColor,draw opacity=0.20,line width= 0.4pt,line join=round,line cap=round,fill=fillColor,fill opacity=0.20] (108.20, 54.81) circle (  1.96);

\path[draw=drawColor,draw opacity=0.20,line width= 0.4pt,line join=round,line cap=round,fill=fillColor,fill opacity=0.20] (108.20, 93.12) circle (  1.96);
\definecolor{drawColor}{RGB}{78,155,133}

\path[draw=drawColor,line width= 0.6pt,line join=round] (108.20, 72.38) -- (108.20, 73.52);

\path[draw=drawColor,line width= 0.6pt,line join=round] (108.20, 68.36) -- (108.20, 64.61);

\path[draw=drawColor,line width= 0.6pt,fill=fillColor,fill opacity=0.20] ( 98.03, 72.38) --
	( 98.03, 68.36) --
	(118.38, 68.36) --
	(118.38, 72.38) --
	( 98.03, 72.38) --
	cycle;

\path[draw=drawColor,line width= 1.1pt] ( 98.03, 69.97) -- (118.38, 69.97);
\definecolor{drawColor}{RGB}{78,155,133}

\path[draw=drawColor,draw opacity=0.20,line width= 0.4pt,line join=round,line cap=round,fill=fillColor,fill opacity=0.20] (135.33, 36.01) circle (  1.96);

\path[draw=drawColor,draw opacity=0.20,line width= 0.4pt,line join=round,line cap=round,fill=fillColor,fill opacity=0.20] (135.33, 93.75) circle (  1.96);

\path[draw=drawColor,draw opacity=0.20,line width= 0.4pt,line join=round,line cap=round,fill=fillColor,fill opacity=0.20] (135.33, 46.66) circle (  1.96);

\path[draw=drawColor,draw opacity=0.20,line width= 0.4pt,line join=round,line cap=round,fill=fillColor,fill opacity=0.20] (135.33, 86.78) circle (  1.96);

\path[draw=drawColor,draw opacity=0.20,line width= 0.4pt,line join=round,line cap=round,fill=fillColor,fill opacity=0.20] (135.33, 92.92) circle (  1.96);

\path[draw=drawColor,draw opacity=0.20,line width= 0.4pt,line join=round,line cap=round,fill=fillColor,fill opacity=0.20] (135.33, 41.15) circle (  1.96);

\path[draw=drawColor,draw opacity=0.20,line width= 0.4pt,line join=round,line cap=round,fill=fillColor,fill opacity=0.20] (135.33, 54.81) circle (  1.96);

\path[draw=drawColor,draw opacity=0.20,line width= 0.4pt,line join=round,line cap=round,fill=fillColor,fill opacity=0.20] (135.33, 93.12) circle (  1.96);
\definecolor{drawColor}{RGB}{78,155,133}

\path[draw=drawColor,line width= 0.6pt,line join=round] (135.33, 72.43) -- (135.33, 74.71);

\path[draw=drawColor,line width= 0.6pt,line join=round] (135.33, 67.13) -- (135.33, 63.65);

\path[draw=drawColor,line width= 0.6pt,fill=fillColor,fill opacity=0.20] (125.16, 72.43) --
	(125.16, 67.13) --
	(145.50, 67.13) --
	(145.50, 72.43) --
	(125.16, 72.43) --
	cycle;

\path[draw=drawColor,line width= 1.1pt] (125.16, 69.97) -- (145.50, 69.97);
\definecolor{drawColor}{RGB}{78,155,133}

\path[draw=drawColor,draw opacity=0.20,line width= 0.4pt,line join=round,line cap=round,fill=fillColor,fill opacity=0.20] (162.46, 43.73) circle (  1.96);

\path[draw=drawColor,draw opacity=0.20,line width= 0.4pt,line join=round,line cap=round,fill=fillColor,fill opacity=0.20] (162.46, 88.34) circle (  1.96);

\path[draw=drawColor,draw opacity=0.20,line width= 0.4pt,line join=round,line cap=round,fill=fillColor,fill opacity=0.20] (162.46, 29.45) circle (  1.96);

\path[draw=drawColor,draw opacity=0.20,line width= 0.4pt,line join=round,line cap=round,fill=fillColor,fill opacity=0.20] (162.46, 99.20) circle (  1.96);

\path[draw=drawColor,draw opacity=0.20,line width= 0.4pt,line join=round,line cap=round,fill=fillColor,fill opacity=0.20] (162.46, 91.35) circle (  1.96);

\path[draw=drawColor,draw opacity=0.20,line width= 0.4pt,line join=round,line cap=round,fill=fillColor,fill opacity=0.20] (162.46, 57.14) circle (  1.96);

\path[draw=drawColor,draw opacity=0.20,line width= 0.4pt,line join=round,line cap=round,fill=fillColor,fill opacity=0.20] (162.46, 89.57) circle (  1.96);

\path[draw=drawColor,draw opacity=0.20,line width= 0.4pt,line join=round,line cap=round,fill=fillColor,fill opacity=0.20] (162.46, 56.23) circle (  1.96);
\definecolor{drawColor}{RGB}{78,155,133}

\path[draw=drawColor,line width= 0.6pt,line join=round] (162.46, 72.12) -- (162.46, 75.99);

\path[draw=drawColor,line width= 0.6pt,line join=round] (162.46, 67.73) -- (162.46, 63.16);

\path[draw=drawColor,line width= 0.6pt,fill=fillColor,fill opacity=0.20] (152.29, 72.12) --
	(152.29, 67.73) --
	(172.63, 67.73) --
	(172.63, 72.12) --
	(152.29, 72.12) --
	cycle;

\path[draw=drawColor,line width= 1.1pt] (152.29, 70.04) -- (172.63, 70.04);
\definecolor{drawColor}{RGB}{78,155,133}

\path[draw=drawColor,draw opacity=0.20,line width= 0.4pt,line join=round,line cap=round,fill=fillColor,fill opacity=0.20] (189.59, 42.26) circle (  1.96);

\path[draw=drawColor,draw opacity=0.20,line width= 0.4pt,line join=round,line cap=round,fill=fillColor,fill opacity=0.20] (189.59, 37.75) circle (  1.96);

\path[draw=drawColor,draw opacity=0.20,line width= 0.4pt,line join=round,line cap=round,fill=fillColor,fill opacity=0.20] (189.59, 93.21) circle (  1.96);

\path[draw=drawColor,draw opacity=0.20,line width= 0.4pt,line join=round,line cap=round,fill=fillColor,fill opacity=0.20] (189.59, 92.37) circle (  1.96);

\path[draw=drawColor,draw opacity=0.20,line width= 0.4pt,line join=round,line cap=round,fill=fillColor,fill opacity=0.20] (189.59, 46.99) circle (  1.96);

\path[draw=drawColor,draw opacity=0.20,line width= 0.4pt,line join=round,line cap=round,fill=fillColor,fill opacity=0.20] (189.59,105.07) circle (  1.96);

\path[draw=drawColor,draw opacity=0.20,line width= 0.4pt,line join=round,line cap=round,fill=fillColor,fill opacity=0.20] (189.59,-35.68) circle (  1.96);

\path[draw=drawColor,draw opacity=0.20,line width= 0.4pt,line join=round,line cap=round,fill=fillColor,fill opacity=0.20] (189.59,130.96) circle (  1.96);
\definecolor{drawColor}{RGB}{78,155,133}

\path[draw=drawColor,line width= 0.6pt,line join=round] (189.59, 77.05) -- (189.59, 89.37);

\path[draw=drawColor,line width= 0.6pt,line join=round] (189.59, 67.67) -- (189.59, 61.15);

\path[draw=drawColor,line width= 0.6pt,fill=fillColor,fill opacity=0.20] (179.41, 77.05) --
	(179.41, 67.67) --
	(199.76, 67.67) --
	(199.76, 77.05) --
	(179.41, 77.05) --
	cycle;

\path[draw=drawColor,line width= 1.1pt] (179.41, 69.97) -- (199.76, 69.97);
\definecolor{drawColor}{RGB}{78,155,133}

\path[draw=drawColor,draw opacity=0.20,line width= 0.4pt,line join=round,line cap=round,fill=fillColor,fill opacity=0.20] (216.71, 24.89) circle (  1.96);

\path[draw=drawColor,draw opacity=0.20,line width= 0.4pt,line join=round,line cap=round,fill=fillColor,fill opacity=0.20] (216.71,151.47) circle (  1.96);
\definecolor{drawColor}{RGB}{78,155,133}

\path[draw=drawColor,line width= 0.6pt,line join=round] (216.71, 92.24) -- (216.71,110.68);

\path[draw=drawColor,line width= 0.6pt,line join=round] (216.71, 67.62) -- (216.71, 31.95);

\path[draw=drawColor,line width= 0.6pt,fill=fillColor,fill opacity=0.20] (206.54, 92.24) --
	(206.54, 67.62) --
	(226.89, 67.62) --
	(226.89, 92.24) --
	(206.54, 92.24) --
	cycle;

\path[draw=drawColor,line width= 1.1pt] (206.54, 70.88) -- (226.89, 70.88);
\end{scope}
\begin{scope}
\path[clip] (  0.00,  0.00) rectangle (238.49,115.63);
\definecolor{drawColor}{RGB}{0,0,0}

\path[draw=drawColor,line width= 0.6pt,line join=round] ( 37.67, 29.80) --
	( 37.67,110.13);

\path[draw=drawColor,line width= 0.6pt,line join=round] ( 39.09,107.67) --
	( 37.67,110.13) --
	( 36.25,107.67);
\end{scope}
\begin{scope}
\path[clip] (  0.00,  0.00) rectangle (238.49,115.63);
\definecolor{drawColor}{gray}{0.30}

\node[text=drawColor,anchor=base east,inner sep=0pt, outer sep=0pt, scale=  0.88] at ( 32.72, 42.59) {0.98};

\node[text=drawColor,anchor=base east,inner sep=0pt, outer sep=0pt, scale=  0.88] at ( 32.72, 66.94) {1.00};

\node[text=drawColor,anchor=base east,inner sep=0pt, outer sep=0pt, scale=  0.88] at ( 32.72, 91.28) {1.02};
\end{scope}
\begin{scope}
\path[clip] (  0.00,  0.00) rectangle (238.49,115.63);
\definecolor{drawColor}{gray}{0.20}

\path[draw=drawColor,line width= 0.6pt,line join=round] ( 34.92, 45.63) --
	( 37.67, 45.63);

\path[draw=drawColor,line width= 0.6pt,line join=round] ( 34.92, 69.97) --
	( 37.67, 69.97);

\path[draw=drawColor,line width= 0.6pt,line join=round] ( 34.92, 94.31) --
	( 37.67, 94.31);
\end{scope}
\begin{scope}
\path[clip] (  0.00,  0.00) rectangle (238.49,115.63);
\definecolor{drawColor}{RGB}{0,0,0}

\path[draw=drawColor,line width= 0.6pt,line join=round] ( 37.67, 29.80) --
	(232.99, 29.80);

\path[draw=drawColor,line width= 0.6pt,line join=round] (230.53, 28.38) --
	(232.99, 29.80) --
	(230.53, 31.23);
\end{scope}
\begin{scope}
\path[clip] (  0.00,  0.00) rectangle (238.49,115.63);
\definecolor{drawColor}{gray}{0.20}

\path[draw=drawColor,line width= 0.6pt,line join=round] ( 53.95, 27.05) --
	( 53.95, 29.80);

\path[draw=drawColor,line width= 0.6pt,line join=round] ( 81.08, 27.05) --
	( 81.08, 29.80);

\path[draw=drawColor,line width= 0.6pt,line join=round] (108.20, 27.05) --
	(108.20, 29.80);

\path[draw=drawColor,line width= 0.6pt,line join=round] (135.33, 27.05) --
	(135.33, 29.80);

\path[draw=drawColor,line width= 0.6pt,line join=round] (162.46, 27.05) --
	(162.46, 29.80);

\path[draw=drawColor,line width= 0.6pt,line join=round] (189.59, 27.05) --
	(189.59, 29.80);

\path[draw=drawColor,line width= 0.6pt,line join=round] (216.71, 27.05) --
	(216.71, 29.80);
\end{scope}
\begin{scope}
\path[clip] (  0.00,  0.00) rectangle (238.49,115.63);
\definecolor{drawColor}{gray}{0.30}

\node[text=drawColor,anchor=base,inner sep=0pt, outer sep=0pt, scale=  0.88] at ( 53.95, 18.79) {2};

\node[text=drawColor,anchor=base,inner sep=0pt, outer sep=0pt, scale=  0.88] at ( 81.08, 18.79) {4};

\node[text=drawColor,anchor=base,inner sep=0pt, outer sep=0pt, scale=  0.88] at (108.20, 18.79) {8};

\node[text=drawColor,anchor=base,inner sep=0pt, outer sep=0pt, scale=  0.88] at (135.33, 18.79) {16};

\node[text=drawColor,anchor=base,inner sep=0pt, outer sep=0pt, scale=  0.88] at (162.46, 18.79) {32};

\node[text=drawColor,anchor=base,inner sep=0pt, outer sep=0pt, scale=  0.88] at (189.59, 18.79) {64};

\node[text=drawColor,anchor=base,inner sep=0pt, outer sep=0pt, scale=  0.88] at (216.71, 18.79) {128};
\end{scope}
\begin{scope}
\path[clip] (  0.00,  0.00) rectangle (238.49,115.63);
\definecolor{drawColor}{RGB}{0,0,0}

\node[text=drawColor,anchor=base,inner sep=0pt, outer sep=0pt, scale=  1.00] at (135.33,  7.44) {domain size};
\end{scope}
\begin{scope}
\path[clip] (  0.00,  0.00) rectangle (238.49,115.63);
\definecolor{drawColor}{RGB}{0,0,0}

\node[text=drawColor,rotate= 90.00,anchor=base,inner sep=0pt, outer sep=0pt, scale=  1.00] at ( 12.39, 69.97) {$p' \mathbin{/} p$};
\end{scope}
\end{tikzpicture}

%% file: files/plot-quots-p=1-eps=0.001.tex
\begin{tikzpicture}[x=1pt,y=1pt]
\definecolor{fillColor}{RGB}{255,255,255}
\path[use as bounding box,fill=fillColor,fill opacity=0.00] (0,0) rectangle (238.49,115.63);
\begin{scope}
\path[clip] (  0.00,  0.00) rectangle (238.49,115.63);
\definecolor{drawColor}{RGB}{255,255,255}
\definecolor{fillColor}{RGB}{255,255,255}

\path[draw=drawColor,line width= 0.6pt,line join=round,line cap=round,fill=fillColor] (  0.00,  0.00) rectangle (238.49,115.63);
\end{scope}
\begin{scope}
\path[clip] ( 46.47, 29.80) rectangle (232.99,110.13);
\definecolor{fillColor}{RGB}{255,255,255}

\path[fill=fillColor] ( 46.47, 29.80) rectangle (232.99,110.13);
\definecolor{drawColor}{RGB}{78,155,133}
\definecolor{fillColor}{RGB}{78,155,133}

\path[draw=drawColor,draw opacity=0.20,line width= 0.4pt,line join=round,line cap=round,fill=fillColor,fill opacity=0.20] ( 62.01, 63.57) circle (  1.96);

\path[draw=drawColor,draw opacity=0.20,line width= 0.4pt,line join=round,line cap=round,fill=fillColor,fill opacity=0.20] ( 62.01, 75.59) circle (  1.96);

\path[draw=drawColor,draw opacity=0.20,line width= 0.4pt,line join=round,line cap=round,fill=fillColor,fill opacity=0.20] ( 62.01, 71.26) circle (  1.96);

\path[draw=drawColor,draw opacity=0.20,line width= 0.4pt,line join=round,line cap=round,fill=fillColor,fill opacity=0.20] ( 62.01, 68.46) circle (  1.96);

\path[draw=drawColor,draw opacity=0.20,line width= 0.4pt,line join=round,line cap=round,fill=fillColor,fill opacity=0.20] ( 62.01, 55.77) circle (  1.96);

\path[draw=drawColor,draw opacity=0.20,line width= 0.4pt,line join=round,line cap=round,fill=fillColor,fill opacity=0.20] ( 62.01, 82.69) circle (  1.96);

\path[draw=drawColor,draw opacity=0.20,line width= 0.4pt,line join=round,line cap=round,fill=fillColor,fill opacity=0.20] ( 62.01, 66.01) circle (  1.96);

\path[draw=drawColor,draw opacity=0.20,line width= 0.4pt,line join=round,line cap=round,fill=fillColor,fill opacity=0.20] ( 62.01, 76.05) circle (  1.96);

\path[draw=drawColor,draw opacity=0.20,line width= 0.4pt,line join=round,line cap=round,fill=fillColor,fill opacity=0.20] ( 62.01, 78.96) circle (  1.96);

\path[draw=drawColor,draw opacity=0.20,line width= 0.4pt,line join=round,line cap=round,fill=fillColor,fill opacity=0.20] ( 62.01, 65.35) circle (  1.96);
\definecolor{drawColor}{RGB}{78,155,133}

\path[draw=drawColor,line width= 0.6pt,line join=round] ( 62.01, 70.28) -- ( 62.01, 71.09);

\path[draw=drawColor,line width= 0.6pt,line join=round] ( 62.01, 69.67) -- ( 62.01, 69.03);

\path[draw=drawColor,line width= 0.6pt,fill=fillColor,fill opacity=0.20] ( 52.30, 70.28) --
	( 52.30, 69.67) --
	( 71.73, 69.67) --
	( 71.73, 70.28) --
	( 52.30, 70.28) --
	cycle;

\path[draw=drawColor,line width= 1.1pt] ( 52.30, 69.97) -- ( 71.73, 69.97);
\definecolor{drawColor}{RGB}{78,155,133}

\path[draw=drawColor,draw opacity=0.20,line width= 0.4pt,line join=round,line cap=round,fill=fillColor,fill opacity=0.20] ( 87.92, 61.93) circle (  1.96);

\path[draw=drawColor,draw opacity=0.20,line width= 0.4pt,line join=round,line cap=round,fill=fillColor,fill opacity=0.20] ( 87.92, 75.62) circle (  1.96);

\path[draw=drawColor,draw opacity=0.20,line width= 0.4pt,line join=round,line cap=round,fill=fillColor,fill opacity=0.20] ( 87.92, 71.74) circle (  1.96);

\path[draw=drawColor,draw opacity=0.20,line width= 0.4pt,line join=round,line cap=round,fill=fillColor,fill opacity=0.20] ( 87.92, 67.86) circle (  1.96);

\path[draw=drawColor,draw opacity=0.20,line width= 0.4pt,line join=round,line cap=round,fill=fillColor,fill opacity=0.20] ( 87.92, 52.82) circle (  1.96);

\path[draw=drawColor,draw opacity=0.20,line width= 0.4pt,line join=round,line cap=round,fill=fillColor,fill opacity=0.20] ( 87.92, 82.27) circle (  1.96);

\path[draw=drawColor,draw opacity=0.20,line width= 0.4pt,line join=round,line cap=round,fill=fillColor,fill opacity=0.20] ( 87.92, 71.77) circle (  1.96);

\path[draw=drawColor,draw opacity=0.20,line width= 0.4pt,line join=round,line cap=round,fill=fillColor,fill opacity=0.20] ( 87.92, 78.46) circle (  1.96);

\path[draw=drawColor,draw opacity=0.20,line width= 0.4pt,line join=round,line cap=round,fill=fillColor,fill opacity=0.20] ( 87.92, 64.55) circle (  1.96);
\definecolor{drawColor}{RGB}{78,155,133}

\path[draw=drawColor,line width= 0.6pt,line join=round] ( 87.92, 70.33) -- ( 87.92, 70.47);

\path[draw=drawColor,line width= 0.6pt,line join=round] ( 87.92, 69.62) -- ( 87.92, 68.90);

\path[draw=drawColor,line width= 0.6pt,fill=fillColor,fill opacity=0.20] ( 78.20, 70.33) --
	( 78.20, 69.62) --
	( 97.63, 69.62) --
	( 97.63, 70.33) --
	( 78.20, 70.33) --
	cycle;

\path[draw=drawColor,line width= 1.1pt] ( 78.20, 69.97) -- ( 97.63, 69.97);
\definecolor{drawColor}{RGB}{78,155,133}

\path[draw=drawColor,draw opacity=0.20,line width= 0.4pt,line join=round,line cap=round,fill=fillColor,fill opacity=0.20] (113.82, 48.35) circle (  1.96);

\path[draw=drawColor,draw opacity=0.20,line width= 0.4pt,line join=round,line cap=round,fill=fillColor,fill opacity=0.20] (113.82, 39.16) circle (  1.96);

\path[draw=drawColor,draw opacity=0.20,line width= 0.4pt,line join=round,line cap=round,fill=fillColor,fill opacity=0.20] (113.82, 89.56) circle (  1.96);

\path[draw=drawColor,draw opacity=0.20,line width= 0.4pt,line join=round,line cap=round,fill=fillColor,fill opacity=0.20] (113.82, 94.28) circle (  1.96);

\path[draw=drawColor,draw opacity=0.20,line width= 0.4pt,line join=round,line cap=round,fill=fillColor,fill opacity=0.20] (113.82, 40.76) circle (  1.96);

\path[draw=drawColor,draw opacity=0.20,line width= 0.4pt,line join=round,line cap=round,fill=fillColor,fill opacity=0.20] (113.82, 51.18) circle (  1.96);

\path[draw=drawColor,draw opacity=0.20,line width= 0.4pt,line join=round,line cap=round,fill=fillColor,fill opacity=0.20] (113.82, 95.86) circle (  1.96);
\definecolor{drawColor}{RGB}{78,155,133}

\path[draw=drawColor,line width= 0.6pt,line join=round] (113.82, 73.47) -- (113.82, 83.47);

\path[draw=drawColor,line width= 0.6pt,line join=round] (113.82, 66.50) -- (113.82, 61.11);

\path[draw=drawColor,line width= 0.6pt,fill=fillColor,fill opacity=0.20] (104.11, 73.47) --
	(104.11, 66.50) --
	(123.54, 66.50) --
	(123.54, 73.47) --
	(104.11, 73.47) --
	cycle;

\path[draw=drawColor,line width= 1.1pt] (104.11, 69.97) -- (123.54, 69.97);
\definecolor{drawColor}{RGB}{78,155,133}

\path[draw=drawColor,draw opacity=0.20,line width= 0.4pt,line join=round,line cap=round,fill=fillColor,fill opacity=0.20] (139.73, 42.39) circle (  1.96);

\path[draw=drawColor,draw opacity=0.20,line width= 0.4pt,line join=round,line cap=round,fill=fillColor,fill opacity=0.20] (139.73, 86.96) circle (  1.96);

\path[draw=drawColor,draw opacity=0.20,line width= 0.4pt,line join=round,line cap=round,fill=fillColor,fill opacity=0.20] (139.73, 30.55) circle (  1.96);

\path[draw=drawColor,draw opacity=0.20,line width= 0.4pt,line join=round,line cap=round,fill=fillColor,fill opacity=0.20] (139.73, 94.68) circle (  1.96);

\path[draw=drawColor,draw opacity=0.20,line width= 0.4pt,line join=round,line cap=round,fill=fillColor,fill opacity=0.20] (139.73, 83.84) circle (  1.96);

\path[draw=drawColor,draw opacity=0.20,line width= 0.4pt,line join=round,line cap=round,fill=fillColor,fill opacity=0.20] (139.73, 53.30) circle (  1.96);

\path[draw=drawColor,draw opacity=0.20,line width= 0.4pt,line join=round,line cap=round,fill=fillColor,fill opacity=0.20] (139.73, 49.61) circle (  1.96);

\path[draw=drawColor,draw opacity=0.20,line width= 0.4pt,line join=round,line cap=round,fill=fillColor,fill opacity=0.20] (139.73, 98.04) circle (  1.96);
\definecolor{drawColor}{RGB}{78,155,133}

\path[draw=drawColor,line width= 0.6pt,line join=round] (139.73, 72.47) -- (139.73, 75.57);

\path[draw=drawColor,line width= 0.6pt,line join=round] (139.73, 66.70) -- (139.73, 63.24);

\path[draw=drawColor,line width= 0.6pt,fill=fillColor,fill opacity=0.20] (130.02, 72.47) --
	(130.02, 66.70) --
	(149.45, 66.70) --
	(149.45, 72.47) --
	(130.02, 72.47) --
	cycle;

\path[draw=drawColor,line width= 1.1pt] (130.02, 69.97) -- (149.45, 69.97);
\definecolor{drawColor}{RGB}{78,155,133}

\path[draw=drawColor,draw opacity=0.20,line width= 0.4pt,line join=round,line cap=round,fill=fillColor,fill opacity=0.20] (165.64, 39.34) circle (  1.96);

\path[draw=drawColor,draw opacity=0.20,line width= 0.4pt,line join=round,line cap=round,fill=fillColor,fill opacity=0.20] (165.64, 88.84) circle (  1.96);

\path[draw=drawColor,draw opacity=0.20,line width= 0.4pt,line join=round,line cap=round,fill=fillColor,fill opacity=0.20] (165.64, 29.72) circle (  1.96);

\path[draw=drawColor,draw opacity=0.20,line width= 0.4pt,line join=round,line cap=round,fill=fillColor,fill opacity=0.20] (165.64, 95.20) circle (  1.96);

\path[draw=drawColor,draw opacity=0.20,line width= 0.4pt,line join=round,line cap=round,fill=fillColor,fill opacity=0.20] (165.64, 52.33) circle (  1.96);

\path[draw=drawColor,draw opacity=0.20,line width= 0.4pt,line join=round,line cap=round,fill=fillColor,fill opacity=0.20] (165.64, 94.29) circle (  1.96);
\definecolor{drawColor}{RGB}{78,155,133}

\path[draw=drawColor,line width= 0.6pt,line join=round] (165.64, 72.86) -- (165.64, 75.24);

\path[draw=drawColor,line width= 0.6pt,line join=round] (165.64, 68.02) -- (165.64, 64.35);

\path[draw=drawColor,line width= 0.6pt,fill=fillColor,fill opacity=0.20] (155.92, 72.86) --
	(155.92, 68.02) --
	(175.35, 68.02) --
	(175.35, 72.86) --
	(155.92, 72.86) --
	cycle;

\path[draw=drawColor,line width= 1.1pt] (155.92, 69.97) -- (175.35, 69.97);
\definecolor{drawColor}{RGB}{78,155,133}

\path[draw=drawColor,draw opacity=0.20,line width= 0.4pt,line join=round,line cap=round,fill=fillColor,fill opacity=0.20] (191.54, 41.01) circle (  1.96);

\path[draw=drawColor,draw opacity=0.20,line width= 0.4pt,line join=round,line cap=round,fill=fillColor,fill opacity=0.20] (191.54, 87.81) circle (  1.96);

\path[draw=drawColor,draw opacity=0.20,line width= 0.4pt,line join=round,line cap=round,fill=fillColor,fill opacity=0.20] (191.54, 32.02) circle (  1.96);

\path[draw=drawColor,draw opacity=0.20,line width= 0.4pt,line join=round,line cap=round,fill=fillColor,fill opacity=0.20] (191.54, 93.76) circle (  1.96);

\path[draw=drawColor,draw opacity=0.20,line width= 0.4pt,line join=round,line cap=round,fill=fillColor,fill opacity=0.20] (191.54, 48.30) circle (  1.96);

\path[draw=drawColor,draw opacity=0.20,line width= 0.4pt,line join=round,line cap=round,fill=fillColor,fill opacity=0.20] (191.54, 99.84) circle (  1.96);

\path[draw=drawColor,draw opacity=0.20,line width= 0.4pt,line join=round,line cap=round,fill=fillColor,fill opacity=0.20] (191.54,138.37) circle (  1.96);
\definecolor{drawColor}{RGB}{78,155,133}

\path[draw=drawColor,line width= 0.6pt,line join=round] (191.54, 74.99) -- (191.54, 79.79);

\path[draw=drawColor,line width= 0.6pt,line join=round] (191.54, 66.53) -- (191.54, 61.80);

\path[draw=drawColor,line width= 0.6pt,fill=fillColor,fill opacity=0.20] (181.83, 74.99) --
	(181.83, 66.53) --
	(201.26, 66.53) --
	(201.26, 74.99) --
	(181.83, 74.99) --
	cycle;

\path[draw=drawColor,line width= 1.1pt] (181.83, 69.97) -- (201.26, 69.97);
\definecolor{drawColor}{RGB}{78,155,133}

\path[draw=drawColor,draw opacity=0.20,line width= 0.4pt,line join=round,line cap=round,fill=fillColor,fill opacity=0.20] (217.45, 37.08) circle (  1.96);

\path[draw=drawColor,draw opacity=0.20,line width= 0.4pt,line join=round,line cap=round,fill=fillColor,fill opacity=0.20] (217.45, 26.95) circle (  1.96);

\path[draw=drawColor,draw opacity=0.20,line width= 0.4pt,line join=round,line cap=round,fill=fillColor,fill opacity=0.20] (217.45, 96.93) circle (  1.96);

\path[draw=drawColor,draw opacity=0.20,line width= 0.4pt,line join=round,line cap=round,fill=fillColor,fill opacity=0.20] (217.45, 92.26) circle (  1.96);

\path[draw=drawColor,draw opacity=0.20,line width= 0.4pt,line join=round,line cap=round,fill=fillColor,fill opacity=0.20] (217.45, 94.59) circle (  1.96);
\definecolor{drawColor}{RGB}{78,155,133}

\path[draw=drawColor,line width= 0.6pt,line join=round] (217.45, 75.27) -- (217.45, 90.23);

\path[draw=drawColor,line width= 0.6pt,line join=round] (217.45, 64.62) -- (217.45, 51.42);

\path[draw=drawColor,line width= 0.6pt,fill=fillColor,fill opacity=0.20] (207.73, 75.27) --
	(207.73, 64.62) --
	(227.16, 64.62) --
	(227.16, 75.27) --
	(207.73, 75.27) --
	cycle;

\path[draw=drawColor,line width= 1.1pt] (207.73, 69.98) -- (227.16, 69.98);
\end{scope}
\begin{scope}
\path[clip] (  0.00,  0.00) rectangle (238.49,115.63);
\definecolor{drawColor}{RGB}{0,0,0}

\path[draw=drawColor,line width= 0.6pt,line join=round] ( 46.47, 29.80) --
	( 46.47,110.13);

\path[draw=drawColor,line width= 0.6pt,line join=round] ( 47.89,107.67) --
	( 46.47,110.13) --
	( 45.05,107.67);
\end{scope}
\begin{scope}
\path[clip] (  0.00,  0.00) rectangle (238.49,115.63);
\definecolor{drawColor}{gray}{0.30}

\node[text=drawColor,anchor=base east,inner sep=0pt, outer sep=0pt, scale=  0.88] at ( 41.52, 42.59) {0.9998};

\node[text=drawColor,anchor=base east,inner sep=0pt, outer sep=0pt, scale=  0.88] at ( 41.52, 66.94) {1.0000};

\node[text=drawColor,anchor=base east,inner sep=0pt, outer sep=0pt, scale=  0.88] at ( 41.52, 91.28) {1.0002};
\end{scope}
\begin{scope}
\path[clip] (  0.00,  0.00) rectangle (238.49,115.63);
\definecolor{drawColor}{gray}{0.20}

\path[draw=drawColor,line width= 0.6pt,line join=round] ( 43.72, 45.63) --
	( 46.47, 45.63);

\path[draw=drawColor,line width= 0.6pt,line join=round] ( 43.72, 69.97) --
	( 46.47, 69.97);

\path[draw=drawColor,line width= 0.6pt,line join=round] ( 43.72, 94.31) --
	( 46.47, 94.31);
\end{scope}
\begin{scope}
\path[clip] (  0.00,  0.00) rectangle (238.49,115.63);
\definecolor{drawColor}{RGB}{0,0,0}

\path[draw=drawColor,line width= 0.6pt,line join=round] ( 46.47, 29.80) --
	(232.99, 29.80);

\path[draw=drawColor,line width= 0.6pt,line join=round] (230.53, 28.38) --
	(232.99, 29.80) --
	(230.53, 31.23);
\end{scope}
\begin{scope}
\path[clip] (  0.00,  0.00) rectangle (238.49,115.63);
\definecolor{drawColor}{gray}{0.20}

\path[draw=drawColor,line width= 0.6pt,line join=round] ( 62.01, 27.05) --
	( 62.01, 29.80);

\path[draw=drawColor,line width= 0.6pt,line join=round] ( 87.92, 27.05) --
	( 87.92, 29.80);

\path[draw=drawColor,line width= 0.6pt,line join=round] (113.82, 27.05) --
	(113.82, 29.80);

\path[draw=drawColor,line width= 0.6pt,line join=round] (139.73, 27.05) --
	(139.73, 29.80);

\path[draw=drawColor,line width= 0.6pt,line join=round] (165.64, 27.05) --
	(165.64, 29.80);

\path[draw=drawColor,line width= 0.6pt,line join=round] (191.54, 27.05) --
	(191.54, 29.80);

\path[draw=drawColor,line width= 0.6pt,line join=round] (217.45, 27.05) --
	(217.45, 29.80);
\end{scope}
\begin{scope}
\path[clip] (  0.00,  0.00) rectangle (238.49,115.63);
\definecolor{drawColor}{gray}{0.30}

\node[text=drawColor,anchor=base,inner sep=0pt, outer sep=0pt, scale=  0.88] at ( 62.01, 18.79) {2};

\node[text=drawColor,anchor=base,inner sep=0pt, outer sep=0pt, scale=  0.88] at ( 87.92, 18.79) {4};

\node[text=drawColor,anchor=base,inner sep=0pt, outer sep=0pt, scale=  0.88] at (113.82, 18.79) {8};

\node[text=drawColor,anchor=base,inner sep=0pt, outer sep=0pt, scale=  0.88] at (139.73, 18.79) {16};

\node[text=drawColor,anchor=base,inner sep=0pt, outer sep=0pt, scale=  0.88] at (165.64, 18.79) {32};

\node[text=drawColor,anchor=base,inner sep=0pt, outer sep=0pt, scale=  0.88] at (191.54, 18.79) {64};

\node[text=drawColor,anchor=base,inner sep=0pt, outer sep=0pt, scale=  0.88] at (217.45, 18.79) {128};
\end{scope}
\begin{scope}
\path[clip] (  0.00,  0.00) rectangle (238.49,115.63);
\definecolor{drawColor}{RGB}{0,0,0}

\node[text=drawColor,anchor=base,inner sep=0pt, outer sep=0pt, scale=  1.00] at (139.73,  7.44) {domain size};
\end{scope}
\begin{scope}
\path[clip] (  0.00,  0.00) rectangle (238.49,115.63);
\definecolor{drawColor}{RGB}{0,0,0}

\node[text=drawColor,rotate= 90.00,anchor=base,inner sep=0pt, outer sep=0pt, scale=  1.00] at ( 12.39, 69.97) {$p' \mathbin{/} p$};
\end{scope}
\end{tikzpicture}

%% file: files/plot-quots-p=1-eps=0.1.tex
\begin{tikzpicture}[x=1pt,y=1pt]
\definecolor{fillColor}{RGB}{255,255,255}
\path[use as bounding box,fill=fillColor,fill opacity=0.00] (0,0) rectangle (238.49,115.63);
\begin{scope}
\path[clip] (  0.00,  0.00) rectangle (238.49,115.63);
\definecolor{drawColor}{RGB}{255,255,255}
\definecolor{fillColor}{RGB}{255,255,255}

\path[draw=drawColor,line width= 0.6pt,line join=round,line cap=round,fill=fillColor] (  0.00,  0.00) rectangle (238.49,115.63);
\end{scope}
\begin{scope}
\path[clip] ( 37.67, 29.80) rectangle (232.99,110.13);
\definecolor{fillColor}{RGB}{255,255,255}

\path[fill=fillColor] ( 37.67, 29.80) rectangle (232.99,110.13);
\definecolor{drawColor}{RGB}{78,155,133}
\definecolor{fillColor}{RGB}{78,155,133}

\path[draw=drawColor,draw opacity=0.20,line width= 0.4pt,line join=round,line cap=round,fill=fillColor,fill opacity=0.20] ( 53.95, 61.59) circle (  1.96);

\path[draw=drawColor,draw opacity=0.20,line width= 0.4pt,line join=round,line cap=round,fill=fillColor,fill opacity=0.20] ( 53.95, 77.80) circle (  1.96);

\path[draw=drawColor,draw opacity=0.20,line width= 0.4pt,line join=round,line cap=round,fill=fillColor,fill opacity=0.20] ( 53.95, 57.08) circle (  1.96);

\path[draw=drawColor,draw opacity=0.20,line width= 0.4pt,line join=round,line cap=round,fill=fillColor,fill opacity=0.20] ( 53.95, 82.39) circle (  1.96);

\path[draw=drawColor,draw opacity=0.20,line width= 0.4pt,line join=round,line cap=round,fill=fillColor,fill opacity=0.20] ( 53.95, 66.21) circle (  1.96);

\path[draw=drawColor,draw opacity=0.20,line width= 0.4pt,line join=round,line cap=round,fill=fillColor,fill opacity=0.20] ( 53.95, 75.54) circle (  1.96);

\path[draw=drawColor,draw opacity=0.20,line width= 0.4pt,line join=round,line cap=round,fill=fillColor,fill opacity=0.20] ( 53.95, 77.79) circle (  1.96);

\path[draw=drawColor,draw opacity=0.20,line width= 0.4pt,line join=round,line cap=round,fill=fillColor,fill opacity=0.20] ( 53.95, 66.28) circle (  1.96);
\definecolor{drawColor}{RGB}{78,155,133}

\path[draw=drawColor,line width= 0.6pt,line join=round] ( 53.95, 70.19) -- ( 53.95, 70.91);

\path[draw=drawColor,line width= 0.6pt,line join=round] ( 53.95, 69.54) -- ( 53.95, 69.21);

\path[draw=drawColor,line width= 0.6pt,fill=fillColor,fill opacity=0.20] ( 43.78, 70.19) --
	( 43.78, 69.54) --
	( 64.12, 69.54) --
	( 64.12, 70.19) --
	( 43.78, 70.19) --
	cycle;

\path[draw=drawColor,line width= 1.1pt] ( 43.78, 69.97) -- ( 64.12, 69.97);
\definecolor{drawColor}{RGB}{78,155,133}

\path[draw=drawColor,draw opacity=0.20,line width= 0.4pt,line join=round,line cap=round,fill=fillColor,fill opacity=0.20] ( 81.08, 60.14) circle (  1.96);

\path[draw=drawColor,draw opacity=0.20,line width= 0.4pt,line join=round,line cap=round,fill=fillColor,fill opacity=0.20] ( 81.08, 77.62) circle (  1.96);

\path[draw=drawColor,draw opacity=0.20,line width= 0.4pt,line join=round,line cap=round,fill=fillColor,fill opacity=0.20] ( 81.08, 55.02) circle (  1.96);

\path[draw=drawColor,draw opacity=0.20,line width= 0.4pt,line join=round,line cap=round,fill=fillColor,fill opacity=0.20] ( 81.08, 81.98) circle (  1.96);

\path[draw=drawColor,draw opacity=0.20,line width= 0.4pt,line join=round,line cap=round,fill=fillColor,fill opacity=0.20] ( 81.08, 77.81) circle (  1.96);

\path[draw=drawColor,draw opacity=0.20,line width= 0.4pt,line join=round,line cap=round,fill=fillColor,fill opacity=0.20] ( 81.08, 65.86) circle (  1.96);
\definecolor{drawColor}{RGB}{78,155,133}

\path[draw=drawColor,line width= 0.6pt,line join=round] ( 81.08, 70.28) -- ( 81.08, 71.38);

\path[draw=drawColor,line width= 0.6pt,line join=round] ( 81.08, 69.47) -- ( 81.08, 68.86);

\path[draw=drawColor,line width= 0.6pt,fill=fillColor,fill opacity=0.20] ( 70.90, 70.28) --
	( 70.90, 69.47) --
	( 91.25, 69.47) --
	( 91.25, 70.28) --
	( 70.90, 70.28) --
	cycle;

\path[draw=drawColor,line width= 1.1pt] ( 70.90, 69.97) -- ( 91.25, 69.97);
\definecolor{drawColor}{RGB}{78,155,133}

\path[draw=drawColor,draw opacity=0.20,line width= 0.4pt,line join=round,line cap=round,fill=fillColor,fill opacity=0.20] (108.20, 49.88) circle (  1.96);

\path[draw=drawColor,draw opacity=0.20,line width= 0.4pt,line join=round,line cap=round,fill=fillColor,fill opacity=0.20] (108.20, 84.18) circle (  1.96);

\path[draw=drawColor,draw opacity=0.20,line width= 0.4pt,line join=round,line cap=round,fill=fillColor,fill opacity=0.20] (108.20, 44.68) circle (  1.96);

\path[draw=drawColor,draw opacity=0.20,line width= 0.4pt,line join=round,line cap=round,fill=fillColor,fill opacity=0.20] (108.20, 88.41) circle (  1.96);

\path[draw=drawColor,draw opacity=0.20,line width= 0.4pt,line join=round,line cap=round,fill=fillColor,fill opacity=0.20] (108.20, 92.92) circle (  1.96);

\path[draw=drawColor,draw opacity=0.20,line width= 0.4pt,line join=round,line cap=round,fill=fillColor,fill opacity=0.20] (108.20, 41.15) circle (  1.96);

\path[draw=drawColor,draw opacity=0.20,line width= 0.4pt,line join=round,line cap=round,fill=fillColor,fill opacity=0.20] (108.20, 54.81) circle (  1.96);

\path[draw=drawColor,draw opacity=0.20,line width= 0.4pt,line join=round,line cap=round,fill=fillColor,fill opacity=0.20] (108.20, 93.12) circle (  1.96);
\definecolor{drawColor}{RGB}{78,155,133}

\path[draw=drawColor,line width= 0.6pt,line join=round] (108.20, 72.32) -- (108.20, 73.61);

\path[draw=drawColor,line width= 0.6pt,line join=round] (108.20, 68.48) -- (108.20, 64.48);

\path[draw=drawColor,line width= 0.6pt,fill=fillColor,fill opacity=0.20] ( 98.03, 72.32) --
	( 98.03, 68.48) --
	(118.38, 68.48) --
	(118.38, 72.32) --
	( 98.03, 72.32) --
	cycle;

\path[draw=drawColor,line width= 1.1pt] ( 98.03, 69.97) -- (118.38, 69.97);
\definecolor{drawColor}{RGB}{78,155,133}

\path[draw=drawColor,draw opacity=0.20,line width= 0.4pt,line join=round,line cap=round,fill=fillColor,fill opacity=0.20] (135.33, 44.27) circle (  1.96);

\path[draw=drawColor,draw opacity=0.20,line width= 0.4pt,line join=round,line cap=round,fill=fillColor,fill opacity=0.20] (135.33, 87.97) circle (  1.96);

\path[draw=drawColor,draw opacity=0.20,line width= 0.4pt,line join=round,line cap=round,fill=fillColor,fill opacity=0.20] (135.33, 38.31) circle (  1.96);

\path[draw=drawColor,draw opacity=0.20,line width= 0.4pt,line join=round,line cap=round,fill=fillColor,fill opacity=0.20] (135.33, 92.81) circle (  1.96);

\path[draw=drawColor,draw opacity=0.20,line width= 0.4pt,line join=round,line cap=round,fill=fillColor,fill opacity=0.20] (135.33, 83.72) circle (  1.96);

\path[draw=drawColor,draw opacity=0.20,line width= 0.4pt,line join=round,line cap=round,fill=fillColor,fill opacity=0.20] (135.33, 52.70) circle (  1.96);

\path[draw=drawColor,draw opacity=0.20,line width= 0.4pt,line join=round,line cap=round,fill=fillColor,fill opacity=0.20] (135.33, 53.29) circle (  1.96);

\path[draw=drawColor,draw opacity=0.20,line width= 0.4pt,line join=round,line cap=round,fill=fillColor,fill opacity=0.20] (135.33, 95.44) circle (  1.96);
\definecolor{drawColor}{RGB}{78,155,133}

\path[draw=drawColor,line width= 0.6pt,line join=round] (135.33, 71.65) -- (135.33, 74.77);

\path[draw=drawColor,line width= 0.6pt,line join=round] (135.33, 67.91) -- (135.33, 63.58);

\path[draw=drawColor,line width= 0.6pt,fill=fillColor,fill opacity=0.20] (125.16, 71.65) --
	(125.16, 67.91) --
	(145.50, 67.91) --
	(145.50, 71.65) --
	(125.16, 71.65) --
	cycle;

\path[draw=drawColor,line width= 1.1pt] (125.16, 69.97) -- (145.50, 69.97);
\definecolor{drawColor}{RGB}{78,155,133}

\path[draw=drawColor,draw opacity=0.20,line width= 0.4pt,line join=round,line cap=round,fill=fillColor,fill opacity=0.20] (162.46, 41.70) circle (  1.96);

\path[draw=drawColor,draw opacity=0.20,line width= 0.4pt,line join=round,line cap=round,fill=fillColor,fill opacity=0.20] (162.46, 89.76) circle (  1.96);

\path[draw=drawColor,draw opacity=0.20,line width= 0.4pt,line join=round,line cap=round,fill=fillColor,fill opacity=0.20] (162.46, 37.10) circle (  1.96);

\path[draw=drawColor,draw opacity=0.20,line width= 0.4pt,line join=round,line cap=round,fill=fillColor,fill opacity=0.20] (162.46, 93.68) circle (  1.96);

\path[draw=drawColor,draw opacity=0.20,line width= 0.4pt,line join=round,line cap=round,fill=fillColor,fill opacity=0.20] (162.46, 80.51) circle (  1.96);

\path[draw=drawColor,draw opacity=0.20,line width= 0.4pt,line join=round,line cap=round,fill=fillColor,fill opacity=0.20] (162.46, 55.41) circle (  1.96);

\path[draw=drawColor,draw opacity=0.20,line width= 0.4pt,line join=round,line cap=round,fill=fillColor,fill opacity=0.20] (162.46, 92.21) circle (  1.96);
\definecolor{drawColor}{RGB}{78,155,133}

\path[draw=drawColor,line width= 0.6pt,line join=round] (162.46, 72.15) -- (162.46, 76.01);

\path[draw=drawColor,line width= 0.6pt,line join=round] (162.46, 67.90) -- (162.46, 64.01);

\path[draw=drawColor,line width= 0.6pt,fill=fillColor,fill opacity=0.20] (152.29, 72.15) --
	(152.29, 67.90) --
	(172.63, 67.90) --
	(172.63, 72.15) --
	(152.29, 72.15) --
	cycle;

\path[draw=drawColor,line width= 1.1pt] (152.29, 69.98) -- (172.63, 69.98);
\definecolor{drawColor}{RGB}{78,155,133}

\path[draw=drawColor,draw opacity=0.20,line width= 0.4pt,line join=round,line cap=round,fill=fillColor,fill opacity=0.20] (189.59, 43.50) circle (  1.96);

\path[draw=drawColor,draw opacity=0.20,line width= 0.4pt,line join=round,line cap=round,fill=fillColor,fill opacity=0.20] (189.59, 39.45) circle (  1.96);

\path[draw=drawColor,draw opacity=0.20,line width= 0.4pt,line join=round,line cap=round,fill=fillColor,fill opacity=0.20] (189.59, 52.67) circle (  1.96);

\path[draw=drawColor,draw opacity=0.20,line width= 0.4pt,line join=round,line cap=round,fill=fillColor,fill opacity=0.20] (189.59, 96.40) circle (  1.96);

\path[draw=drawColor,draw opacity=0.20,line width= 0.4pt,line join=round,line cap=round,fill=fillColor,fill opacity=0.20] (189.59,  2.93) circle (  1.96);

\path[draw=drawColor,draw opacity=0.20,line width= 0.4pt,line join=round,line cap=round,fill=fillColor,fill opacity=0.20] (189.59,129.11) circle (  1.96);
\definecolor{drawColor}{RGB}{78,155,133}

\path[draw=drawColor,line width= 0.6pt,line join=round] (189.59, 77.76) -- (189.59, 91.99);

\path[draw=drawColor,line width= 0.6pt,line join=round] (189.59, 67.77) -- (189.59, 63.60);

\path[draw=drawColor,line width= 0.6pt,fill=fillColor,fill opacity=0.20] (179.41, 77.76) --
	(179.41, 67.77) --
	(199.76, 67.77) --
	(199.76, 77.76) --
	(179.41, 77.76) --
	cycle;

\path[draw=drawColor,line width= 1.1pt] (179.41, 69.97) -- (199.76, 69.97);
\definecolor{drawColor}{RGB}{78,155,133}

\path[draw=drawColor,draw opacity=0.20,line width= 0.4pt,line join=round,line cap=round,fill=fillColor,fill opacity=0.20] (216.71, 40.14) circle (  1.96);

\path[draw=drawColor,draw opacity=0.20,line width= 0.4pt,line join=round,line cap=round,fill=fillColor,fill opacity=0.20] (216.71, 35.69) circle (  1.96);
\definecolor{drawColor}{RGB}{78,155,133}

\path[draw=drawColor,line width= 0.6pt,line join=round] (216.71, 87.04) -- (216.71,101.40);

\path[draw=drawColor,line width= 0.6pt,line join=round] (216.71, 68.31) -- (216.71, 53.91);

\path[draw=drawColor,line width= 0.6pt,fill=fillColor,fill opacity=0.20] (206.54, 87.04) --
	(206.54, 68.31) --
	(226.89, 68.31) --
	(226.89, 87.04) --
	(206.54, 87.04) --
	cycle;

\path[draw=drawColor,line width= 1.1pt] (206.54, 70.50) -- (226.89, 70.50);
\end{scope}
\begin{scope}
\path[clip] (  0.00,  0.00) rectangle (238.49,115.63);
\definecolor{drawColor}{RGB}{0,0,0}

\path[draw=drawColor,line width= 0.6pt,line join=round] ( 37.67, 29.80) --
	( 37.67,110.13);

\path[draw=drawColor,line width= 0.6pt,line join=round] ( 39.09,107.67) --
	( 37.67,110.13) --
	( 36.25,107.67);
\end{scope}
\begin{scope}
\path[clip] (  0.00,  0.00) rectangle (238.49,115.63);
\definecolor{drawColor}{gray}{0.30}

\node[text=drawColor,anchor=base east,inner sep=0pt, outer sep=0pt, scale=  0.88] at ( 32.72, 42.59) {0.98};

\node[text=drawColor,anchor=base east,inner sep=0pt, outer sep=0pt, scale=  0.88] at ( 32.72, 66.94) {1.00};

\node[text=drawColor,anchor=base east,inner sep=0pt, outer sep=0pt, scale=  0.88] at ( 32.72, 91.28) {1.02};
\end{scope}
\begin{scope}
\path[clip] (  0.00,  0.00) rectangle (238.49,115.63);
\definecolor{drawColor}{gray}{0.20}

\path[draw=drawColor,line width= 0.6pt,line join=round] ( 34.92, 45.63) --
	( 37.67, 45.63);

\path[draw=drawColor,line width= 0.6pt,line join=round] ( 34.92, 69.97) --
	( 37.67, 69.97);

\path[draw=drawColor,line width= 0.6pt,line join=round] ( 34.92, 94.31) --
	( 37.67, 94.31);
\end{scope}
\begin{scope}
\path[clip] (  0.00,  0.00) rectangle (238.49,115.63);
\definecolor{drawColor}{RGB}{0,0,0}

\path[draw=drawColor,line width= 0.6pt,line join=round] ( 37.67, 29.80) --
	(232.99, 29.80);

\path[draw=drawColor,line width= 0.6pt,line join=round] (230.53, 28.38) --
	(232.99, 29.80) --
	(230.53, 31.23);
\end{scope}
\begin{scope}
\path[clip] (  0.00,  0.00) rectangle (238.49,115.63);
\definecolor{drawColor}{gray}{0.20}

\path[draw=drawColor,line width= 0.6pt,line join=round] ( 53.95, 27.05) --
	( 53.95, 29.80);

\path[draw=drawColor,line width= 0.6pt,line join=round] ( 81.08, 27.05) --
	( 81.08, 29.80);

\path[draw=drawColor,line width= 0.6pt,line join=round] (108.20, 27.05) --
	(108.20, 29.80);

\path[draw=drawColor,line width= 0.6pt,line join=round] (135.33, 27.05) --
	(135.33, 29.80);

\path[draw=drawColor,line width= 0.6pt,line join=round] (162.46, 27.05) --
	(162.46, 29.80);

\path[draw=drawColor,line width= 0.6pt,line join=round] (189.59, 27.05) --
	(189.59, 29.80);

\path[draw=drawColor,line width= 0.6pt,line join=round] (216.71, 27.05) --
	(216.71, 29.80);
\end{scope}
\begin{scope}
\path[clip] (  0.00,  0.00) rectangle (238.49,115.63);
\definecolor{drawColor}{gray}{0.30}

\node[text=drawColor,anchor=base,inner sep=0pt, outer sep=0pt, scale=  0.88] at ( 53.95, 18.79) {2};

\node[text=drawColor,anchor=base,inner sep=0pt, outer sep=0pt, scale=  0.88] at ( 81.08, 18.79) {4};

\node[text=drawColor,anchor=base,inner sep=0pt, outer sep=0pt, scale=  0.88] at (108.20, 18.79) {8};

\node[text=drawColor,anchor=base,inner sep=0pt, outer sep=0pt, scale=  0.88] at (135.33, 18.79) {16};

\node[text=drawColor,anchor=base,inner sep=0pt, outer sep=0pt, scale=  0.88] at (162.46, 18.79) {32};

\node[text=drawColor,anchor=base,inner sep=0pt, outer sep=0pt, scale=  0.88] at (189.59, 18.79) {64};

\node[text=drawColor,anchor=base,inner sep=0pt, outer sep=0pt, scale=  0.88] at (216.71, 18.79) {128};
\end{scope}
\begin{scope}
\path[clip] (  0.00,  0.00) rectangle (238.49,115.63);
\definecolor{drawColor}{RGB}{0,0,0}

\node[text=drawColor,anchor=base,inner sep=0pt, outer sep=0pt, scale=  1.00] at (135.33,  7.44) {domain size};
\end{scope}
\begin{scope}
\path[clip] (  0.00,  0.00) rectangle (238.49,115.63);
\definecolor{drawColor}{RGB}{0,0,0}

\node[text=drawColor,rotate= 90.00,anchor=base,inner sep=0pt, outer sep=0pt, scale=  1.00] at ( 12.39, 69.97) {$p' \mathbin{/} p$};
\end{scope}
\end{tikzpicture}

%% file: files/plot-offline-p=0.1.tex
\begin{tikzpicture}[x=1pt,y=1pt]
\definecolor{fillColor}{RGB}{255,255,255}
\path[use as bounding box,fill=fillColor,fill opacity=0.00] (0,0) rectangle (238.49,115.63);
\begin{scope}
\path[clip] (  0.00,  0.00) rectangle (238.49,115.63);
\definecolor{drawColor}{RGB}{255,255,255}
\definecolor{fillColor}{RGB}{255,255,255}

\path[draw=drawColor,line width= 0.6pt,line join=round,line cap=round,fill=fillColor] (  0.00,  0.00) rectangle (238.49,115.63);
\end{scope}
\begin{scope}
\path[clip] ( 46.47, 29.80) rectangle (232.99,110.13);
\definecolor{fillColor}{RGB}{255,255,255}

\path[fill=fillColor] ( 46.47, 29.80) rectangle (232.99,110.13);
\definecolor{drawColor}{RGB}{247,192,26}
\definecolor{fillColor}{RGB}{247,192,26}

\path[draw=drawColor,draw opacity=0.20,line width= 0.4pt,line join=round,line cap=round,fill=fillColor,fill opacity=0.20] ( 55.54, 51.22) circle (  1.96);
\definecolor{drawColor}{RGB}{247,192,26}

\path[draw=drawColor,line width= 0.6pt,line join=round] ( 55.54, 38.67) -- ( 55.54, 40.66);

\path[draw=drawColor,line width= 0.6pt,line join=round] ( 55.54, 33.45) -- ( 55.54, 33.45);
\definecolor{fillColor}{RGB}{255,255,255}

\path[draw=drawColor,line width= 0.6pt,fill=fillColor,fill opacity=0.20] ( 52.62, 38.67) --
	( 52.62, 33.45) --
	( 58.45, 33.45) --
	( 58.45, 38.67) --
	( 52.62, 38.67) --
	cycle;

\path[draw=drawColor,line width= 1.1pt] ( 52.62, 35.73) -- ( 58.45, 35.73);
\definecolor{drawColor}{RGB}{37,122,164}

\path[draw=drawColor,line width= 0.6pt,line join=round] ( 62.01, 44.54) -- ( 62.01, 45.21);

\path[draw=drawColor,line width= 0.6pt,line join=round] ( 62.01, 33.45) -- ( 62.01, 33.45);

\path[draw=drawColor,line width= 0.6pt,fill=fillColor,fill opacity=0.20] ( 59.10, 44.54) --
	( 59.10, 33.45) --
	( 64.93, 33.45) --
	( 64.93, 44.54) --
	( 59.10, 44.54) --
	cycle;

\path[draw=drawColor,line width= 1.1pt] ( 59.10, 38.00) -- ( 64.93, 38.00);
\definecolor{drawColor}{RGB}{78,155,133}

\path[draw=drawColor,line width= 0.6pt,line join=round] ( 68.49, 61.23) -- ( 68.49, 61.23);

\path[draw=drawColor,line width= 0.6pt,line join=round] ( 68.49, 61.23) -- ( 68.49, 61.23);

\path[draw=drawColor,line width= 0.6pt,fill=fillColor,fill opacity=0.20] ( 65.58, 61.23) --
	( 65.58, 61.23) --
	( 71.40, 61.23) --
	( 71.40, 61.23) --
	( 65.58, 61.23) --
	cycle;

\path[draw=drawColor,line width= 1.1pt] ( 65.58, 61.23) -- ( 71.40, 61.23);
\definecolor{drawColor}{RGB}{247,192,26}
\definecolor{fillColor}{RGB}{247,192,26}

\path[draw=drawColor,draw opacity=0.20,line width= 0.4pt,line join=round,line cap=round,fill=fillColor,fill opacity=0.20] ( 81.44, 47.10) circle (  1.96);
\definecolor{drawColor}{RGB}{247,192,26}

\path[draw=drawColor,line width= 0.6pt,line join=round] ( 81.44, 41.50) -- ( 81.44, 44.01);

\path[draw=drawColor,line width= 0.6pt,line join=round] ( 81.44, 38.00) -- ( 81.44, 38.00);
\definecolor{fillColor}{RGB}{255,255,255}

\path[draw=drawColor,line width= 0.6pt,fill=fillColor,fill opacity=0.20] ( 78.53, 41.50) --
	( 78.53, 38.00) --
	( 84.36, 38.00) --
	( 84.36, 41.50) --
	( 78.53, 41.50) --
	cycle;

\path[draw=drawColor,line width= 1.1pt] ( 78.53, 39.33) -- ( 84.36, 39.33);
\definecolor{drawColor}{RGB}{37,122,164}

\path[draw=drawColor,line width= 0.6pt,line join=round] ( 87.92, 45.21) -- ( 87.92, 47.87);

\path[draw=drawColor,line width= 0.6pt,line join=round] ( 87.92, 38.00) -- ( 87.92, 33.45);

\path[draw=drawColor,line width= 0.6pt,fill=fillColor,fill opacity=0.20] ( 85.00, 45.21) --
	( 85.00, 38.00) --
	( 90.83, 38.00) --
	( 90.83, 45.21) --
	( 85.00, 45.21) --
	cycle;

\path[draw=drawColor,line width= 1.1pt] ( 85.00, 42.55) -- ( 90.83, 42.55);
\definecolor{drawColor}{RGB}{78,155,133}

\path[draw=drawColor,line width= 0.6pt,line join=round] ( 94.40, 47.10) -- ( 94.40, 49.76);

\path[draw=drawColor,line width= 0.6pt,line join=round] ( 94.40, 40.66) -- ( 94.40, 33.45);

\path[draw=drawColor,line width= 0.6pt,fill=fillColor,fill opacity=0.20] ( 91.48, 47.10) --
	( 91.48, 40.66) --
	( 97.31, 40.66) --
	( 97.31, 47.10) --
	( 91.48, 47.10) --
	cycle;

\path[draw=drawColor,line width= 1.1pt] ( 91.48, 42.55) -- ( 97.31, 42.55);
\definecolor{drawColor}{RGB}{247,192,26}
\definecolor{fillColor}{RGB}{247,192,26}

\path[draw=drawColor,draw opacity=0.20,line width= 0.4pt,line join=round,line cap=round,fill=fillColor,fill opacity=0.20] (107.35, 60.32) circle (  1.96);

\path[draw=drawColor,draw opacity=0.20,line width= 0.4pt,line join=round,line cap=round,fill=fillColor,fill opacity=0.20] (107.35, 65.92) circle (  1.96);

\path[draw=drawColor,draw opacity=0.20,line width= 0.4pt,line join=round,line cap=round,fill=fillColor,fill opacity=0.20] (107.35, 66.50) circle (  1.96);

\path[draw=drawColor,draw opacity=0.20,line width= 0.4pt,line join=round,line cap=round,fill=fillColor,fill opacity=0.20] (107.35, 33.45) circle (  1.96);

\path[draw=drawColor,draw opacity=0.20,line width= 0.4pt,line join=round,line cap=round,fill=fillColor,fill opacity=0.20] (107.35, 33.45) circle (  1.96);

\path[draw=drawColor,draw opacity=0.20,line width= 0.4pt,line join=round,line cap=round,fill=fillColor,fill opacity=0.20] (107.35, 33.45) circle (  1.96);
\definecolor{drawColor}{RGB}{247,192,26}

\path[draw=drawColor,line width= 0.6pt,line join=round] (107.35, 47.10) -- (107.35, 47.10);

\path[draw=drawColor,line width= 0.6pt,line join=round] (107.35, 44.01) -- (107.35, 44.01);
\definecolor{fillColor}{RGB}{255,255,255}

\path[draw=drawColor,line width= 0.6pt,fill=fillColor,fill opacity=0.20] (104.43, 47.10) --
	(104.43, 44.01) --
	(110.26, 44.01) --
	(110.26, 47.10) --
	(104.43, 47.10) --
	cycle;

\path[draw=drawColor,line width= 1.1pt] (104.43, 44.01) -- (110.26, 44.01);
\definecolor{drawColor}{RGB}{37,122,164}
\definecolor{fillColor}{RGB}{37,122,164}

\path[draw=drawColor,draw opacity=0.20,line width= 0.4pt,line join=round,line cap=round,fill=fillColor,fill opacity=0.20] (113.82, 60.09) circle (  1.96);

\path[draw=drawColor,draw opacity=0.20,line width= 0.4pt,line join=round,line cap=round,fill=fillColor,fill opacity=0.20] (113.82, 68.63) circle (  1.96);

\path[draw=drawColor,draw opacity=0.20,line width= 0.4pt,line join=round,line cap=round,fill=fillColor,fill opacity=0.20] (113.82, 66.33) circle (  1.96);

\path[draw=drawColor,draw opacity=0.20,line width= 0.4pt,line join=round,line cap=round,fill=fillColor,fill opacity=0.20] (113.82, 33.45) circle (  1.96);

\path[draw=drawColor,draw opacity=0.20,line width= 0.4pt,line join=round,line cap=round,fill=fillColor,fill opacity=0.20] (113.82, 33.45) circle (  1.96);
\definecolor{drawColor}{RGB}{37,122,164}

\path[draw=drawColor,line width= 0.6pt,line join=round] (113.82, 47.10) -- (113.82, 47.10);

\path[draw=drawColor,line width= 0.6pt,line join=round] (113.82, 42.55) -- (113.82, 38.00);
\definecolor{fillColor}{RGB}{255,255,255}

\path[draw=drawColor,line width= 0.6pt,fill=fillColor,fill opacity=0.20] (110.91, 47.10) --
	(110.91, 42.55) --
	(116.74, 42.55) --
	(116.74, 47.10) --
	(110.91, 47.10) --
	cycle;

\path[draw=drawColor,line width= 1.1pt] (110.91, 44.01) -- (116.74, 44.01);
\definecolor{drawColor}{RGB}{78,155,133}
\definecolor{fillColor}{RGB}{78,155,133}

\path[draw=drawColor,draw opacity=0.20,line width= 0.4pt,line join=round,line cap=round,fill=fillColor,fill opacity=0.20] (120.30, 71.11) circle (  1.96);

\path[draw=drawColor,draw opacity=0.20,line width= 0.4pt,line join=round,line cap=round,fill=fillColor,fill opacity=0.20] (120.30, 81.98) circle (  1.96);
\definecolor{drawColor}{RGB}{78,155,133}

\path[draw=drawColor,line width= 0.6pt,line join=round] (120.30, 50.79) -- (120.30, 61.87);

\path[draw=drawColor,line width= 0.6pt,line join=round] (120.30, 42.55) -- (120.30, 33.45);
\definecolor{fillColor}{RGB}{255,255,255}

\path[draw=drawColor,line width= 0.6pt,fill=fillColor,fill opacity=0.20] (117.39, 50.79) --
	(117.39, 42.55) --
	(123.22, 42.55) --
	(123.22, 50.79) --
	(117.39, 50.79) --
	cycle;

\path[draw=drawColor,line width= 1.1pt] (117.39, 44.61) -- (123.22, 44.61);
\definecolor{drawColor}{RGB}{247,192,26}

\path[draw=drawColor,line width= 0.6pt,line join=round] (133.25, 64.96) -- (133.25, 82.52);

\path[draw=drawColor,line width= 0.6pt,line join=round] (133.25, 44.64) -- (133.25, 38.00);

\path[draw=drawColor,line width= 0.6pt,fill=fillColor,fill opacity=0.20] (130.34, 64.96) --
	(130.34, 44.64) --
	(136.17, 44.64) --
	(136.17, 64.96) --
	(130.34, 64.96) --
	cycle;

\path[draw=drawColor,line width= 1.1pt] (130.34, 59.53) -- (136.17, 59.53);
\definecolor{drawColor}{RGB}{37,122,164}

\path[draw=drawColor,line width= 0.6pt,line join=round] (139.73, 67.05) -- (139.73, 79.65);

\path[draw=drawColor,line width= 0.6pt,line join=round] (139.73, 45.14) -- (139.73, 38.00);

\path[draw=drawColor,line width= 0.6pt,fill=fillColor,fill opacity=0.20] (136.82, 67.05) --
	(136.82, 45.14) --
	(142.64, 45.14) --
	(142.64, 67.05) --
	(136.82, 67.05) --
	cycle;

\path[draw=drawColor,line width= 1.1pt] (136.82, 59.36) -- (142.64, 59.36);
\definecolor{drawColor}{RGB}{78,155,133}

\path[draw=drawColor,line width= 0.6pt,line join=round] (146.21, 67.09) -- (146.21, 81.50);

\path[draw=drawColor,line width= 0.6pt,line join=round] (146.21, 44.69) -- (146.21, 38.00);

\path[draw=drawColor,line width= 0.6pt,fill=fillColor,fill opacity=0.20] (143.29, 67.09) --
	(143.29, 44.69) --
	(149.12, 44.69) --
	(149.12, 67.09) --
	(143.29, 67.09) --
	cycle;

\path[draw=drawColor,line width= 1.1pt] (143.29, 59.56) -- (149.12, 59.56);
\definecolor{drawColor}{RGB}{247,192,26}

\path[draw=drawColor,line width= 0.6pt,line join=round] (159.16, 75.96) -- (159.16, 85.17);

\path[draw=drawColor,line width= 0.6pt,line join=round] (159.16, 56.23) -- (159.16, 54.03);

\path[draw=drawColor,line width= 0.6pt,fill=fillColor,fill opacity=0.20] (156.25, 75.96) --
	(156.25, 56.23) --
	(162.07, 56.23) --
	(162.07, 75.96) --
	(156.25, 75.96) --
	cycle;

\path[draw=drawColor,line width= 1.1pt] (156.25, 59.74) -- (162.07, 59.74);
\definecolor{drawColor}{RGB}{37,122,164}

\path[draw=drawColor,line width= 0.6pt,line join=round] (165.64, 83.65) -- (165.64, 91.35);

\path[draw=drawColor,line width= 0.6pt,line join=round] (165.64, 59.50) -- (165.64, 55.32);

\path[draw=drawColor,line width= 0.6pt,fill=fillColor,fill opacity=0.20] (162.72, 83.65) --
	(162.72, 59.50) --
	(168.55, 59.50) --
	(168.55, 83.65) --
	(162.72, 83.65) --
	cycle;

\path[draw=drawColor,line width= 1.1pt] (162.72, 70.70) -- (168.55, 70.70);
\definecolor{drawColor}{RGB}{78,155,133}

\path[draw=drawColor,line width= 0.6pt,line join=round] (172.11, 81.33) -- (172.11, 87.65);

\path[draw=drawColor,line width= 0.6pt,line join=round] (172.11, 58.96) -- (172.11, 55.32);

\path[draw=drawColor,line width= 0.6pt,fill=fillColor,fill opacity=0.20] (169.20, 81.33) --
	(169.20, 58.96) --
	(175.03, 58.96) --
	(175.03, 81.33) --
	(169.20, 81.33) --
	cycle;

\path[draw=drawColor,line width= 1.1pt] (169.20, 67.28) -- (175.03, 67.28);
\definecolor{drawColor}{RGB}{247,192,26}

\path[draw=drawColor,line width= 0.6pt,line join=round] (185.07, 94.96) -- (185.07,105.48);

\path[draw=drawColor,line width= 0.6pt,line join=round] (185.07, 62.90) -- (185.07, 61.23);

\path[draw=drawColor,line width= 0.6pt,fill=fillColor,fill opacity=0.20] (182.15, 94.96) --
	(182.15, 62.90) --
	(187.98, 62.90) --
	(187.98, 94.96) --
	(182.15, 94.96) --
	cycle;

\path[draw=drawColor,line width= 1.1pt] (182.15, 72.51) -- (187.98, 72.51);
\definecolor{drawColor}{RGB}{37,122,164}

\path[draw=drawColor,line width= 0.6pt,line join=round] (191.54, 82.06) -- (191.54,106.48);

\path[draw=drawColor,line width= 0.6pt,line join=round] (191.54, 62.47) -- (191.54, 60.74);

\path[draw=drawColor,line width= 0.6pt,fill=fillColor,fill opacity=0.20] (188.63, 82.06) --
	(188.63, 62.47) --
	(194.46, 62.47) --
	(194.46, 82.06) --
	(188.63, 82.06) --
	cycle;

\path[draw=drawColor,line width= 1.1pt] (188.63, 69.01) -- (194.46, 69.01);
\definecolor{drawColor}{RGB}{78,155,133}
\definecolor{fillColor}{RGB}{78,155,133}

\path[draw=drawColor,draw opacity=0.20,line width= 0.4pt,line join=round,line cap=round,fill=fillColor,fill opacity=0.20] (198.02, 98.42) circle (  1.96);

\path[draw=drawColor,draw opacity=0.20,line width= 0.4pt,line join=round,line cap=round,fill=fillColor,fill opacity=0.20] (198.02, 96.58) circle (  1.96);
\definecolor{drawColor}{RGB}{78,155,133}

\path[draw=drawColor,line width= 0.6pt,line join=round] (198.02, 75.34) -- (198.02, 77.02);

\path[draw=drawColor,line width= 0.6pt,line join=round] (198.02, 63.67) -- (198.02, 61.69);
\definecolor{fillColor}{RGB}{255,255,255}

\path[draw=drawColor,line width= 0.6pt,fill=fillColor,fill opacity=0.20] (195.10, 75.34) --
	(195.10, 63.67) --
	(200.93, 63.67) --
	(200.93, 75.34) --
	(195.10, 75.34) --
	cycle;

\path[draw=drawColor,line width= 1.1pt] (195.10, 70.33) -- (200.93, 70.33);
\definecolor{drawColor}{RGB}{247,192,26}

\path[draw=drawColor,line width= 0.6pt,line join=round] (210.97, 74.00) -- (210.97, 74.69);

\path[draw=drawColor,line width= 0.6pt,line join=round] (210.97, 66.54) -- (210.97, 62.04);

\path[draw=drawColor,line width= 0.6pt,fill=fillColor,fill opacity=0.20] (208.06, 74.00) --
	(208.06, 66.54) --
	(213.89, 66.54) --
	(213.89, 74.00) --
	(208.06, 74.00) --
	cycle;

\path[draw=drawColor,line width= 1.1pt] (208.06, 71.46) -- (213.89, 71.46);
\definecolor{drawColor}{RGB}{37,122,164}

\path[draw=drawColor,line width= 0.6pt,line join=round] (217.45, 73.43) -- (217.45, 76.43);

\path[draw=drawColor,line width= 0.6pt,line join=round] (217.45, 66.28) -- (217.45, 61.33);

\path[draw=drawColor,line width= 0.6pt,fill=fillColor,fill opacity=0.20] (214.53, 73.43) --
	(214.53, 66.28) --
	(220.36, 66.28) --
	(220.36, 73.43) --
	(214.53, 73.43) --
	cycle;

\path[draw=drawColor,line width= 1.1pt] (214.53, 72.05) -- (220.36, 72.05);
\definecolor{drawColor}{RGB}{78,155,133}

\path[draw=drawColor,line width= 0.6pt,line join=round] (223.92, 73.18) -- (223.92, 73.71);

\path[draw=drawColor,line width= 0.6pt,line join=round] (223.92, 66.41) -- (223.92, 61.78);

\path[draw=drawColor,line width= 0.6pt,fill=fillColor,fill opacity=0.20] (221.01, 73.18) --
	(221.01, 66.41) --
	(226.84, 66.41) --
	(226.84, 73.18) --
	(221.01, 73.18) --
	cycle;

\path[draw=drawColor,line width= 1.1pt] (221.01, 69.49) -- (226.84, 69.49);
\end{scope}
\begin{scope}
\path[clip] (  0.00,  0.00) rectangle (238.49,115.63);
\definecolor{drawColor}{RGB}{0,0,0}

\path[draw=drawColor,line width= 0.6pt,line join=round] ( 46.47, 29.80) --
	( 46.47,110.13);

\path[draw=drawColor,line width= 0.6pt,line join=round] ( 47.89,107.67) --
	( 46.47,110.13) --
	( 45.05,107.67);
\end{scope}
\begin{scope}
\path[clip] (  0.00,  0.00) rectangle (238.49,115.63);
\definecolor{drawColor}{gray}{0.30}

\node[text=drawColor,anchor=base east,inner sep=0pt, outer sep=0pt, scale=  0.88] at ( 41.52, 45.53) {   0.1};

\node[text=drawColor,anchor=base east,inner sep=0pt, outer sep=0pt, scale=  0.88] at ( 41.52, 75.74) {  10.0};

\node[text=drawColor,anchor=base east,inner sep=0pt, outer sep=0pt, scale=  0.88] at ( 41.52,105.96) {1000.0};
\end{scope}
\begin{scope}
\path[clip] (  0.00,  0.00) rectangle (238.49,115.63);
\definecolor{drawColor}{gray}{0.20}

\path[draw=drawColor,line width= 0.6pt,line join=round] ( 43.72, 48.56) --
	( 46.47, 48.56);

\path[draw=drawColor,line width= 0.6pt,line join=round] ( 43.72, 78.77) --
	( 46.47, 78.77);

\path[draw=drawColor,line width= 0.6pt,line join=round] ( 43.72,108.99) --
	( 46.47,108.99);
\end{scope}
\begin{scope}
\path[clip] (  0.00,  0.00) rectangle (238.49,115.63);
\definecolor{drawColor}{RGB}{0,0,0}

\path[draw=drawColor,line width= 0.6pt,line join=round] ( 46.47, 29.80) --
	(232.99, 29.80);

\path[draw=drawColor,line width= 0.6pt,line join=round] (230.53, 28.38) --
	(232.99, 29.80) --
	(230.53, 31.23);
\end{scope}
\begin{scope}
\path[clip] (  0.00,  0.00) rectangle (238.49,115.63);
\definecolor{drawColor}{gray}{0.20}

\path[draw=drawColor,line width= 0.6pt,line join=round] ( 62.01, 27.05) --
	( 62.01, 29.80);

\path[draw=drawColor,line width= 0.6pt,line join=round] ( 87.92, 27.05) --
	( 87.92, 29.80);

\path[draw=drawColor,line width= 0.6pt,line join=round] (113.82, 27.05) --
	(113.82, 29.80);

\path[draw=drawColor,line width= 0.6pt,line join=round] (139.73, 27.05) --
	(139.73, 29.80);

\path[draw=drawColor,line width= 0.6pt,line join=round] (165.64, 27.05) --
	(165.64, 29.80);

\path[draw=drawColor,line width= 0.6pt,line join=round] (191.54, 27.05) --
	(191.54, 29.80);

\path[draw=drawColor,line width= 0.6pt,line join=round] (217.45, 27.05) --
	(217.45, 29.80);
\end{scope}
\begin{scope}
\path[clip] (  0.00,  0.00) rectangle (238.49,115.63);
\definecolor{drawColor}{gray}{0.30}

\node[text=drawColor,anchor=base,inner sep=0pt, outer sep=0pt, scale=  0.88] at ( 62.01, 18.79) {2};

\node[text=drawColor,anchor=base,inner sep=0pt, outer sep=0pt, scale=  0.88] at ( 87.92, 18.79) {4};

\node[text=drawColor,anchor=base,inner sep=0pt, outer sep=0pt, scale=  0.88] at (113.82, 18.79) {8};

\node[text=drawColor,anchor=base,inner sep=0pt, outer sep=0pt, scale=  0.88] at (139.73, 18.79) {16};

\node[text=drawColor,anchor=base,inner sep=0pt, outer sep=0pt, scale=  0.88] at (165.64, 18.79) {32};

\node[text=drawColor,anchor=base,inner sep=0pt, outer sep=0pt, scale=  0.88] at (191.54, 18.79) {64};

\node[text=drawColor,anchor=base,inner sep=0pt, outer sep=0pt, scale=  0.88] at (217.45, 18.79) {128};
\end{scope}
\begin{scope}
\path[clip] (  0.00,  0.00) rectangle (238.49,115.63);
\definecolor{drawColor}{RGB}{0,0,0}

\node[text=drawColor,anchor=base,inner sep=0pt, outer sep=0pt, scale=  1.00] at (139.73,  7.44) {domain size};
\end{scope}
\begin{scope}
\path[clip] (  0.00,  0.00) rectangle (238.49,115.63);
\definecolor{drawColor}{RGB}{0,0,0}

\node[text=drawColor,rotate= 90.00,anchor=base,inner sep=0pt, outer sep=0pt, scale=  1.00] at ( 12.39, 69.97) {$\alpha$};
\end{scope}
\begin{scope}
\path[clip] (  0.00,  0.00) rectangle (238.49,115.63);

\path[] ( 41.93, 97.41) rectangle (215.15,122.86);
\end{scope}
\begin{scope}
\path[clip] (  0.00,  0.00) rectangle (238.49,115.63);
\definecolor{drawColor}{RGB}{247,192,26}

\path[draw=drawColor,line width= 0.6pt] ( 60.16,104.35) --
	( 60.16,106.52);

\path[draw=drawColor,line width= 0.6pt] ( 60.16,113.75) --
	( 60.16,115.91);
\definecolor{fillColor}{RGB}{255,255,255}

\path[draw=drawColor,line width= 0.6pt,fill=fillColor,fill opacity=0.20] ( 54.74,106.52) rectangle ( 65.58,113.75);

\path[draw=drawColor,line width= 0.6pt] ( 54.74,110.13) --
	( 65.58,110.13);
\end{scope}
\begin{scope}
\path[clip] (  0.00,  0.00) rectangle (238.49,115.63);
\definecolor{drawColor}{RGB}{37,122,164}

\path[draw=drawColor,line width= 0.6pt] (118.23,104.35) --
	(118.23,106.52);

\path[draw=drawColor,line width= 0.6pt] (118.23,113.75) --
	(118.23,115.91);
\definecolor{fillColor}{RGB}{255,255,255}

\path[draw=drawColor,line width= 0.6pt,fill=fillColor,fill opacity=0.20] (112.81,106.52) rectangle (123.65,113.75);

\path[draw=drawColor,line width= 0.6pt] (112.81,110.13) --
	(123.65,110.13);
\end{scope}
\begin{scope}
\path[clip] (  0.00,  0.00) rectangle (238.49,115.63);
\definecolor{drawColor}{RGB}{78,155,133}

\path[draw=drawColor,line width= 0.6pt] (172.30,104.35) --
	(172.30,106.52);

\path[draw=drawColor,line width= 0.6pt] (172.30,113.75) --
	(172.30,115.91);
\definecolor{fillColor}{RGB}{255,255,255}

\path[draw=drawColor,line width= 0.6pt,fill=fillColor,fill opacity=0.20] (166.88,106.52) rectangle (177.72,113.75);

\path[draw=drawColor,line width= 0.6pt] (166.88,110.13) --
	(177.72,110.13);
\end{scope}
\begin{scope}
\path[clip] (  0.00,  0.00) rectangle (238.49,115.63);
\definecolor{drawColor}{RGB}{0,0,0}

\node[text=drawColor,anchor=base west,inner sep=0pt, outer sep=0pt, scale=  0.80] at ( 72.88,107.38) {$\varepsilon=0.001$};
\end{scope}
\begin{scope}
\path[clip] (  0.00,  0.00) rectangle (238.49,115.63);
\definecolor{drawColor}{RGB}{0,0,0}

\node[text=drawColor,anchor=base west,inner sep=0pt, outer sep=0pt, scale=  0.80] at (130.96,107.38) {$\varepsilon=0.01$};
\end{scope}
\begin{scope}
\path[clip] (  0.00,  0.00) rectangle (238.49,115.63);
\definecolor{drawColor}{RGB}{0,0,0}

\node[text=drawColor,anchor=base west,inner sep=0pt, outer sep=0pt, scale=  0.80] at (185.03,107.38) {$\varepsilon=0.1$};
\end{scope}
\end{tikzpicture}

%% file: files/plot-offline-p=0.3.tex
\begin{tikzpicture}[x=1pt,y=1pt]
\definecolor{fillColor}{RGB}{255,255,255}
\path[use as bounding box,fill=fillColor,fill opacity=0.00] (0,0) rectangle (238.49,115.63);
\begin{scope}
\path[clip] (  0.00,  0.00) rectangle (238.49,115.63);
\definecolor{drawColor}{RGB}{255,255,255}
\definecolor{fillColor}{RGB}{255,255,255}

\path[draw=drawColor,line width= 0.6pt,line join=round,line cap=round,fill=fillColor] (  0.00,  0.00) rectangle (238.49,115.63);
\end{scope}
\begin{scope}
\path[clip] ( 46.47, 29.80) rectangle (232.99,110.13);
\definecolor{fillColor}{RGB}{255,255,255}

\path[fill=fillColor] ( 46.47, 29.80) rectangle (232.99,110.13);
\definecolor{drawColor}{RGB}{247,192,26}

\path[draw=drawColor,line width= 0.6pt,line join=round] ( 55.54, 46.30) -- ( 55.54, 51.22);

\path[draw=drawColor,line width= 0.6pt,line join=round] ( 55.54, 33.45) -- ( 55.54, 33.45);
\definecolor{fillColor}{RGB}{255,255,255}

\path[draw=drawColor,line width= 0.6pt,fill=fillColor,fill opacity=0.20] ( 52.62, 46.30) --
	( 52.62, 33.45) --
	( 58.45, 33.45) --
	( 58.45, 46.30) --
	( 52.62, 46.30) --
	cycle;

\path[draw=drawColor,line width= 1.1pt] ( 52.62, 39.73) -- ( 58.45, 39.73);
\definecolor{drawColor}{RGB}{37,122,164}
\definecolor{fillColor}{RGB}{37,122,164}

\path[draw=drawColor,draw opacity=0.20,line width= 0.4pt,line join=round,line cap=round,fill=fillColor,fill opacity=0.20] ( 62.01, 50.18) circle (  1.96);
\definecolor{drawColor}{RGB}{37,122,164}

\path[draw=drawColor,line width= 0.6pt,line join=round] ( 62.01, 39.73) -- ( 62.01, 41.82);

\path[draw=drawColor,line width= 0.6pt,line join=round] ( 62.01, 33.45) -- ( 62.01, 33.45);
\definecolor{fillColor}{RGB}{255,255,255}

\path[draw=drawColor,line width= 0.6pt,fill=fillColor,fill opacity=0.20] ( 59.10, 39.73) --
	( 59.10, 33.45) --
	( 64.93, 33.45) --
	( 64.93, 39.73) --
	( 59.10, 39.73) --
	cycle;

\path[draw=drawColor,line width= 1.1pt] ( 59.10, 33.45) -- ( 64.93, 33.45);
\definecolor{drawColor}{RGB}{78,155,133}

\path[draw=drawColor,line width= 0.6pt,line join=round] ( 68.49, 48.31) -- ( 68.49, 51.22);

\path[draw=drawColor,line width= 0.6pt,line join=round] ( 68.49, 46.00) -- ( 68.49, 46.00);

\path[draw=drawColor,line width= 0.6pt,fill=fillColor,fill opacity=0.20] ( 65.58, 48.31) --
	( 65.58, 46.00) --
	( 71.40, 46.00) --
	( 71.40, 48.31) --
	( 65.58, 48.31) --
	cycle;

\path[draw=drawColor,line width= 1.1pt] ( 65.58, 46.67) -- ( 71.40, 46.67);
\definecolor{drawColor}{RGB}{247,192,26}

\path[draw=drawColor,line width= 0.6pt,line join=round] ( 81.44, 46.79) -- ( 81.44, 55.40);

\path[draw=drawColor,line width= 0.6pt,line join=round] ( 81.44, 40.08) -- ( 81.44, 37.64);

\path[draw=drawColor,line width= 0.6pt,fill=fillColor,fill opacity=0.20] ( 78.53, 46.79) --
	( 78.53, 40.08) --
	( 84.36, 40.08) --
	( 84.36, 46.79) --
	( 78.53, 46.79) --
	cycle;

\path[draw=drawColor,line width= 1.1pt] ( 78.53, 41.82) -- ( 84.36, 41.82);
\definecolor{drawColor}{RGB}{37,122,164}

\path[draw=drawColor,line width= 0.6pt,line join=round] ( 87.92, 53.46) -- ( 87.92, 63.92);

\path[draw=drawColor,line width= 0.6pt,line join=round] ( 87.92, 40.08) -- ( 87.92, 37.64);

\path[draw=drawColor,line width= 0.6pt,fill=fillColor,fill opacity=0.20] ( 85.00, 53.46) --
	( 85.00, 40.08) --
	( 90.83, 40.08) --
	( 90.83, 53.46) --
	( 85.00, 53.46) --
	cycle;

\path[draw=drawColor,line width= 1.1pt] ( 85.00, 41.82) -- ( 90.83, 41.82);
\definecolor{drawColor}{RGB}{78,155,133}

\path[draw=drawColor,line width= 0.6pt,line join=round] ( 94.40, 49.18) -- ( 94.40, 58.15);

\path[draw=drawColor,line width= 0.6pt,line join=round] ( 94.40, 35.11) -- ( 94.40, 33.45);

\path[draw=drawColor,line width= 0.6pt,fill=fillColor,fill opacity=0.20] ( 91.48, 49.18) --
	( 91.48, 35.11) --
	( 97.31, 35.11) --
	( 97.31, 49.18) --
	( 91.48, 49.18) --
	cycle;

\path[draw=drawColor,line width= 1.1pt] ( 91.48, 43.16) -- ( 97.31, 43.16);
\definecolor{drawColor}{RGB}{247,192,26}

\path[draw=drawColor,line width= 0.6pt,line join=round] (107.35, 53.62) -- (107.35, 58.45);

\path[draw=drawColor,line width= 0.6pt,line join=round] (107.35, 39.73) -- (107.35, 33.45);

\path[draw=drawColor,line width= 0.6pt,fill=fillColor,fill opacity=0.20] (104.43, 53.62) --
	(104.43, 39.73) --
	(110.26, 39.73) --
	(110.26, 53.62) --
	(104.43, 53.62) --
	cycle;

\path[draw=drawColor,line width= 1.1pt] (104.43, 47.03) -- (110.26, 47.03);
\definecolor{drawColor}{RGB}{37,122,164}

\path[draw=drawColor,line width= 0.6pt,line join=round] (113.82, 50.28) -- (113.82, 55.55);

\path[draw=drawColor,line width= 0.6pt,line join=round] (113.82, 41.36) -- (113.82, 37.64);

\path[draw=drawColor,line width= 0.6pt,fill=fillColor,fill opacity=0.20] (110.91, 50.28) --
	(110.91, 41.36) --
	(116.74, 41.36) --
	(116.74, 50.28) --
	(110.91, 50.28) --
	cycle;

\path[draw=drawColor,line width= 1.1pt] (110.91, 47.22) -- (116.74, 47.22);
\definecolor{drawColor}{RGB}{78,155,133}

\path[draw=drawColor,line width= 0.6pt,line join=round] (120.30, 52.80) -- (120.30, 57.95);

\path[draw=drawColor,line width= 0.6pt,line join=round] (120.30, 41.90) -- (120.30, 37.64);

\path[draw=drawColor,line width= 0.6pt,fill=fillColor,fill opacity=0.20] (117.39, 52.80) --
	(117.39, 41.90) --
	(123.22, 41.90) --
	(123.22, 52.80) --
	(117.39, 52.80) --
	cycle;

\path[draw=drawColor,line width= 1.1pt] (117.39, 47.63) -- (123.22, 47.63);
\definecolor{drawColor}{RGB}{247,192,26}

\path[draw=drawColor,line width= 0.6pt,line join=round] (133.25, 63.72) -- (133.25, 70.03);

\path[draw=drawColor,line width= 0.6pt,line join=round] (133.25, 47.34) -- (133.25, 46.00);

\path[draw=drawColor,line width= 0.6pt,fill=fillColor,fill opacity=0.20] (130.34, 63.72) --
	(130.34, 47.34) --
	(136.17, 47.34) --
	(136.17, 63.72) --
	(130.34, 63.72) --
	cycle;

\path[draw=drawColor,line width= 1.1pt] (130.34, 53.34) -- (136.17, 53.34);
\definecolor{drawColor}{RGB}{37,122,164}

\path[draw=drawColor,line width= 0.6pt,line join=round] (139.73, 62.68) -- (139.73, 68.89);

\path[draw=drawColor,line width= 0.6pt,line join=round] (139.73, 47.34) -- (139.73, 43.16);

\path[draw=drawColor,line width= 0.6pt,fill=fillColor,fill opacity=0.20] (136.82, 62.68) --
	(136.82, 47.34) --
	(142.64, 47.34) --
	(142.64, 62.68) --
	(136.82, 62.68) --
	cycle;

\path[draw=drawColor,line width= 1.1pt] (136.82, 52.87) -- (142.64, 52.87);
\definecolor{drawColor}{RGB}{78,155,133}

\path[draw=drawColor,line width= 0.6pt,line join=round] (146.21, 61.97) -- (146.21, 66.34);

\path[draw=drawColor,line width= 0.6pt,line join=round] (146.21, 46.00) -- (146.21, 45.19);

\path[draw=drawColor,line width= 0.6pt,fill=fillColor,fill opacity=0.20] (143.29, 61.97) --
	(143.29, 46.00) --
	(149.12, 46.00) --
	(149.12, 61.97) --
	(143.29, 61.97) --
	cycle;

\path[draw=drawColor,line width= 1.1pt] (143.29, 51.53) -- (149.12, 51.53);
\definecolor{drawColor}{RGB}{247,192,26}

\path[draw=drawColor,line width= 0.6pt,line join=round] (159.16, 61.47) -- (159.16, 62.72);

\path[draw=drawColor,line width= 0.6pt,line join=round] (159.16, 55.40) -- (159.16, 49.37);

\path[draw=drawColor,line width= 0.6pt,fill=fillColor,fill opacity=0.20] (156.25, 61.47) --
	(156.25, 55.40) --
	(162.07, 55.40) --
	(162.07, 61.47) --
	(156.25, 61.47) --
	cycle;

\path[draw=drawColor,line width= 1.1pt] (156.25, 59.89) -- (162.07, 59.89);
\definecolor{drawColor}{RGB}{37,122,164}

\path[draw=drawColor,line width= 0.6pt,line join=round] (165.64, 60.80) -- (165.64, 63.68);

\path[draw=drawColor,line width= 0.6pt,line join=round] (165.64, 55.40) -- (165.64, 48.93);

\path[draw=drawColor,line width= 0.6pt,fill=fillColor,fill opacity=0.20] (162.72, 60.80) --
	(162.72, 55.40) --
	(168.55, 55.40) --
	(168.55, 60.80) --
	(162.72, 60.80) --
	cycle;

\path[draw=drawColor,line width= 1.1pt] (162.72, 60.04) -- (168.55, 60.04);
\definecolor{drawColor}{RGB}{78,155,133}

\path[draw=drawColor,line width= 0.6pt,line join=round] (172.11, 62.28) -- (172.11, 64.61);

\path[draw=drawColor,line width= 0.6pt,line join=round] (172.11, 56.68) -- (172.11, 49.79);

\path[draw=drawColor,line width= 0.6pt,fill=fillColor,fill opacity=0.20] (169.20, 62.28) --
	(169.20, 56.68) --
	(175.03, 56.68) --
	(175.03, 62.28) --
	(169.20, 62.28) --
	cycle;

\path[draw=drawColor,line width= 1.1pt] (169.20, 59.66) -- (175.03, 59.66);
\definecolor{drawColor}{RGB}{247,192,26}
\definecolor{fillColor}{RGB}{247,192,26}

\path[draw=drawColor,draw opacity=0.20,line width= 0.4pt,line join=round,line cap=round,fill=fillColor,fill opacity=0.20] (185.07, 87.04) circle (  1.96);

\path[draw=drawColor,draw opacity=0.20,line width= 0.4pt,line join=round,line cap=round,fill=fillColor,fill opacity=0.20] (185.07, 88.57) circle (  1.96);

\path[draw=drawColor,draw opacity=0.20,line width= 0.4pt,line join=round,line cap=round,fill=fillColor,fill opacity=0.20] (185.07, 87.04) circle (  1.96);
\definecolor{drawColor}{RGB}{247,192,26}

\path[draw=drawColor,line width= 0.6pt,line join=round] (185.07, 66.01) -- (185.07, 66.01);

\path[draw=drawColor,line width= 0.6pt,line join=round] (185.07, 54.07) -- (185.07, 52.37);
\definecolor{fillColor}{RGB}{255,255,255}

\path[draw=drawColor,line width= 0.6pt,fill=fillColor,fill opacity=0.20] (182.15, 66.01) --
	(182.15, 54.07) --
	(187.98, 54.07) --
	(187.98, 66.01) --
	(182.15, 66.01) --
	cycle;

\path[draw=drawColor,line width= 1.1pt] (182.15, 57.04) -- (187.98, 57.04);
\definecolor{drawColor}{RGB}{37,122,164}

\path[draw=drawColor,line width= 0.6pt,line join=round] (191.54, 90.51) -- (191.54,101.49);

\path[draw=drawColor,line width= 0.6pt,line join=round] (191.54, 55.24) -- (191.54, 52.37);

\path[draw=drawColor,line width= 0.6pt,fill=fillColor,fill opacity=0.20] (188.63, 90.51) --
	(188.63, 55.24) --
	(194.46, 55.24) --
	(194.46, 90.51) --
	(188.63, 90.51) --
	cycle;

\path[draw=drawColor,line width= 1.1pt] (188.63, 59.34) -- (194.46, 59.34);
\definecolor{drawColor}{RGB}{78,155,133}

\path[draw=drawColor,line width= 0.6pt,line join=round] (198.02, 87.92) -- (198.02,106.48);

\path[draw=drawColor,line width= 0.6pt,line join=round] (198.02, 55.86) -- (198.02, 52.63);

\path[draw=drawColor,line width= 0.6pt,fill=fillColor,fill opacity=0.20] (195.10, 87.92) --
	(195.10, 55.86) --
	(200.93, 55.86) --
	(200.93, 87.92) --
	(195.10, 87.92) --
	cycle;

\path[draw=drawColor,line width= 1.1pt] (195.10, 58.25) -- (200.93, 58.25);
\definecolor{drawColor}{RGB}{247,192,26}

\path[draw=drawColor,line width= 0.6pt,line join=round] (210.97, 68.36) -- (210.97, 69.82);

\path[draw=drawColor,line width= 0.6pt,line join=round] (210.97, 54.73) -- (210.97, 49.79);

\path[draw=drawColor,line width= 0.6pt,fill=fillColor,fill opacity=0.20] (208.06, 68.36) --
	(208.06, 54.73) --
	(213.89, 54.73) --
	(213.89, 68.36) --
	(208.06, 68.36) --
	cycle;

\path[draw=drawColor,line width= 1.1pt] (208.06, 58.82) -- (213.89, 58.82);
\definecolor{drawColor}{RGB}{37,122,164}

\path[draw=drawColor,line width= 0.6pt,line join=round] (217.45, 68.29) -- (217.45, 69.35);

\path[draw=drawColor,line width= 0.6pt,line join=round] (217.45, 54.17) -- (217.45, 50.18);

\path[draw=drawColor,line width= 0.6pt,fill=fillColor,fill opacity=0.20] (214.53, 68.29) --
	(214.53, 54.17) --
	(220.36, 54.17) --
	(220.36, 68.29) --
	(214.53, 68.29) --
	cycle;

\path[draw=drawColor,line width= 1.1pt] (214.53, 59.25) -- (220.36, 59.25);
\definecolor{drawColor}{RGB}{78,155,133}

\path[draw=drawColor,line width= 0.6pt,line join=round] (223.92, 68.19) -- (223.92, 69.38);

\path[draw=drawColor,line width= 0.6pt,line join=round] (223.92, 54.36) -- (223.92, 49.79);

\path[draw=drawColor,line width= 0.6pt,fill=fillColor,fill opacity=0.20] (221.01, 68.19) --
	(221.01, 54.36) --
	(226.84, 54.36) --
	(226.84, 68.19) --
	(221.01, 68.19) --
	cycle;

\path[draw=drawColor,line width= 1.1pt] (221.01, 58.73) -- (226.84, 58.73);
\end{scope}
\begin{scope}
\path[clip] (  0.00,  0.00) rectangle (238.49,115.63);
\definecolor{drawColor}{RGB}{0,0,0}

\path[draw=drawColor,line width= 0.6pt,line join=round] ( 46.47, 29.80) --
	( 46.47,110.13);

\path[draw=drawColor,line width= 0.6pt,line join=round] ( 47.89,107.67) --
	( 46.47,110.13) --
	( 45.05,107.67);
\end{scope}
\begin{scope}
\path[clip] (  0.00,  0.00) rectangle (238.49,115.63);
\definecolor{drawColor}{gray}{0.30}

\node[text=drawColor,anchor=base east,inner sep=0pt, outer sep=0pt, scale=  0.88] at ( 41.52, 44.31) {   0.1};

\node[text=drawColor,anchor=base east,inner sep=0pt, outer sep=0pt, scale=  0.88] at ( 41.52, 72.09) {  10.0};

\node[text=drawColor,anchor=base east,inner sep=0pt, outer sep=0pt, scale=  0.88] at ( 41.52, 99.88) {1000.0};
\end{scope}
\begin{scope}
\path[clip] (  0.00,  0.00) rectangle (238.49,115.63);
\definecolor{drawColor}{gray}{0.20}

\path[draw=drawColor,line width= 0.6pt,line join=round] ( 43.72, 47.34) --
	( 46.47, 47.34);

\path[draw=drawColor,line width= 0.6pt,line join=round] ( 43.72, 75.13) --
	( 46.47, 75.13);

\path[draw=drawColor,line width= 0.6pt,line join=round] ( 43.72,102.91) --
	( 46.47,102.91);
\end{scope}
\begin{scope}
\path[clip] (  0.00,  0.00) rectangle (238.49,115.63);
\definecolor{drawColor}{RGB}{0,0,0}

\path[draw=drawColor,line width= 0.6pt,line join=round] ( 46.47, 29.80) --
	(232.99, 29.80);

\path[draw=drawColor,line width= 0.6pt,line join=round] (230.53, 28.38) --
	(232.99, 29.80) --
	(230.53, 31.23);
\end{scope}
\begin{scope}
\path[clip] (  0.00,  0.00) rectangle (238.49,115.63);
\definecolor{drawColor}{gray}{0.20}

\path[draw=drawColor,line width= 0.6pt,line join=round] ( 62.01, 27.05) --
	( 62.01, 29.80);

\path[draw=drawColor,line width= 0.6pt,line join=round] ( 87.92, 27.05) --
	( 87.92, 29.80);

\path[draw=drawColor,line width= 0.6pt,line join=round] (113.82, 27.05) --
	(113.82, 29.80);

\path[draw=drawColor,line width= 0.6pt,line join=round] (139.73, 27.05) --
	(139.73, 29.80);

\path[draw=drawColor,line width= 0.6pt,line join=round] (165.64, 27.05) --
	(165.64, 29.80);

\path[draw=drawColor,line width= 0.6pt,line join=round] (191.54, 27.05) --
	(191.54, 29.80);

\path[draw=drawColor,line width= 0.6pt,line join=round] (217.45, 27.05) --
	(217.45, 29.80);
\end{scope}
\begin{scope}
\path[clip] (  0.00,  0.00) rectangle (238.49,115.63);
\definecolor{drawColor}{gray}{0.30}

\node[text=drawColor,anchor=base,inner sep=0pt, outer sep=0pt, scale=  0.88] at ( 62.01, 18.79) {2};

\node[text=drawColor,anchor=base,inner sep=0pt, outer sep=0pt, scale=  0.88] at ( 87.92, 18.79) {4};

\node[text=drawColor,anchor=base,inner sep=0pt, outer sep=0pt, scale=  0.88] at (113.82, 18.79) {8};

\node[text=drawColor,anchor=base,inner sep=0pt, outer sep=0pt, scale=  0.88] at (139.73, 18.79) {16};

\node[text=drawColor,anchor=base,inner sep=0pt, outer sep=0pt, scale=  0.88] at (165.64, 18.79) {32};

\node[text=drawColor,anchor=base,inner sep=0pt, outer sep=0pt, scale=  0.88] at (191.54, 18.79) {64};

\node[text=drawColor,anchor=base,inner sep=0pt, outer sep=0pt, scale=  0.88] at (217.45, 18.79) {128};
\end{scope}
\begin{scope}
\path[clip] (  0.00,  0.00) rectangle (238.49,115.63);
\definecolor{drawColor}{RGB}{0,0,0}

\node[text=drawColor,anchor=base,inner sep=0pt, outer sep=0pt, scale=  1.00] at (139.73,  7.44) {domain size};
\end{scope}
\begin{scope}
\path[clip] (  0.00,  0.00) rectangle (238.49,115.63);
\definecolor{drawColor}{RGB}{0,0,0}

\node[text=drawColor,rotate= 90.00,anchor=base,inner sep=0pt, outer sep=0pt, scale=  1.00] at ( 12.39, 69.97) {$\alpha$};
\end{scope}
\begin{scope}
\path[clip] (  0.00,  0.00) rectangle (238.49,115.63);

\path[] ( 41.93, 97.41) rectangle (215.15,122.86);
\end{scope}
\begin{scope}
\path[clip] (  0.00,  0.00) rectangle (238.49,115.63);
\definecolor{drawColor}{RGB}{247,192,26}

\path[draw=drawColor,line width= 0.6pt] ( 60.16,104.35) --
	( 60.16,106.52);

\path[draw=drawColor,line width= 0.6pt] ( 60.16,113.75) --
	( 60.16,115.91);
\definecolor{fillColor}{RGB}{255,255,255}

\path[draw=drawColor,line width= 0.6pt,fill=fillColor,fill opacity=0.20] ( 54.74,106.52) rectangle ( 65.58,113.75);

\path[draw=drawColor,line width= 0.6pt] ( 54.74,110.13) --
	( 65.58,110.13);
\end{scope}
\begin{scope}
\path[clip] (  0.00,  0.00) rectangle (238.49,115.63);
\definecolor{drawColor}{RGB}{37,122,164}

\path[draw=drawColor,line width= 0.6pt] (118.23,104.35) --
	(118.23,106.52);

\path[draw=drawColor,line width= 0.6pt] (118.23,113.75) --
	(118.23,115.91);
\definecolor{fillColor}{RGB}{255,255,255}

\path[draw=drawColor,line width= 0.6pt,fill=fillColor,fill opacity=0.20] (112.81,106.52) rectangle (123.65,113.75);

\path[draw=drawColor,line width= 0.6pt] (112.81,110.13) --
	(123.65,110.13);
\end{scope}
\begin{scope}
\path[clip] (  0.00,  0.00) rectangle (238.49,115.63);
\definecolor{drawColor}{RGB}{78,155,133}

\path[draw=drawColor,line width= 0.6pt] (172.30,104.35) --
	(172.30,106.52);

\path[draw=drawColor,line width= 0.6pt] (172.30,113.75) --
	(172.30,115.91);
\definecolor{fillColor}{RGB}{255,255,255}

\path[draw=drawColor,line width= 0.6pt,fill=fillColor,fill opacity=0.20] (166.88,106.52) rectangle (177.72,113.75);

\path[draw=drawColor,line width= 0.6pt] (166.88,110.13) --
	(177.72,110.13);
\end{scope}
\begin{scope}
\path[clip] (  0.00,  0.00) rectangle (238.49,115.63);
\definecolor{drawColor}{RGB}{0,0,0}

\node[text=drawColor,anchor=base west,inner sep=0pt, outer sep=0pt, scale=  0.80] at ( 72.88,107.38) {$\varepsilon=0.001$};
\end{scope}
\begin{scope}
\path[clip] (  0.00,  0.00) rectangle (238.49,115.63);
\definecolor{drawColor}{RGB}{0,0,0}

\node[text=drawColor,anchor=base west,inner sep=0pt, outer sep=0pt, scale=  0.80] at (130.96,107.38) {$\varepsilon=0.01$};
\end{scope}
\begin{scope}
\path[clip] (  0.00,  0.00) rectangle (238.49,115.63);
\definecolor{drawColor}{RGB}{0,0,0}

\node[text=drawColor,anchor=base west,inner sep=0pt, outer sep=0pt, scale=  0.80] at (185.03,107.38) {$\varepsilon=0.1$};
\end{scope}
\end{tikzpicture}

%% file: files/plot-offline-p=0.5.tex
\begin{tikzpicture}[x=1pt,y=1pt]
\definecolor{fillColor}{RGB}{255,255,255}
\path[use as bounding box,fill=fillColor,fill opacity=0.00] (0,0) rectangle (238.49,115.63);
\begin{scope}
\path[clip] (  0.00,  0.00) rectangle (238.49,115.63);
\definecolor{drawColor}{RGB}{255,255,255}
\definecolor{fillColor}{RGB}{255,255,255}

\path[draw=drawColor,line width= 0.6pt,line join=round,line cap=round,fill=fillColor] (  0.00,  0.00) rectangle (238.49,115.63);
\end{scope}
\begin{scope}
\path[clip] ( 46.47, 29.80) rectangle (232.99,110.13);
\definecolor{fillColor}{RGB}{255,255,255}

\path[fill=fillColor] ( 46.47, 29.80) rectangle (232.99,110.13);
\definecolor{drawColor}{RGB}{247,192,26}

\path[draw=drawColor,line width= 0.6pt,line join=round] ( 55.54, 42.90) -- ( 55.54, 44.00);

\path[draw=drawColor,line width= 0.6pt,line join=round] ( 55.54, 33.45) -- ( 55.54, 33.45);
\definecolor{fillColor}{RGB}{255,255,255}

\path[draw=drawColor,line width= 0.6pt,fill=fillColor,fill opacity=0.20] ( 52.62, 42.90) --
	( 52.62, 33.45) --
	( 58.45, 33.45) --
	( 58.45, 42.90) --
	( 52.62, 42.90) --
	cycle;

\path[draw=drawColor,line width= 1.1pt] ( 52.62, 37.99) -- ( 58.45, 37.99);
\definecolor{drawColor}{RGB}{37,122,164}

\path[draw=drawColor,line width= 0.6pt,line join=round] ( 62.01, 47.74) -- ( 62.01, 49.73);

\path[draw=drawColor,line width= 0.6pt,line join=round] ( 62.01, 33.45) -- ( 62.01, 33.45);

\path[draw=drawColor,line width= 0.6pt,fill=fillColor,fill opacity=0.20] ( 59.10, 47.74) --
	( 59.10, 33.45) --
	( 64.93, 33.45) --
	( 64.93, 47.74) --
	( 59.10, 47.74) --
	cycle;

\path[draw=drawColor,line width= 1.1pt] ( 59.10, 40.26) -- ( 64.93, 40.26);
\definecolor{drawColor}{RGB}{78,155,133}
\definecolor{fillColor}{RGB}{78,155,133}

\path[draw=drawColor,draw opacity=0.20,line width= 0.4pt,line join=round,line cap=round,fill=fillColor,fill opacity=0.20] ( 68.49, 49.73) circle (  1.96);
\definecolor{drawColor}{RGB}{78,155,133}

\path[draw=drawColor,line width= 0.6pt,line join=round] ( 68.49, 37.99) -- ( 68.49, 37.99);

\path[draw=drawColor,line width= 0.6pt,line join=round] ( 68.49, 33.45) -- ( 68.49, 33.45);
\definecolor{fillColor}{RGB}{255,255,255}

\path[draw=drawColor,line width= 0.6pt,fill=fillColor,fill opacity=0.20] ( 65.58, 37.99) --
	( 65.58, 33.45) --
	( 71.40, 33.45) --
	( 71.40, 37.99) --
	( 65.58, 37.99) --
	cycle;

\path[draw=drawColor,line width= 1.1pt] ( 65.58, 33.45) -- ( 71.40, 33.45);
\definecolor{drawColor}{RGB}{247,192,26}

\path[draw=drawColor,line width= 0.6pt,line join=round] ( 81.44, 51.19) -- ( 81.44, 55.73);

\path[draw=drawColor,line width= 0.6pt,line join=round] ( 81.44, 37.99) -- ( 81.44, 33.45);

\path[draw=drawColor,line width= 0.6pt,fill=fillColor,fill opacity=0.20] ( 78.53, 51.19) --
	( 78.53, 37.99) --
	( 84.36, 37.99) --
	( 84.36, 51.19) --
	( 78.53, 51.19) --
	cycle;

\path[draw=drawColor,line width= 1.1pt] ( 78.53, 48.54) -- ( 84.36, 48.54);
\definecolor{drawColor}{RGB}{37,122,164}

\path[draw=drawColor,line width= 0.6pt,line join=round] ( 87.92, 51.90) -- ( 87.92, 58.81);

\path[draw=drawColor,line width= 0.6pt,line join=round] ( 87.92, 41.40) -- ( 87.92, 33.45);

\path[draw=drawColor,line width= 0.6pt,fill=fillColor,fill opacity=0.20] ( 85.00, 51.90) --
	( 85.00, 41.40) --
	( 90.83, 41.40) --
	( 90.83, 51.90) --
	( 85.00, 51.90) --
	cycle;

\path[draw=drawColor,line width= 1.1pt] ( 85.00, 45.19) -- ( 90.83, 45.19);
\definecolor{drawColor}{RGB}{78,155,133}

\path[draw=drawColor,line width= 0.6pt,line join=round] ( 94.40, 53.23) -- ( 94.40, 58.09);

\path[draw=drawColor,line width= 0.6pt,line join=round] ( 94.40, 42.06) -- ( 94.40, 37.99);

\path[draw=drawColor,line width= 0.6pt,fill=fillColor,fill opacity=0.20] ( 91.48, 53.23) --
	( 91.48, 42.06) --
	( 97.31, 42.06) --
	( 97.31, 53.23) --
	( 91.48, 53.23) --
	cycle;

\path[draw=drawColor,line width= 1.1pt] ( 91.48, 47.18) -- ( 97.31, 47.18);
\definecolor{drawColor}{RGB}{247,192,26}

\path[draw=drawColor,line width= 0.6pt,line join=round] (107.35, 55.73) -- (107.35, 62.93);

\path[draw=drawColor,line width= 0.6pt,line join=round] (107.35, 50.26) -- (107.35, 46.20);

\path[draw=drawColor,line width= 0.6pt,fill=fillColor,fill opacity=0.20] (104.43, 55.73) --
	(104.43, 50.26) --
	(110.26, 50.26) --
	(110.26, 55.73) --
	(104.43, 55.73) --
	cycle;

\path[draw=drawColor,line width= 1.1pt] (104.43, 53.08) -- (110.26, 53.08);
\definecolor{drawColor}{RGB}{37,122,164}

\path[draw=drawColor,line width= 0.6pt,line join=round] (113.82, 56.16) -- (113.82, 58.95);

\path[draw=drawColor,line width= 0.6pt,line join=round] (113.82, 49.73) -- (113.82, 46.20);

\path[draw=drawColor,line width= 0.6pt,fill=fillColor,fill opacity=0.20] (110.91, 56.16) --
	(110.91, 49.73) --
	(116.74, 49.73) --
	(116.74, 56.16) --
	(110.91, 56.16) --
	cycle;

\path[draw=drawColor,line width= 1.1pt] (110.91, 52.01) -- (116.74, 52.01);
\definecolor{drawColor}{RGB}{78,155,133}

\path[draw=drawColor,line width= 0.6pt,line join=round] (120.30, 53.25) -- (120.30, 58.09);

\path[draw=drawColor,line width= 0.6pt,line join=round] (120.30, 49.43) -- (120.30, 47.85);

\path[draw=drawColor,line width= 0.6pt,fill=fillColor,fill opacity=0.20] (117.39, 53.25) --
	(117.39, 49.43) --
	(123.22, 49.43) --
	(123.22, 53.25) --
	(117.39, 53.25) --
	cycle;

\path[draw=drawColor,line width= 1.1pt] (117.39, 51.81) -- (123.22, 51.81);
\definecolor{drawColor}{RGB}{247,192,26}

\path[draw=drawColor,line width= 0.6pt,line join=round] (133.25, 63.60) -- (133.25, 77.70);

\path[draw=drawColor,line width= 0.6pt,line join=round] (133.25, 51.51) -- (133.25, 50.74);

\path[draw=drawColor,line width= 0.6pt,fill=fillColor,fill opacity=0.20] (130.34, 63.60) --
	(130.34, 51.51) --
	(136.17, 51.51) --
	(136.17, 63.60) --
	(130.34, 63.60) --
	cycle;

\path[draw=drawColor,line width= 1.1pt] (130.34, 54.45) -- (136.17, 54.45);
\definecolor{drawColor}{RGB}{37,122,164}
\definecolor{fillColor}{RGB}{37,122,164}

\path[draw=drawColor,draw opacity=0.20,line width= 0.4pt,line join=round,line cap=round,fill=fillColor,fill opacity=0.20] (139.73, 81.26) circle (  1.96);
\definecolor{drawColor}{RGB}{37,122,164}

\path[draw=drawColor,line width= 0.6pt,line join=round] (139.73, 61.66) -- (139.73, 70.62);

\path[draw=drawColor,line width= 0.6pt,line join=round] (139.73, 49.99) -- (139.73, 48.54);
\definecolor{fillColor}{RGB}{255,255,255}

\path[draw=drawColor,line width= 0.6pt,fill=fillColor,fill opacity=0.20] (136.82, 61.66) --
	(136.82, 49.99) --
	(142.64, 49.99) --
	(142.64, 61.66) --
	(136.82, 61.66) --
	cycle;

\path[draw=drawColor,line width= 1.1pt] (136.82, 54.27) -- (142.64, 54.27);
\definecolor{drawColor}{RGB}{78,155,133}

\path[draw=drawColor,line width= 0.6pt,line join=round] (146.21, 60.37) -- (146.21, 72.72);

\path[draw=drawColor,line width= 0.6pt,line join=round] (146.21, 50.50) -- (146.21, 48.54);

\path[draw=drawColor,line width= 0.6pt,fill=fillColor,fill opacity=0.20] (143.29, 60.37) --
	(143.29, 50.50) --
	(149.12, 50.50) --
	(149.12, 60.37) --
	(143.29, 60.37) --
	cycle;

\path[draw=drawColor,line width= 1.1pt] (143.29, 51.62) -- (149.12, 51.62);
\definecolor{drawColor}{RGB}{247,192,26}
\definecolor{fillColor}{RGB}{247,192,26}

\path[draw=drawColor,draw opacity=0.20,line width= 0.4pt,line join=round,line cap=round,fill=fillColor,fill opacity=0.20] (159.16, 82.60) circle (  1.96);
\definecolor{drawColor}{RGB}{247,192,26}

\path[draw=drawColor,line width= 0.6pt,line join=round] (159.16, 57.90) -- (159.16, 59.08);

\path[draw=drawColor,line width= 0.6pt,line join=round] (159.16, 53.84) -- (159.16, 50.26);
\definecolor{fillColor}{RGB}{255,255,255}

\path[draw=drawColor,line width= 0.6pt,fill=fillColor,fill opacity=0.20] (156.25, 57.90) --
	(156.25, 53.84) --
	(162.07, 53.84) --
	(162.07, 57.90) --
	(156.25, 57.90) --
	cycle;

\path[draw=drawColor,line width= 1.1pt] (156.25, 54.92) -- (162.07, 54.92);
\definecolor{drawColor}{RGB}{37,122,164}
\definecolor{fillColor}{RGB}{37,122,164}

\path[draw=drawColor,draw opacity=0.20,line width= 0.4pt,line join=round,line cap=round,fill=fillColor,fill opacity=0.20] (165.64, 86.19) circle (  1.96);

\path[draw=drawColor,draw opacity=0.20,line width= 0.4pt,line join=round,line cap=round,fill=fillColor,fill opacity=0.20] (165.64, 92.05) circle (  1.96);

\path[draw=drawColor,draw opacity=0.20,line width= 0.4pt,line join=round,line cap=round,fill=fillColor,fill opacity=0.20] (165.64, 94.28) circle (  1.96);
\definecolor{drawColor}{RGB}{37,122,164}

\path[draw=drawColor,line width= 0.6pt,line join=round] (165.64, 66.14) -- (165.64, 66.14);

\path[draw=drawColor,line width= 0.6pt,line join=round] (165.64, 53.99) -- (165.64, 49.73);
\definecolor{fillColor}{RGB}{255,255,255}

\path[draw=drawColor,line width= 0.6pt,fill=fillColor,fill opacity=0.20] (162.72, 66.14) --
	(162.72, 53.99) --
	(168.55, 53.99) --
	(168.55, 66.14) --
	(162.72, 66.14) --
	cycle;

\path[draw=drawColor,line width= 1.1pt] (162.72, 56.08) -- (168.55, 56.08);
\definecolor{drawColor}{RGB}{78,155,133}
\definecolor{fillColor}{RGB}{78,155,133}

\path[draw=drawColor,draw opacity=0.20,line width= 0.4pt,line join=round,line cap=round,fill=fillColor,fill opacity=0.20] (172.11, 86.90) circle (  1.96);

\path[draw=drawColor,draw opacity=0.20,line width= 0.4pt,line join=round,line cap=round,fill=fillColor,fill opacity=0.20] (172.11, 99.15) circle (  1.96);

\path[draw=drawColor,draw opacity=0.20,line width= 0.4pt,line join=round,line cap=round,fill=fillColor,fill opacity=0.20] (172.11,100.60) circle (  1.96);
\definecolor{drawColor}{RGB}{78,155,133}

\path[draw=drawColor,line width= 0.6pt,line join=round] (172.11, 66.50) -- (172.11, 66.50);

\path[draw=drawColor,line width= 0.6pt,line join=round] (172.11, 53.63) -- (172.11, 49.73);
\definecolor{fillColor}{RGB}{255,255,255}

\path[draw=drawColor,line width= 0.6pt,fill=fillColor,fill opacity=0.20] (169.20, 66.50) --
	(169.20, 53.63) --
	(175.03, 53.63) --
	(175.03, 66.50) --
	(169.20, 66.50) --
	cycle;

\path[draw=drawColor,line width= 1.1pt] (169.20, 55.40) -- (175.03, 55.40);
\definecolor{drawColor}{RGB}{247,192,26}

\path[draw=drawColor,line width= 0.6pt,line join=round] (185.07, 86.08) -- (185.07, 96.88);

\path[draw=drawColor,line width= 0.6pt,line join=round] (185.07, 52.74) -- (185.07, 48.54);

\path[draw=drawColor,line width= 0.6pt,fill=fillColor,fill opacity=0.20] (182.15, 86.08) --
	(182.15, 52.74) --
	(187.98, 52.74) --
	(187.98, 86.08) --
	(182.15, 86.08) --
	cycle;

\path[draw=drawColor,line width= 1.1pt] (182.15, 55.51) -- (187.98, 55.51);
\definecolor{drawColor}{RGB}{37,122,164}

\path[draw=drawColor,line width= 0.6pt,line join=round] (191.54, 91.57) -- (191.54, 93.94);

\path[draw=drawColor,line width= 0.6pt,line join=round] (191.54, 53.08) -- (191.54, 49.73);

\path[draw=drawColor,line width= 0.6pt,fill=fillColor,fill opacity=0.20] (188.63, 91.57) --
	(188.63, 53.08) --
	(194.46, 53.08) --
	(194.46, 91.57) --
	(188.63, 91.57) --
	cycle;

\path[draw=drawColor,line width= 1.1pt] (188.63, 55.51) -- (194.46, 55.51);
\definecolor{drawColor}{RGB}{78,155,133}

\path[draw=drawColor,line width= 0.6pt,line join=round] (198.02, 89.04) -- (198.02, 91.95);

\path[draw=drawColor,line width= 0.6pt,line join=round] (198.02, 53.40) -- (198.02, 49.16);

\path[draw=drawColor,line width= 0.6pt,fill=fillColor,fill opacity=0.20] (195.10, 89.04) --
	(195.10, 53.40) --
	(200.93, 53.40) --
	(200.93, 89.04) --
	(195.10, 89.04) --
	cycle;

\path[draw=drawColor,line width= 1.1pt] (195.10, 53.99) -- (200.93, 53.99);
\definecolor{drawColor}{RGB}{247,192,26}

\path[draw=drawColor,line width= 0.6pt,line join=round] (210.97,101.04) -- (210.97,104.98);

\path[draw=drawColor,line width= 0.6pt,line join=round] (210.97, 54.27) -- (210.97, 49.16);

\path[draw=drawColor,line width= 0.6pt,fill=fillColor,fill opacity=0.20] (208.06,101.04) --
	(208.06, 54.27) --
	(213.89, 54.27) --
	(213.89,101.04) --
	(208.06,101.04) --
	cycle;

\path[draw=drawColor,line width= 1.1pt] (208.06, 57.11) -- (213.89, 57.11);
\definecolor{drawColor}{RGB}{37,122,164}

\path[draw=drawColor,line width= 0.6pt,line join=round] (217.45, 99.64) -- (217.45,104.76);

\path[draw=drawColor,line width= 0.6pt,line join=round] (217.45, 53.99) -- (217.45, 47.85);

\path[draw=drawColor,line width= 0.6pt,fill=fillColor,fill opacity=0.20] (214.53, 99.64) --
	(214.53, 53.99) --
	(220.36, 53.99) --
	(220.36, 99.64) --
	(214.53, 99.64) --
	cycle;

\path[draw=drawColor,line width= 1.1pt] (214.53, 57.28) -- (220.36, 57.28);
\definecolor{drawColor}{RGB}{78,155,133}

\path[draw=drawColor,line width= 0.6pt,line join=round] (223.92, 99.05) -- (223.92,106.48);

\path[draw=drawColor,line width= 0.6pt,line join=round] (223.92, 53.40) -- (223.92, 47.08);

\path[draw=drawColor,line width= 0.6pt,fill=fillColor,fill opacity=0.20] (221.01, 99.05) --
	(221.01, 53.40) --
	(226.84, 53.40) --
	(226.84, 99.05) --
	(221.01, 99.05) --
	cycle;

\path[draw=drawColor,line width= 1.1pt] (221.01, 56.74) -- (226.84, 56.74);
\end{scope}
\begin{scope}
\path[clip] (  0.00,  0.00) rectangle (238.49,115.63);
\definecolor{drawColor}{RGB}{0,0,0}

\path[draw=drawColor,line width= 0.6pt,line join=round] ( 46.47, 29.80) --
	( 46.47,110.13);

\path[draw=drawColor,line width= 0.6pt,line join=round] ( 47.89,107.67) --
	( 46.47,110.13) --
	( 45.05,107.67);
\end{scope}
\begin{scope}
\path[clip] (  0.00,  0.00) rectangle (238.49,115.63);
\definecolor{drawColor}{gray}{0.30}

\node[text=drawColor,anchor=base east,inner sep=0pt, outer sep=0pt, scale=  0.88] at ( 41.52, 45.51) {   0.1};

\node[text=drawColor,anchor=base east,inner sep=0pt, outer sep=0pt, scale=  0.88] at ( 41.52, 75.67) {  10.0};

\node[text=drawColor,anchor=base east,inner sep=0pt, outer sep=0pt, scale=  0.88] at ( 41.52,105.84) {1000.0};
\end{scope}
\begin{scope}
\path[clip] (  0.00,  0.00) rectangle (238.49,115.63);
\definecolor{drawColor}{gray}{0.20}

\path[draw=drawColor,line width= 0.6pt,line join=round] ( 43.72, 48.54) --
	( 46.47, 48.54);

\path[draw=drawColor,line width= 0.6pt,line join=round] ( 43.72, 78.70) --
	( 46.47, 78.70);

\path[draw=drawColor,line width= 0.6pt,line join=round] ( 43.72,108.87) --
	( 46.47,108.87);
\end{scope}
\begin{scope}
\path[clip] (  0.00,  0.00) rectangle (238.49,115.63);
\definecolor{drawColor}{RGB}{0,0,0}

\path[draw=drawColor,line width= 0.6pt,line join=round] ( 46.47, 29.80) --
	(232.99, 29.80);

\path[draw=drawColor,line width= 0.6pt,line join=round] (230.53, 28.38) --
	(232.99, 29.80) --
	(230.53, 31.23);
\end{scope}
\begin{scope}
\path[clip] (  0.00,  0.00) rectangle (238.49,115.63);
\definecolor{drawColor}{gray}{0.20}

\path[draw=drawColor,line width= 0.6pt,line join=round] ( 62.01, 27.05) --
	( 62.01, 29.80);

\path[draw=drawColor,line width= 0.6pt,line join=round] ( 87.92, 27.05) --
	( 87.92, 29.80);

\path[draw=drawColor,line width= 0.6pt,line join=round] (113.82, 27.05) --
	(113.82, 29.80);

\path[draw=drawColor,line width= 0.6pt,line join=round] (139.73, 27.05) --
	(139.73, 29.80);

\path[draw=drawColor,line width= 0.6pt,line join=round] (165.64, 27.05) --
	(165.64, 29.80);

\path[draw=drawColor,line width= 0.6pt,line join=round] (191.54, 27.05) --
	(191.54, 29.80);

\path[draw=drawColor,line width= 0.6pt,line join=round] (217.45, 27.05) --
	(217.45, 29.80);
\end{scope}
\begin{scope}
\path[clip] (  0.00,  0.00) rectangle (238.49,115.63);
\definecolor{drawColor}{gray}{0.30}

\node[text=drawColor,anchor=base,inner sep=0pt, outer sep=0pt, scale=  0.88] at ( 62.01, 18.79) {2};

\node[text=drawColor,anchor=base,inner sep=0pt, outer sep=0pt, scale=  0.88] at ( 87.92, 18.79) {4};

\node[text=drawColor,anchor=base,inner sep=0pt, outer sep=0pt, scale=  0.88] at (113.82, 18.79) {8};

\node[text=drawColor,anchor=base,inner sep=0pt, outer sep=0pt, scale=  0.88] at (139.73, 18.79) {16};

\node[text=drawColor,anchor=base,inner sep=0pt, outer sep=0pt, scale=  0.88] at (165.64, 18.79) {32};

\node[text=drawColor,anchor=base,inner sep=0pt, outer sep=0pt, scale=  0.88] at (191.54, 18.79) {64};

\node[text=drawColor,anchor=base,inner sep=0pt, outer sep=0pt, scale=  0.88] at (217.45, 18.79) {128};
\end{scope}
\begin{scope}
\path[clip] (  0.00,  0.00) rectangle (238.49,115.63);
\definecolor{drawColor}{RGB}{0,0,0}

\node[text=drawColor,anchor=base,inner sep=0pt, outer sep=0pt, scale=  1.00] at (139.73,  7.44) {domain size};
\end{scope}
\begin{scope}
\path[clip] (  0.00,  0.00) rectangle (238.49,115.63);
\definecolor{drawColor}{RGB}{0,0,0}

\node[text=drawColor,rotate= 90.00,anchor=base,inner sep=0pt, outer sep=0pt, scale=  1.00] at ( 12.39, 69.97) {$\alpha$};
\end{scope}
\begin{scope}
\path[clip] (  0.00,  0.00) rectangle (238.49,115.63);

\path[] ( 41.93, 97.41) rectangle (215.15,122.86);
\end{scope}
\begin{scope}
\path[clip] (  0.00,  0.00) rectangle (238.49,115.63);
\definecolor{drawColor}{RGB}{247,192,26}

\path[draw=drawColor,line width= 0.6pt] ( 60.16,104.35) --
	( 60.16,106.52);

\path[draw=drawColor,line width= 0.6pt] ( 60.16,113.75) --
	( 60.16,115.91);
\definecolor{fillColor}{RGB}{255,255,255}

\path[draw=drawColor,line width= 0.6pt,fill=fillColor,fill opacity=0.20] ( 54.74,106.52) rectangle ( 65.58,113.75);

\path[draw=drawColor,line width= 0.6pt] ( 54.74,110.13) --
	( 65.58,110.13);
\end{scope}
\begin{scope}
\path[clip] (  0.00,  0.00) rectangle (238.49,115.63);
\definecolor{drawColor}{RGB}{37,122,164}

\path[draw=drawColor,line width= 0.6pt] (118.23,104.35) --
	(118.23,106.52);

\path[draw=drawColor,line width= 0.6pt] (118.23,113.75) --
	(118.23,115.91);
\definecolor{fillColor}{RGB}{255,255,255}

\path[draw=drawColor,line width= 0.6pt,fill=fillColor,fill opacity=0.20] (112.81,106.52) rectangle (123.65,113.75);

\path[draw=drawColor,line width= 0.6pt] (112.81,110.13) --
	(123.65,110.13);
\end{scope}
\begin{scope}
\path[clip] (  0.00,  0.00) rectangle (238.49,115.63);
\definecolor{drawColor}{RGB}{78,155,133}

\path[draw=drawColor,line width= 0.6pt] (172.30,104.35) --
	(172.30,106.52);

\path[draw=drawColor,line width= 0.6pt] (172.30,113.75) --
	(172.30,115.91);
\definecolor{fillColor}{RGB}{255,255,255}

\path[draw=drawColor,line width= 0.6pt,fill=fillColor,fill opacity=0.20] (166.88,106.52) rectangle (177.72,113.75);

\path[draw=drawColor,line width= 0.6pt] (166.88,110.13) --
	(177.72,110.13);
\end{scope}
\begin{scope}
\path[clip] (  0.00,  0.00) rectangle (238.49,115.63);
\definecolor{drawColor}{RGB}{0,0,0}

\node[text=drawColor,anchor=base west,inner sep=0pt, outer sep=0pt, scale=  0.80] at ( 72.88,107.38) {$\varepsilon=0.001$};
\end{scope}
\begin{scope}
\path[clip] (  0.00,  0.00) rectangle (238.49,115.63);
\definecolor{drawColor}{RGB}{0,0,0}

\node[text=drawColor,anchor=base west,inner sep=0pt, outer sep=0pt, scale=  0.80] at (130.96,107.38) {$\varepsilon=0.01$};
\end{scope}
\begin{scope}
\path[clip] (  0.00,  0.00) rectangle (238.49,115.63);
\definecolor{drawColor}{RGB}{0,0,0}

\node[text=drawColor,anchor=base west,inner sep=0pt, outer sep=0pt, scale=  0.80] at (185.03,107.38) {$\varepsilon=0.1$};
\end{scope}
\end{tikzpicture}

%% file: files/plot-offline-p=0.7.tex
\begin{tikzpicture}[x=1pt,y=1pt]
\definecolor{fillColor}{RGB}{255,255,255}
\path[use as bounding box,fill=fillColor,fill opacity=0.00] (0,0) rectangle (238.49,115.63);
\begin{scope}
\path[clip] (  0.00,  0.00) rectangle (238.49,115.63);
\definecolor{drawColor}{RGB}{255,255,255}
\definecolor{fillColor}{RGB}{255,255,255}

\path[draw=drawColor,line width= 0.6pt,line join=round,line cap=round,fill=fillColor] (  0.00,  0.00) rectangle (238.49,115.63);
\end{scope}
\begin{scope}
\path[clip] ( 46.47, 29.80) rectangle (232.99,110.13);
\definecolor{fillColor}{RGB}{255,255,255}

\path[fill=fillColor] ( 46.47, 29.80) rectangle (232.99,110.13);
\definecolor{drawColor}{RGB}{247,192,26}
\definecolor{fillColor}{RGB}{247,192,26}

\path[draw=drawColor,draw opacity=0.20,line width= 0.4pt,line join=round,line cap=round,fill=fillColor,fill opacity=0.20] ( 55.54, 56.31) circle (  1.96);
\definecolor{drawColor}{RGB}{247,192,26}

\path[draw=drawColor,line width= 0.6pt,line join=round] ( 55.54, 45.17) -- ( 55.54, 45.17);

\path[draw=drawColor,line width= 0.6pt,line join=round] ( 55.54, 37.99) -- ( 55.54, 33.45);
\definecolor{fillColor}{RGB}{255,255,255}

\path[draw=drawColor,line width= 0.6pt,fill=fillColor,fill opacity=0.20] ( 52.62, 45.17) --
	( 52.62, 37.99) --
	( 58.45, 37.99) --
	( 58.45, 45.17) --
	( 52.62, 45.17) --
	cycle;

\path[draw=drawColor,line width= 1.1pt] ( 52.62, 37.99) -- ( 58.45, 37.99);
\definecolor{drawColor}{RGB}{37,122,164}

\path[draw=drawColor,line width= 0.6pt,line join=round] ( 62.01, 41.47) -- ( 62.01, 43.98);

\path[draw=drawColor,line width= 0.6pt,line join=round] ( 62.01, 33.45) -- ( 62.01, 33.45);

\path[draw=drawColor,line width= 0.6pt,fill=fillColor,fill opacity=0.20] ( 59.10, 41.47) --
	( 59.10, 33.45) --
	( 64.93, 33.45) --
	( 64.93, 41.47) --
	( 59.10, 41.47) --
	cycle;

\path[draw=drawColor,line width= 1.1pt] ( 59.10, 37.05) -- ( 64.93, 37.05);
\definecolor{drawColor}{RGB}{78,155,133}

\path[draw=drawColor,line width= 0.6pt,line join=round] ( 68.49, 51.16) -- ( 68.49, 54.23);

\path[draw=drawColor,line width= 0.6pt,line join=round] ( 68.49, 37.99) -- ( 68.49, 33.45);

\path[draw=drawColor,line width= 0.6pt,fill=fillColor,fill opacity=0.20] ( 65.58, 51.16) --
	( 65.58, 37.99) --
	( 71.40, 37.99) --
	( 71.40, 51.16) --
	( 65.58, 51.16) --
	cycle;

\path[draw=drawColor,line width= 1.1pt] ( 65.58, 37.99) -- ( 71.40, 37.99);
\definecolor{drawColor}{RGB}{247,192,26}

\path[draw=drawColor,line width= 0.6pt,line join=round] ( 81.44, 51.16) -- ( 81.44, 60.54);

\path[draw=drawColor,line width= 0.6pt,line join=round] ( 81.44, 40.64) -- ( 81.44, 37.99);

\path[draw=drawColor,line width= 0.6pt,fill=fillColor,fill opacity=0.20] ( 78.53, 51.16) --
	( 78.53, 40.64) --
	( 84.36, 40.64) --
	( 84.36, 51.16) --
	( 78.53, 51.16) --
	cycle;

\path[draw=drawColor,line width= 1.1pt] ( 78.53, 43.98) -- ( 84.36, 43.98);
\definecolor{drawColor}{RGB}{37,122,164}

\path[draw=drawColor,line width= 0.6pt,line join=round] ( 87.92, 56.12) -- ( 87.92, 68.53);

\path[draw=drawColor,line width= 0.6pt,line join=round] ( 87.92, 40.64) -- ( 87.92, 37.99);

\path[draw=drawColor,line width= 0.6pt,fill=fillColor,fill opacity=0.20] ( 85.00, 56.12) --
	( 85.00, 40.64) --
	( 90.83, 40.64) --
	( 90.83, 56.12) --
	( 85.00, 56.12) --
	cycle;

\path[draw=drawColor,line width= 1.1pt] ( 85.00, 42.31) -- ( 90.83, 42.31);
\definecolor{drawColor}{RGB}{78,155,133}
\definecolor{fillColor}{RGB}{78,155,133}

\path[draw=drawColor,draw opacity=0.20,line width= 0.4pt,line join=round,line cap=round,fill=fillColor,fill opacity=0.20] ( 94.40, 94.61) circle (  1.96);
\definecolor{drawColor}{RGB}{78,155,133}

\path[draw=drawColor,line width= 0.6pt,line join=round] ( 94.40, 51.98) -- ( 94.40, 66.71);

\path[draw=drawColor,line width= 0.6pt,line join=round] ( 94.40, 40.64) -- ( 94.40, 37.99);
\definecolor{fillColor}{RGB}{255,255,255}

\path[draw=drawColor,line width= 0.6pt,fill=fillColor,fill opacity=0.20] ( 91.48, 51.98) --
	( 91.48, 40.64) --
	( 97.31, 40.64) --
	( 97.31, 51.98) --
	( 91.48, 51.98) --
	cycle;

\path[draw=drawColor,line width= 1.1pt] ( 91.48, 43.98) -- ( 97.31, 43.98);
\definecolor{drawColor}{RGB}{247,192,26}
\definecolor{fillColor}{RGB}{247,192,26}

\path[draw=drawColor,draw opacity=0.20,line width= 0.4pt,line join=round,line cap=round,fill=fillColor,fill opacity=0.20] (107.35, 82.34) circle (  1.96);
\definecolor{drawColor}{RGB}{247,192,26}

\path[draw=drawColor,line width= 0.6pt,line join=round] (107.35, 59.00) -- (107.35, 62.57);

\path[draw=drawColor,line width= 0.6pt,line join=round] (107.35, 50.71) -- (107.35, 50.22);
\definecolor{fillColor}{RGB}{255,255,255}

\path[draw=drawColor,line width= 0.6pt,fill=fillColor,fill opacity=0.20] (104.43, 59.00) --
	(104.43, 50.71) --
	(110.26, 50.71) --
	(110.26, 59.00) --
	(104.43, 59.00) --
	cycle;

\path[draw=drawColor,line width= 1.1pt] (104.43, 53.04) -- (110.26, 53.04);
\definecolor{drawColor}{RGB}{37,122,164}

\path[draw=drawColor,line width= 0.6pt,line join=round] (113.82, 59.22) -- (113.82, 62.02);

\path[draw=drawColor,line width= 0.6pt,line join=round] (113.82, 50.46) -- (113.82, 49.13);

\path[draw=drawColor,line width= 0.6pt,fill=fillColor,fill opacity=0.20] (110.91, 59.22) --
	(110.91, 50.46) --
	(116.74, 50.46) --
	(116.74, 59.22) --
	(110.91, 59.22) --
	cycle;

\path[draw=drawColor,line width= 1.1pt] (110.91, 51.16) -- (116.74, 51.16);
\definecolor{drawColor}{RGB}{78,155,133}

\path[draw=drawColor,line width= 0.6pt,line join=round] (120.30, 55.16) -- (120.30, 60.44);

\path[draw=drawColor,line width= 0.6pt,line join=round] (120.30, 49.27) -- (120.30, 46.18);

\path[draw=drawColor,line width= 0.6pt,fill=fillColor,fill opacity=0.20] (117.39, 55.16) --
	(117.39, 49.27) --
	(123.22, 49.27) --
	(123.22, 55.16) --
	(117.39, 55.16) --
	cycle;

\path[draw=drawColor,line width= 1.1pt] (117.39, 50.22) -- (123.22, 50.22);
\definecolor{drawColor}{RGB}{247,192,26}

\path[draw=drawColor,line width= 0.6pt,line join=round] (133.25, 62.65) -- (133.25, 64.12);

\path[draw=drawColor,line width= 0.6pt,line join=round] (133.25, 53.95) -- (133.25, 50.71);

\path[draw=drawColor,line width= 0.6pt,fill=fillColor,fill opacity=0.20] (130.34, 62.65) --
	(130.34, 53.95) --
	(136.17, 53.95) --
	(136.17, 62.65) --
	(130.34, 62.65) --
	cycle;

\path[draw=drawColor,line width= 1.1pt] (130.34, 56.88) -- (136.17, 56.88);
\definecolor{drawColor}{RGB}{37,122,164}

\path[draw=drawColor,line width= 0.6pt,line join=round] (139.73, 62.65) -- (139.73, 65.94);

\path[draw=drawColor,line width= 0.6pt,line join=round] (139.73, 53.04) -- (139.73, 51.16);

\path[draw=drawColor,line width= 0.6pt,fill=fillColor,fill opacity=0.20] (136.82, 62.65) --
	(136.82, 53.04) --
	(142.64, 53.04) --
	(142.64, 62.65) --
	(136.82, 62.65) --
	cycle;

\path[draw=drawColor,line width= 1.1pt] (136.82, 58.19) -- (142.64, 58.19);
\definecolor{drawColor}{RGB}{78,155,133}

\path[draw=drawColor,line width= 0.6pt,line join=round] (146.21, 63.09) -- (146.21, 65.94);

\path[draw=drawColor,line width= 0.6pt,line join=round] (146.21, 51.16) -- (146.21, 49.70);

\path[draw=drawColor,line width= 0.6pt,fill=fillColor,fill opacity=0.20] (143.29, 63.09) --
	(143.29, 51.16) --
	(149.12, 51.16) --
	(149.12, 63.09) --
	(143.29, 63.09) --
	cycle;

\path[draw=drawColor,line width= 1.1pt] (143.29, 58.19) -- (149.12, 58.19);
\definecolor{drawColor}{RGB}{247,192,26}

\path[draw=drawColor,line width= 0.6pt,line join=round] (159.16, 57.73) -- (159.16, 59.03);

\path[draw=drawColor,line width= 0.6pt,line join=round] (159.16, 51.16) -- (159.16, 50.22);

\path[draw=drawColor,line width= 0.6pt,fill=fillColor,fill opacity=0.20] (156.25, 57.73) --
	(156.25, 51.16) --
	(162.07, 51.16) --
	(162.07, 57.73) --
	(156.25, 57.73) --
	cycle;

\path[draw=drawColor,line width= 1.1pt] (156.25, 53.66) -- (162.07, 53.66);
\definecolor{drawColor}{RGB}{37,122,164}
\definecolor{fillColor}{RGB}{37,122,164}

\path[draw=drawColor,draw opacity=0.20,line width= 0.4pt,line join=round,line cap=round,fill=fillColor,fill opacity=0.20] (165.64,106.48) circle (  1.96);

\path[draw=drawColor,draw opacity=0.20,line width= 0.4pt,line join=round,line cap=round,fill=fillColor,fill opacity=0.20] (165.64, 95.59) circle (  1.96);
\definecolor{drawColor}{RGB}{37,122,164}

\path[draw=drawColor,line width= 0.6pt,line join=round] (165.64, 58.89) -- (165.64, 59.29);

\path[draw=drawColor,line width= 0.6pt,line join=round] (165.64, 51.87) -- (165.64, 50.71);
\definecolor{fillColor}{RGB}{255,255,255}

\path[draw=drawColor,line width= 0.6pt,fill=fillColor,fill opacity=0.20] (162.72, 58.89) --
	(162.72, 51.87) --
	(168.55, 51.87) --
	(168.55, 58.89) --
	(162.72, 58.89) --
	cycle;

\path[draw=drawColor,line width= 1.1pt] (162.72, 54.75) -- (168.55, 54.75);
\definecolor{drawColor}{RGB}{78,155,133}
\definecolor{fillColor}{RGB}{78,155,133}

\path[draw=drawColor,draw opacity=0.20,line width= 0.4pt,line join=round,line cap=round,fill=fillColor,fill opacity=0.20] (172.11, 93.10) circle (  1.96);
\definecolor{drawColor}{RGB}{78,155,133}

\path[draw=drawColor,line width= 0.6pt,line join=round] (172.11, 58.71) -- (172.11, 59.16);

\path[draw=drawColor,line width= 0.6pt,line join=round] (172.11, 51.45) -- (172.11, 47.05);
\definecolor{fillColor}{RGB}{255,255,255}

\path[draw=drawColor,line width= 0.6pt,fill=fillColor,fill opacity=0.20] (169.20, 58.71) --
	(169.20, 51.45) --
	(175.03, 51.45) --
	(175.03, 58.71) --
	(169.20, 58.71) --
	cycle;

\path[draw=drawColor,line width= 1.1pt] (169.20, 55.11) -- (175.03, 55.11);
\definecolor{drawColor}{RGB}{247,192,26}

\path[draw=drawColor,line width= 0.6pt,line join=round] (185.07, 58.19) -- (185.07, 59.77);

\path[draw=drawColor,line width= 0.6pt,line join=round] (185.07, 47.05) -- (185.07, 45.17);

\path[draw=drawColor,line width= 0.6pt,fill=fillColor,fill opacity=0.20] (182.15, 58.19) --
	(182.15, 47.05) --
	(187.98, 47.05) --
	(187.98, 58.19) --
	(182.15, 58.19) --
	cycle;

\path[draw=drawColor,line width= 1.1pt] (182.15, 50.71) -- (187.98, 50.71);
\definecolor{drawColor}{RGB}{37,122,164}

\path[draw=drawColor,line width= 0.6pt,line join=round] (191.54, 57.41) -- (191.54, 59.03);

\path[draw=drawColor,line width= 0.6pt,line join=round] (191.54, 50.71) -- (191.54, 47.05);

\path[draw=drawColor,line width= 0.6pt,fill=fillColor,fill opacity=0.20] (188.63, 57.41) --
	(188.63, 50.71) --
	(194.46, 50.71) --
	(194.46, 57.41) --
	(188.63, 57.41) --
	cycle;

\path[draw=drawColor,line width= 1.1pt] (188.63, 52.70) -- (194.46, 52.70);
\definecolor{drawColor}{RGB}{78,155,133}

\path[draw=drawColor,line width= 0.6pt,line join=round] (198.02, 58.34) -- (198.02, 59.41);

\path[draw=drawColor,line width= 0.6pt,line join=round] (198.02, 50.71) -- (198.02, 47.05);

\path[draw=drawColor,line width= 0.6pt,fill=fillColor,fill opacity=0.20] (195.10, 58.34) --
	(195.10, 50.71) --
	(200.93, 50.71) --
	(200.93, 58.34) --
	(195.10, 58.34) --
	cycle;

\path[draw=drawColor,line width= 1.1pt] (195.10, 52.70) -- (200.93, 52.70);
\definecolor{drawColor}{RGB}{247,192,26}

\path[draw=drawColor,line width= 0.6pt,line join=round] (210.97, 59.41) -- (210.97, 60.33);

\path[draw=drawColor,line width= 0.6pt,line join=round] (210.97, 50.71) -- (210.97, 46.18);

\path[draw=drawColor,line width= 0.6pt,fill=fillColor,fill opacity=0.20] (208.06, 59.41) --
	(208.06, 50.71) --
	(213.89, 50.71) --
	(213.89, 59.41) --
	(208.06, 59.41) --
	cycle;

\path[draw=drawColor,line width= 1.1pt] (208.06, 52.70) -- (213.89, 52.70);
\definecolor{drawColor}{RGB}{37,122,164}

\path[draw=drawColor,line width= 0.6pt,line join=round] (217.45, 59.41) -- (217.45, 62.80);

\path[draw=drawColor,line width= 0.6pt,line join=round] (217.45, 51.16) -- (217.45, 46.18);

\path[draw=drawColor,line width= 0.6pt,fill=fillColor,fill opacity=0.20] (214.53, 59.41) --
	(214.53, 51.16) --
	(220.36, 51.16) --
	(220.36, 59.41) --
	(214.53, 59.41) --
	cycle;

\path[draw=drawColor,line width= 1.1pt] (214.53, 52.35) -- (220.36, 52.35);
\definecolor{drawColor}{RGB}{78,155,133}

\path[draw=drawColor,line width= 0.6pt,line join=round] (223.92, 59.16) -- (223.92, 60.22);

\path[draw=drawColor,line width= 0.6pt,line join=round] (223.92, 49.70) -- (223.92, 47.05);

\path[draw=drawColor,line width= 0.6pt,fill=fillColor,fill opacity=0.20] (221.01, 59.16) --
	(221.01, 49.70) --
	(226.84, 49.70) --
	(226.84, 59.16) --
	(221.01, 59.16) --
	cycle;

\path[draw=drawColor,line width= 1.1pt] (221.01, 52.35) -- (226.84, 52.35);
\end{scope}
\begin{scope}
\path[clip] (  0.00,  0.00) rectangle (238.49,115.63);
\definecolor{drawColor}{RGB}{0,0,0}

\path[draw=drawColor,line width= 0.6pt,line join=round] ( 46.47, 29.80) --
	( 46.47,110.13);

\path[draw=drawColor,line width= 0.6pt,line join=round] ( 47.89,107.67) --
	( 46.47,110.13) --
	( 45.05,107.67);
\end{scope}
\begin{scope}
\path[clip] (  0.00,  0.00) rectangle (238.49,115.63);
\definecolor{drawColor}{gray}{0.30}

\node[text=drawColor,anchor=base east,inner sep=0pt, outer sep=0pt, scale=  0.88] at ( 41.52, 45.48) {   0.1};

\node[text=drawColor,anchor=base east,inner sep=0pt, outer sep=0pt, scale=  0.88] at ( 41.52, 75.58) {  10.0};

\node[text=drawColor,anchor=base east,inner sep=0pt, outer sep=0pt, scale=  0.88] at ( 41.52,105.69) {1000.0};
\end{scope}
\begin{scope}
\path[clip] (  0.00,  0.00) rectangle (238.49,115.63);
\definecolor{drawColor}{gray}{0.20}

\path[draw=drawColor,line width= 0.6pt,line join=round] ( 43.72, 48.51) --
	( 46.47, 48.51);

\path[draw=drawColor,line width= 0.6pt,line join=round] ( 43.72, 78.61) --
	( 46.47, 78.61);

\path[draw=drawColor,line width= 0.6pt,line join=round] ( 43.72,108.72) --
	( 46.47,108.72);
\end{scope}
\begin{scope}
\path[clip] (  0.00,  0.00) rectangle (238.49,115.63);
\definecolor{drawColor}{RGB}{0,0,0}

\path[draw=drawColor,line width= 0.6pt,line join=round] ( 46.47, 29.80) --
	(232.99, 29.80);

\path[draw=drawColor,line width= 0.6pt,line join=round] (230.53, 28.38) --
	(232.99, 29.80) --
	(230.53, 31.23);
\end{scope}
\begin{scope}
\path[clip] (  0.00,  0.00) rectangle (238.49,115.63);
\definecolor{drawColor}{gray}{0.20}

\path[draw=drawColor,line width= 0.6pt,line join=round] ( 62.01, 27.05) --
	( 62.01, 29.80);

\path[draw=drawColor,line width= 0.6pt,line join=round] ( 87.92, 27.05) --
	( 87.92, 29.80);

\path[draw=drawColor,line width= 0.6pt,line join=round] (113.82, 27.05) --
	(113.82, 29.80);

\path[draw=drawColor,line width= 0.6pt,line join=round] (139.73, 27.05) --
	(139.73, 29.80);

\path[draw=drawColor,line width= 0.6pt,line join=round] (165.64, 27.05) --
	(165.64, 29.80);

\path[draw=drawColor,line width= 0.6pt,line join=round] (191.54, 27.05) --
	(191.54, 29.80);

\path[draw=drawColor,line width= 0.6pt,line join=round] (217.45, 27.05) --
	(217.45, 29.80);
\end{scope}
\begin{scope}
\path[clip] (  0.00,  0.00) rectangle (238.49,115.63);
\definecolor{drawColor}{gray}{0.30}

\node[text=drawColor,anchor=base,inner sep=0pt, outer sep=0pt, scale=  0.88] at ( 62.01, 18.79) {2};

\node[text=drawColor,anchor=base,inner sep=0pt, outer sep=0pt, scale=  0.88] at ( 87.92, 18.79) {4};

\node[text=drawColor,anchor=base,inner sep=0pt, outer sep=0pt, scale=  0.88] at (113.82, 18.79) {8};

\node[text=drawColor,anchor=base,inner sep=0pt, outer sep=0pt, scale=  0.88] at (139.73, 18.79) {16};

\node[text=drawColor,anchor=base,inner sep=0pt, outer sep=0pt, scale=  0.88] at (165.64, 18.79) {32};

\node[text=drawColor,anchor=base,inner sep=0pt, outer sep=0pt, scale=  0.88] at (191.54, 18.79) {64};

\node[text=drawColor,anchor=base,inner sep=0pt, outer sep=0pt, scale=  0.88] at (217.45, 18.79) {128};
\end{scope}
\begin{scope}
\path[clip] (  0.00,  0.00) rectangle (238.49,115.63);
\definecolor{drawColor}{RGB}{0,0,0}

\node[text=drawColor,anchor=base,inner sep=0pt, outer sep=0pt, scale=  1.00] at (139.73,  7.44) {domain size};
\end{scope}
\begin{scope}
\path[clip] (  0.00,  0.00) rectangle (238.49,115.63);
\definecolor{drawColor}{RGB}{0,0,0}

\node[text=drawColor,rotate= 90.00,anchor=base,inner sep=0pt, outer sep=0pt, scale=  1.00] at ( 12.39, 69.97) {$\alpha$};
\end{scope}
\begin{scope}
\path[clip] (  0.00,  0.00) rectangle (238.49,115.63);

\path[] ( 41.93, 97.41) rectangle (215.15,122.86);
\end{scope}
\begin{scope}
\path[clip] (  0.00,  0.00) rectangle (238.49,115.63);
\definecolor{drawColor}{RGB}{247,192,26}

\path[draw=drawColor,line width= 0.6pt] ( 60.16,104.35) --
	( 60.16,106.52);

\path[draw=drawColor,line width= 0.6pt] ( 60.16,113.75) --
	( 60.16,115.91);
\definecolor{fillColor}{RGB}{255,255,255}

\path[draw=drawColor,line width= 0.6pt,fill=fillColor,fill opacity=0.20] ( 54.74,106.52) rectangle ( 65.58,113.75);

\path[draw=drawColor,line width= 0.6pt] ( 54.74,110.13) --
	( 65.58,110.13);
\end{scope}
\begin{scope}
\path[clip] (  0.00,  0.00) rectangle (238.49,115.63);
\definecolor{drawColor}{RGB}{37,122,164}

\path[draw=drawColor,line width= 0.6pt] (118.23,104.35) --
	(118.23,106.52);

\path[draw=drawColor,line width= 0.6pt] (118.23,113.75) --
	(118.23,115.91);
\definecolor{fillColor}{RGB}{255,255,255}

\path[draw=drawColor,line width= 0.6pt,fill=fillColor,fill opacity=0.20] (112.81,106.52) rectangle (123.65,113.75);

\path[draw=drawColor,line width= 0.6pt] (112.81,110.13) --
	(123.65,110.13);
\end{scope}
\begin{scope}
\path[clip] (  0.00,  0.00) rectangle (238.49,115.63);
\definecolor{drawColor}{RGB}{78,155,133}

\path[draw=drawColor,line width= 0.6pt] (172.30,104.35) --
	(172.30,106.52);

\path[draw=drawColor,line width= 0.6pt] (172.30,113.75) --
	(172.30,115.91);
\definecolor{fillColor}{RGB}{255,255,255}

\path[draw=drawColor,line width= 0.6pt,fill=fillColor,fill opacity=0.20] (166.88,106.52) rectangle (177.72,113.75);

\path[draw=drawColor,line width= 0.6pt] (166.88,110.13) --
	(177.72,110.13);
\end{scope}
\begin{scope}
\path[clip] (  0.00,  0.00) rectangle (238.49,115.63);
\definecolor{drawColor}{RGB}{0,0,0}

\node[text=drawColor,anchor=base west,inner sep=0pt, outer sep=0pt, scale=  0.80] at ( 72.88,107.38) {$\varepsilon=0.001$};
\end{scope}
\begin{scope}
\path[clip] (  0.00,  0.00) rectangle (238.49,115.63);
\definecolor{drawColor}{RGB}{0,0,0}

\node[text=drawColor,anchor=base west,inner sep=0pt, outer sep=0pt, scale=  0.80] at (130.96,107.38) {$\varepsilon=0.01$};
\end{scope}
\begin{scope}
\path[clip] (  0.00,  0.00) rectangle (238.49,115.63);
\definecolor{drawColor}{RGB}{0,0,0}

\node[text=drawColor,anchor=base west,inner sep=0pt, outer sep=0pt, scale=  0.80] at (185.03,107.38) {$\varepsilon=0.1$};
\end{scope}
\end{tikzpicture}

%% file: files/plot-offline-p=0.9.tex
\begin{tikzpicture}[x=1pt,y=1pt]
\definecolor{fillColor}{RGB}{255,255,255}
\path[use as bounding box,fill=fillColor,fill opacity=0.00] (0,0) rectangle (238.49,115.63);
\begin{scope}
\path[clip] (  0.00,  0.00) rectangle (238.49,115.63);
\definecolor{drawColor}{RGB}{255,255,255}
\definecolor{fillColor}{RGB}{255,255,255}

\path[draw=drawColor,line width= 0.6pt,line join=round,line cap=round,fill=fillColor] (  0.00,  0.00) rectangle (238.49,115.63);
\end{scope}
\begin{scope}
\path[clip] ( 42.07, 29.80) rectangle (232.99,110.13);
\definecolor{fillColor}{RGB}{255,255,255}

\path[fill=fillColor] ( 42.07, 29.80) rectangle (232.99,110.13);
\definecolor{drawColor}{RGB}{247,192,26}

\path[draw=drawColor,line width= 0.6pt,line join=round] ( 51.35, 61.38) -- ( 51.35, 63.33);

\path[draw=drawColor,line width= 0.6pt,line join=round] ( 51.35, 34.93) -- ( 51.35, 33.45);
\definecolor{fillColor}{RGB}{255,255,255}

\path[draw=drawColor,line width= 0.6pt,fill=fillColor,fill opacity=0.20] ( 48.37, 61.38) --
	( 48.37, 34.93) --
	( 54.33, 34.93) --
	( 54.33, 61.38) --
	( 48.37, 61.38) --
	cycle;

\path[draw=drawColor,line width= 1.1pt] ( 48.37, 48.26) -- ( 54.33, 48.26);
\definecolor{drawColor}{RGB}{37,122,164}

\path[draw=drawColor,line width= 0.6pt,line join=round] ( 57.98, 42.84) -- ( 57.98, 45.30);

\path[draw=drawColor,line width= 0.6pt,line join=round] ( 57.98, 33.45) -- ( 57.98, 33.45);

\path[draw=drawColor,line width= 0.6pt,fill=fillColor,fill opacity=0.20] ( 55.00, 42.84) --
	( 55.00, 33.45) --
	( 60.96, 33.45) --
	( 60.96, 42.84) --
	( 55.00, 42.84) --
	cycle;

\path[draw=drawColor,line width= 1.1pt] ( 55.00, 39.38) -- ( 60.96, 39.38);
\definecolor{drawColor}{RGB}{78,155,133}

\path[draw=drawColor,line width= 0.6pt,line join=round] ( 64.61, 55.33) -- ( 64.61, 59.47);

\path[draw=drawColor,line width= 0.6pt,line join=round] ( 64.61, 37.90) -- ( 64.61, 33.45);

\path[draw=drawColor,line width= 0.6pt,fill=fillColor,fill opacity=0.20] ( 61.63, 55.33) --
	( 61.63, 37.90) --
	( 67.59, 37.90) --
	( 67.59, 55.33) --
	( 61.63, 55.33) --
	cycle;

\path[draw=drawColor,line width= 1.1pt] ( 61.63, 46.66) -- ( 67.59, 46.66);
\definecolor{drawColor}{RGB}{247,192,26}

\path[draw=drawColor,line width= 0.6pt,line join=round] ( 77.87, 54.69) -- ( 77.87, 68.30);

\path[draw=drawColor,line width= 0.6pt,line join=round] ( 77.87, 44.07) -- ( 77.87, 39.38);

\path[draw=drawColor,line width= 0.6pt,fill=fillColor,fill opacity=0.20] ( 74.89, 54.69) --
	( 74.89, 44.07) --
	( 80.85, 44.07) --
	( 80.85, 54.69) --
	( 74.89, 54.69) --
	cycle;

\path[draw=drawColor,line width= 1.1pt] ( 74.89, 47.21) -- ( 80.85, 47.21);
\definecolor{drawColor}{RGB}{37,122,164}

\path[draw=drawColor,line width= 0.6pt,line join=round] ( 84.50, 54.72) -- ( 84.50, 67.54);

\path[draw=drawColor,line width= 0.6pt,line join=round] ( 84.50, 42.84) -- ( 84.50, 39.38);

\path[draw=drawColor,line width= 0.6pt,fill=fillColor,fill opacity=0.20] ( 81.51, 54.72) --
	( 81.51, 42.84) --
	( 87.48, 42.84) --
	( 87.48, 54.72) --
	( 81.51, 54.72) --
	cycle;

\path[draw=drawColor,line width= 1.1pt] ( 81.51, 45.30) -- ( 87.48, 45.30);
\definecolor{drawColor}{RGB}{78,155,133}
\definecolor{fillColor}{RGB}{78,155,133}

\path[draw=drawColor,draw opacity=0.20,line width= 0.4pt,line join=round,line cap=round,fill=fillColor,fill opacity=0.20] ( 91.13, 62.52) circle (  1.96);
\definecolor{drawColor}{RGB}{78,155,133}

\path[draw=drawColor,line width= 0.6pt,line join=round] ( 91.13, 53.13) -- ( 91.13, 58.61);

\path[draw=drawColor,line width= 0.6pt,line join=round] ( 91.13, 47.21) -- ( 91.13, 42.84);
\definecolor{fillColor}{RGB}{255,255,255}

\path[draw=drawColor,line width= 0.6pt,fill=fillColor,fill opacity=0.20] ( 88.14, 53.13) --
	( 88.14, 47.21) --
	( 94.11, 47.21) --
	( 94.11, 53.13) --
	( 88.14, 53.13) --
	cycle;

\path[draw=drawColor,line width= 1.1pt] ( 88.14, 52.23) -- ( 94.11, 52.23);
\definecolor{drawColor}{RGB}{247,192,26}
\definecolor{fillColor}{RGB}{247,192,26}

\path[draw=drawColor,draw opacity=0.20,line width= 0.4pt,line join=round,line cap=round,fill=fillColor,fill opacity=0.20] (104.39, 92.58) circle (  1.96);
\definecolor{drawColor}{RGB}{247,192,26}

\path[draw=drawColor,line width= 0.6pt,line join=round] (104.39, 65.18) -- (104.39, 71.81);

\path[draw=drawColor,line width= 0.6pt,line join=round] (104.39, 53.13) -- (104.39, 50.08);
\definecolor{fillColor}{RGB}{255,255,255}

\path[draw=drawColor,line width= 0.6pt,fill=fillColor,fill opacity=0.20] (101.40, 65.18) --
	(101.40, 53.13) --
	(107.37, 53.13) --
	(107.37, 65.18) --
	(101.40, 65.18) --
	cycle;

\path[draw=drawColor,line width= 1.1pt] (101.40, 58.60) -- (107.37, 58.60);
\definecolor{drawColor}{RGB}{37,122,164}
\definecolor{fillColor}{RGB}{37,122,164}

\path[draw=drawColor,draw opacity=0.20,line width= 0.4pt,line join=round,line cap=round,fill=fillColor,fill opacity=0.20] (111.01, 81.82) circle (  1.96);
\definecolor{drawColor}{RGB}{37,122,164}

\path[draw=drawColor,line width= 0.6pt,line join=round] (111.01, 61.66) -- (111.01, 64.31);

\path[draw=drawColor,line width= 0.6pt,line join=round] (111.01, 52.68) -- (111.01, 52.23);
\definecolor{fillColor}{RGB}{255,255,255}

\path[draw=drawColor,line width= 0.6pt,fill=fillColor,fill opacity=0.20] (108.03, 61.66) --
	(108.03, 52.68) --
	(114.00, 52.68) --
	(114.00, 61.66) --
	(108.03, 61.66) --
	cycle;

\path[draw=drawColor,line width= 1.1pt] (108.03, 56.01) -- (114.00, 56.01);
\definecolor{drawColor}{RGB}{78,155,133}
\definecolor{fillColor}{RGB}{78,155,133}

\path[draw=drawColor,draw opacity=0.20,line width= 0.4pt,line join=round,line cap=round,fill=fillColor,fill opacity=0.20] (117.64, 74.57) circle (  1.96);
\definecolor{drawColor}{RGB}{78,155,133}

\path[draw=drawColor,line width= 0.6pt,line join=round] (117.64, 61.56) -- (117.64, 68.58);

\path[draw=drawColor,line width= 0.6pt,line join=round] (117.64, 53.13) -- (117.64, 47.21);
\definecolor{fillColor}{RGB}{255,255,255}

\path[draw=drawColor,line width= 0.6pt,fill=fillColor,fill opacity=0.20] (114.66, 61.56) --
	(114.66, 53.13) --
	(120.63, 53.13) --
	(120.63, 61.56) --
	(114.66, 61.56) --
	cycle;

\path[draw=drawColor,line width= 1.1pt] (114.66, 53.94) -- (120.63, 53.94);
\definecolor{drawColor}{RGB}{247,192,26}

\path[draw=drawColor,line width= 0.6pt,line join=round] (130.90, 60.61) -- (130.90, 68.30);

\path[draw=drawColor,line width= 0.6pt,line join=round] (130.90, 54.69) -- (130.90, 51.22);

\path[draw=drawColor,line width= 0.6pt,fill=fillColor,fill opacity=0.20] (127.92, 60.61) --
	(127.92, 54.69) --
	(133.88, 54.69) --
	(133.88, 60.61) --
	(127.92, 60.61) --
	cycle;

\path[draw=drawColor,line width= 1.1pt] (127.92, 55.37) -- (133.88, 55.37);
\definecolor{drawColor}{RGB}{37,122,164}

\path[draw=drawColor,line width= 0.6pt,line join=round] (137.53, 66.17) -- (137.53, 68.44);

\path[draw=drawColor,line width= 0.6pt,line join=round] (137.53, 53.94) -- (137.53, 51.22);

\path[draw=drawColor,line width= 0.6pt,fill=fillColor,fill opacity=0.20] (134.55, 66.17) --
	(134.55, 53.94) --
	(140.51, 53.94) --
	(140.51, 66.17) --
	(134.55, 66.17) --
	cycle;

\path[draw=drawColor,line width= 1.1pt] (134.55, 54.69) -- (140.51, 54.69);
\definecolor{drawColor}{RGB}{78,155,133}
\definecolor{fillColor}{RGB}{78,155,133}

\path[draw=drawColor,draw opacity=0.20,line width= 0.4pt,line join=round,line cap=round,fill=fillColor,fill opacity=0.20] (144.16, 94.48) circle (  1.96);
\definecolor{drawColor}{RGB}{78,155,133}

\path[draw=drawColor,line width= 0.6pt,line join=round] (144.16, 65.39) -- (144.16, 66.53);

\path[draw=drawColor,line width= 0.6pt,line join=round] (144.16, 53.13) -- (144.16, 51.22);
\definecolor{fillColor}{RGB}{255,255,255}

\path[draw=drawColor,line width= 0.6pt,fill=fillColor,fill opacity=0.20] (141.18, 65.39) --
	(141.18, 53.13) --
	(147.14, 53.13) --
	(147.14, 65.39) --
	(141.18, 65.39) --
	cycle;

\path[draw=drawColor,line width= 1.1pt] (141.18, 55.37) -- (147.14, 55.37);
\definecolor{drawColor}{RGB}{247,192,26}

\path[draw=drawColor,line width= 0.6pt,line join=round] (157.42, 84.63) -- (157.42,106.48);

\path[draw=drawColor,line width= 0.6pt,line join=round] (157.42, 54.69) -- (157.42, 53.94);

\path[draw=drawColor,line width= 0.6pt,fill=fillColor,fill opacity=0.20] (154.44, 84.63) --
	(154.44, 54.69) --
	(160.40, 54.69) --
	(160.40, 84.63) --
	(154.44, 84.63) --
	cycle;

\path[draw=drawColor,line width= 1.1pt] (154.44, 57.15) -- (160.40, 57.15);
\definecolor{drawColor}{RGB}{37,122,164}

\path[draw=drawColor,line width= 0.6pt,line join=round] (164.05, 88.40) -- (164.05, 93.34);

\path[draw=drawColor,line width= 0.6pt,line join=round] (164.05, 55.37) -- (164.05, 52.23);

\path[draw=drawColor,line width= 0.6pt,fill=fillColor,fill opacity=0.20] (161.06, 88.40) --
	(161.06, 55.37) --
	(167.03, 55.37) --
	(167.03, 88.40) --
	(161.06, 88.40) --
	cycle;

\path[draw=drawColor,line width= 1.1pt] (161.06, 59.05) -- (167.03, 59.05);
\definecolor{drawColor}{RGB}{78,155,133}

\path[draw=drawColor,line width= 0.6pt,line join=round] (170.68, 85.83) -- (170.68, 91.25);

\path[draw=drawColor,line width= 0.6pt,line join=round] (170.68, 53.94) -- (170.68, 53.13);

\path[draw=drawColor,line width= 0.6pt,fill=fillColor,fill opacity=0.20] (167.69, 85.83) --
	(167.69, 53.94) --
	(173.66, 53.94) --
	(173.66, 85.83) --
	(167.69, 85.83) --
	cycle;

\path[draw=drawColor,line width= 1.1pt] (167.69, 57.66) -- (173.66, 57.66);
\definecolor{drawColor}{RGB}{247,192,26}

\path[draw=drawColor,line width= 0.6pt,line join=round] (183.94, 55.37) -- (183.94, 59.47);

\path[draw=drawColor,line width= 0.6pt,line join=round] (183.94, 52.23) -- (183.94, 50.08);

\path[draw=drawColor,line width= 0.6pt,fill=fillColor,fill opacity=0.20] (180.95, 55.37) --
	(180.95, 52.23) --
	(186.92, 52.23) --
	(186.92, 55.37) --
	(180.95, 55.37) --
	cycle;

\path[draw=drawColor,line width= 1.1pt] (180.95, 53.13) -- (186.92, 53.13);
\definecolor{drawColor}{RGB}{37,122,164}

\path[draw=drawColor,line width= 0.6pt,line join=round] (190.56, 55.37) -- (190.56, 58.15);

\path[draw=drawColor,line width= 0.6pt,line join=round] (190.56, 52.23) -- (190.56, 50.08);

\path[draw=drawColor,line width= 0.6pt,fill=fillColor,fill opacity=0.20] (187.58, 55.37) --
	(187.58, 52.23) --
	(193.55, 52.23) --
	(193.55, 55.37) --
	(187.58, 55.37) --
	cycle;

\path[draw=drawColor,line width= 1.1pt] (187.58, 53.13) -- (193.55, 53.13);
\definecolor{drawColor}{RGB}{78,155,133}
\definecolor{fillColor}{RGB}{78,155,133}

\path[draw=drawColor,draw opacity=0.20,line width= 0.4pt,line join=round,line cap=round,fill=fillColor,fill opacity=0.20] (197.19, 56.01) circle (  1.96);

\path[draw=drawColor,draw opacity=0.20,line width= 0.4pt,line join=round,line cap=round,fill=fillColor,fill opacity=0.20] (197.19, 56.59) circle (  1.96);

\path[draw=drawColor,draw opacity=0.20,line width= 0.4pt,line join=round,line cap=round,fill=fillColor,fill opacity=0.20] (197.19, 50.08) circle (  1.96);

\path[draw=drawColor,draw opacity=0.20,line width= 0.4pt,line join=round,line cap=round,fill=fillColor,fill opacity=0.20] (197.19, 51.22) circle (  1.96);

\path[draw=drawColor,draw opacity=0.20,line width= 0.4pt,line join=round,line cap=round,fill=fillColor,fill opacity=0.20] (197.19, 58.61) circle (  1.96);
\definecolor{drawColor}{RGB}{78,155,133}

\path[draw=drawColor,line width= 0.6pt,line join=round] (197.19, 53.94) -- (197.19, 53.94);

\path[draw=drawColor,line width= 0.6pt,line join=round] (197.19, 53.13) -- (197.19, 52.23);
\definecolor{fillColor}{RGB}{255,255,255}

\path[draw=drawColor,line width= 0.6pt,fill=fillColor,fill opacity=0.20] (194.21, 53.94) --
	(194.21, 53.13) --
	(200.18, 53.13) --
	(200.18, 53.94) --
	(194.21, 53.94) --
	cycle;

\path[draw=drawColor,line width= 1.1pt] (194.21, 53.13) -- (200.18, 53.13);
\definecolor{drawColor}{RGB}{247,192,26}

\path[draw=drawColor,line width= 0.6pt,line join=round] (210.45, 58.15) -- (210.45, 60.96);

\path[draw=drawColor,line width= 0.6pt,line join=round] (210.45, 52.23) -- (210.45, 51.22);

\path[draw=drawColor,line width= 0.6pt,fill=fillColor,fill opacity=0.20] (207.47, 58.15) --
	(207.47, 52.23) --
	(213.43, 52.23) --
	(213.43, 58.15) --
	(207.47, 58.15) --
	cycle;

\path[draw=drawColor,line width= 1.1pt] (207.47, 56.01) -- (213.43, 56.01);
\definecolor{drawColor}{RGB}{37,122,164}

\path[draw=drawColor,line width= 0.6pt,line join=round] (217.08, 56.01) -- (217.08, 61.29);

\path[draw=drawColor,line width= 0.6pt,line join=round] (217.08, 52.23) -- (217.08, 51.22);

\path[draw=drawColor,line width= 0.6pt,fill=fillColor,fill opacity=0.20] (214.10, 56.01) --
	(214.10, 52.23) --
	(220.06, 52.23) --
	(220.06, 56.01) --
	(214.10, 56.01) --
	cycle;

\path[draw=drawColor,line width= 1.1pt] (214.10, 53.94) -- (220.06, 53.94);
\definecolor{drawColor}{RGB}{78,155,133}

\path[draw=drawColor,line width= 0.6pt,line join=round] (223.71, 58.15) -- (223.71, 60.96);

\path[draw=drawColor,line width= 0.6pt,line join=round] (223.71, 53.94) -- (223.71, 52.23);

\path[draw=drawColor,line width= 0.6pt,fill=fillColor,fill opacity=0.20] (220.73, 58.15) --
	(220.73, 53.94) --
	(226.69, 53.94) --
	(226.69, 58.15) --
	(220.73, 58.15) --
	cycle;

\path[draw=drawColor,line width= 1.1pt] (220.73, 54.69) -- (226.69, 54.69);
\end{scope}
\begin{scope}
\path[clip] (  0.00,  0.00) rectangle (238.49,115.63);
\definecolor{drawColor}{RGB}{0,0,0}

\path[draw=drawColor,line width= 0.6pt,line join=round] ( 42.07, 29.80) --
	( 42.07,110.13);

\path[draw=drawColor,line width= 0.6pt,line join=round] ( 43.49,107.67) --
	( 42.07,110.13) --
	( 40.65,107.67);
\end{scope}
\begin{scope}
\path[clip] (  0.00,  0.00) rectangle (238.49,115.63);
\definecolor{drawColor}{gray}{0.30}

\node[text=drawColor,anchor=base east,inner sep=0pt, outer sep=0pt, scale=  0.88] at ( 37.12, 30.42) { 0.01};

\node[text=drawColor,anchor=base east,inner sep=0pt, outer sep=0pt, scale=  0.88] at ( 37.12, 50.10) { 0.10};

\node[text=drawColor,anchor=base east,inner sep=0pt, outer sep=0pt, scale=  0.88] at ( 37.12, 69.78) { 1.00};

\node[text=drawColor,anchor=base east,inner sep=0pt, outer sep=0pt, scale=  0.88] at ( 37.12, 89.45) {10.00};
\end{scope}
\begin{scope}
\path[clip] (  0.00,  0.00) rectangle (238.49,115.63);
\definecolor{drawColor}{gray}{0.20}

\path[draw=drawColor,line width= 0.6pt,line join=round] ( 39.32, 33.45) --
	( 42.07, 33.45);

\path[draw=drawColor,line width= 0.6pt,line join=round] ( 39.32, 53.13) --
	( 42.07, 53.13);

\path[draw=drawColor,line width= 0.6pt,line join=round] ( 39.32, 72.81) --
	( 42.07, 72.81);

\path[draw=drawColor,line width= 0.6pt,line join=round] ( 39.32, 92.48) --
	( 42.07, 92.48);
\end{scope}
\begin{scope}
\path[clip] (  0.00,  0.00) rectangle (238.49,115.63);
\definecolor{drawColor}{RGB}{0,0,0}

\path[draw=drawColor,line width= 0.6pt,line join=round] ( 42.07, 29.80) --
	(232.99, 29.80);

\path[draw=drawColor,line width= 0.6pt,line join=round] (230.53, 28.38) --
	(232.99, 29.80) --
	(230.53, 31.23);
\end{scope}
\begin{scope}
\path[clip] (  0.00,  0.00) rectangle (238.49,115.63);
\definecolor{drawColor}{gray}{0.20}

\path[draw=drawColor,line width= 0.6pt,line join=round] ( 57.98, 27.05) --
	( 57.98, 29.80);

\path[draw=drawColor,line width= 0.6pt,line join=round] ( 84.50, 27.05) --
	( 84.50, 29.80);

\path[draw=drawColor,line width= 0.6pt,line join=round] (111.01, 27.05) --
	(111.01, 29.80);

\path[draw=drawColor,line width= 0.6pt,line join=round] (137.53, 27.05) --
	(137.53, 29.80);

\path[draw=drawColor,line width= 0.6pt,line join=round] (164.05, 27.05) --
	(164.05, 29.80);

\path[draw=drawColor,line width= 0.6pt,line join=round] (190.56, 27.05) --
	(190.56, 29.80);

\path[draw=drawColor,line width= 0.6pt,line join=round] (217.08, 27.05) --
	(217.08, 29.80);
\end{scope}
\begin{scope}
\path[clip] (  0.00,  0.00) rectangle (238.49,115.63);
\definecolor{drawColor}{gray}{0.30}

\node[text=drawColor,anchor=base,inner sep=0pt, outer sep=0pt, scale=  0.88] at ( 57.98, 18.79) {2};

\node[text=drawColor,anchor=base,inner sep=0pt, outer sep=0pt, scale=  0.88] at ( 84.50, 18.79) {4};

\node[text=drawColor,anchor=base,inner sep=0pt, outer sep=0pt, scale=  0.88] at (111.01, 18.79) {8};

\node[text=drawColor,anchor=base,inner sep=0pt, outer sep=0pt, scale=  0.88] at (137.53, 18.79) {16};

\node[text=drawColor,anchor=base,inner sep=0pt, outer sep=0pt, scale=  0.88] at (164.05, 18.79) {32};

\node[text=drawColor,anchor=base,inner sep=0pt, outer sep=0pt, scale=  0.88] at (190.56, 18.79) {64};

\node[text=drawColor,anchor=base,inner sep=0pt, outer sep=0pt, scale=  0.88] at (217.08, 18.79) {128};
\end{scope}
\begin{scope}
\path[clip] (  0.00,  0.00) rectangle (238.49,115.63);
\definecolor{drawColor}{RGB}{0,0,0}

\node[text=drawColor,anchor=base,inner sep=0pt, outer sep=0pt, scale=  1.00] at (137.53,  7.44) {domain size};
\end{scope}
\begin{scope}
\path[clip] (  0.00,  0.00) rectangle (238.49,115.63);
\definecolor{drawColor}{RGB}{0,0,0}

\node[text=drawColor,rotate= 90.00,anchor=base,inner sep=0pt, outer sep=0pt, scale=  1.00] at ( 12.39, 69.97) {$\alpha$};
\end{scope}
\begin{scope}
\path[clip] (  0.00,  0.00) rectangle (238.49,115.63);

\path[] ( 39.47, 97.41) rectangle (212.69,122.86);
\end{scope}
\begin{scope}
\path[clip] (  0.00,  0.00) rectangle (238.49,115.63);
\definecolor{drawColor}{RGB}{247,192,26}

\path[draw=drawColor,line width= 0.6pt] ( 57.69,104.35) --
	( 57.69,106.52);

\path[draw=drawColor,line width= 0.6pt] ( 57.69,113.75) --
	( 57.69,115.91);
\definecolor{fillColor}{RGB}{255,255,255}

\path[draw=drawColor,line width= 0.6pt,fill=fillColor,fill opacity=0.20] ( 52.27,106.52) rectangle ( 63.11,113.75);

\path[draw=drawColor,line width= 0.6pt] ( 52.27,110.13) --
	( 63.11,110.13);
\end{scope}
\begin{scope}
\path[clip] (  0.00,  0.00) rectangle (238.49,115.63);
\definecolor{drawColor}{RGB}{37,122,164}

\path[draw=drawColor,line width= 0.6pt] (115.77,104.35) --
	(115.77,106.52);

\path[draw=drawColor,line width= 0.6pt] (115.77,113.75) --
	(115.77,115.91);
\definecolor{fillColor}{RGB}{255,255,255}

\path[draw=drawColor,line width= 0.6pt,fill=fillColor,fill opacity=0.20] (110.35,106.52) rectangle (121.19,113.75);

\path[draw=drawColor,line width= 0.6pt] (110.35,110.13) --
	(121.19,110.13);
\end{scope}
\begin{scope}
\path[clip] (  0.00,  0.00) rectangle (238.49,115.63);
\definecolor{drawColor}{RGB}{78,155,133}

\path[draw=drawColor,line width= 0.6pt] (169.84,104.35) --
	(169.84,106.52);

\path[draw=drawColor,line width= 0.6pt] (169.84,113.75) --
	(169.84,115.91);
\definecolor{fillColor}{RGB}{255,255,255}

\path[draw=drawColor,line width= 0.6pt,fill=fillColor,fill opacity=0.20] (164.42,106.52) rectangle (175.26,113.75);

\path[draw=drawColor,line width= 0.6pt] (164.42,110.13) --
	(175.26,110.13);
\end{scope}
\begin{scope}
\path[clip] (  0.00,  0.00) rectangle (238.49,115.63);
\definecolor{drawColor}{RGB}{0,0,0}

\node[text=drawColor,anchor=base west,inner sep=0pt, outer sep=0pt, scale=  0.80] at ( 70.42,107.38) {$\varepsilon=0.001$};
\end{scope}
\begin{scope}
\path[clip] (  0.00,  0.00) rectangle (238.49,115.63);
\definecolor{drawColor}{RGB}{0,0,0}

\node[text=drawColor,anchor=base west,inner sep=0pt, outer sep=0pt, scale=  0.80] at (128.49,107.38) {$\varepsilon=0.01$};
\end{scope}
\begin{scope}
\path[clip] (  0.00,  0.00) rectangle (238.49,115.63);
\definecolor{drawColor}{RGB}{0,0,0}

\node[text=drawColor,anchor=base west,inner sep=0pt, outer sep=0pt, scale=  0.80] at (182.57,107.38) {$\varepsilon=0.1$};
\end{scope}
\end{tikzpicture}

%% file: files/plot-offline-p=1.tex
\begin{tikzpicture}[x=1pt,y=1pt]
\definecolor{fillColor}{RGB}{255,255,255}
\path[use as bounding box,fill=fillColor,fill opacity=0.00] (0,0) rectangle (238.49,115.63);
\begin{scope}
\path[clip] (  0.00,  0.00) rectangle (238.49,115.63);
\definecolor{drawColor}{RGB}{255,255,255}
\definecolor{fillColor}{RGB}{255,255,255}

\path[draw=drawColor,line width= 0.6pt,line join=round,line cap=round,fill=fillColor] (  0.00,  0.00) rectangle (238.49,115.63);
\end{scope}
\begin{scope}
\path[clip] ( 46.47, 29.80) rectangle (232.99,110.13);
\definecolor{fillColor}{RGB}{255,255,255}

\path[fill=fillColor] ( 46.47, 29.80) rectangle (232.99,110.13);
\definecolor{drawColor}{RGB}{247,192,26}

\path[draw=drawColor,line width= 0.6pt,line join=round] ( 55.54, 37.76) -- ( 55.54, 40.28);

\path[draw=drawColor,line width= 0.6pt,line join=round] ( 55.54, 33.45) -- ( 55.54, 33.45);
\definecolor{fillColor}{RGB}{255,255,255}

\path[draw=drawColor,line width= 0.6pt,fill=fillColor,fill opacity=0.20] ( 52.62, 37.76) --
	( 52.62, 33.45) --
	( 58.45, 33.45) --
	( 58.45, 37.76) --
	( 52.62, 37.76) --
	cycle;

\path[draw=drawColor,line width= 1.1pt] ( 52.62, 35.61) -- ( 58.45, 35.61);
\definecolor{drawColor}{RGB}{37,122,164}

\path[draw=drawColor,line width= 0.6pt,line join=round] ( 62.01, 46.57) -- ( 62.01, 62.50);

\path[draw=drawColor,line width= 0.6pt,line join=round] ( 62.01, 35.61) -- ( 62.01, 33.45);

\path[draw=drawColor,line width= 0.6pt,fill=fillColor,fill opacity=0.20] ( 59.10, 46.57) --
	( 59.10, 35.61) --
	( 64.93, 35.61) --
	( 64.93, 46.57) --
	( 59.10, 46.57) --
	cycle;

\path[draw=drawColor,line width= 1.1pt] ( 59.10, 37.76) -- ( 64.93, 37.76);
\definecolor{drawColor}{RGB}{78,155,133}

\path[draw=drawColor,line width= 0.6pt,line join=round] ( 68.49, 49.27) -- ( 68.49, 55.73);

\path[draw=drawColor,line width= 0.6pt,line join=round] ( 68.49, 38.58) -- ( 68.49, 33.45);

\path[draw=drawColor,line width= 0.6pt,fill=fillColor,fill opacity=0.20] ( 65.58, 49.27) --
	( 65.58, 38.58) --
	( 71.40, 38.58) --
	( 71.40, 49.27) --
	( 65.58, 49.27) --
	cycle;

\path[draw=drawColor,line width= 1.1pt] ( 65.58, 43.70) -- ( 71.40, 43.70);
\definecolor{drawColor}{RGB}{247,192,26}
\definecolor{fillColor}{RGB}{247,192,26}

\path[draw=drawColor,draw opacity=0.20,line width= 0.4pt,line join=round,line cap=round,fill=fillColor,fill opacity=0.20] ( 81.44,106.48) circle (  1.96);
\definecolor{drawColor}{RGB}{247,192,26}

\path[draw=drawColor,line width= 0.6pt,line join=round] ( 81.44, 43.96) -- ( 81.44, 44.59);

\path[draw=drawColor,line width= 0.6pt,line join=round] ( 81.44, 40.28) -- ( 81.44, 37.76);
\definecolor{fillColor}{RGB}{255,255,255}

\path[draw=drawColor,line width= 0.6pt,fill=fillColor,fill opacity=0.20] ( 78.53, 43.96) --
	( 78.53, 40.28) --
	( 84.36, 40.28) --
	( 84.36, 43.96) --
	( 78.53, 43.96) --
	cycle;

\path[draw=drawColor,line width= 1.1pt] ( 78.53, 41.18) -- ( 84.36, 41.18);
\definecolor{drawColor}{RGB}{37,122,164}
\definecolor{fillColor}{RGB}{37,122,164}

\path[draw=drawColor,draw opacity=0.20,line width= 0.4pt,line join=round,line cap=round,fill=fillColor,fill opacity=0.20] ( 87.92, 54.60) circle (  1.96);
\definecolor{drawColor}{RGB}{37,122,164}

\path[draw=drawColor,line width= 0.6pt,line join=round] ( 87.92, 46.18) -- ( 87.92, 47.77);

\path[draw=drawColor,line width= 0.6pt,line join=round] ( 87.92, 41.18) -- ( 87.92, 37.76);
\definecolor{fillColor}{RGB}{255,255,255}

\path[draw=drawColor,line width= 0.6pt,fill=fillColor,fill opacity=0.20] ( 85.00, 46.18) --
	( 85.00, 41.18) --
	( 90.83, 41.18) --
	( 90.83, 46.18) --
	( 85.00, 46.18) --
	cycle;

\path[draw=drawColor,line width= 1.1pt] ( 85.00, 43.46) -- ( 90.83, 43.46);
\definecolor{drawColor}{RGB}{78,155,133}
\definecolor{fillColor}{RGB}{78,155,133}

\path[draw=drawColor,draw opacity=0.20,line width= 0.4pt,line join=round,line cap=round,fill=fillColor,fill opacity=0.20] ( 94.40, 37.76) circle (  1.96);
\definecolor{drawColor}{RGB}{78,155,133}

\path[draw=drawColor,line width= 0.6pt,line join=round] ( 94.40, 44.31) -- ( 94.40, 45.55);

\path[draw=drawColor,line width= 0.6pt,line join=round] ( 94.40, 43.46) -- ( 94.40, 43.46);
\definecolor{fillColor}{RGB}{255,255,255}

\path[draw=drawColor,line width= 0.6pt,fill=fillColor,fill opacity=0.20] ( 91.48, 44.31) --
	( 91.48, 43.46) --
	( 97.31, 43.46) --
	( 97.31, 44.31) --
	( 91.48, 44.31) --
	cycle;

\path[draw=drawColor,line width= 1.1pt] ( 91.48, 43.46) -- ( 97.31, 43.46);
\definecolor{drawColor}{RGB}{247,192,26}

\path[draw=drawColor,line width= 0.6pt,line join=round] (107.35, 57.43) -- (107.35, 72.53);

\path[draw=drawColor,line width= 0.6pt,line join=round] (107.35, 47.28) -- (107.35, 46.38);

\path[draw=drawColor,line width= 0.6pt,fill=fillColor,fill opacity=0.20] (104.43, 57.43) --
	(104.43, 47.28) --
	(110.26, 47.28) --
	(110.26, 57.43) --
	(104.43, 57.43) --
	cycle;

\path[draw=drawColor,line width= 1.1pt] (104.43, 47.77) -- (110.26, 47.77);
\definecolor{drawColor}{RGB}{37,122,164}
\definecolor{fillColor}{RGB}{37,122,164}

\path[draw=drawColor,draw opacity=0.20,line width= 0.4pt,line join=round,line cap=round,fill=fillColor,fill opacity=0.20] (113.82, 60.70) circle (  1.96);
\definecolor{drawColor}{RGB}{37,122,164}

\path[draw=drawColor,line width= 0.6pt,line join=round] (113.82, 52.22) -- (113.82, 56.69);

\path[draw=drawColor,line width= 0.6pt,line join=round] (113.82, 47.77) -- (113.82, 46.38);
\definecolor{fillColor}{RGB}{255,255,255}

\path[draw=drawColor,line width= 0.6pt,fill=fillColor,fill opacity=0.20] (110.91, 52.22) --
	(110.91, 47.77) --
	(116.74, 47.77) --
	(116.74, 52.22) --
	(110.91, 52.22) --
	cycle;

\path[draw=drawColor,line width= 1.1pt] (110.91, 48.36) -- (116.74, 48.36);
\definecolor{drawColor}{RGB}{78,155,133}

\path[draw=drawColor,line width= 0.6pt,line join=round] (120.30, 56.26) -- (120.30, 68.02);

\path[draw=drawColor,line width= 0.6pt,line join=round] (120.30, 48.36) -- (120.30, 45.55);

\path[draw=drawColor,line width= 0.6pt,fill=fillColor,fill opacity=0.20] (117.39, 56.26) --
	(117.39, 48.36) --
	(123.22, 48.36) --
	(123.22, 56.26) --
	(117.39, 56.26) --
	cycle;

\path[draw=drawColor,line width= 1.1pt] (117.39, 49.63) -- (123.22, 49.63);
\definecolor{drawColor}{RGB}{247,192,26}

\path[draw=drawColor,line width= 0.6pt,line join=round] (133.25, 53.94) -- (133.25, 56.39);

\path[draw=drawColor,line width= 0.6pt,line join=round] (133.25, 46.17) -- (133.25, 43.46);

\path[draw=drawColor,line width= 0.6pt,fill=fillColor,fill opacity=0.20] (130.34, 53.94) --
	(130.34, 46.17) --
	(136.17, 46.17) --
	(136.17, 53.94) --
	(130.34, 53.94) --
	cycle;

\path[draw=drawColor,line width= 1.1pt] (130.34, 48.07) -- (136.17, 48.07);
\definecolor{drawColor}{RGB}{37,122,164}
\definecolor{fillColor}{RGB}{37,122,164}

\path[draw=drawColor,draw opacity=0.20,line width= 0.4pt,line join=round,line cap=round,fill=fillColor,fill opacity=0.20] (139.73, 37.76) circle (  1.96);

\path[draw=drawColor,draw opacity=0.20,line width= 0.4pt,line join=round,line cap=round,fill=fillColor,fill opacity=0.20] (139.73, 37.76) circle (  1.96);

\path[draw=drawColor,draw opacity=0.20,line width= 0.4pt,line join=round,line cap=round,fill=fillColor,fill opacity=0.20] (139.73, 57.77) circle (  1.96);

\path[draw=drawColor,draw opacity=0.20,line width= 0.4pt,line join=round,line cap=round,fill=fillColor,fill opacity=0.20] (139.73, 58.14) circle (  1.96);

\path[draw=drawColor,draw opacity=0.20,line width= 0.4pt,line join=round,line cap=round,fill=fillColor,fill opacity=0.20] (139.73, 58.59) circle (  1.96);
\definecolor{drawColor}{RGB}{37,122,164}

\path[draw=drawColor,line width= 0.6pt,line join=round] (139.73, 50.29) -- (139.73, 50.29);

\path[draw=drawColor,line width= 0.6pt,line join=round] (139.73, 45.55) -- (139.73, 40.28);
\definecolor{fillColor}{RGB}{255,255,255}

\path[draw=drawColor,line width= 0.6pt,fill=fillColor,fill opacity=0.20] (136.82, 50.29) --
	(136.82, 45.55) --
	(142.64, 45.55) --
	(142.64, 50.29) --
	(136.82, 50.29) --
	cycle;

\path[draw=drawColor,line width= 1.1pt] (136.82, 47.77) -- (142.64, 47.77);
\definecolor{drawColor}{RGB}{78,155,133}

\path[draw=drawColor,line width= 0.6pt,line join=round] (146.21, 55.43) -- (146.21, 58.25);

\path[draw=drawColor,line width= 0.6pt,line join=round] (146.21, 47.92) -- (146.21, 46.38);

\path[draw=drawColor,line width= 0.6pt,fill=fillColor,fill opacity=0.20] (143.29, 55.43) --
	(143.29, 47.92) --
	(149.12, 47.92) --
	(149.12, 55.43) --
	(143.29, 55.43) --
	cycle;

\path[draw=drawColor,line width= 1.1pt] (143.29, 48.63) -- (149.12, 48.63);
\definecolor{drawColor}{RGB}{247,192,26}

\path[draw=drawColor,line width= 0.6pt,line join=round] (159.16, 48.36) -- (159.16, 52.38);

\path[draw=drawColor,line width= 0.6pt,line join=round] (159.16, 45.55) -- (159.16, 42.07);

\path[draw=drawColor,line width= 0.6pt,fill=fillColor,fill opacity=0.20] (156.25, 48.36) --
	(156.25, 45.55) --
	(162.07, 45.55) --
	(162.07, 48.36) --
	(156.25, 48.36) --
	cycle;

\path[draw=drawColor,line width= 1.1pt] (156.25, 47.11) -- (162.07, 47.11);
\definecolor{drawColor}{RGB}{37,122,164}
\definecolor{fillColor}{RGB}{37,122,164}

\path[draw=drawColor,draw opacity=0.20,line width= 0.4pt,line join=round,line cap=round,fill=fillColor,fill opacity=0.20] (165.64, 40.28) circle (  1.96);

\path[draw=drawColor,draw opacity=0.20,line width= 0.4pt,line join=round,line cap=round,fill=fillColor,fill opacity=0.20] (165.64, 52.08) circle (  1.96);

\path[draw=drawColor,draw opacity=0.20,line width= 0.4pt,line join=round,line cap=round,fill=fillColor,fill opacity=0.20] (165.64, 51.42) circle (  1.96);
\definecolor{drawColor}{RGB}{37,122,164}

\path[draw=drawColor,line width= 0.6pt,line join=round] (165.64, 48.36) -- (165.64, 48.90);

\path[draw=drawColor,line width= 0.6pt,line join=round] (165.64, 46.38) -- (165.64, 46.38);
\definecolor{fillColor}{RGB}{255,255,255}

\path[draw=drawColor,line width= 0.6pt,fill=fillColor,fill opacity=0.20] (162.72, 48.36) --
	(162.72, 46.38) --
	(168.55, 46.38) --
	(168.55, 48.36) --
	(162.72, 48.36) --
	cycle;

\path[draw=drawColor,line width= 1.1pt] (162.72, 47.77) -- (168.55, 47.77);
\definecolor{drawColor}{RGB}{78,155,133}
\definecolor{fillColor}{RGB}{78,155,133}

\path[draw=drawColor,draw opacity=0.20,line width= 0.4pt,line join=round,line cap=round,fill=fillColor,fill opacity=0.20] (172.11, 43.46) circle (  1.96);

\path[draw=drawColor,draw opacity=0.20,line width= 0.4pt,line join=round,line cap=round,fill=fillColor,fill opacity=0.20] (172.11, 51.07) circle (  1.96);

\path[draw=drawColor,draw opacity=0.20,line width= 0.4pt,line join=round,line cap=round,fill=fillColor,fill opacity=0.20] (172.11, 52.08) circle (  1.96);
\definecolor{drawColor}{RGB}{78,155,133}

\path[draw=drawColor,line width= 0.6pt,line join=round] (172.11, 47.77) -- (172.11, 47.77);

\path[draw=drawColor,line width= 0.6pt,line join=round] (172.11, 46.38) -- (172.11, 45.55);
\definecolor{fillColor}{RGB}{255,255,255}

\path[draw=drawColor,line width= 0.6pt,fill=fillColor,fill opacity=0.20] (169.20, 47.77) --
	(169.20, 46.38) --
	(175.03, 46.38) --
	(175.03, 47.77) --
	(169.20, 47.77) --
	cycle;

\path[draw=drawColor,line width= 1.1pt] (169.20, 47.11) -- (175.03, 47.11);
\definecolor{drawColor}{RGB}{247,192,26}

\path[draw=drawColor,line width= 0.6pt,line join=round] (185.07, 49.86) -- (185.07, 51.42);

\path[draw=drawColor,line width= 0.6pt,line join=round] (185.07, 46.38) -- (185.07, 44.59);

\path[draw=drawColor,line width= 0.6pt,fill=fillColor,fill opacity=0.20] (182.15, 49.86) --
	(182.15, 46.38) --
	(187.98, 46.38) --
	(187.98, 49.86) --
	(182.15, 49.86) --
	cycle;

\path[draw=drawColor,line width= 1.1pt] (182.15, 48.90) -- (187.98, 48.90);
\definecolor{drawColor}{RGB}{37,122,164}

\path[draw=drawColor,line width= 0.6pt,line join=round] (191.54, 49.86) -- (191.54, 50.69);

\path[draw=drawColor,line width= 0.6pt,line join=round] (191.54, 47.11) -- (191.54, 43.46);

\path[draw=drawColor,line width= 0.6pt,fill=fillColor,fill opacity=0.20] (188.63, 49.86) --
	(188.63, 47.11) --
	(194.46, 47.11) --
	(194.46, 49.86) --
	(188.63, 49.86) --
	cycle;

\path[draw=drawColor,line width= 1.1pt] (188.63, 47.77) -- (194.46, 47.77);
\definecolor{drawColor}{RGB}{78,155,133}

\path[draw=drawColor,line width= 0.6pt,line join=round] (198.02, 48.90) -- (198.02, 49.40);

\path[draw=drawColor,line width= 0.6pt,line join=round] (198.02, 47.11) -- (198.02, 44.59);

\path[draw=drawColor,line width= 0.6pt,fill=fillColor,fill opacity=0.20] (195.10, 48.90) --
	(195.10, 47.11) --
	(200.93, 47.11) --
	(200.93, 48.90) --
	(195.10, 48.90) --
	cycle;

\path[draw=drawColor,line width= 1.1pt] (195.10, 47.77) -- (200.93, 47.77);
\definecolor{drawColor}{RGB}{247,192,26}
\definecolor{fillColor}{RGB}{247,192,26}

\path[draw=drawColor,draw opacity=0.20,line width= 0.4pt,line join=round,line cap=round,fill=fillColor,fill opacity=0.20] (210.97, 52.38) circle (  1.96);

\path[draw=drawColor,draw opacity=0.20,line width= 0.4pt,line join=round,line cap=round,fill=fillColor,fill opacity=0.20] (210.97, 51.42) circle (  1.96);

\path[draw=drawColor,draw opacity=0.20,line width= 0.4pt,line join=round,line cap=round,fill=fillColor,fill opacity=0.20] (210.97, 51.07) circle (  1.96);

\path[draw=drawColor,draw opacity=0.20,line width= 0.4pt,line join=round,line cap=round,fill=fillColor,fill opacity=0.20] (210.97, 43.46) circle (  1.96);
\definecolor{drawColor}{RGB}{247,192,26}

\path[draw=drawColor,line width= 0.6pt,line join=round] (210.97, 48.36) -- (210.97, 48.36);

\path[draw=drawColor,line width= 0.6pt,line join=round] (210.97, 47.11) -- (210.97, 46.38);
\definecolor{fillColor}{RGB}{255,255,255}

\path[draw=drawColor,line width= 0.6pt,fill=fillColor,fill opacity=0.20] (208.06, 48.36) --
	(208.06, 47.11) --
	(213.89, 47.11) --
	(213.89, 48.36) --
	(208.06, 48.36) --
	cycle;

\path[draw=drawColor,line width= 1.1pt] (208.06, 48.36) -- (213.89, 48.36);
\definecolor{drawColor}{RGB}{37,122,164}
\definecolor{fillColor}{RGB}{37,122,164}

\path[draw=drawColor,draw opacity=0.20,line width= 0.4pt,line join=round,line cap=round,fill=fillColor,fill opacity=0.20] (217.45, 52.08) circle (  1.96);

\path[draw=drawColor,draw opacity=0.20,line width= 0.4pt,line join=round,line cap=round,fill=fillColor,fill opacity=0.20] (217.45, 52.08) circle (  1.96);

\path[draw=drawColor,draw opacity=0.20,line width= 0.4pt,line join=round,line cap=round,fill=fillColor,fill opacity=0.20] (217.45, 43.46) circle (  1.96);
\definecolor{drawColor}{RGB}{37,122,164}

\path[draw=drawColor,line width= 0.6pt,line join=round] (217.45, 49.40) -- (217.45, 49.86);

\path[draw=drawColor,line width= 0.6pt,line join=round] (217.45, 47.77) -- (217.45, 45.55);
\definecolor{fillColor}{RGB}{255,255,255}

\path[draw=drawColor,line width= 0.6pt,fill=fillColor,fill opacity=0.20] (214.53, 49.40) --
	(214.53, 47.77) --
	(220.36, 47.77) --
	(220.36, 49.40) --
	(214.53, 49.40) --
	cycle;

\path[draw=drawColor,line width= 1.1pt] (214.53, 47.77) -- (220.36, 47.77);
\definecolor{drawColor}{RGB}{78,155,133}

\path[draw=drawColor,line width= 0.6pt,line join=round] (223.92, 50.29) -- (223.92, 52.38);

\path[draw=drawColor,line width= 0.6pt,line join=round] (223.92, 47.77) -- (223.92, 44.59);

\path[draw=drawColor,line width= 0.6pt,fill=fillColor,fill opacity=0.20] (221.01, 50.29) --
	(221.01, 47.77) --
	(226.84, 47.77) --
	(226.84, 50.29) --
	(221.01, 50.29) --
	cycle;

\path[draw=drawColor,line width= 1.1pt] (221.01, 47.77) -- (226.84, 47.77);
\end{scope}
\begin{scope}
\path[clip] (  0.00,  0.00) rectangle (238.49,115.63);
\definecolor{drawColor}{RGB}{0,0,0}

\path[draw=drawColor,line width= 0.6pt,line join=round] ( 46.47, 29.80) --
	( 46.47,110.13);

\path[draw=drawColor,line width= 0.6pt,line join=round] ( 47.89,107.67) --
	( 46.47,110.13) --
	( 45.05,107.67);
\end{scope}
\begin{scope}
\path[clip] (  0.00,  0.00) rectangle (238.49,115.63);
\definecolor{drawColor}{gray}{0.30}

\node[text=drawColor,anchor=base east,inner sep=0pt, outer sep=0pt, scale=  0.88] at ( 41.52, 44.74) {   0.1};

\node[text=drawColor,anchor=base east,inner sep=0pt, outer sep=0pt, scale=  0.88] at ( 41.52, 73.37) {  10.0};

\node[text=drawColor,anchor=base east,inner sep=0pt, outer sep=0pt, scale=  0.88] at ( 41.52,102.00) {1000.0};
\end{scope}
\begin{scope}
\path[clip] (  0.00,  0.00) rectangle (238.49,115.63);
\definecolor{drawColor}{gray}{0.20}

\path[draw=drawColor,line width= 0.6pt,line join=round] ( 43.72, 47.77) --
	( 46.47, 47.77);

\path[draw=drawColor,line width= 0.6pt,line join=round] ( 43.72, 76.40) --
	( 46.47, 76.40);

\path[draw=drawColor,line width= 0.6pt,line join=round] ( 43.72,105.03) --
	( 46.47,105.03);
\end{scope}
\begin{scope}
\path[clip] (  0.00,  0.00) rectangle (238.49,115.63);
\definecolor{drawColor}{RGB}{0,0,0}

\path[draw=drawColor,line width= 0.6pt,line join=round] ( 46.47, 29.80) --
	(232.99, 29.80);

\path[draw=drawColor,line width= 0.6pt,line join=round] (230.53, 28.38) --
	(232.99, 29.80) --
	(230.53, 31.23);
\end{scope}
\begin{scope}
\path[clip] (  0.00,  0.00) rectangle (238.49,115.63);
\definecolor{drawColor}{gray}{0.20}

\path[draw=drawColor,line width= 0.6pt,line join=round] ( 62.01, 27.05) --
	( 62.01, 29.80);

\path[draw=drawColor,line width= 0.6pt,line join=round] ( 87.92, 27.05) --
	( 87.92, 29.80);

\path[draw=drawColor,line width= 0.6pt,line join=round] (113.82, 27.05) --
	(113.82, 29.80);

\path[draw=drawColor,line width= 0.6pt,line join=round] (139.73, 27.05) --
	(139.73, 29.80);

\path[draw=drawColor,line width= 0.6pt,line join=round] (165.64, 27.05) --
	(165.64, 29.80);

\path[draw=drawColor,line width= 0.6pt,line join=round] (191.54, 27.05) --
	(191.54, 29.80);

\path[draw=drawColor,line width= 0.6pt,line join=round] (217.45, 27.05) --
	(217.45, 29.80);
\end{scope}
\begin{scope}
\path[clip] (  0.00,  0.00) rectangle (238.49,115.63);
\definecolor{drawColor}{gray}{0.30}

\node[text=drawColor,anchor=base,inner sep=0pt, outer sep=0pt, scale=  0.88] at ( 62.01, 18.79) {2};

\node[text=drawColor,anchor=base,inner sep=0pt, outer sep=0pt, scale=  0.88] at ( 87.92, 18.79) {4};

\node[text=drawColor,anchor=base,inner sep=0pt, outer sep=0pt, scale=  0.88] at (113.82, 18.79) {8};

\node[text=drawColor,anchor=base,inner sep=0pt, outer sep=0pt, scale=  0.88] at (139.73, 18.79) {16};

\node[text=drawColor,anchor=base,inner sep=0pt, outer sep=0pt, scale=  0.88] at (165.64, 18.79) {32};

\node[text=drawColor,anchor=base,inner sep=0pt, outer sep=0pt, scale=  0.88] at (191.54, 18.79) {64};

\node[text=drawColor,anchor=base,inner sep=0pt, outer sep=0pt, scale=  0.88] at (217.45, 18.79) {128};
\end{scope}
\begin{scope}
\path[clip] (  0.00,  0.00) rectangle (238.49,115.63);
\definecolor{drawColor}{RGB}{0,0,0}

\node[text=drawColor,anchor=base,inner sep=0pt, outer sep=0pt, scale=  1.00] at (139.73,  7.44) {domain size};
\end{scope}
\begin{scope}
\path[clip] (  0.00,  0.00) rectangle (238.49,115.63);
\definecolor{drawColor}{RGB}{0,0,0}

\node[text=drawColor,rotate= 90.00,anchor=base,inner sep=0pt, outer sep=0pt, scale=  1.00] at ( 12.39, 69.97) {$\alpha$};
\end{scope}
\begin{scope}
\path[clip] (  0.00,  0.00) rectangle (238.49,115.63);

\path[] ( 41.93, 97.41) rectangle (215.15,122.86);
\end{scope}
\begin{scope}
\path[clip] (  0.00,  0.00) rectangle (238.49,115.63);
\definecolor{drawColor}{RGB}{247,192,26}

\path[draw=drawColor,line width= 0.6pt] ( 60.16,104.35) --
	( 60.16,106.52);

\path[draw=drawColor,line width= 0.6pt] ( 60.16,113.75) --
	( 60.16,115.91);
\definecolor{fillColor}{RGB}{255,255,255}

\path[draw=drawColor,line width= 0.6pt,fill=fillColor,fill opacity=0.20] ( 54.74,106.52) rectangle ( 65.58,113.75);

\path[draw=drawColor,line width= 0.6pt] ( 54.74,110.13) --
	( 65.58,110.13);
\end{scope}
\begin{scope}
\path[clip] (  0.00,  0.00) rectangle (238.49,115.63);
\definecolor{drawColor}{RGB}{37,122,164}

\path[draw=drawColor,line width= 0.6pt] (118.23,104.35) --
	(118.23,106.52);

\path[draw=drawColor,line width= 0.6pt] (118.23,113.75) --
	(118.23,115.91);
\definecolor{fillColor}{RGB}{255,255,255}

\path[draw=drawColor,line width= 0.6pt,fill=fillColor,fill opacity=0.20] (112.81,106.52) rectangle (123.65,113.75);

\path[draw=drawColor,line width= 0.6pt] (112.81,110.13) --
	(123.65,110.13);
\end{scope}
\begin{scope}
\path[clip] (  0.00,  0.00) rectangle (238.49,115.63);
\definecolor{drawColor}{RGB}{78,155,133}

\path[draw=drawColor,line width= 0.6pt] (172.30,104.35) --
	(172.30,106.52);

\path[draw=drawColor,line width= 0.6pt] (172.30,113.75) --
	(172.30,115.91);
\definecolor{fillColor}{RGB}{255,255,255}

\path[draw=drawColor,line width= 0.6pt,fill=fillColor,fill opacity=0.20] (166.88,106.52) rectangle (177.72,113.75);

\path[draw=drawColor,line width= 0.6pt] (166.88,110.13) --
	(177.72,110.13);
\end{scope}
\begin{scope}
\path[clip] (  0.00,  0.00) rectangle (238.49,115.63);
\definecolor{drawColor}{RGB}{0,0,0}

\node[text=drawColor,anchor=base west,inner sep=0pt, outer sep=0pt, scale=  0.80] at ( 72.88,107.38) {$\varepsilon=0.001$};
\end{scope}
\begin{scope}
\path[clip] (  0.00,  0.00) rectangle (238.49,115.63);
\definecolor{drawColor}{RGB}{0,0,0}

\node[text=drawColor,anchor=base west,inner sep=0pt, outer sep=0pt, scale=  0.80] at (130.96,107.38) {$\varepsilon=0.01$};
\end{scope}
\begin{scope}
\path[clip] (  0.00,  0.00) rectangle (238.49,115.63);
\definecolor{drawColor}{RGB}{0,0,0}

\node[text=drawColor,anchor=base west,inner sep=0pt, outer sep=0pt, scale=  0.80] at (185.03,107.38) {$\varepsilon=0.1$};
\end{scope}
\end{tikzpicture}